%% file: PhDThesis.tex
\documentclass[a4paper, 11pt, twoside]{report}  % Use the "Thesis" style, based on the ECS Thesis style by Steve Gunn
\usepackage[style=numeric-comp,sorting=nty,doi=false,isbn=false,url=false,eprint=false,firstinits=true]{biblatex}
\bibliography{../library} % Where journals.bib and phd-references.bib are BibTeX databases

\graphicspath{{Figures/}}  % Location of the graphics files (set up for graphics to be in PDF format)

\usepackage{utopia}
\usepackage{charter}
\usepackage{amsfonts}
\usepackage[T1]{fontenc}        % Use Type 1 encoding
\usepackage{enumitem}
\usepackage{verbatim}   % Needed for the "comment" environment to make LaTeX comments
\usepackage{vector}     % Allows "\bvec{}" and "\buvec{}" for "blackboard" style bold vectors in maths
\usepackage{listings}
\usepackage{hyperref}
\usepackage{breakurl}
\usepackage{tabularx}
\usepackage{array}
\usepackage[table]{xcolor}
\usepackage{multirow}
\usepackage{comment}

\usepackage[unbalanced,totoc]{idxlayout}
\usepackage{makeidx}
\makeindex

\usepackage{anyfontsize}
\usepackage{titlesec}
\titleformat{\chapter}[display]
{\normalfont\Huge\filleft}
{\titlerule[2pt]%
\vspace{0.5pc}%
\bfseries\rmfamily\fontsize{30}{30}\selectfont\chaptertitlename \hspace{0.75cm} \bfseries\rmfamily\fontsize{42}{42}\selectfont\textbf{\thechapter}}
{1pc}
{\titlerule[2pt]
\vspace{0.5pc}%
\raggedright\bfseries\fontsize{19}{19}\selectfont\thispagestyle{empty}}

\usepackage[lined,ruled,linesnumbered]{algorithm2e}

\SetAlFnt{\normalsize}
\SetAlCapFnt{\small}
\SetAlCapNameFnt{\small}
\SetAlCapHSkip{0pt}
\IncMargin{-\parindent}
\makeatletter
\def\hlinewd#1{%
\noalign{\ifnum0=`}\fi\hrule \@height #1 %
\futurelet\reserved@a\@xhline}
\makeatother


\hypersetup{pdfstartview={XYZ null null 1.00}, linkcolor=black, citecolor=red, urlcolor=blue, colorlinks=true}  % Colours hyperlinks in blue
\newcommand{\ie}{{\it i.e. }}
\newcommand{\etal}{{\it et al. }}
\newcommand{\eg}{{\it e.g. }}

\DeclareMathOperator*{\argmin}{arg\,min}

\begin{document}
%\addtotoc{Cover}
\mainmatter
%\frontmatter	  % Begin Roman style (i, ii, iii, iv...) page numbering

% Set up the Title Page
%\addresses  {\deptname\\\univname}
\title{Detecting Events and Patterns in Large-Scale User Generated Textual Streams with Statistical Learning Methods}
\authors{\textbf{Vasileios Lampos}}

\maketitle
%% ----------------------------------------------------------------
\setstretch{1.0}  % It is better to have smaller font and larger line spacing than the other way round

% Define the page headers using the FancyHdr package and set up for one-sided printing
%\fancyhead{}        % Clears all page headers and footers
%\renewcommand{\headrulewidth}{2pt}
%\rhead{\thepage}    % Sets the right side header to show the page number
%\lhead{\thesection}            % Clears the left side page header

%% ----------------------------------------------------------------
% Quotes page
%\addtotoc{Quotes}
\pagestyle{empty}  % No headers or footers for the following pages
\null\vfill
\begin{center}
\large
``\emph{The lunatic is in the hall,\\
the lunatics are in my hall,\\
the paper holds their folded faces to the floor,\\
and every day the paper boy brings more.}\\
\medskip
\emph{And if the dam breaks open many years too soon,\\
and if there is no room upon the hill,\\
and if your head explodes with dark forebodings too,\\
I'll see you on the dark side of the moon.}''\\
\bigskip
\bigskip
\textbf{Pink Floyd} $\sim$ \textbf{Brain Damage}
%\emph{Welcome my son, welcome to the machine.\\
%What did you dream?\\
%It's alright we told you what to dream.}
\end{center}
\vfill\vfill\vfill\vfill\vfill\vfill\null
\clearpage

%% ----------------------------------------------------------------
%% Signed declaration page
\declarationPage{
\thispagestyle{empty}
I declare that the work in this thesis was carried out in accordance with the requirements of the University's Regulations and Code of Practice for Research Degree Programmes and that it has not been submitted for any other academic award.

Except where indicated by specific reference in the text, the work is the candidate's own work. Work done in collaboration with, or with the assistance of, others, is indicated as such.

Any views expressed in the thesis are those of the author.

\begin{flushright}
\textbf{Vasileios Lampos}
\end{flushright}
\bigskip
\medskip
\begin{tabbing}
Signature: \=   .............................. \\ \\ \\
Date: \>        ..............................\\
\end{tabbing}
}
\clearpage  % End of declaration
%% ----------------------------------------------------------------

%% ----------------------------------------------------------------
% The Abstract Page
\setstretch{1.3}
%\addtotoc{Abstract}  % Add the "Abstract" page entry to the Contents
\abstractXXX{
\thispagestyle{empty}
\addtocontents{toc}{\vspace{1em}}  % Add a gap in the Contents, for aesthetics
\emph{A vast amount of textual web streams is influenced by events or phenomena emerging in the real world. The social web forms an excellent modern paradigm, where unstructured user generated content is published on a regular basis and in most occasions is freely distributed. The present Ph.D. Thesis deals with the problem of inferring information -- or patterns in general -- about events emerging in real life based on the contents of this textual stream. We show that it is possible to extract valuable information about social phenomena, such as an epidemic or even rainfall rates, by automatic analysis of the content published in Social Media, and in particular Twitter, using Statistical Machine Learning methods. An important intermediate task regards the formation and identification of features which characterise a target event; we select and use those textual features in several linear, non-linear and hybrid inference approaches achieving a significantly good performance in terms of the applied loss function. By examining further this rich data set, we also propose methods for extracting various types of mood signals revealing how affective norms -- at least within the social web's population -- evolve during the day and how significant events emerging in the real world are influencing them. Lastly, we present some preliminary findings showing several spatiotemporal characteristics of this textual information as well as the potential of using it to tackle tasks such as the prediction of voting intentions.}
}
\clearpage  % Abstract ended, start a new page
%% ----------------------------------------------------------------
%\setstretch{1.3}  % Reset the line-spacing to 1.3 for body text (if it has changed)

%% ----------------------------------------------------------------
%% The Acknowledgements page, for thanking everyone
\acknowledgements{
\thispagestyle{empty}
\addtocontents{toc}{\vspace{1em}}  % Add a gap in the Contents, for aesthetics
I leave this Ph.D. `significantly' older. This is a matter of fact. All other matters -- abstract or not -- surrounding this script as well as the time behind it are going to be products of subjective analysis.

I am grateful to my parents, \textbf{Ioannis Lampos} and \textbf{Sofia Kousidou}, and my brother \textbf{Spyros Lampos}. Every person in this world is bound to its `family' in a healthy and limitless way. Well, `limitless' might sometimes be -- by definition -- unhealthy, but diving properly into this primitive relationship may need another Ph.D. Thesis and surely no `funding' exists for such an adventure.

Next, and, of course, with a significant importance measured, however, on a relatively different scale, I would like to thank \textbf{Prof. Nello Cristianini}, my Ph.D. advisor. Keeping all the good moments in mind, it has been fun working with him; it was as if we were not really working.

I would like to express my gratitude to \textbf{Prof. Peter Flach} -- my internal Ph.D. progress reviewer -- for his quite useful advice and remarks in every single progress review.

I am also indebted to \textbf{Dr. Steve Gregory} for his help in several occasions during and right after the completion of my M.Sc. studies.

I am grateful to \textbf{Dr. Tijl De Bie} and \textbf{Prof. Ricardo Araya} for their contribution in my work. I would also like to thank \textbf{Thomas Lansdall-Welfare} and \textbf{Elena Hensinger} for our short collaboration.

During my time in the university, I came across many interesting people such as the \textbf{porters} of the Merchant Venturers Building. I thank them for being polite and positive every time I was in and out from the building. This might sound strange, but positive energy -- for some people who are by default negative -- can be extremely important.

I would like to thank \textbf{Omar Ali}, \textbf{Syed Rahman} and \textbf{Stefanos Vatsikas} for being honest and helpful in many occasions. I am also grateful to \textbf{Christopher Mims}, the journalist who wrote the first article on our work for MIT's Technology Review. A special `thank you' to \textbf{Dr. Philip J. Naylor} for maintaining our servers and databases; he is the person who once literally saved this project.

This Ph.D. project would not have been possible, if \textbf{EPSRC} did not grant me a DTA scholarship (DTA/SB1826) to cover my tuition fees and without \textbf{NOKIA Research}'s financial support. On the same ground, I would also like to thank the Computer Science Department of the University of Bristol, and its Head at the time \textbf{Prof. Nishan Canagarajah}, for covering parts of my funding in the third year of my Ph.D.; I have to thank my family -- again -- for covering the rest.
%and state my overall disappointment on the `funding issue' as it could have -- without much effort -- been a nonexistent one.

My friends -- amongst other more important things -- were also part of my Ph.D. studies and I `have' to thank them. Therefore, \textbf{Charis Fanourakis}, \textbf{Chlo\'{e} Vastesaeger}, \textbf{Costas Mimis}, \textbf{Leonidas Kapsokalivas}, \textbf{Nick Mavrogiannakis}, \textbf{Takis Tsiatsis} and \textbf{Thanos Ntzoumanis} there you go, you got your names in my `book'.

Finally, I would like to thank all those \textbf{Twitter people} I have been interacting with for their indirect help in making me understand several aspects of this platform better and, of course, for their invaluable company during the endless nights I spent in writing up. I wish them `success' in their life quests, if and only if.
}
\clearpage  % End of the Acknowledgements
%% ----------------------------------------------------------------

\pagestyle{fancy}  %The page style headers have been "empty" all this time, now use the "fancy" headers as defined before to bring them back
%\lhead{\emph{Contents}}  % Set the left side page header to "Contents"
\lhead[Contents]{\textbf{\thepage}}
\rhead[\textbf{\thepage}]{Vasileios Lampos, Ph.D. Thesis}
{
\small
\tableofcontents  % Write out the Table of Contents
}

%% ----------------------------------------------------------------
%\lhead{\emph{List of Figures}}  % Set the left side page header to "List of Figures"
\lhead[List of Figures]{\textbf{\thepage}}
\rhead[\textbf{\thepage}]{Vasileios Lampos, Ph.D. Thesis}
{
\small
\listoffigures
}  % Write out the List of Figures

%% ----------------------------------------------------------------
%\lhead{\emph{List of Tables}}  % Set the left side page header to "List of Tables"
\lhead[List of Tables]{\textbf{\thepage}}
\rhead[\textbf{\thepage}]{Vasileios Lampos, Ph.D. Thesis}
{
\small
\listoftables
}  % Write out the List of Tables

%% ----------------------------------------------------------------
%\setstretch{1.5}  % Set the line spacing to 1.5, this makes the following tables easier to read
\clearpage  % Start a new page
\lhead[Abbreviations]{\textbf{\thepage}}
\rhead[\textbf{\thepage}]{Vasileios Lampos, Ph.D. Thesis}
{
\small
\setstretch{1.3}
\listofsymbols{ll}  % Include a list of Abbreviations (a table of two columns)
{
{\large\textbf{Acronym}}    & {\large\textbf{W}hat (it) \textbf{S}tands \textbf{F}or}\\
\textbf{API}        & \textbf{A}pplication \textbf{P}rogramming \textbf{I}nterface \\
\textbf{Bagging}    & \textbf{B}ootstrap \textbf{Agg}regat\textbf{ing} \\
\textbf{Bolasso}    & \textbf{Bo}otstrap \textbf{L}east \textbf{A}bsolute \textbf{S}hrinkage and \textbf{S}election \textbf{O}perator\\
\textbf{CART}       & \textbf{C}lassification \textbf{A}nd \textbf{R}egression \textbf{T}ree\\
\textbf{CDC}        & \textbf{C}enters for \textbf{D}isease \textbf{C}ontrol (and Prevention)\\
\textbf{CDF}        & \textbf{C}umulative \textbf{D}ensity \textbf{F}unction\\
\textbf{CI}         & \textbf{C}onfidence \textbf{I}nterval\\
\textbf{CT}         & \textbf{C}onsensus \textbf{T}hreshold\\
\textbf{DSC}        & \textbf{D}ominant \textbf{S}entiment \textbf{C}lass\\
\textbf{GP}         & \textbf{G}eneral \textbf{P}ractitioner\\
\textbf{i.i.d.}     & \textbf{I}ndependent and \textbf{I}dentically \textbf{D}istributed\\
\textbf{ILI}        & \textbf{I}nfluenza-\textbf{l}ike \textbf{I}llness\\
\textbf{IR}         & \textbf{I}nformation \textbf{R}etrieval \\
\textbf{IS}         & \textbf{I}mpact \textbf{S}core\\
\textbf{JD}         & \textbf{J}accard \textbf{D}istance\\
\textbf{LAR}        & \textbf{L}east \textbf{A}ngle \textbf{R}egression \\
\textbf{LASSO}      & \textbf{L}east \textbf{A}bsolute \textbf{S}hrinkage and \textbf{S}election \textbf{O}perator\\
\textbf{Lowess}     & \textbf{Lo}cally \textbf{we}ighted \textbf{s}catterplot \textbf{s}moothing\\
\textbf{MAE}        & \textbf{M}ean \textbf{A}bsolute \textbf{E}rror\\
\textbf{MDS}        & \textbf{M}ulti\textbf{d}imensional \textbf{S}caling\\
\textbf{MFMS}       & \textbf{M}ean \textbf{F}requency \textbf{M}ood \textbf{S}coring\\
\textbf{MRE}        & \textbf{M}ean \textbf{R}anking \textbf{E}rror\\
\textbf{MSE}        & \textbf{M}ean \textbf{S}quare \textbf{E}rror\\
\textbf{MSFMS}      & \textbf{M}ean \textbf{S}tandardised \textbf{F}requency \textbf{M}ood \textbf{S}coring\\
\textbf{MTS}        & \textbf{M}ean \textbf{T}hresholded \textbf{S}entiment\\
\textbf{NA}         & \textbf{N}egative \textbf{A}ffectivity\\
\textbf{NLP}        & \textbf{N}atural \textbf{L}anguage \textbf{P}rocessing\\
\textbf{OLS}        & \textbf{O}rdinary \textbf{L}east \textbf{S}quares\\
\textbf{PA}         & \textbf{P}ositive \textbf{A}ffectivity\\
\textbf{PCA}        & \textbf{P}rincipal \textbf{C}omponent \textbf{A}nalysis\\
\textbf{PDF}        & \textbf{P}robability \textbf{D}ensity \textbf{F}unction\\
\textbf{POS}        & \textbf{P}art \textbf{O}f \textbf{S}peech\\
\textbf{RCGP}       & \textbf{R}oyal \textbf{C}ollege of \textbf{G}eneral \textbf{P}ractitioners\\
\textbf{RMSE}       & \textbf{R}oot \textbf{M}ean \textbf{S}quare \textbf{E}rror\\
\textbf{RSS}        & \textbf{R}esidual \textbf{S}um of \textbf{S}quares \\
\textbf{SE}         & \textbf{S}tandard \textbf{E}rror\\
\textbf{SnPOS}      & \textbf{S}entiment \textbf{n}o \textbf{P}art \textbf{O}f \textbf{S}peech (tagging)\\
\textbf{SPOS}       & \textbf{S}entiment \textbf{w}ith \textbf{P}art \textbf{O}f \textbf{S}peech (tagging)\\
\textbf{SPOSW}      & \textbf{S}entiment \textbf{w}ith \textbf{P}art \textbf{O}f \textbf{S}peech (tagging) and \textbf{W}ordNet's core terms\\
\textbf{SS}         & \textbf{S}imilarity \textbf{S}core\\
\textbf{TF-IDF}     & \textbf{T}erm \textbf{F}requency \textbf{I}nverse \textbf{D}ocument \textbf{F}requency \\
\textbf{VSM}        & \textbf{V}ector \textbf{S}pace \textbf{M}odel\\
\textbf{VSR}        & \textbf{V}ector \textbf{S}pace \textbf{R}epresentation\\
\textbf{XML}        & e\textbf{X}tensive \textbf{M}arkup \textbf{L}anguage\\
}
}

%% ----------------------------------------------------------------
%%\clearpage  % Start a new page
%%\lhead{\emph{Physical Constants}}  % Set the left side page header to "Physical Constants"
%%\listofconstants{lrcl}  % Include a list of Physical Constants (a four column table)
%%{
%% Constant Name & Symbol & = & Constant Value (with units) \\
%%Speed of Light & $c$ & $=$ & $2.997\ 924\ 58\times10^{8}\ \mbox{ms}^{-\mbox{s}}$ (exact)\\
%%
%%}

%% ----------------------------------------------------------------
%%\clearpage  %Start a new page
%%\lhead{\emph{Symbols}}  % Set the left side page header to "Symbols"
%%\listofnomenclature{lll}  % Include a list of Symbols (a three column table)
%%{
%% symbol & name & unit \\
%%$a$ & distance & m \\
%%$P$ & power & W (Js$^{-1}$) \\
%%& & \\ % Gap to separate the Roman symbols from the Greek
%%$\omega$ & angular frequency & rads$^{-1}$ \\
%%}
%% ----------------------------------------------------------------
% End of the pre-able, contents and lists of things
% Begin the Dedication page
%%
%\setstretch{1.1}  % Return the line spacing back to 1.3
%%
%\pagestyle{empty}  % Page style needs to be empty for this page
%\dedicatory{Dedicated to my dog with love and passion}
%%
%\addtocontents{toc}{\vspace{2em}}  % Add a gap in the Contents, for aesthetics

%% ----------------------------------------------------------------
%\mainmatter	  % Begin normal, numeric (1,2,3...) page numbering
\pagestyle{fancy}  % Return the page headers back to the "fancy" style

% Include the chapters of the thesis, as separate files
% Just uncomment the lines as you write the chapters
\input{Chapters/Chapter1}   % Introduction
\input{Chapters/Chapter2}   % Background
\input{Chapters/Chapter3}   % Data
\input{Chapters/Chapter4}   % First Steps
\input{Chapters/Chapter5}   % Journal Paper
\input{Chapters/Chapter6}   % Mood
\input{Chapters/Chapter7}   % Patterns
\input{Chapters/Chapter8}   % Applications
\input{Chapters/Chapter9}   % Conclusions

%% ----------------------------------------------------------------
% Now begin the Appendices, including them as separate files

\addtocontents{toc}{\vspace{2em}} % Add a gap in the Contents, for aesthetics
\appendix % Cue to tell LaTeX that the following 'chapters' are Appendices
\input{Appendices/AppendixA}

\input{Appendices/AppendixB}

\addtocontents{toc}{\vspace{2em}}  % Add a gap in the Contents, for aesthetics
\backmatter

%% ----------------------------------------------------------------
\label{Bibliography}
\lhead[Bibliography]{\textbf{\thepage}}
\rhead[\textbf{\thepage}]{Vasileios Lampos, Ph.D. Thesis}
%\lhead{\emph{Bibliography}}  % Change the left side page header to "Bibliography"
%\bibliographystyle{unsrtnat}  % Use the "unsrtnat" BibTeX style for formatting the Bibliography
%\bibliographystyle{alpha}
%\bibliographystyle{ieeetr}
%\bibliographystyle{is-plain}
%\bibliographystyle{amsplain}
%\bibliographystyle{is-abbrv}
%\bibliographystyle{elsart-num}
{
\small
\setstretch{1.5}
%\bibliography{../library}  % The references (bibliography) information are stored in the file named "Bibliography.bib"
\printbibliography
}

\lhead[Index]{\textbf{\thepage}}
\rhead[\textbf{\thepage}]{Vasileios Lampos, Ph.D. Thesis}
\addtocontents{toc}{\vspace{2em}}
{
\setstretch{1.2}
\small
\printindex
}
% \addcontentsline{toc}{chapter}{Index}
\addtocontents{toc}{\vspace{2em}}
\backmatter

\end{document}

%% file: Chapters/Chapter1.tex
\chapter{Introduction}
\label{Chapter:Introduction}
\lhead[\leftmark]{\textbf{\thepage}}
\rhead[\textbf{\thepage}]{Vasileios Lampos, Ph.D. Thesis}

In July 2008, Google reported that the World Wide Web is formed by at least one trillion of unique web pages,\footnote{ Official Google blog, ``We knew the web was big'', July 2008, \url{http://googleblog.blogspot.com/2008/07/we-knew-web-was-big.html}.} approximately one hundred times more than their estimated number in 2005 \cite{gulli2005indexable}. What seemed obvious to everybody was proven in numbers; the World Wide Web has become immense.

Web, also known as `the internet', is definitely one of the most important tools in our everyday life. Numerous of our activities, including education, work, socialising and entertainment are now assisted by or transferred entirely to the `virtual' online space handled by numerous pieces of hardware and software. Together with internet's evolution, the way users interact with it has been evolving as well. For example, while in the past it would have been considered as an abnormal behaviour to give up privacy for socialising in an online medium, nowadays, this is becoming a mainstream way of conducting life \cite{Debatin2009}.

Social networks, such as Twitter or Facebook, apart from boosting the web usage by individuals,\footnote{ Nielsen News, ``Social Networks/Blogs Now Account for One in Every Four and a Half Minutes Online'', June 2010, \url{http://goo.gl/DWOe0}.} also increased significantly the overall amount of information stored on or linked via the web. This fast expansion of the Social Web -- which still is under way -- means that large numbers of people can publish their thoughts at no cost. Current estimates put the number of Facebook users at 845 million \cite{Inc.2012}\footnote{ As of 31/12/2011.} and of Twitter active users at 100 million.\footnote{Twitter Blog, ``One hundred million voices'', 08/09/2011. Link: \url{http://blog.twitter.com/2011/09/one-hundred-million-voices.html}.} The result is a massive stream of digital text that has attracted the attention of marketers \cite{Jansen2009a}, politicians \cite{Diakopoulos2010} and social scientists \cite{Huberman2008}. By analysing the stream of communications in an unmediated way, without relying on questionnaires or interviews, many scientists are having direct access to people's opinions and observations for the first time. Perhaps equally important is that they have access -- although indirectly -- to situations on the ground that affect the web users, such as a spreading infectious disease or extreme weather conditions, as long as these are mentioned in the messages being published.
%%%%%%%

The reader might have noticed that the references in the previous paragraph cite fairly recent work. When we started this project, in September 2008, we were ourselves one of the research groups with the desire to explore this rich amount of information and the aforementioned pieces of work were also on the development phase. Our first and very preliminary findings, presented in ``Weather talk''\index{Weather Talk} \cite{Lampos2008a},\footnote{ ``Weather talk'' is the title of the author's M.Sc. Thesis, written right before the beginning of his Ph.D. project. That work provided interesting insights, but the developed methods described or the data used in the M.Sc. project were not considered in the project presented by this Thesis.} were pointing to the conclusion that even web resources, such as blogs, which are not frequently updated and usually do not reflect real time events, enclosed information that under the right modelling could be used to make inferences about moments in real life; in this particular case, we were making predictions about the weather state of several UK locations based on blog posts and news articles.

One could argue that web information might be unreliable, because it is not always ``officially verified'', and badly structured due to the great variety of ways for organising and publishing content and thus, the task of automatic interpretation is -- by definition -- hard. By considering that we are also dealing with large-scale amounts of information, this interpretation becomes a non human-feasible process.

During this project, we are forming ways for understanding and explaining the textual content that lies on the web. Understanding covers the notion of identifying key properties in the text, which might in turn express domains of knowledge emerging in the real world. Explaining defines the set of functions, which are in the position to map those key properties to proper inferences that relate to events occurring in those domains. Encapsulated in this process is the discovery of patterns, which simplify and merge those large amounts of data into simpler and representable conclusions.

\section{Summary of questions, aims and goals}
\label{section_intro_summary_of_aims}
This research project is applied on the overlapping scientific fields of Artificial Intelligence, Machine Learning, Data Mining, Pattern Recognition and Information Retrieval. Those fields offer a plethora of well developed techniques; we are using the state-of-the-start, but also try to achieve adaptations or improvements in methods, algorithms or frameworks relevant to the context of our work, wherever is needed. In this section, we made an effort to break up the main purpose of this project -- already expressed in the previous section -- into sub-questions or tasks ending up with a more specific and constrained list of aims\index{aims} and goals.

\textbf{Does useful information exist in the `open' web text streams?} While this question might seem simple to address, every answer must be followed by proper justification. A simplistic hypothesis would, of course, favour the case that in such large amounts of textual information, the probability of not encountering subsets with a decreased uncertainty and hence, of high informational value should be very small. Indeed, several works -- including ours -- proved that statement in practice during the early stages of this Ph.D. project. Those findings reshaped the initial question to the following one. % entropy has been removed

\textbf{Supposing that such information does exist, how can it be successfully extracted from the text stream?} The answer to this question is essentially the main point of interest for this Ph.D. project as a substantial part of it has been dedicated in developing methods for deriving conclusions or conducting inferences by processing text. Methodologies are formed by combining or extending models from the fields of Machine Learning, Information Retrieval\index{Information Retrieval|see{IR}} (\textbf{IR}\index{IR}) and Natural Language Processing\index{Natural Language Processing|see{NLP}} (\textbf{NLP}\index{NLP}). The fact that we will be handling massive amounts of data, makes almost certain the application of techniques that aim to reduce the dimensionality of the problem. `Useful' information might consider specific events or more general patterns arising in the population.

\textbf{Is this extracted information a good representation of one -- at least -- fraction of real life?} Measuring the performance of a method is a mainstream task in statistical learning performed by defining one or more loss functions. In this case though, it is necessary to numerically quantify this fraction of real life; it might consider an event that can be measured in a variety of units or an even more abstract notion, such as a public reaction, emotion or opinion on real-time happenings. In a supervised learning framework, the existence of ground truth is essential not only for learning the parameters of a model, but also for testing the inferred learning function. On the other hand, other more abstract scenarios, enclosed in pattern discovery tasks, usually seek confirmation by either proving their statistical significance or by querying `common sense'.

\textbf{Formulate generic methodologies.} In this project we aim to discover generalised methodologies that are not solving only one specific problem, but are applicable to a class of problems. This could be achieved through an experimental process which investigates case studies with different characteristics. A general method should be applicable to a class of problems and thus, such a class must be carefully defined.

\textbf{Improve various subtasks.} Depending on the success in developing methods able to address the aforementioned questions and aims, we should also consider how the latter can be improved. Our methods may be multilayered since they will not only consider text processing, but also learning and inference schemes. The extraction and selection of textual features or the algorithms for learning and inference might play an important role in this process and hence, improvements should always be a part of our research.

\section{Work dissemination}
\label{section_work_dissemination}
Several parts of the work presented in this Thesis have attracted the interest of mainstream, technology or health news media. In this section, we refer to some significant articles that disseminated our research to the general public as they provide the reader with well-shaped introductory information about it as well as with a perspective different from the academic one.
\begin{itemize}
  \item ``How Twitter could better predict disease outbreaks'', in MIT Technology Review by Christopher Mims (14/07/2010). Link: \url{http://www.technologyreview.com/blog/mimssbits/25480/} -- Article featuring our paper \cite{Lampos2010}.
  \item ``Engines of the future: The cyber crystal ball'', in New Scientist by Phil McKenna (27/11/2010). Link: \url{http://goo.gl/roBa2}.
  \item ``Could Social Media Be Used to Detect Disease Outbreaks?'', in Science Daily. Press release from the University of Bristol (1/11/2011). Link: \url{http://www.sciencedaily.com/releases/2011/11/111101125812.htm} -- Article featuring our paper \cite{Lampos2011a}.
  \item ``2011: the year in review'', EPSRC's annual review (23/12/2011). Link: \url{http://www.epsrc.ac.uk/newsevents/news/2011/Pages/yearinreview.aspx\#may}.
  \item ``Brits study forecasting flu outbreaks using Twitter'', in Examiner by Linda Chalmer Zemel (27/12/2011). Link: \url{http://goo.gl/7auob} -- Article featuring our paper \cite{Lampos2011a}.
  \item ``Diagnosing flu symptoms with social media'', in Natural Hazards Observer (Vol. XXXVI(4), pp. 7--9) by Vasileios Lampos and Nello Cristianini (March 2012) \cite{Lampos2011c}. Link: \url{http://www.colorado.edu/hazards/o/archives/2012/mar12_observerweb.pdf}.
  \item ``Can Twitter Tell Us What We Are Feeling?'', in Mashable by Sonia Paul (17/04/2012). Link: \url{http://mashable.com/2012/04/17/twitter-feeling/} -- Article featuring our paper \cite{Lansdall-Welfare2012}.
\end{itemize}

Our work has also had some impact on current research as several recent publications from various research groups build on or cite it; we refer to some of them in the last chapter of the Thesis.

\section{Peer-reviewed publications}
\label{section:intro_publications}
In this section, we list all relevant peer-reviewed publications\index{Publications} in chronological order co-authored by Vasileios Lampos during his Ph.D. project. We have defined three levels of contribution and tagged all listed publications according to them:\\
--- Written the paper (\textbf{W})\\
--- Contributed in ideas, methods, or algorithms (\textbf{M})\\
--- Provided data (\textbf{D})

\textbf{Tracking the flu pandemic by monitoring the Social Web}. \textbf{Vasileios Lampos} and Nello Cristianini. IAPR Cognitive Information Processing (CIP 2010), Elba Island, Italy. IEEE Press, 2010. --- \textbf{W}, \textbf{M}, \textbf{D}\\
In this paper, we presented the first -- though preliminary -- method for using Twitter content to infer flu rates in a population (see Chapter \ref{Chapter_first_steps}).

\textbf{Flu Detector -- Tracking Epidemics on Twitter}. \textbf{Vasileios Lampos}, Tijl De Bie and Nello Cristianini. European Conference on Machine Learning and Principles and Practice of Knowledge Discovery in Databases (ECML PKDD 2010), Barcelona, Spain. Springer, 2010. --- \textbf{W}, \textbf{M}, \textbf{D}\\
This paper presented Flu Detector (\url{http://geopatterns.enm.bris.ac.uk/epidemics/}), a tool for detecting flu rates in several regions in the UK by exploiting Twitter content. The method described in this paper is an improvement of the one presented in the previous publication (see Chapter \ref{chapter:theory_in_practice}).

\textbf{Nowcasting Events from the Social Web with Statistical Learning}. \textbf{Vasileios Lampos} and Nello Cristianini. ACM Transactions on Intelligent Systems and Technology (TIST), Vol. 3(4), 2011. --- \textbf{W}, \textbf{M}, \textbf{D}\\
In this paper, we present a general methodology for nowcasting the occurrence and magnitude of an event or phenomenon by exploring the rich amount of unstructured textual information on the Social Web. The paper supports its findings with two case studies: a) inference of flu rates for several UK regions and b) inference of rainfall rates for several locations in the UK (see Chapter \ref{Chapter_Nowcasting_Events_From_The_Social_Web}).

\textbf{Effects of the Recession on Public Mood in the UK}. Thomas Lansdall-Welfare, \textbf{Vasileios Lampos}, and Nello Cristianini. Mining Social Network Dynamics (MSND) session on Social Media Applications in News and Entertainment (SMANE) at WWW '12, Lyon, France in February 2012. --- \textbf{M}, \textbf{D}\\
In this paper, we associate mood scores with real-life events in the UK. Amongst other interesting findings, we show that there exists evidence which connects recession in the UK with increased levels of negative mood in the Social Media (see some relevant findings in Section \ref{section:detecting_daily_mood_patterns}).

\textbf{Detecting Temporal Mood Patterns in the UK by Analysis of Twitter Content}. \textbf{Vasileios Lampos}, Thomas Lansdall-Welfare, Ricardo Araya, and Nello Cristianini. Under submission process to the British Journal of Psychiatry (April 2012). --- \textbf{W}, \textbf{M}, \textbf{D}\\
In this paper, we analyse seasonal circadian affective mood patterns based on Twitter content (see Section \ref{section:detecting_circadian_patterns}).

\section{Summary of the Thesis}
\label{section_intro_chapter_summary}
After this introductory section, in \textbf{Chapter \ref{chapter:theoretical_background}} we define the theoretical background on which our work has been based on. This includes basic notions from Machine Learning and Information Retrieval; some even more fundamental notions -- usually considered as common knowledge -- have been included in \textbf{Appendix \ref{AppendixA}}; \textbf{Appendix \ref{AppendixB}} contains tables with input data used in several experimental processes of ours. In \textbf{Chapter \ref{chapter:data_characterisation_collection}}, we refer to the data collection methods used to crawl the social network of Twitter, as well as provide the reader with a characterisation of this Social Network. \textbf{Chapter \ref{Chapter_first_steps}} describes the first steps of our work, where we developed a supervised learning method able to infer flu rates by using the content of Twitter as input information \cite{Lampos2010f}. \textbf{Chapter \ref{Chapter_Nowcasting_Events_From_The_Social_Web}} builds on the previously presented methodology, trying to resolve its limitations and also prove its generalised capacity by applying it to another task: the inference of rainfall rates from the content of Social Media \cite{Lampos2011a}. In \textbf{Chapter \ref{chapter:detecting_temporal_mood_patterns}}, we are aiming to extract affective mood patterns from the Social Web. We focus on two tasks: the extraction of circadian and daily mood patterns and we look into ways able to interpret them \cite{Lampos2012p,Lansdall-Welfare2012}. \textbf{Chapter \ref{chapter:pattern_discovery}} presents some preliminary and unpublished work regarding three Pattern Discovery tasks on Twitter content; the first one investigates content-based relations among locations and proposes a method for forming geo-temporal content similarity networks, the second analyses users' posting time patterns and the third one presents methodologies which aim to model and infer voting intention polls \cite{Lampos2012b}. \textbf{Chapter \ref{chapter:theory_in_practice}} presents two tools we developed that showcased our work to the general public and the academic community: Flu Detector \cite{Lampos2010} and Mood of the Nation. We conclude this Thesis in \textbf{Chapter \ref{Chapter_Conclusions}}, referring also to possible future work.

%% file: Chapters/Chapter2.tex
\chapter{Theoretical Background}
\label{chapter:theoretical_background}
%\lhead{Chapter 2. Theoretical Foundations and Related Work}

\rule{\linewidth}{0.5mm}
In this chapter we go through some fundamental theoretical notions which serve as a basis for our work. We start by defining the general field of this research project, that is Machine Learning. Regression and Classification, the most common supervised learning tasks, are discussed in short; we draw our interest into regularised regression as well as classification and regression trees. Bootstrap and Bagging, two methods for enhancing the learning process, are also presented and discussed. We move on by presenting some basic notions and methods from the field of Information Retrieval, such as the Vector Space Model and stemming. Finally, we define feature extraction and selection, two important tasks that have been performed quite often in our work.
\newline \rule{\linewidth}{0.5mm}
\newpage

\section{Defining the general field of research}
\label{section:Introduction_Chapter2}

``\emph{Learning is remembering}'', says Socratis$^{1}$ but Platonas$^{1}$ makes the inverse statement, ``\emph{remembering is learning}.'' Aristotelis\footnote{ Also -- commonly -- known as Socrates, Pluto and Aristotle respectively.} expands the notion of learning by claiming that ``\emph{each form of teaching or mental learning begins from the knowledge that already exists}.''

\textbf{Learning}\index{learning}, therefore, could be conceived as a process of acquiring new knowledge or other characteristics, such as a specific skill, by fusing input information with experience and previous knowledge.\footnote{ A modified version of the definition given in Wikipedia. Link: \url{http://en.wikipedia.org/wiki/Learning}.} The Greek philosophers were probably making statements primarily referring to humans or animals, but nowadays the notion of learning could also refer to a machine or a computer program.

Several definitions have been given to \textbf{Machine Learning}\index{Machine Learning} in the past years. Arthur Samuel -- broadly known for his pioneering work on developing computer programs able to play the game of checkers \cite{Samuel1959,Samuel1967} -- defined it as a branch of Artificial Intelligence focused on the task of making computers learn without being explicitly programmed. Tom Mitchell in \cite{Mitchell2006} pointed out that Machine Learning is a discipline that addresses the following general question: ``\emph{How can we build computer systems that automatically improve with experience, and what are the fundamental laws that govern all learning processes?}'' In his previous work \cite{Mitchell1997} he had already formalised the main concept of Machine Learning with the following statement: ``\emph{A computer program is said to learn from experience \textbf{E} with respect to some class of tasks \textbf{T} and performance measure \textbf{P}, if its performance at tasks in \textbf{T}, as measured by \textbf{P}, improves with experience \textbf{E}.}'' Consequently, in contrast to animals, the level of learning in machines can be quantified in a non abstract manner; actually, improving based on a performance metric serves as the main proof of learning.

In several Machine Learning tasks one tries to identify and exploit patterns in data using algorithms primarily driven by the statistical approach. The roots of Machine Learning come from the general field of Computer Science, whereas some of its most popular theoretical tools derive from Statistics, and therefore it is related or even quite often also referred to as \textbf{Statistical Learning}\index{Statistical Learning} \cite{hastie2005elements, Bishop2006}. Over the past years, three basic categories of learning have been defined: supervised\index{supervised learning}, unsupervised\index{unsupervised learning} and reinforcement learning.

In a supervised learning scenario we are provided with a set of inputs $\mathcal{X}$ (also known as observations or evidence), a corresponding set $\mathcal{Y}$ of targets or responses and the main task is to learn a function $f$ which maps the inputs to the targets, $f: \mathcal{X} \rightarrow \mathcal{Y}$. In unsupervised learning there exist no specific targets; the main task there could be generally described as trying to discover a pattern enclosed in the input data by exploiting their content. Finally, reinforcement learning aims to learn from interaction, for instance by discovering which actions yield higher rewards after trying them \cite{Sutton1998}.

In this work, we apply or derive methodologies which mainly belong to the supervised learning class. We also apply basic methods from unsupervised learning or other generic statistical algorithms in our effort to extract patterns from data. In the next sections, we give a short description of the theory and methods that have been used as a basis throughout this project.

%%%%%%%%%%%%%%%%%%%%%%%%%%%%%%%%%%%%%%%%%%%%%%% regression
\section{Regression}
\label{section:Regression}
Regression\index{regression} is a supervised learning task, where the unknown variable $y$, also known as the target variable, is real valued, that is $y \in \mathbb{R}$. In some cases, the target variable can be $n$-dimensional, $y \in \mathbb{R}^{n}$ for some $n \in \mathbb{N}$, though with loss of important information, we can break up this task to $n$ one-dimensional independent regression scenarios.

\subsection{Linear Least Squares}
\label{section:Linear_Least_Squares}
Suppose that knowing the mapping function $f: \mathcal{X} \rightarrow \mathcal{Y}$, we are given an input set in the form of a vector $X = [x_1, x_2, ..., x_n]$, $X \in \mathcal{X}$ and we want to predict a real-valued output $y \in \mathcal{Y}$. The linear regression model takes the following form:
\begin{equation}
y = f(X, w_{0}, w) = w_0 + \sum_{i=1}^{n}x_{i}w_{i},
\end{equation}
where $f$ is defined by $n$ coefficients or weights $(w_{1}$, ...,$w_{n})$ one for each dimension of $\mathcal{X}$ and one global intercept or bias, denoted by $w_0$. The target variable (also known as the response) $y$ is a linear combination of the observations and the elements of the mapping function.

Suppose now that we have a set of $m$ $n$-dimensional observations held in a $m \times n$ matrix $X$, where $x_{ij}$ is the $j$-th element of the $i$-th row, and their corresponding $m$ targets as contents of a vector $y$. The most common way to learn the mapping function $f$, known as Ordinary Least Squares\index{ordinary least squares} (OLS) regression\index{regression!ordinary least squares}, is by trying to minimise the Residual Sum of Squares (RSS):
\begin{equation}
\label{eq_RSS}
\text{RSS}(w_{0}, w) = \sum_{i=1}^{m}\left(y_i - w_0 - \sum_{j=1}^{n}x_{ij}w_{j}\right)^{2}.
\end{equation}
Therefore, the solution is given by
\begin{equation}
[w_{0}^{*}, w^{*}] = \argmin_{w_{0}, w}\text{RSS}(w_{0}, w).
\end{equation}

In the equivalent vector notation and by adding a column of 1's to the left side of $X$ (thus the first element of $w$ is the intercept), we have that:
\begin{equation}
\text{RSS}(w) = (y - Xw)^{T}(y-Xw).
\end{equation}
By differentiating RSS with respect to $w$ and setting the derivative equal to zero we can derive a closed-form solution for the ordinary least squares:
\begin{equation}
[w_0^{*}, w^{*}] = w = (X^{T}X)^{-1}X^{T}y.
\end{equation}

This closed-form solution reveals an obvious limitation of the OLS regression: $X^{T}X$ may be singular and thus difficult to invert, resulting in inconsistent solutions \cite{Hoerl1970}. Further analysis and interpretation of OLS indicates that least squares estimates often have low bias but large variance which affects negatively their prediction accuracy (see Appendix \ref{Ap:variance_bias_predictor} for the definitions of a predictor's variance and bias). Additionally, when the input is high-dimensional, OLS generates a large number of predictors (often as much as the dimensions), something that complicates interpretation \cite{hastie2005elements}.

% say OLS favours overfitting

%%%%%%%%%%%%%%%%%%%%%%%%%%%%%%%%%%%%%%%%
\subsection{Regularised Regression}
\label{Regularised_Regression}
%%%%% revision %%%%%%%%%%%%%%%%%%%%%%
Regularisation\index{regression!regularised} is a common practice to encounter the limitations of OLS regression. It can be seen as adding a penalty term to the original error function (Equation \ref{eq_RSS}) of OLS regression:
\begin{eqnarray}
\label{eq_ridge_regression}
\text{E}(w_0, w) & = & \text{RSS}(w_0, w) + \lambda \|w\|_{q}\\
& = & \sum_{i=1}^{m}\left(y_i - w_0 - \sum_{j=1}^{n}x_{ij}w_{j}\right)^{2} + \lambda\sum_{j=1}^{n}|w_{j}|_{q},
\end{eqnarray}
where $\lambda$ $(\geq 0)$ is known as the regularisation parameter and $q$ decides which norm of $w$ is being used. Parameter $\lambda$ can be interpreted as a coefficient that controls the relative importance between the data dependent error (RSS) and the regularised one. The amount of regularisation is increased as $\lambda$ increases. In literature, regularisation is also referred to as shrinkage since by minimising $\text{E}(w_0, w)$ one also aims to `shrink' the regression coefficients $w_{j}$. It is not common to regularise the bias term $w_0$ as this would remove its main purpose which is to provide a prior knowledge for the target variable.
%%%%%%%%%%%%%%%%%%%%%%%%%%%%%%%%%%%%%

In a high-dimensional linear regression model many variables might be correlated. Consequently, a large positive coefficient on a variable $x_j$ could be cancelled by a large negative coefficient on $x_{j'}$ given that $x_j$ and $x_{j'}$ are correlated. By shrinking regression's coefficients and therefore imposing a size constraint on them, one can restrict this effect \cite{hastie2005elements}.

%%% subset selection
By setting $q = 0$ we are performing a variable \textbf{subset selection}\index{regression!subset selection} -- that is finding a subset of predictors which gives the smallest RSS -- since penalty's sum will count the number of nonzero coefficients. This approach provides sparse solutions, but is not computationally feasible for problems of high dimensionality. Still, there exist algorithms, which can handle a small number of predictors \cite{Furnival1974, Branch1977} or hybrid approaches, such as Forward or Backward Stepwise Selection \cite{Draper1998}. The common disadvantage of all those learning models is that being discrete processes, they exhibit high variance and quite often do not reduce the prediction error of the full OLS regression model.

%%% ridge regression
For $q = 2$, Equation \ref{eq_ridge_regression}\index{regression!ridge} defines the minimisation target of \textbf{ridge regression}, also referred to as L2-norm regularisation given that the L2-norm of $w$ is minimised \cite{Hoerl1970}. Similarly to OLS, ridge regression has a closed form solution given by:
\begin{equation}
[w^{*}, w_0^{*}] = (X^{T}X + \lambda C)^{-1}X^{T}y,
\end{equation}
where a column of 1's has been added to the left side of $X$ to accommodate $w_0^{*}$ and $C$ is a an identity matrix with a 0 as its top diagonal element. Another way to look at ridge regression is to observe that a positive constant $\lambda$ has been added to the diagonal of $X^{T}X$ to deal with singularity problems of the original matrix \cite{boyd2004convex}, a process also known as Tikhonov regularisation \cite{Tikhonov1977}. However, regularising the L2-norm does not encourage sparsity, \ie regression's coefficients are nonzero, and therefore ridge regression does not offer an easily interpretable model.

%%%%%%%%%%%%%%%%%%%%%%%%%%%% LASSO
\subsection{Least Absolute Shrinkage and Selection Operator (LASSO)}
\label{section_LASSO}
On the other hand, for $q = 1$, \ie by regularising the L1-norm, the advantages of regularisation are maintained and at the same time sparse solutions are retrieved. This regularisation is known as \emph{Least Absolute Shrinkage and Selection Operator} or the \textbf{LASSO} \cite{tibshirani1996regression}. The original formulation of LASSO\index{regression!LASSO} is the following optimisation problem:
\begin{equation}
\label{eq_lasso}
\centering
[w^{*}_0, w^{*}] = \argmin_{w}\sum_{i=1}^{m}\left(y_i - w_0 - \sum_{j=1}^{n}x_{ij}w_j\right)^{2}
\mbox{subject to } \sum_{j=1}^{n}|w_j| \leq t,
\end{equation}
where $t$ controls the amount of regularisation (or shrinking) of the weight vector's L1-norm. This can be easily adapted to the previous formulation,
\begin{equation}
\label{eq_lasso_alternative}
\centering
[w^{*}_0, w^{*}] = \argmin_{w}\left\{\sum_{i=1}^{m}\left(y_i - w_0 - \sum_{j=1}^{n}x_{ij}w_j\right)^{2} + \lambda \sum_{j=1}^{n}|w_j|\right\}.
\end{equation}
As in the previous cases, $\lambda$ controls the amount of regularisation. Setting $\lambda = 0$ transforms LASSO to OLS, whereas using a very large $\lambda$ will completely shrink $w$ to zero leading to a null model. A moderate choice of $\lambda$ will result to the cancellation of some variables of the initial model by setting their corresponding regression coefficients to zero.

In contrast to ridge regression, LASSO does not have a closed form solution. Computing the solution of LASSO is a quadratic programming problem and estimations of the solution can be derived within this framework \cite{Bertsekas1995}. A more efficient way to compute LASSO is achieved by enforcing a modification on the Least Angle Regression (LAR)\index{regression!LAR} algorithm \cite{efron2004least}.

\begin{algorithm}[tp]
\caption{Least Angle Regression (LAR) modification for LASSO}
\label{algorithm_LAR}
\begin{enumerate}
  \item Standardise $X$ so that each predictor (or column) $X_{:,j}$ has a zero mean and variance equal to 1. Start with the residual set to $r = y - \bar{y}$ and all regression coefficients to zero, $w_1$, $w_2$, ..., $w_p = 0$.
  \item Find $X_{:,c}$ the most correlated predictor with $r$.
  \item Move $w_c$ from zero towards its least-squares coefficient $\langle X_{:,c},r \rangle$ until some other predictor $X_{:,i}$ has as much correlation with the current (active) residual $r_{a}$ as $X_{:,c}$.
  \item Move $(w_c,w_i)$ in the direction defined by their joint least squares coefficient of the current residual on $(X_{:,c},X_{:,i})$, until some other predictor $X_{:,\ell}$ has as much correlation with the current residual. If a nonzero coefficient becomes zero, then drop its corresponding variable and recompute the current joint least squares direction.
  \item Iterate through this process until all $p$ predictors have been entered. If the predictors are more or equal to the observations ($p \geq N$), then LAR completes after $N-1$ steps.
\end{enumerate}
\end{algorithm}

The modified LAR (see algorithm \ref{algorithm_LAR}) computes the entire regularisation path of LASSO, \ie all LASSO solutions for different choices of the regularisation parameter $\lambda$. In a nutshell, LAR firstly sets all regression coefficients to zero and searches for the predictor (or observation) $X_{:,c}$ that has the highest correlation with the residual $r = y - \bar{y}$, where $\bar{y}$ denotes the mean of $y$. Then explores the direction of this predictor until an other predictor $X_{:,i}$ has as much correlation with the current residual. It proceeds in a direction equiangular between the two predictors until a third variable $X_{:,\ell}$ enters the active set of most correlated predictors. Then again it continues along the least angle direction of $X_{:,c}$, $X_{:,i}$ and $X_{:,\ell}$ until a fourth variable enters the most correlated set and so on. LASSO modification proposes dropping variables when their corresponding coefficients arrive to zero from a nonzero previous value during this process \cite{hastie2005elements}.

%%%%%%%%%%%%%% revision %%%%%%%%%%%%%%%%%%%%%
LASSO has a unique solution assuming that there exists no pair of predictors which are perfectly collinear \cite{tibshirani2005sparsity}. Recent work has also shown that LASSO has a unique solution (with probability one), if the predictor variables are drawn from a continuous probability distribution \cite{Tibshirani2012}. LASSO is shown to be sign-consistent -- meaning that sign as well as zeros in the inferred $w$ are consistent --, if there are low correlations between relevant and irrelevant variables \cite{Zhao2006, Wainwright2009, Zou2006, Yuan2006}. However, it is not model-consistent as in many settings (for example when predictors are strongly correlated) it does not recover the true model, \ie errors are made in the assignment of nonzero weights to variables and therefore, the inferred sparsity pattern differs from the true one \cite{Lv2009, Zhao2006}. There exist various proposed ways to fix the model inconsistency of LASSO, such as the adaptive \cite{Zou2006} or relaxed LASSO \cite{Meinshausen2007}, thresholding \cite{Lounici2008}, bootstrapping \cite{Bach2008} and others \cite{Meinshausen2010, Wasserman2007}.
%%%%%%%%%%%%%%%%%%%%%%%%%%%%%%%%%%%%%%%%%%%%%

%%%%%%%%%%%%%%%%%%%%%%%%%%%%%%%%%%%%%%%%%%%%%%%%%%%%%%%% classification
\section{Classification}
\textbf{Classification}\index{classification} is a supervised learning task where a model, known as the \emph{classifier}, is constructed to predict categorical labels. Recall that the prediction target in regression is a numeric value. In classification, we aim to infer the label of a class, for example ``Category A'' or ``Category B'' and so on. Those categories can be represented by discrete numerical values, \ie 1 and 2 for Categories A and B respectively \cite{Han2011}. Using the notation of section \ref{section:Regression}, the target variable $y$ is a set of $m$ labels $\ell$, $y = \{\ell_{1}, \text{...}, \ell_{m}\}$, $m\in\mathbb{N}$, and by mapping labels to different integers, it can also take a numeric representation, $y\in\mathbb{Z}$. We are aiming to learn a function $f: \mathcal{X} \rightarrow y$, where $\mathcal{X}$ denotes the set of observations. In this work, we do not directly solve classification problems; however, we are using structures -- such as the Classification and Regression Trees (see Section \ref{section_tb_CART}) -- that were primarily built to solve such problems, but later on were extended to include regression problems as well.

%%%%%%%%%%%%%%%%%%%%%%%%%%%%%%%%%%%%%%%%%%%% Decision Trees
\subsection{Decision Trees}
\label{section:decision_trees}
A decision tree\index{classification!decision tree} is a structure that classifies instances of observable data by sorting them down the tree using a top-down approach, starting from the root and finishing at a leaf node. Leaf nodes are labels indicating the result of the classification for this instance. A node of the decision tree targets a specific attribute of the instance and depending on the value of this attribute, it indicates which descending branch the decision making process should follow \cite{Mitchell1997}.

\begin{figure*}[tp]
\begin{center}
\includegraphics[width=4.5in]{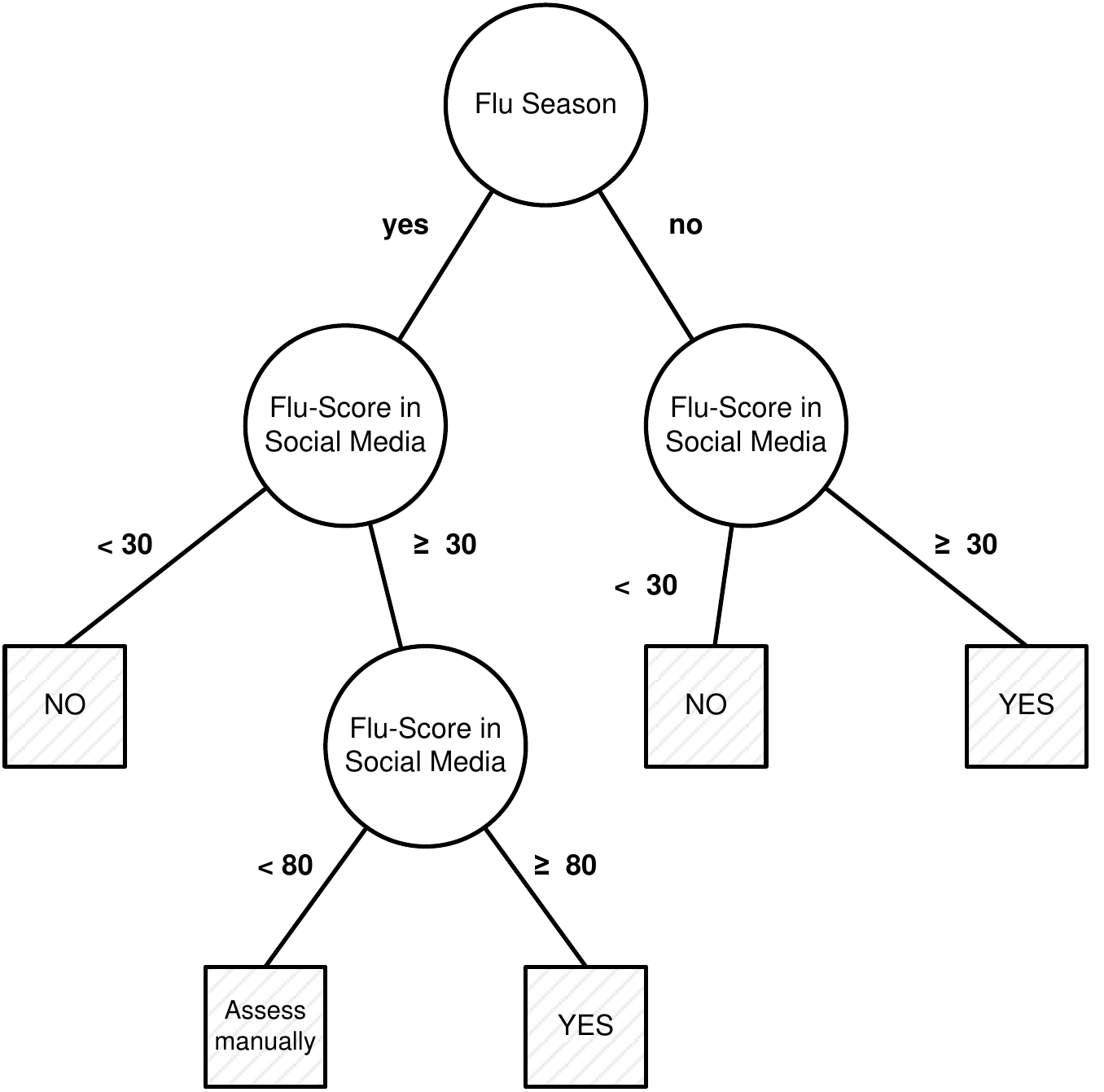}
\end{center}
\caption{A simplified decision tree for answering the following question: ``Based on the Social Media content, should we issue a warning for a possible flu epidemic?''}
\label{fig_decisionTree_Flu}
\end{figure*}

Figure \ref{fig_decisionTree_Flu} shows an example of a very basic decision tree. Supposing that we have a function able to compute a flu-score from the Social Media content on the web -- the higher the flu-score, the bigger the risk for an epidemic --, and at the same time no other official information from health authorities exists about the actual flu rates in the population, the purpose of this decision tree is to figure out whether a warning for a spreading flu epidemic should be issued or not. Of course, this is a very simplified example to help the reader understand how decision trees function.

Suppose that we are within a typical `flu season' and from our data we compute a flu score equal to 90. The instance
\begin{equation*}
\langle \text{\textbf{Flu Season}: yes, \textbf{Flu-Score in Social Media}} = 90 \rangle
\end{equation*}
would be sorted down the middle leaf of the decision tree and would therefore be classified as `YES', meaning that a warning should be issued. Likewise, the instance
\begin{equation*}
\langle \text{\textbf{Flu Season}: yes, \textbf{Flu-Score in Social Media}} = 75 \rangle
\end{equation*}
would be classified as `Assess manually', which in practice means that human supervision is needed in order to make a decision.

Decision trees have found application in text mining \cite{Apte1998}, computational linguistics \cite{Schmid1994}, failure diagnosis \cite{Chen2004}, the medical domain \cite{Kokol1994,Sims2000} and so on. Many scientific fields such as Statistics, Artificial Intelligence (Machine Learning) or Decision Theory have developed methodologies for learning/constructing automatically decisions trees from data \cite{Murthy1998}. Over the past years, several approaches (and the respective tools) have been developed for decision tree learning, such as \textbf{CHAID} (CHi-squared Automatic Interaction Detector) \cite{Kass1980}, the well-known \textbf{ID3} \cite{Quinlan1986} and its descendant \textbf{C4.5} \cite{Quinlan1993}, \textbf{QUEST} (Quick Unbiased Efficient Statistical Tree) \cite{Loh1997} and many more. In the next section, we will focus briefly on one approach, the Classification and Regression Tree (\textbf{CART}) \cite{Breiman1984} as we have chosen to use this one in our work (see Chapter \ref{Chapter_Nowcasting_Events_From_The_Social_Web}).

The main advantage that decision trees have is their interpretability. When their size is reasonable, they do create self-explanatory models that can be easily understood by non-experts. Furthermore, decision tree algorithms have been developed able to handle discrete, continuous variables as well as missing values and perform classification or regression. In addition, decision trees are a non-parametric method, which means that they do not rely on assumptions made about the probability distribution that creates the data \cite{Rokach2008}. However, the construction of optimal decision trees is an NP-complete problem \cite{Hyafil1976} and therefore the approximate result of learning might not always be optimal. Apart from the non-optimality, decision trees also suffer from instability; minor changes in the training set, might result to the construction of a very different model \cite{Breiman1984,Quinlan1993}.

%%%%%%%%%%%%%%%%%%%%%%%%%%%%%%%%%%%%%%%%%%%%%%%%%%%% CART
\subsection{Classification and Regression Tree}
\label{section_tb_CART}
In this section, we explain briefly the main operations of Classification and Regression Tree (CART)\index{CART}\index{classification and regression tree|see{CART}}, which is a sophisticated program for constructing trees from data developed by Breiman \etal and presented in \cite{Breiman1984}. In addition to the original script, this section was also based on review papers about CART \cite{Sutton2005,Lewis2000}.

CART is a non-parametric method capable of solving classification and regression problems, which not only has had a significant coverage in the literature (of several academic fields), but also a great number of applications (see for example \cite{Riley1989,Fonarow2005,Lee2006,Prasad2006} and many more). Since it is a well established method, there also exist off-the-shelf software packages implementing variations of it, such as the one in MATLAB \cite{Martinez2002}.

\begin{figure*}[tp]
\begin{center}
\includegraphics[width=4in]{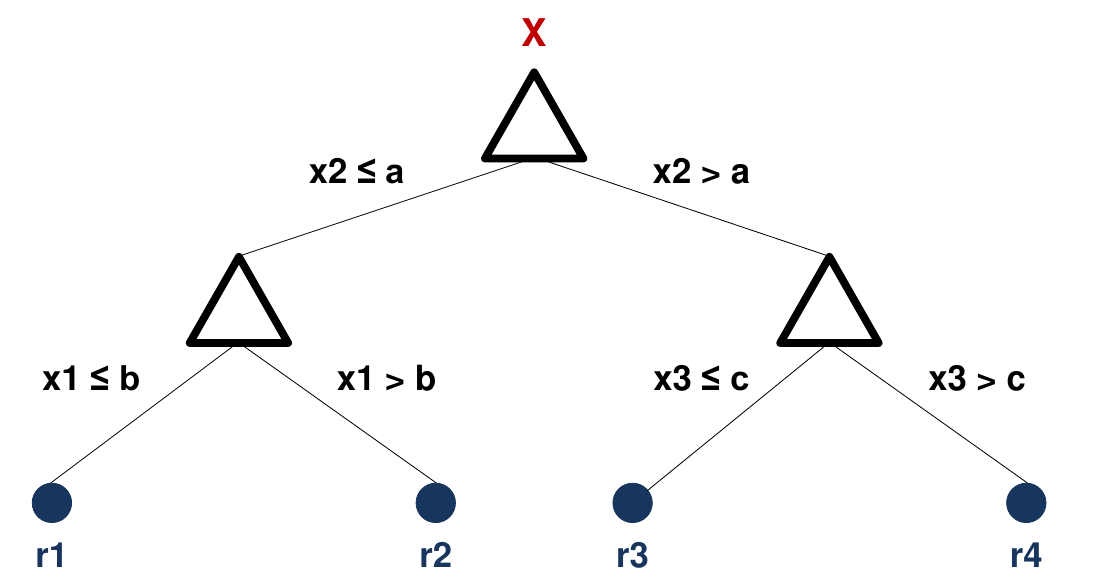}
\end{center}
\caption{An example of a classification and regression tree (CART)}
\label{fig_CART_example}
\end{figure*}

Figure \ref{fig_CART_example} shows a very simple example of a CART with four terminal nodes. $\mathcal{X}$ denotes the entire space of input variables; specific variables (such as $x_1$, $x_2$ or $x_3$) are used in the decision making process similarly to the previous section's decision tree example. Given that the terminal nodes of the tree are real-valued numbers, this tree is used to perform regression; a similar tree could perform classification if the terminal nodes were class labels.

CART is constructed by splitting $\mathcal{X}$ into two subsets and then by recursively splitting $\mathcal{X}$'s subsets and so on. By forming two splits of each subset, CART constructs a binary tree. This should not be considered as a limitation due to the fact that binary trees can be used to represent any other type of tree structures. During the tree construction process, when for example $\mathcal{X}' \subset \mathcal{X}$ is divided into two subsets, there exists no rule forcing to divide those two new subsets using the same variable. This is made more clear by observing Figure \ref{fig_CART_example}, where the left subset is divided by $x_1$ and the right one by $x_3$. There is also no limitation in using the same variable for creating subsets in different branches of the tree as well in the same branch for more than one time.

The core question about the construction of a CART is how to best split the subsets of $\mathcal{X}$. Supposing that $y$ is the set of response values for the training set $\mathcal{X}$, each split of a subset $\mathcal{X}' \subset \mathcal{X}$ to a left ($\mathcal{L}$) and a right ($\mathcal{R}$) sub-subset is decided by solving the following optimisation problem:
\begin{equation}
\{\mathcal{L},\mathcal{R}\} = \argmin_{\mathcal{L},\mathcal{R}} \left(\sum_{i\text{: }x_i \in \mathcal{L}} (y_i - \bar{y}_{\mathcal{L}})^2 + \sum_{i\text{: }x_i \in \mathcal{R}} (y_i - \bar{y}_{\mathcal{R}})^2 \right),
\end{equation}
where $\bar{y}_{\mathcal{L}}$ and $\bar{y}_{\mathcal{R}}$ are the sample means of the response values in $y$ corresponding to $\mathcal{L}$ and $\mathcal{R}$ respectively. The value of a terminal node $t$ of the regression tree is set as the mean response value $y_{\mathcal{X}_t}$ that corresponds to the subset $\mathcal{X}_t \subset \mathcal{X}$.

Deciding the optimal size of the tree is controlled by an error-complexity metric $C_{\alpha}(T)$ defined as:
\begin{equation}
C_{\alpha}(\mathcal{T}) = C(\mathcal{T}) + \alpha |\mathcal{T}|,
\end{equation}
where $|\mathcal{T}|$ denotes the size of the regression tree, $C(\mathcal{T})$ is sum of squared errors (or of any other loss function applied) and $\alpha$ is a parameter that controls the ratio between the size of the tree and its loss on the learning sample. Similarly to LASSO, $\alpha |\mathcal{T}|$ can be seen as a regularisation parameter that penalises overfitting. By using an independent validation set and a loss function -- usually Mean Square Error (\textbf{MSE}\index{Mean Square Error|see{MSE}}\index{MSE}) --, one can further prune the tree and decide its optimal and final version.

\section{Bootstrap}
\textbf{Bootstrap}\index{Bootstrap} was introduced as a method for assessing the accuracy of a prediction but has also found applications in improving the prediction itself \cite{Efron1979, Efron1986, Efron1993}. Suppose that we have a sample of $n$ observations $X = [x_1, x_2, ..., x_n]$, all of the same probability $1/n$. Based on $X$, as well as on a predictor function $f$, we are able to calculate an estimate $\hat{y} = f(X)$ of the actual response $y$. The primary aim of bootstrap was to estimate how accurate the prediction $\hat{y}$ is.

A bootstrap sample $B = [x_{1}^{*}, x_{2}^{*}, ..., x_{n}^{*}]$ is a random sample of size $n$ drawn with replacement from $X$. $x_{i}^{*}$'s notation denotes that it is different from the original $x_{i}$; the bootstrap data set $B$ consists of members of $X$, some appearing zero times, some appearing once, some appearing more than once and so on. Algorithm \ref{algorithm_Bootstrap} shows how Bootstrap is used to estimate the Standard Error (\textbf{SE})\index{SE} of a predictor $f$, one of its primary applications.

%... do a short mention on bolasso here to justify the existence of this section ...
%... bolasso makes lasso stable or consistent ...

\begin{algorithm}[tp]
\caption{Bootstrap algorithm for estimating standard error}
\label{algorithm_Bootstrap}
\begin{enumerate}
\item Form $M$ independent bootstrap samples $B_{m}$ each consisting of $n$ elements drawn with replacement from $X$.
\item Evaluate the bootstrap replication corresponding to each bootstrap sample
        \begin{equation}
        \hat{y}_{m} = f(B_{m})\text{, } m = 1, 2,\text{ ..., } M.
        \end{equation}
\item Estimate the SE$(\hat{y})$\index{Standard Error|see{SE}}\index{SE} by computing the standard deviation of the $M$ replications
        \begin{equation}
        \text{SE}(\hat{y}) = \sqrt{\left(\displaystyle\sum_{m = 1}^{M} \left( \hat{y}_{m} - \text{E}(\hat{y}) \right)^{2}\right)/(M-1)}.
        \end{equation}
\end{enumerate}
\end{algorithm}

\subsection{Bagging}
\label{section_bagging}
\textbf{B}ootstrap \textbf{Agg}regat\textbf{ing}\index{Bagging}\index{Bootstrap Aggregating|see{Bagging}}, known as \textbf{Bagging}, is a method for generating multiple versions of a predictor and then using them to get an aggregated and possibly better -- in terms of performance -- predictor. Those versions are the result of carrying out learning on bootstrap replicates of the original learning set. When predicting a numerical value, Bagging averages over the versions of the predictor, whereas when the prediction target is categorical, it uses a plurality vote to decide the outcome of the prediction \cite{Breiman1996}. The main operations of Bagging are also also described in Algorithm \ref{algorithm_Bagging}.

Bagging improves the MSE\index{MSE} for unstable predictors, but does not have a significant impact on stable schemes; a predictor is unstable, if small changes in the data can cause large changes in the predicted values \cite{Breiman1996a}. Bagging might also improve MSE for sufficiently large sample sizes, increases the squared bias \cite{Buja2000} and reduces variance for higher order terms leaving the linear term unaffected \cite{Friedman2007}. Those two properties of reducing MSE and variance have an important effect on hard decision problems, \eg an estimation of a regression function after testing in regression and classification trees; bagging is shown to smooth such hard decisions \cite{Buhlmann2002}. Interestingly, its properties do not necessarily depend on the dimensionality of the problem \cite{Buja2000a, Buhlmann2002}.

%... do a short mention on the ensemble of carts here to justify the existence of this section ...
%... bagging sorts out the instability problems of cart .....................

\begin{algorithm}[tp]
\caption{Bootstrap Aggregating (Bagging) for estimating a better predictor function}
\label{algorithm_Bagging}
\begin{enumerate}
\item Suppose we are given a learning set $\mathcal{L} = \{X, y\}$, where $X$ and $y$ denote observations and responses respectively.
\item Take $M$ bootstrap samples from $\mathcal{L}$, $\mathcal{L}_{m} = \{X_m, y_m\}$, $m = 1, 2,\text{ ..., } M$.
\item Learn $M$ predictor functions $f_m$ using $\mathcal{L}_{m}$.
\item Compute $M$ inference sets $y^{(t)}_m$ using $f_m$ on a set of unseen observations $X^{(t)}$, $y^{(t)}_m = f_{m}(X^{(t)})$.
\item
    \begin{enumerate}
      \item If $y$ is a categorical variable, then use a voting scheme to decide the dominant class among $y^{(t)}_m$.
      \item If $y$ is a numerical variable, then $y^{(t)}$ = $\text{E}(y^{(t)}_m)$.
    \end{enumerate}
\end{enumerate}
\end{algorithm}

\section{Feature Extraction and Selection}
\label{section_feature_extraction_selection}

\textbf{Feature extraction}\index{feature extraction} aims to create new features based on transformations or combinations of an original set of raw attributes. Given a feature space $\Omega$ of dimensionality $d$, feature extraction methods determine an appropriate subspace $\Omega'$ of dimensionality $m \leq d$, in a linear or non-linear manner \cite{jain2000statistical}. Feature extraction is also a special form of dimensionality reduction, and finds use in many applications in the areas of bioinformatics, text processing, speech processing and so on \cite{guyon2006feature}.

\textbf{Feature selection}\index{feature selection}, on the other hand, is a dimensionality reduction method which aims to select a subset of features or dimensions in an effort to overcome high dimensionality and overfitting, improve the interpretability of the inferred model, and in some learning scenarios speed up the learning and inference process. Feature selection may follow feature extraction; feature extraction is used to define an original set of dimensions or attributes and then feature selection specifies which ones are the most important.

A very basic form of feature selection can be achieved via correlation analysis. Consider one of the feature candidates as a vector $x$, and the corresponding target as a vector $y$, both of size $n$. One of the most primitive feature selection techniques is to measure the \textbf{correlation coefficient}\index{linear correlation coefficient} $\rho(x,y)$ which is defined as the covariance of $(x, y)$ divided by the product of their standard deviations:
\begin{equation}
\rho(x,y) = \frac{\displaystyle\sum_{i = 1}^n(x_{i} - x_{\mu})(y_{i} - y_{\mu})}{\sqrt{\displaystyle\sum_{i = 1}^n(x_{i} - x_{\mu})^{2}\sum_{i = 1}^n(y_{i} - y_{\mu})^{2}}},
\end{equation}
where $x_{\mu}$ and $y_{\mu}$ are the first moments of $x$ and $y$ respectively. The lower and upper bounds of $\rho$ are -1 and 1, indicating perfect anti-correlation or correlation (in a linear manner) respectively, whereas a $\rho = 0$ indicates the absence of linear correlation. A correlation coefficient can be used to rank candidate features, but only separately; it is a univariate method.

An important part of our work included the optimal selection of textual features for representing a target concept. Recent research has shown that the bootstrapped version of LASSO, conventionally named as \emph{Bolasso}, by intersecting the supports of bootstrap estimates addresses LASSO's model selection inconsistency problems \cite{Bach2008}. We have applied a modification of Bolasso\index{Bolasso} (see Section \ref{section:soft_bolasso_with_Ct_validation}) in our effort to select a consistent subset of textual features. In addition, a bagged version of CART\index{CART} (also referred to as an ensemble of CARTs) has been applied as an alternative nonlinear and nonparametric method, which deals with the instability problems of CART and at the same time is able to -- indirectly -- select a subset of important features from a text stream and then use them to solve a regression problem (see Section \ref{section:applying_different_learning_methods}).

\section{Foundations of Information Retrieval}
\label{section_foundations_of_IR}
Information Retrieval (\textbf{IR})\index{IR} is a scientific field with its main goal set to provide methods for retrieving useful information from large collections of unstructured material (usually text) \cite{Manning2008}. In our context, IR is an automated process based on well defined algorithms and is performed by computer programs. In the following sections, we go through some basic notions of IR used and applied throughout our work.

\subsection{Vector Space Model}
\label{section_vector_space_model}
A corpus is a collection of documents, which in turn are collections of words or n-grams (where n denotes the number of words). To convert all this information into something more structured that can receive queries, an algebraic representation is needed. A Vector Space Model\index{Vector Space Model|see{VSM}}\index{VSM} (\textbf{VSM}) defines such a representation.

For an $m$-sized set of documents $\mathcal{D} = \{d_1,...,d_m\}$, which are formed by an $n$-sized set of n-grams or a vocabulary $\mathcal{V} = \{v_1,...,v_n\}$, the weight of an n-gram $v_i$ in a document $d_j$ is denoted as $w_{ij}$. Therefore a document $d_j$ can be represented as a vector or a set of weights:
\begin{equation}
d_j = \{w_{1j},...,w_{nj}\}.
\end{equation}

There exist several types of VSMs; the simplest one is the Boolean\index{VSM!Boolean}, which sets a weight equal to 1 if an n-gram exists in a document, otherwise sets it to 0:
\begin{equation}
w_{ij} = \left\{
\begin{array}{l l}
1 & \quad \mbox{if $v_i \in d_j$,}\\
0 & \quad \mbox{otherwise.}\\
\end{array} \right.
\end{equation}

Alternatively, one can count the occurrences of the n-gram in the document and normalise this count with the total number of n-grams in the document
\begin{equation}
w_{ij} = \frac{\text{times }v_i\text{ appears in }d_j}{\text{number of }n\text{-grams in }d_j}.
\end{equation}
This weighting scheme is commonly known as Term Frequency (\textbf{TF}) VSM\index{VSM!term frequency}.

The VSMs presented above treat each document as a set of n-grams; they do not exploit possible relationships or semantic connections between the n-grams, but consider them as separate and independent random variables. For this reason they are usually referred to as ``bag of words'' models\index{bag of words}.

\subsection{Term Frequency -- Inverse Document Frequency}
\label{section_TF_IDF}
Term Frequency Inverse Document Frequency (\textbf{TF-IDF})\index{VSM!TF-IDF} is a VSM  for document representation which enables to approximate the actual importance of a word or phrase in a document or several documents \cite{salton1973specification}. Similarly to the previously presented VSMs, TF-IDF is also a ``bag of words'' VSM, but instead of simply counting the terms in a document and assigning frequencies to them, it calculates a normalised frequency over a set of documents in order to define weights of importance.

For the following description we have used \cite{salton1983extended,Liu2007} as additional references. Suppose we have a vocabulary $\mathcal{V}$ with $n$ terms and a set $\mathcal{D}$ with $m$ documents, $\mathcal{V} = \{v_{1}, ..., v_{n}\}$ and $\mathcal{D} = \{d_{1}, ..., d_{m}\}$. For $\forall v_{i} \in \mathcal{V}$, we calculate the number of its appearances in $\forall d_{j} \in \mathcal{D}$. As a result, we derive a set $\mathcal{F}$ of size $n \times m$ which contains the raw frequencies or counts of the vocabulary terms in the documents,
$\mathcal{F} = \{f_{11}, ..., f_{n1}, f_{12}, ..., f_{ij}, ..., f_{nm}\}$.

The frequency $\text{tf}_{ij}$ of a term $v_{i}$ in a document $d_{j}$ is given by
\begin{equation}
\text{tf}_{ij} = \frac{f_{ij}}{\max\{f_{1j}, ..., f_{nj}\}}.
\end{equation}

It is a normalised frequency which depends on the maximum raw frequency of a term in a specific document. For a term $v_{i}$ and a document $d_{j}$, if the raw term frequency is equal to zero, $f_{ij} = 0$ (meaning that the term does not appear in the document), then for the respective normalised frequency holds also that $\text{tf}_{ij} = 0$. At this stage, we have retrieved the set $\mathcal{TF}$ which includes all the normalised term frequencies for the documents,
$\mathcal{TF} = \{\text{tf}_{11}, ..., \text{tf}_{n1}, \text{tf}_{12}, ..., \text{tf}_{ij}, ..., \text{tf}_{nm}\}$.

$\forall v_{i} \in \mathcal{V}$ we calculate the total number of documents in which it appears at least once. This is denoted by the document frequency $df_{i}$. All the document frequencies are represented by the set $\mathcal{DF} = \{\text{df}_{i}, ..., \text{df}_{n}\}$. In order to retrieve the inverse document frequencies set $\mathcal{IDF}$, we use the following formula
\begin{equation}
\text{idf}_{i} = \log\left(\frac{m}{\text{df}_{i}}\right),
\end{equation}
where $m$ is the number of documents. Similarly, the $\mathcal{IDF}$ set has $n$ elements, $\mathcal{IDF} = \{\text{idf}_{i}, ..., \text{idf}_{n}\}$.

In the final stage of the scheme, we build a set $W$ with weights for every term-document pair, $\mathcal{W} = \{w_{11}, ..., w_{n1}, w_{12}, ..., w_{ij}, ..., w_{nm}\}$. Each $w_{ij}$ is the result of
\begin{equation}
w_{ij} = \text{tf}_{ij} \times \text{idf}_{i}.
\end{equation}

\subsection{Text Preprocessing -- Stemming and Stop Words}
\label{section_stemming_stopwords}
It is a quite common principle to preprocess raw text before forming a vector space model. A well-known method for text preprocessing is known as \textbf{stemming}\index{stemming} and it is a process that reduces a word to its stem, that is the basis of the word. In that way, many variants of the same word are merged in one variable, something that helps in reducing significantly the size of the vocabulary index and therefore the dimensionality of a text mining problem.

Several approaches to stemming have been proposed \cite{Frakes1992} with the most popular of them using suffix stripping \cite{Lovins1968,porter1980}. Suffix stripping could be defined as the removal or homogeneous calibration of a word's suffix. In our work, we are applying one of those approaches, the Porter Stemmer\index{stemming!Porter Stemmer}, which has been developed in 1980 and later on has been made available through a stemming-specific language (Snowball) as a software package \cite{Porter2001}. Some examples of stemming using the Porter algorithm are listed below:
\begin{tabbing}
RESEARCH~~~~~~~~~   \= $\longrightarrow$~~~~~ \= RESEARCH~~~~~~~~~~~~ \= HAPPY~~~~~~~~~  \= $\longrightarrow$~~~~~ \= HAPPI\\
RESEARCHES          \> $\longrightarrow$      \> RESEARCH             \> HAPPIER         \> $\longrightarrow$      \> HAPPIER\\
RESEARCHED          \> $\longrightarrow$      \> RESEARCH             \> HAPPIEST        \> $\longrightarrow$      \> HAPPIEST\\
RESEARCHER          \> $\longrightarrow$      \> RESEARCH             \> HAPPINESS       \> $\longrightarrow$      \> HAPPI\\
RESEARCHERS         \> $\longrightarrow$      \> RESEARCH             \>                 \>                        \>\\
RESEARCHING         \> $\longrightarrow$      \> RESEARCH             \>                 \>                        \>
%                    \>\\
%HAPPY               \> $\longrightarrow$ HAPPI\\
%HAPPIER             \> $\longrightarrow$ HAPPIER\\
%HAPPIEST            \> $\longrightarrow$ HAPPIEST\\
%HAPPINESS           \> $\longrightarrow$ HAPPI
\end{tabbing}
In the first group of examples, the stemmer removes any suffix and keeps the core stem `RESEARCH'. In the second series, it converts the last character of `HAPPY' from `Y' to `I' and therefore creates the same stem for `HAPPY' and `HAPPINESS', but keeps the words `HAPPIER' and `HAPPIEST' as they are, \ie the stemmer makes a distinction between the comparative and superlative forms of an adjective. On the other hand, since stemming is an automated method based on a set of rules, there might be occasions where it generates significant information loss. For example, both `SINGULAR' and `SINGULARITY' are stemmed to `SINGULAR' or both `AUTHOR' and `AUTHORITY' are stemmed to `AUTHOR'.

In addition to stemming, it is desirable to remove words that do not have any particular semantic notion mostly because they are used quite often. Those words are commonly referred to as \textbf{stop words}\index{stop words} and some examples are articles, \eg `a', `an', `the', propositions, \eg `about', `after', `on', and so on. For English text, there exist several off-the-shelf lists with stop words with the one proposed by Fox in 1989 being the most popular; this particular list was generated by combining the most frequent English words with a set of manual rules encapsulated in a deterministic finite automaton \cite{Fox1989}. In most IR tasks, stop words are automatically identified by setting a threshold in the maximum allowed frequency of a word -- similarly, spam stop words are decided by setting a threshold on the minimum frequency of a word \cite{Manning2008}. In our work, depending on the task at hand, we either use an off-the-shelf stop word list, or set thresholds for the minimum and maximum frequency of a `legitimate' word or use a combination of both practices. The main motivation behind removing stop words is again the reduction in the dimensionality of the problem, but also the existing experimental proof for increased performance \cite{Silva2003}.

\section{Summary of the chapter}
\label{section:summary_of_chapter_tb}
This chapter gave a summary of the theoretical background behind our research; Appendix \ref{AppendixA} covers some more basic notions. The core scientific field of our work is defined by Artificial Intelligence and more specifically by one of its branches, Machine Learning. Supervised learning, which formulates the scenario where we are given a set of input data $X$ and their corresponding responses $y$ and we have to figure out a function $f$ mapping $X$ to $y$, gathers the main focus in our work, as we apply regularised regression methods throughout it. LASSO is a regularised regressor providing sparse solutions, however it is shown to be inconsistent. CARTs formulate a method able to provide sparse regression solutions which are nonlinear and nonparametric, but are also unstable. Bolasso, the bootstrap version of LASSO, resolves some model inconsistencies of LASSO; likewise, Bagging addresses the instabilities of CARTs. In our project, we have also incorporated some basic tools from the field of Information Retrieval. Feature extraction approaches create attributes from a raw set of unstructured information, whereas vector space models define ways for giving those attributes a consistent algebraic form. When processing textual information, keeping a stem instead of the entire word in addition to removing stop words helps in reducing the initial dimensionality of a problem; feature selection methods are statistical approaches that can further reduce the original dimensions during learning.

%% file: Chapters/Chapter3.tex
\chapter{The Data -- Characterisation, Collection and Processing}
\label{chapter:data_characterisation_collection}

\rule{\linewidth}{0.5mm}
The mission of this chapter is two-fold. First, we give a short description on the Social Network platform which has been used as the input data space for this research project and also explain its value as an information source. Then, we refer to the methods and tools that have been implemented or incorporated in our data collection and processing routines. Of course, this project is not about collecting data; however, our research would have been impossible without it.
\newline \rule{\linewidth}{0.5mm}
\newpage

\section{Twitter: A new pool of information}
\label{section:twitter_pool_of_new_age}
A \textbf{blog}\index{blog}, a term derived from synthesising the words web and log, is a -- or occasionally a part of a -- website supposed to be updated with new content from time to time.\footnote{Definition taken from Wikipedia on 31.10.2011 - \url{http://en.wikipedia.org/wiki/Blog}.} A blog's content is based on posts identified primarily by their unique Uniform Resource Locators (URLs). Those posts can vary from random or opinionated articles to the embedding of songs, videos or other multimedia types.

Microblogging\index{microblogging} is a compact form of blogging, which can be seen as a `technological evolution' of blogs, and surely is one the best complementary tools for the blogosphere. It became very popular when web services like Tumblr\footnote{Tumblr founded in 2007, \url{http://tumblr.com}.} and Twitter\footnote{Twitter launched in 2006, \url{http://twitter.com}.} kicked off. The basic characteristic (or limitation) of microblogs is that authors are restricted to a specific number of characters per post. For the experimental derivations of the work presented in this Thesis, we collect and use Twitter content.

\begin{figure}
\centering
\includegraphics[width=5.25in]{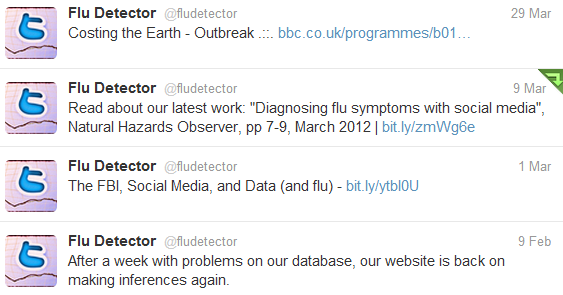}
\caption{This is how the published tweets of the Twitter account (or user) \href{http://twitter.com/fludetector}{@fludetector} look like.}
\label{fig_twitter_website_screenshot}
\end{figure}

Twitter\index{Twitter} was created in March 2006 and by 2011 the number of its users reached 100 million worldwide with a reported new accounts per day ratio equal to 460,000 for February 2011.\footnote{Twitter Blog, $\#$numbers - \url{http://blog.twitter.com/2011/03/numbers.html}, 14.03.2011.} The registered users of Twitter are allowed to post messages, commonly known as tweets (see Figure \ref{fig_twitter_website_screenshot}), which can reach a maximum of 140 characters. By default an account is public not private, \emph{i.e.} everybody can see the tweets published by this account without any authorisation by its owner. Users have also the option to follow other users and therefore see their tweets on their time-line, the main part of Twitter's web interface. For public accounts, becoming a follower does not need any type of approval from the followee; private accounts are able to authorise their followers. In that sense, each user -- with a public or a private account -- has a set of people who he follows (followees) and a set of people who follow him (followers). The only distinction is that connections in public accounts can be seen as one-sided relationships, whereas connected private accounts form a 2-sided -- but hidden -- relationship.

Users can mention or reply to each other in their tweets -- by using the character `@' followed by the username of the user to be mentioned -- and therefore, conduct electronic conversations. They can also reproduce the tweet of another user, an action known as retweet. Twitter incorporates topics in tweets; any word that starts with a hash (\#) is perceived as something that denotes a topic\index{Twitter!topic} (\emph{e.g.} $\#$OlympicGamesLondon2012).

\subsection{Characterisation of Twitter -- Why Twitter content is important}
\label{section:characterisation_twitter}
``\emph{If you think about our society as being a big organism this is just another tool to look inside of it}'', says Noah Smith about Twitter \cite{Savage2011}. In this section, we refer to several works which justify this statement in practice and highlight the importance of Twitter content.

Twitter's social network is comprised by two underlying sub-networks: a very dense one made up by followers and followees and a sparser one, where more close and probably real friends participate. The latter one is more influential as expected, yet the former reveals the great degree of connectivity in Twitter \cite{Huberman2008}. In this space of data, three distinct categories of users have been identified: broadcasters, acquaintances (who are the vast majority) and miscreants (stalkers or spammers) \cite{Krishnamurthy2008}.

Interestingly, Kwak \etal show that Twitter deviates from known human social networks having a non-power law follower distribution, a short effective diameter and low reciprocity\footnote{ \textbf{Reciprocity} is a quantity that characterises directed graphs by determining the degree of mutual connections -- double links with opposite directions -- in the graph \cite{Garlaschelli2004}.} \cite{Kwak2010}. Twitter users (in the United States) are also a highly non-uniform sample of the population and represent a highly non-random sample of the overall race/ethnicity distribution \cite{Mislove2011}.

%Nevertheless, tweets can be considered as public equivalent of a text message (or SMS)...
% public SMS
% justify why twitter data can frame a general case of data like mobile phones or blog posts
% advantages disadvantages compared to other social media
% \subsection{Why Twitter data is important}

An early work of Java \etal has shown that people use microblogging to talk about their daily activities and to seek or share information. An additional derivation presented in this paper was that users with similar intentions are more likely to be connected with each other \cite{Java2007a}. As a result, in this environment, conversationality is also promoted. But Twitter is not only a means for public interaction; it finds use in collaborative scenarios as well \cite{Honeycutt2009}. By posting tweets or conducting conversations on Twitter a feeling of connectedness is sustained in working environments \cite{Zhao2009b} and several aspects of educational schemes are improved \cite{Grosseck2008}. Twitter encourages free-flowing and just-in-time interaction between and among students and faculty \cite{Dunlap2009}, finds great applicability during conferences \cite{Reinhardt2009a} and is a great tool for promoting second language active learning methods \cite{Borau2009}.

The majority of topics on Twitter (approx. 85\% in 2010) are headline news or persistent news in nature \cite{Kwak2010} and methods have been developed enabling the detection of breaking news \cite{Sankaranarayanan2009}. Twitter also facilitates communication between communities with different political orientations. Similarly to real-life, an extremely limited connectivity between left- and right-leaning users is reported \cite{Conover2011}.

Furthermore, Twitter is a rich source for opinion mining and sentiment analysis \cite{Pak2010}. Consumer opinions are public for companies to track and then adapt their overall branding strategies based on this intelligence source \cite{Jansen2009a}. Polls can be easily conducted regarding consumer confidence and political opinion \cite{Connor2010}. TV stations have started to combine their broadcasts with social networks to improve interaction. In particular, Twitter can provide a better understanding of sentiment's temporal dynamics as a reaction to a political debate \cite{Diakopoulos2010}. Predicting the result of elections based on Twitter content is a much harder problem to solve, still some preliminary approaches have already been proposed on this topic \cite{Lui2010,tumasjan2010predicting,Gayo-Avello2011,Metaxas2011a,Lampos2012b}.

Due to the fact that microblogging usually expresses a real-time state of its author, tweets are considered more valuable than other media for connecting this information to personal experiences or situations of the users \cite{Zhao2009b}. Exploiting this fact and Twitter's deviating behaviour during mass convergence and emergency events \cite{Hughes2009a}, one can build applications to improve situational awareness during those events \cite{Vieweg2010a}. Alternatively, one can deploy methodologies that turn tweets to predictions about signals emerging in real life \cite{Lampos2011a}. For example, work is concentrated in detecting the occurrence and magnitude of disease outbreaks in the population \cite{Lampos2010f, Lampos2010, Culotta2010a, Signorini2011} or natural disasters such as an earthquake \cite{Sakaki2010}.

From the above, one may naturally come to a conclusion about the high degree of importance that Twitter content has. The interesting, one-sided nature of relationships, the retweet mechanism which allows for a rapid information spread, the ability to conduct online conversations and, on top of all that, the open Twitter API making this source of information easy to crawl, offer an opportunity to study human behaviour \cite{Kwak2010}.

% to be written elsewhere
%Every second the thoughts and feeling of millions of people across the world are recorded in the form of 140-character tweets \cite{Mislove2011}.
% articles from magazines
%iran protests: twitter the medium of the movement \cite{Grossman2009}
%how twitter will change the way we live \cite{Johnson2009}
% Twitter as medium and message \cite{Savage2011}
% twitter data may help answer sociological questions that are otherwise hard to approach because polling enough people is too expensive and time consuming
% but this is a news article
%``If you think about our society as being a big organism,'' says Noah Smith, ``this is just another tool to look inside of it.''
% discarded
% Use of Twitter even to recommend real-time topical news (based on a collection of RSS feeds)
% using twitter to recommend real-time topical news \cite{Phelan2009}
% novel news recommendation technique that harnesses real time twitter data as the basis for ranking and recommending articles from a collection of rss feeds

\section{Collecting, storing and processing Twitter data}
\label{section:coll_store_process_Twitter_data}
At the early stages of this work, there was no Twitter data set to work with and try our models. This is the main reason which forced us to create a pipeline that is able to collect and store tweets. With hindsight, this was an essential step.\footnote{ In the recent past, Twitter data sets used in academic studies have become publicly available. However, soon after they have been removed possibly because the open distribution of such content is against the company's policy.} In this short section, we describe the data collection process with references to external libraries or tools that have been used.

\begin{figure}[!t]
\centering
\includegraphics[width=6in]{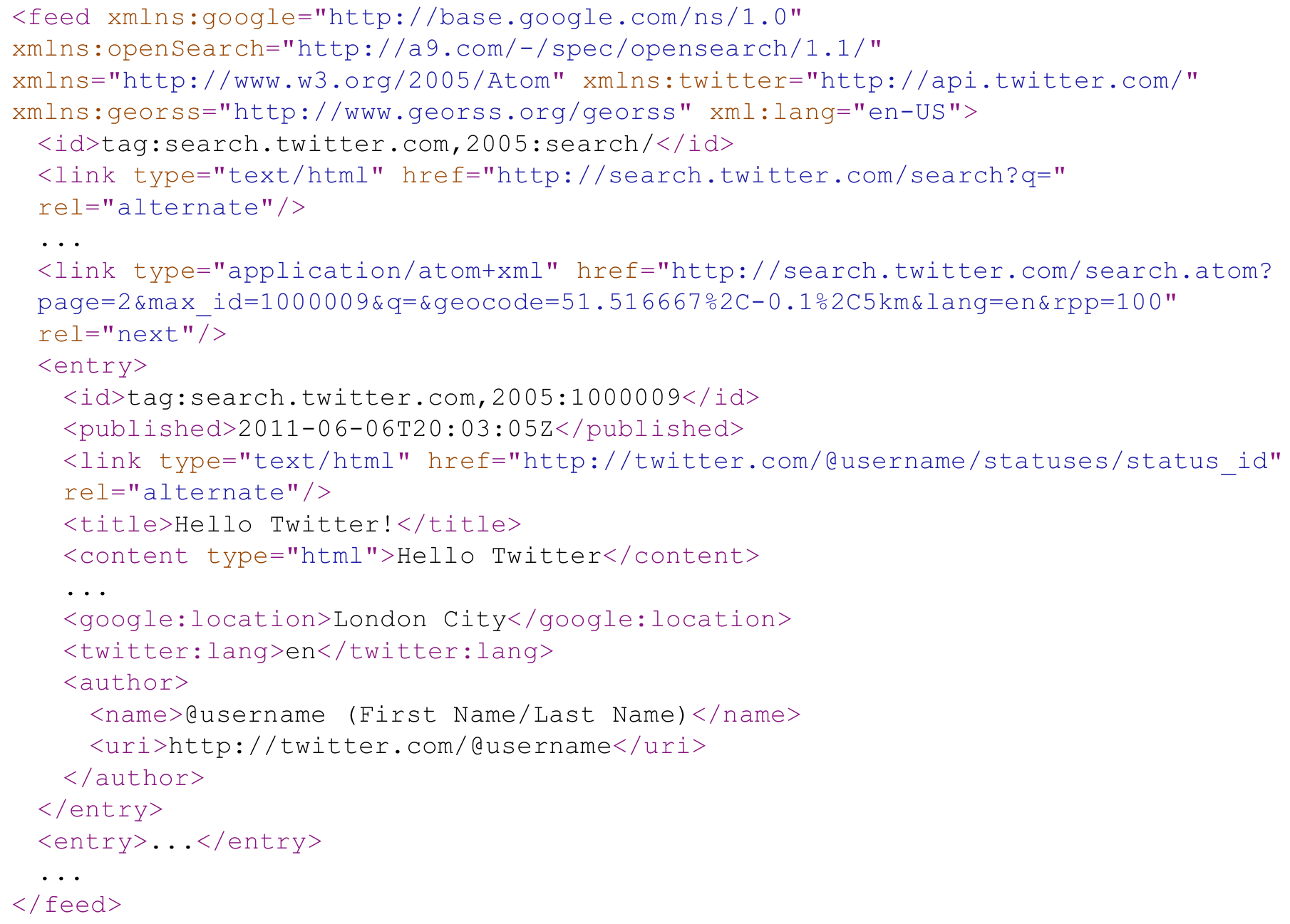}
\caption{A sample Atom feed retrieved by querying Twitter API}
\label{fig_twitter_atom_feed}
\end{figure}

\subsection{RSS and Atom feeds}
Really Simple Syndication\index{RSS feed} (\textbf{RSS})\footnote{ Or originally RDF Site Summary, where RDF stands for Resource Description Framework.} or simply a feed is a family of web formats based on Extensive Markup Language (\textbf{XML}\index{XML}) used for publishing and distributing web content in a standardised manner. A more recent development, the \textbf{Atom} Syndication Format\index{Atom feed}, is an effort to fix the limitations of RSS by supporting, for example, XML namespaces.\footnote{ From Wikipedia: ``XML namespaces are used for providing uniquely named elements and attributes in an XML document'', \url{http://en.wikipedia.org/wiki/XML_namespace}.} Twitter uses Atom feeds to deliver its content in a structured format; one has the option to submit a query on Twitter's API and retrieve Atom feeds which contain sets of tweets matching this query. An example of an Atom feed retrieved from Twitter is shown in Figure \ref{fig_twitter_atom_feed}. The main content -- the actual tweet -- of the feed lies inside the tags \texttt{<entry>} and \texttt{</entry>}. The purpose of the encapsulated sub-tags is self explanatory, for example \texttt{<published>} holds the posting date and time for a tweet.

\subsection{Collecting and storing tweets}
\label{section:crawlers_data_collection_storage}
For data collection, a basic crawler has been implemented in Java\index{Java} programming language; Java was preferred to other object-oriented programming frameworks mainly because it is a well-established programming interface for which many useful methods have already been made available by third-parties. Tweets are collected by querying Twitter's Search API\index{Twitter!Search API}.\footnote{ Twitter Search API, \url{https://dev.twitter.com/docs/api/1/get/search}.} Based on the fact that information geolocation is a key concept in our studies, we are only considering geolocated tweets, \ie tweets for which the author's location is known to Twitter.

In order not to exceed the request limit set by Twitter, something that usually causes suspensions in the data collection process, we limit our data collection to tweets geolocated within a 10Km radius of the 54 most populated urban centres in the UK.\footnote{ The cities considered in alphabetical order are the following: \emph{Aberdeen, Basildon, Belfast, Birmingham, Blackburn, Blackpool, Bolton, Bradford, Brighton, Bristol, Bournemouth, Cardiff, Coventry, Derby, Dundee, Edinburgh, Glasgow, Gloucester, Huddersfield, Hull, Ipswich, Leeds, Leicester, Liverpool, London, Luton, Manchester, Middlesbrough, Newcastle, Newport, Northampton, Norwich, Nottingham, Oldham, Oxford, Peterborough, Plymouth, Poole, Portsmouth, Preston, Reading, Rotherham, Sheffield, Slough, Southampton, Southend, Stockport, Stoke, Sunderland, Swansea, Swindon, Watford, Wolverhampton, York.} More information is also available in Appendix \ref{Ap:Urban_centres_per_region}.}

Here is an example query on Twitter's Search API\index{Twitter!Search API}:

\begin{center}
\href{http://search.twitter.com/search.atom?geocode=51.5166,-0.1,10km&lang=en&rpp=100&result_type=recent}{\textbf{http://search.twitter.com/search.atom?\\geocode=51.5166,-0.1,10km\&\\lang=en\&\\rpp=100\&\\result\_type=recent}}
\end{center}

With this query, we are able to retrieve the most recently published (\texttt{result\_type=recent}) 100 tweets (\texttt{rpp=100}) geolocated in a 10Km radius from a location with latitude $=$ 51.5166 and longitude $=$ $-$0.1 (\texttt{geocode=51.5166,-0.1,10km} -- this is an approximation for the centre of London) written in English language (\texttt{lang=en}). Twitter's response will produce an Atom feed that follows the format shown in Figure \ref{fig_twitter_atom_feed}. We collect and parse the XML\index{XML} content on this feed by using the Java libraries in ROME\footnote{ ROME Atom/RSS Java utilities, \url{http://java.net/projects/rome/}.} and then store its contents -- if they have not crawled before -- in a MySQL\index{MySQL} database.\footnote{ MySQL Relational Database Management System, \url{http://www.mysql.com/}.} Therefore, for each request we make, we can at most retrieve 100 new entries (tweets) to be stored in the database.

From the above, it becomes apparent that a sampling strategy has been followed during data collection. Every $n$ minutes, we form a search query for each one of the 54 locations and retrieve the 100 latest tweets. The value of $n$ varied from 1 to 10 minutes during this project and was mostly dependent on the total number of tweets we aimed to collect on a daily basis. Obviously, the shortest the collection period ($n$), the more tweets are being collected; as the number of Twitter users reached very high figures, we were able to reduce this frequency and still collect a large amount of tweets per day. Nevertheless $n$ is kept the same for all urban centres and therefore, possible sampling biases of this type are reduced.

\begin{figure*}
    \begin{center}
    \subfigure[Daily volume of collected tweets]{\includegraphics[width=3in]{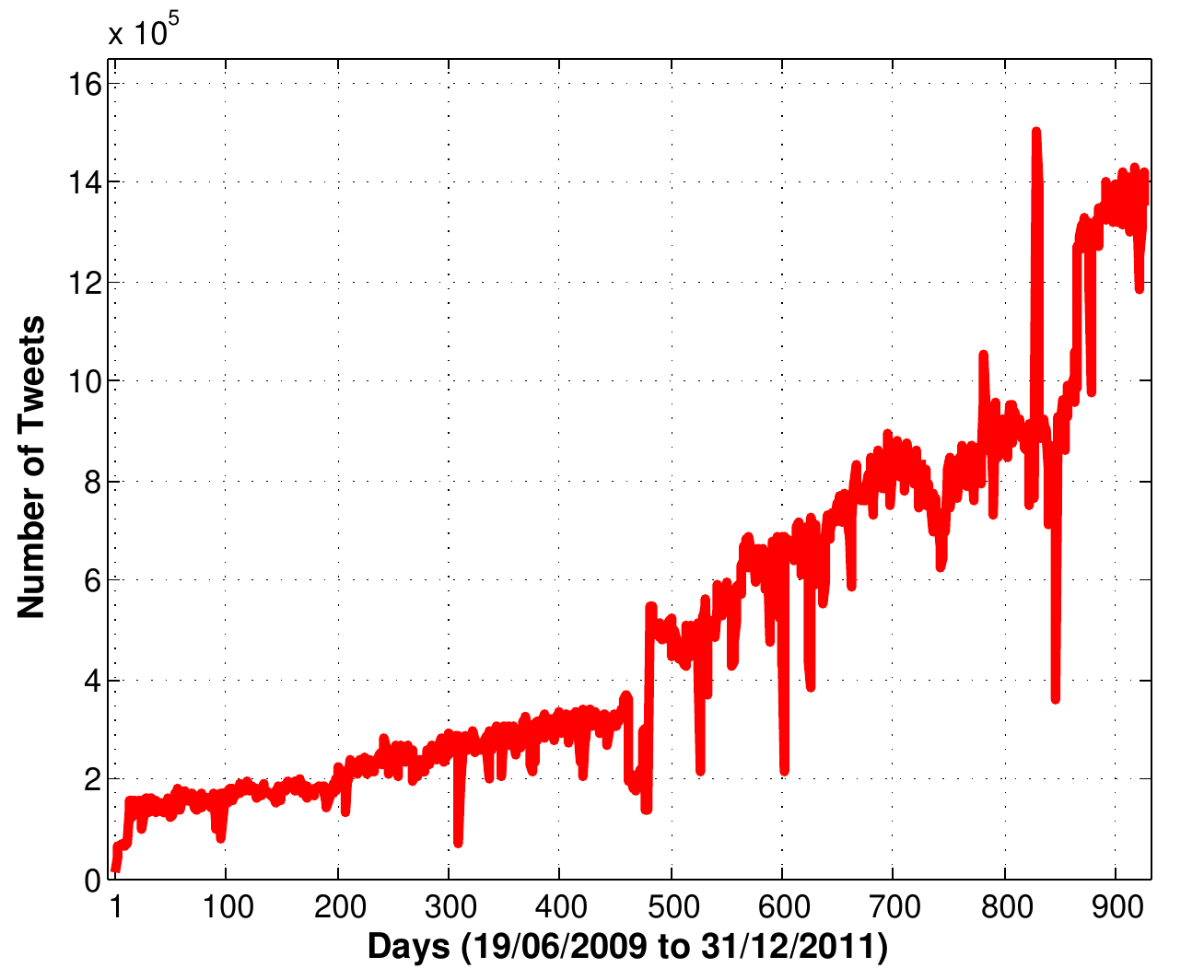}
    \label{fig_twitter_volume}}
    \hfil
    \subfigure[Cumulative volume plot]{\includegraphics[width=3in]{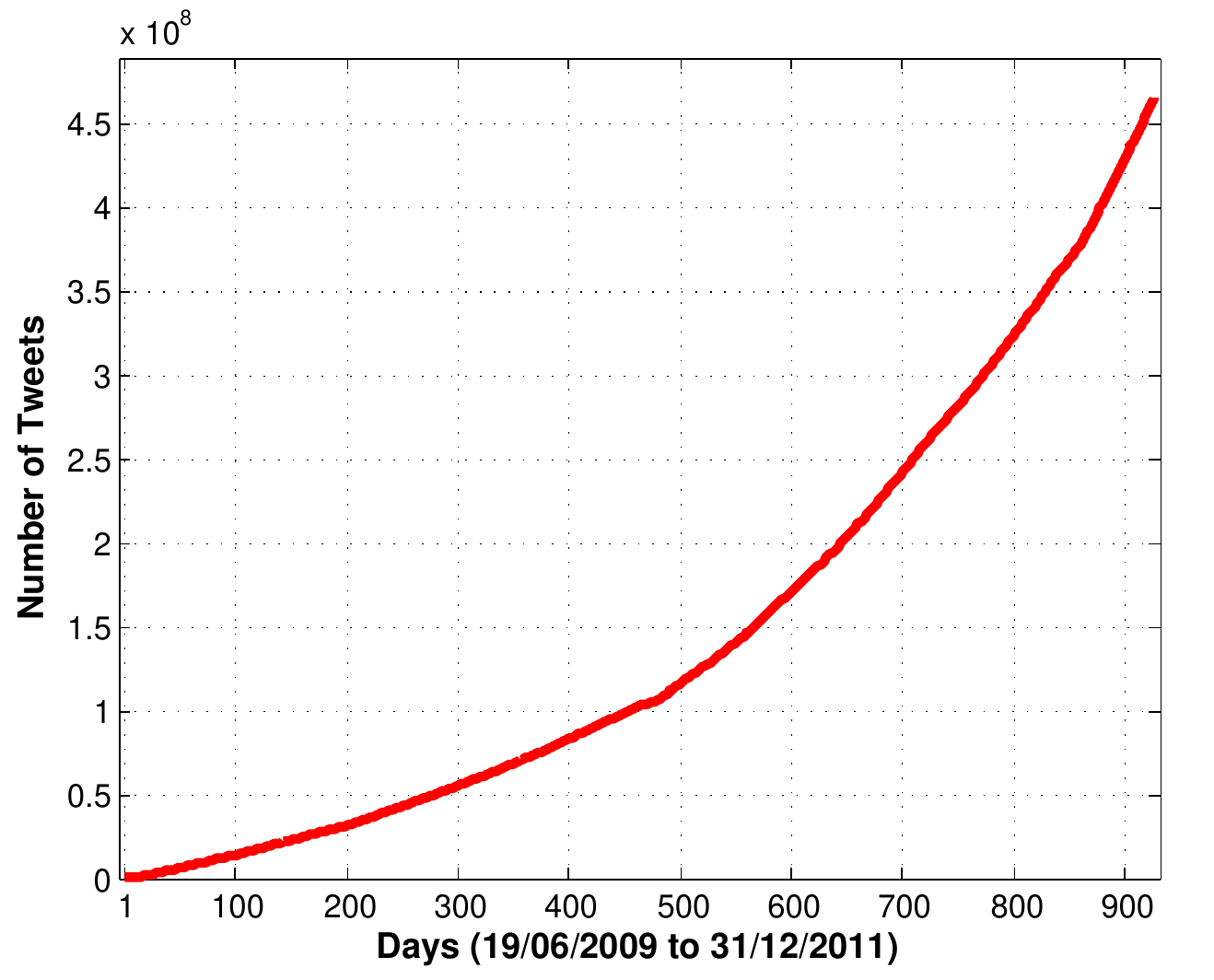}
    \label{fig_twitter_volume_cum}}
    \end{center}
    \caption{Volume and cumulative volume plots for tweets collected from the 19th of June, 2009 to the 31st of December, 2011.}
    \label{fig_twitter_vol}
\end{figure*}

\textbf{465,696,367} tweets (geolocated in the UK) have been collected by the crawler and then stored in the database from the 19th of June 2009 to the 31st of December 2011 (\textbf{926 days}). Those tweets have been published from at most \textbf{9,706,618 unique users}; the latter number is an approximation as we did not track users that might have continued to post from the same Twitter account but with a changed username.\footnote{ The research in this project did not study the behaviour of individual users, but was focused only on sets of published content.} Within those dates, a rough average number of tweets per user is approx. 50; 502,912 tweets have been collected per day on average. However, this is not a very good representation for the daily data collection figures as the number of registered Twitter users has been constantly increasing together with the published volume of content. To provide the reader with a better picture, we have plotted the daily volume of collected tweets in Figure \ref{fig_twitter_volume} as well as its cumulative equivalent in Figure \ref{fig_twitter_volume_cum}. From the figures, we observe that half-way through the considered period the number of collected tweets starts to increase with a higher acceleration and at a point we start collecting more than one million tweets per day. To reduce possible side-effects induced by the increasing volume of tweets and maintain an equal representation of the Twitter corpus between different dates, we tend to introduce a normalisation factor in our vector space models, which is usually equal to the number of tweets considered (see for example the methods in the Chapters \ref{Chapter_first_steps} and \ref{Chapter_Nowcasting_Events_From_The_Social_Web}).

\subsection{Software libraries for processing textual and numerical data}
\label{section_processing_textual_numerical_data}
Several software libraries have been implemented for applying basic methods and algorithms driven by the fields of IR, Data Mining and Machine Learning. A very common software package, used in most projects that deal with the analysis of textual content is Apache Lucene\index{Apache!Lucene},\footnote{ Apache Lucene, \url{http://lucene.apache.org/}.} a set of libraries -- initially programmed in Java -- able to implement a text search engine. We use Lucene to index tweets, \ie to tokenise and identify unique words in the textual stream, but also for counting word frequencies in documents (a document can be a superset of tweets). Optionally, Lucene can also create indices of stemmed text by applying for example Porter's algorithm \cite{porter1980}. Based on the fact that we already use the Java connector (or driver)\footnote{ MySQL Connector/J is the official JDBC driver for MySQL, \url{http://dev.mysql.com/downloads/connector/j/5.0.html}.} to query the database, it is almost straightforward to embed Lucene's libraries in our source code and create several types of indices.

Another useful library, which is again a project of Apache Software Foundation, is Mahout\index{Apache!Mahout}.\footnote{ Apache Mahout, \url{http://mahout.apache.org/}.} Mahout is capable of handling large scale data sets and implements the most common Machine Learning algorithms for clustering and classification or linear algebra operations such as Singular Value Decomposition. The key about Mahout is that its methods are compatible with Apache Hadoop\index{Apache!Hadoop}\footnote{Apache Hadoop, \url{http://hadoop.apache.org/}.} in that they are also using the map-reduce paradigm,\footnote{ About Map Reduce, \url{http://en.wikipedia.org/wiki/MapReduce}.} and therefore, can perform operations using a cluster of computers; when a cluster is not available Mahout can also operate using a single node. We use Mahout to retrieve vector space representations directly from a Lucene index, as there exist already implemented interfaces for this purpose. The VSM can be comprised by term TF or TF-IDF\index{VSM!TF-IDF} weights as well as other neat options which allow us to remove stop words or rarely used words. In some occasions, Mahout is also used for simple operations such as the computation of cosine similarities between pairs of large vectors.

Given the fact that sometimes we needed customised operations with our data which were not -- at least explicitly -- implemented in Lucene or Mahout, we have also created our own software libraries (in Java) which can be used to create a text index, perform stemming, remove stop words, compute vector space representations and finally perform our algorithms. Methods from MATLAB$^{\circledR}$'s native toolboxes as well as from PMTK3\footnote{ Probabilistic Modelling Toolkit, \url{http://code.google.com/p/pmtk3/}.} have been also used in this work.

%\section{Tools for sentiment and mood analysis}
%SentiWordnet \ref{}
%
%WordNet affect
%WordNet core terms
%Stanford POS Tagger

\section{Ground truth}
\label{section_ground_truth}
Ground truth\index{ground truth} is a necessary set of data for our work since many of our proposed methods are based on supervised learning. Hence, ground truth is not only needed for validation, but most importantly for learning the parameters of a model during the training process. We have used three types of ground truth:
\begin{itemize}
  \item Influenza-like Illness (\textbf{ILI})\index{Influenza-like Illness|see{ILI}}\index{ILI} rates from the Health Protection Agency\index{HPA}\index{Health Protection Agency|see{HPA}} (\textbf{HPA})\footnote{ Health Protection Agency, \url{http://www.hpa.org.uk/}.}, the Royal College of General Practitioners\index{RCGP} (\textbf{RCGP})\footnote{ Royal College of General Practitioners, \url{http://www.rcgp.org.uk/}.} \cite{fleming2007lessons} and the QSurveillance scheme (\textbf{QSur}).\footnote{ QSurveillance, University of Nottingham and Egton Medical Information Systems Ltd, \url{http://www.qresearch.org/Public/QSurveillance.aspx}.}
  \item Precipitation (rainfall) measurements from amateur weather stations in the UK. As no official authority could be found to provide us -- timely -- with rainfall indications, we used information collected by independent weather stations in Bristol, London, Middlesbrough, Reading and Stoke-on-Trent.
  \item Voting intention polls prior to the 2010 General Election in the UK from YouGov.\footnote{YouGov Document Archive, \url{http://labs.yougov.co.uk/publicopinion/archive/}.}
\end{itemize}

The first two sets of data (flu and rainfall rates) are used for the experimental process of Chapters \ref{Chapter_first_steps} and \ref{Chapter_Nowcasting_Events_From_The_Social_Web}, whereas voting intention polls are used in the preliminary work presented in Section \ref{section:voting_intentions}.

\section{Summary of the chapter}
\label{section_summary_data_chapter}
In this chapter, we started by defining the terms blog and microblog. We moved on by describing the main operations and characteristics of Twitter, a microblogging service which serves as the main source of information for this project. Twitter has attracted the focus of the academic community in the last few years as it allows researchers to study human behaviour using massive amounts of data. This platform has found also use in many other applications such as opinion mining, event detection and education. Next, we described in detail the way we use Twitter's Search API in order to retrieve tweets geolocated in 54 urban centres in the UK; we also provided the reader with figures reporting the average number of collected tweets per day and their cumulative distribution. Then, we referred to some of the most important software tools that we have used to process text, form vector space representations and carry out statistical operations. Finally, we gave a short description of the various versions of ground truth used throughout this project. 

%% file: Chapters/Chapter4.tex
\chapter{First Steps on Event Detection in Large-Scale Textual Streams}
\label{Chapter_first_steps}
%\lhead{Chapter 4. First Steps on Event Detection in Large-Scale Textual Streams}

\rule{\linewidth}{0.5mm}
Tracking the spread of an epidemic disease like seasonal or pandemic influenza is an important task that can reduce its impact and help authorities plan their response. In particular, early detection and geolocation of an outbreak are important aspects of this monitoring activity. Various methods are routinely employed for this monitoring, such as counting the consultation rates of general practitioners. We report on a monitoring methodology capable of measuring the prevalence of disease in a population by analysing the contents of social networking tools, such as Twitter. Our method is based on the analysis of hundreds of thousands of tweets per day, searching for symptom-related statements or automatically identifying potentially illness-related terms, and turning statistical information into a flu-score. We have tested it in the UK for 24 weeks during the H1N1\index{H1N1} flu pandemic. We compare our flu-score with data from the Health Protection Agency, obtaining on average a statistically significant linear correlation. This preliminary method uses completely independent data to that commonly used for these purposes, and can be used at close time intervals, hence providing inexpensive and timely information about the state of an epidemic. This chapter is an extended version of our paper ``Tracking the flu pandemic by monitoring the Social Web'' \cite{Lampos2010f}, which -- to our knowledge --, is the first one showing that content from the Social Media\index{Social Media} can be used to detect and infer the rate of an illness in the population.
\newline \rule{\linewidth}{0.5mm}
\newpage

% \section{Introduction}
% chapter structure
% chapter purpose

%%%%%%%%%%%%%%%%%%%%%%%%%%%%%%%%%%%%%%%%%%%%%%%%%%%%%%%%%%%%%%%%%%%% no LASSO - paper 1
\section{Introduction}
\label{section_intro_chapter4}
Monitoring the diffusion of an epidemic in a population is an important and challenging task. Information gathered from the general population can provide valuable insight to health authorities about the location, timing and intensity of an epidemic, or even alert the authorities of the existence of a health threat.  Gathering this information, however, is a difficult as well as resource-demanding procedure.

Various methods can be used to estimate the actual number of patients affected by a given illness, from school and workforce absenteeism figures \cite{neuzil2002illness}, to phone calls and visits to doctors and hospitals \cite{elliot2009monitoring}. Other methods include randomised telephone polls, or even sensor networks to detect pathogens in the atmosphere or sewage (\cite{ivnitski1999biosensors}, \cite{metcalf1995environmental}). All of these methodologies require an investment in infrastructure and have various drawbacks, such as the delay due to information aggregation and processing times (for some of them).

Recent research has pointed out that the use of search engine data could detect geographic clusters with a heightened proportion of health-related queries, particularly in the case of ILI\index{ILI} \cite{ginsberg2008detecting}. It demonstrated that timely and  reliable information about the diffusion of an illness can be obtained by examining the content of search engine queries.

The work presented in this chapter extends that concept, by monitoring the content of social-web tools such as Twitter, a micro-blogging website, where users have the option of updating their status with their mobile phone device. These updates, commonly known as tweets, are limited to 140 characters only, similarly to the various schemes for mobile text messaging. We analyse the stream of data generated by Twitter in the UK and extract from it a score that quantifies the diffusion of ILI in various regions of the country. The score, generated by applying Machine Learning technology, is compared with official data from the HPA\index{HPA}, with which it has a statistically significant linear correlation coefficient that is greater than 95\% on average.

The advantage of using Twitter to monitor the diffusion of ILI is that it can reveal the situation on the ground by utilising a stream of data (tweets) created within a few hours, whereas the HPA releases its data with a delay of 1 to 2 weeks. Furthermore, since the source of the data is entirely independent of search engine query-logs or any of the standard approaches, our method can also be used in combination with them, to improve accuracy.

\section{An important observation}
\label{section_an_important_observation}
In early stages of this work, an important observation was made. The idea was simple: ``What if one used the frequencies of a set of words in Twitter to pick signals emerging in real life?'' In the following sections, we describe the first experiment that provided us with evidence for the existence of such signals in the social web.

\subsection{Computing a term-based score from Twitter corpus}
\label{section_computing_a_term_based_score}
We compile a set of textual markers $\mathcal{M} = \{m_i\}$, where $i =$ 1, 2, ..., $k$, and look for them in the Twitter corpus of a day. A textual marker can be a single word (1-gram) or a set of words forming a phrase (or $n$-gram, where $n$ is the number of words\index{n-gram}). The set of markers $\mathcal{M}$ should point to a target topic -- we have chosen an illness, and in particular flu. This choice was not very random; picking flu rates from the social web is an important task that could help in the early detection of epidemics and there is also some ground truth provided from official health authorities, \ie data that could validate the performance of our methodologies.

The daily set of tweets is denoted as $\mathcal{T} = \{t_j\}$, where $j =$ 1, 2, ..., $n$. If a marker $m_i$ appears in a tweet $t_j$, we set $m_i(t_j) = 1$; otherwise $m_i(t_j) = 0$. The score of a tweet $s(t_j)$ is equal to the number of markers it contains divided by the total number of markers used:
\begin{equation}
s(t_j) = \frac{\sum_i m_i(t_j)}{k},
\end{equation}
where $k$ denotes the total number of markers. We compute the target topic's score from the daily Twitter corpus $f(\mathcal{T},\mathcal{M})$, as the sum of all the tweet scores divided by the total number of tweets:
\begin{equation}
f(\mathcal{T},\mathcal{M}) = \frac{\sum_j s(t_j)}{n} = \frac{\sum_j \sum_i m_i(t_j)}{k \times n},
\end{equation}
where $n$ denotes the total number of tweets for one day. The number of tweets is not the same for every day, thus it is a necessary choice to normalise the score by dividing with the daily number of tweets in order to retrieve a comparable time series of scores.

%%%%%%%%%%%%%%%%%%%%%%%%%%%%%%%%%%%%%%%%%%%%%%%%%%%%%%%%%%%%%%%%%%%%%%%%%%% computing correlations
\subsection{Correlations between Twitter flu-scores and HPA flu rates}
\label{subsection:gtruth_expansion}
As our ground truth basis, we use weekly reports from the HPA\index{HPA} related to ILI figures (see also Section \ref{section_ground_truth}). HPA provides weekly, regional statistics for the UK, based on rates gathered by the RCGP\index{RCGP} \cite{fleming2007lessons}, and the more general QSur scheme. RCGP and QSur metrics express the number of GP consultations per $10^5$ citizens, where the result of the diagnosis was ILI.

For this experiment, we are using RCGP data for four UK regions, namely Central England \& Wales (region A), South England (region B), North England (region C), and England \& Wales (region D), whereas QSur covers England, Wales \& Northern Ireland (region E). In order to retrieve an equal representation between the weekly HPA flu rates and the daily Twitter flu-scores, we expand each point of the former over a 7-day period; in fact, each weekly point of an original HPA flu rate is assigned on every day of the respective week. After expanding the HPA flu rates, we perform smoothing on them with a 7-point moving average.\footnote{ $N$-point moving average is explained in Appendix \ref{Ap:MovingAverage}.} Figure \ref{fig_HPAFluScore_smoothed} shows the expanded and smoothed HPA's flu rates for regions A-E during weeks 26-49 in 2009. The early peak in all time series reflects the swine flu (H1N1\index{H1N1}) epidemic that emerged in the UK during June and July, 2009 \cite{Jick2011}.

\begin{figure}[!t]
\centering
\includegraphics[width=5in]{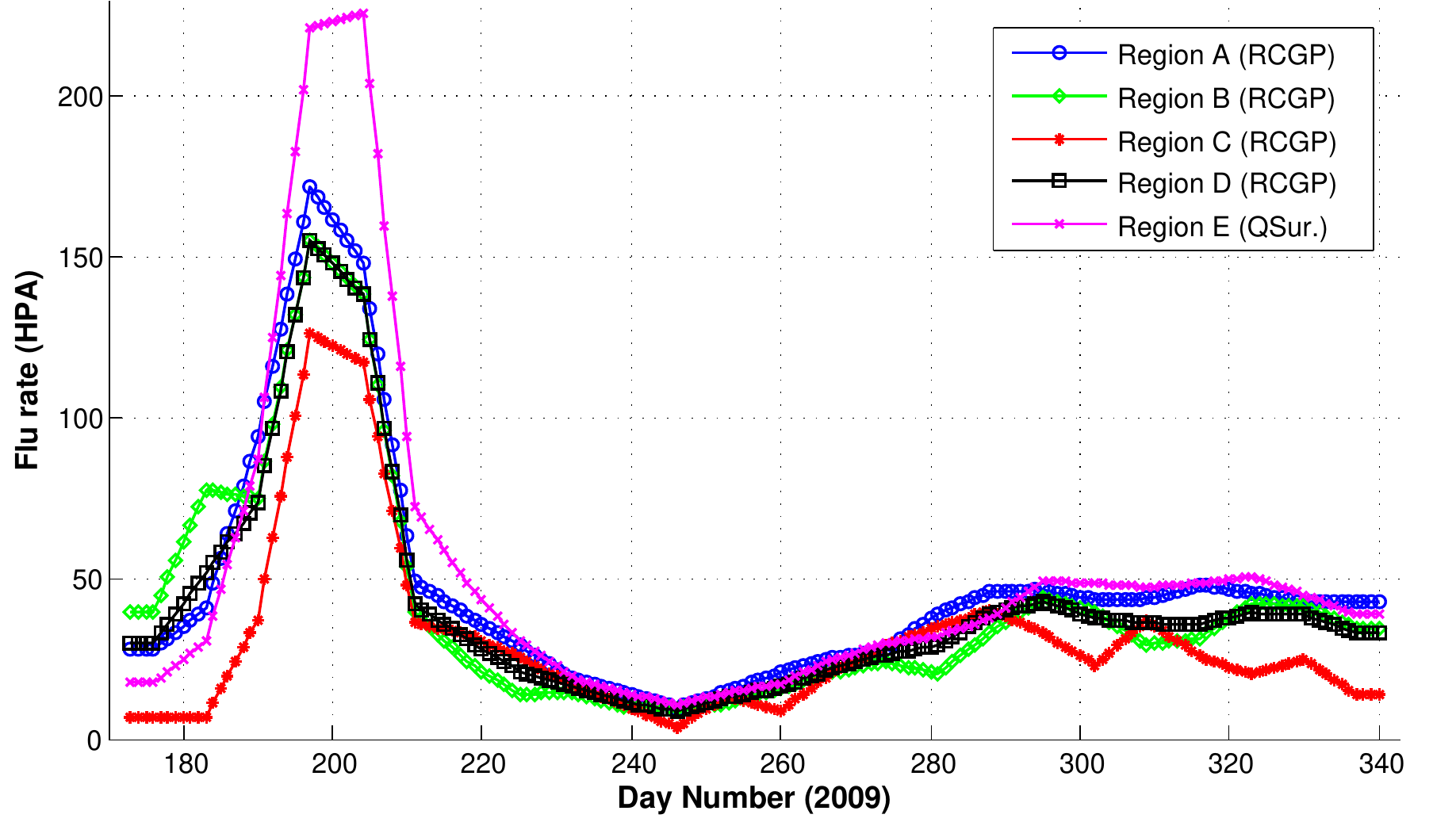}
\caption{Flu rates from the Health Protection Agency (HPA) for regions A-E (weeks 26-49, 2009). The original weekly HPA's flu rates have been expanded and smoothed in order to match with the daily data stream of Twitter.}
\label{fig_HPAFluScore_smoothed}
\end{figure}

\begin{figure}[!t]
\centering
\includegraphics[width=5in]{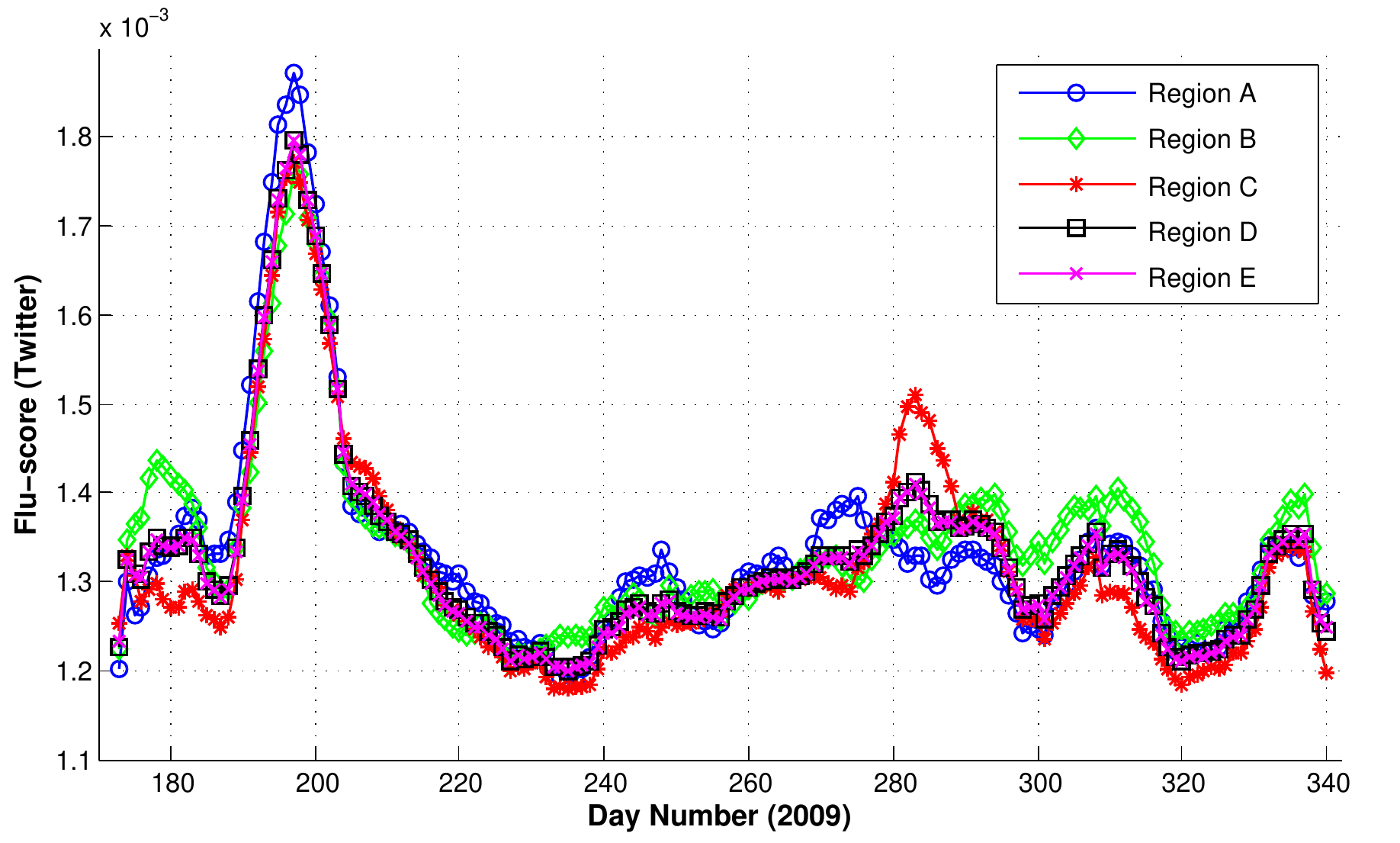}
\caption{Twitter's flu-scores based on our choice of markers for regions A-E (weeks 26-49, 2009). Smoothing with a 7-point moving average (the length of a week) has been applied.}
\label{fig_FluScore_smoothed}
\end{figure}

Using a small set of 41 textual markers expressing illness symptoms or relevant terminology, \emph{e.g.} `fever', `temperature', `sore throat', `infection', `headache', and so on,\footnote{ The entire set of markers is available at Appendix \ref{Ap:41markers}.} we compute the Twitter flu-score time series for regions A--E. We smooth each time series with a 7-point moving average in order to express a weekly tendency in our data. Figure \ref{fig_FluScore_smoothed} shows the time series of Twitter's flu-scores for days 173 to 340 in 2009.

For each one of the five regions, we compute the linear correlation coefficients between Twitter's and HPA's flu-score time series. The correlations and their respective p-values are presented in Table \ref{table_corrcoef1}; the average correlation is equal to 0.8387. The largest correlation of 0.8556 (with a p-value of 2.39e-49) is found for region D, whereas the smallest reported correlation is 0.8178 for region E (with a p-value of 1.1e-41).

%%%% revisions %%%%%%%%%%
Investigating those linear correlations further, we try different window spans for the moving average smoothing function as well as a different approach known as locally weighted scatterplot smoothing (\textbf{lowess})\index{lowess}. Each smoothed value produced by lowess is essentially given by performing a weighted OLS regression across the points of the smoothing window.\footnote{ Lowess algorithm is presented and thoroughly described in \cite{Cleveland1979}. For our purposes we are using its implementation in MATLAB.} The results are presented in Table \ref{table_corrcoef1_dif_smoothing}; we see that increasing the span of the smoothing window produces improved linear correlations under both smoothing functions. Overall moving average smoothing yields better correlations than lowess.
%%%%%%%%%%%%%%%%%%%%%%%%%

\begin{table}[!t]
\renewcommand{\arraystretch}{1.3}
\caption{Linear correlation coefficients between flu-scores derived from Twitter content and HPA's rates based on our choice of markers (for weeks 26-49, 2009) using a 7-point moving average.}
\label{table_corrcoef1}
\centering
\begin{tabular}{ccccc}
\hline
& \textbf{Region} & \textbf{HPA Scheme} & \textbf{Corr. Coef.} & \textbf{P-value}\\\hline
A & Central England \& Wales & RCGP & 0.8471 & 1.95e-47    \\%\hline
B & South England & RCGP & 0.8293 & 8.37e-44    \\%\hline
C & North England & RCGP & 0.8438 & 9.84e-47    \\%\hline
D & England \& Wales & RCGP & 0.8556 & 2.39e-49    \\%\hline
E & England, Wales \& N. Ireland & QSur & 0.8178 & 1.10e-41    \\\hline
\end{tabular}
\end{table}

%%%%% revisions %%%%%%%%%%%%
\begin{table}[!t]
\renewcommand{\arraystretch}{1.3}
\caption{Average linear correlation coefficients across all the considered regions between flu-scores derived from Twitter content and HPA's rates for two smoothing functions (n-point moving average and lowess) and various smoothing window sizes. The p-value for all the correlations is $<$ 1.0e-21.}
\label{table_corrcoef1_dif_smoothing}
\centering
\begin{tabular}{lcccccccc}
\hline
                            & --     & \textbf{3} & \textbf{5} & \textbf{7} & \textbf{9} & \textbf{11} & \textbf{13} & \textbf{15}\\\hline
\textbf{\emph{n}-point MA}  & 0.6933 & 0.7629     & 0.8096     & 0.8387     & 0.8559     & 0.8728      & 0.8863      & 0.8926\\
\textbf{lowess}             & 0.6933 & 0.6933     & 0.7576     & 0.7899     & 0.8149     & 0.8356      & 0.8506      & 0.8625\\\hline
\end{tabular}
\end{table}
%%%%%%%%%%%%%%%%%%%%%%%%%%%%

In Figure \ref{fig_compare_eng_wales}, we have plotted the z-scores\footnote{ See Appendix \ref{Ap:Mean_Std_Zscore} for the definition of a z-score.} of Twitter flu score against the corresponding official flu rates from HPA for region D (England \& Wales). The high correlation of those two signals can directly be noticed without the need of any statistical measure. This was our first important observation. In the following sections, we explain -- in detail -- the steps we took in order to actually build on this observation.

\section{Learning HPA's flu rates from Twitter flu-scores}
We extend our previous scheme in order to form a model for predicting a regional HPA flu rate by observing its corresponding flu-score on Twitter. In the new scheme, we attach a weight $w_i$ to each textual marker $m_i$. The weighted flu-score of a tweet is equal to:
\begin{equation}
s_{w}(t_j) = \frac{\sum_i w_i \times m_i(t_j)}{k},
\end{equation}
where $k$ denotes the number of markers. Similarly, the weighted flu-score based on Twitter's daily corpus $\mathcal{T}$ is computed by:
\begin{equation}
f_{w}(\mathcal{T},\mathcal{M}) = \frac{\sum_j s_{w}(t_j)}{n} = \frac{\sum_j \sum_i w_i \times m_i(t_j)}{k \times n},
\end{equation}
where $n$ denotes the total number of tweets for this day. The contribution of each marker $m_i$ in $f_{w}$ can be considered as a flu-subscore and is equal to:
\begin{equation}
f_{w_i}(\mathcal{T}, m_i) = w_i \times \frac{\sum_j m_i(t_j)}{k \times n}.
\end{equation}
Therefore, a daily flu-score derived from Twitter content can be represented as a vector $\mathcal{F}_w$ of $k$ elements $\mathcal{F}_w = [f_{w_1}(\mathcal{T}, m_1)\text{, ..., }f_{w_k}(\mathcal{T}, m_k)]^T$ each one corresponding to Twitter's flu-subscore for marker $m_i$.

\begin{figure}[!t]
\centering
\includegraphics[width=5in]{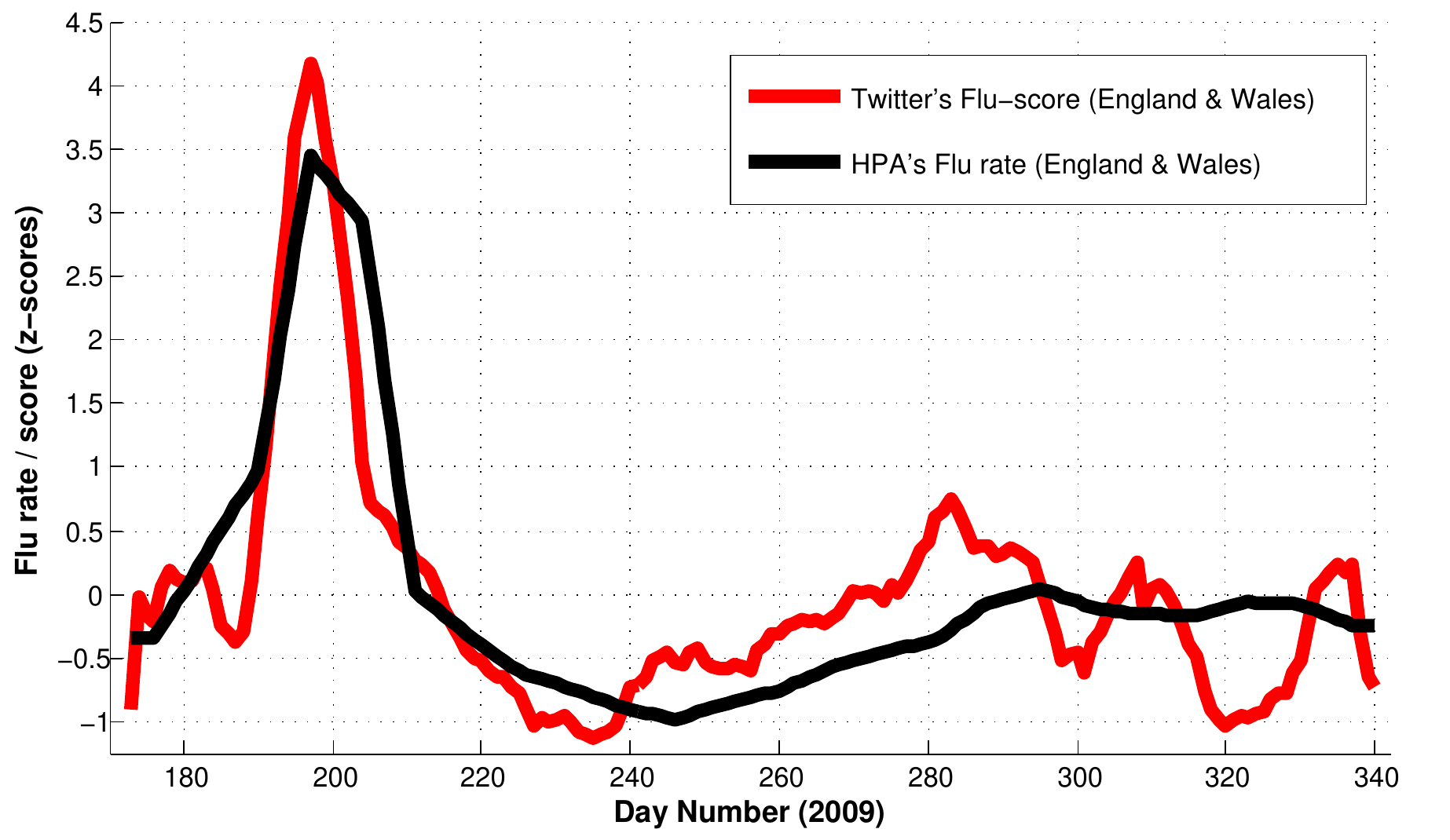}
\caption{Comparison of the unweighted Twitter's flu score (based on our choice of markers) and the respective HPA rates for region D (England \& Wales) using their z-scores. Their linear correlation is equal to 0.8556.}
\label{fig_compare_eng_wales}
\end{figure}

Initially, we retrieve from the Twitter corpus of a day an unweighted flu-score vector $\mathcal{F} = [f(\mathcal{T}, m_1)\text{, ..., }f(\mathcal{T}, m_k)]^T$. The unweighted time series of each term's flu-subscores ($f(\mathcal{T}, m_i)$ for all the days) is smoothed with a 7-point moving average. We perform OLS regression between the time series of $\mathcal{F}$'s smoothed version and the expanded and smoothed HPA's flu rates in order to learn the weights $w_i$ for the terms $m_i$. We use as a training set the data that correspond to one region, and then we test the predictability of the inferred weights on the remaining four regions. We perform this training/testing method for all possible (five) training choices.

The linear correlation coefficient between the inferred and the official time series for the HPA flu rates is used as the performance indicator. The results are presented in Table \ref{table_lg_results}; the correlation coefficients that were retrieved after training on a region A-E are presented in the row A-E respectively. The average performance over all possible training and testing choices is equal to 0.8994 and the p-values of all the presented correlations indicate strong statistical significance (all of them are $<$ 10e-32). We achieve the maximum average performance (0.9314) when tweets from region A (Central England \& Wales) are used for training; the linear correlation of the flu-scores' time series and the HPA flu rates for region E by applying the weights learnt from region A is equal to 0.9723.%In Figure \ref{fig_compare_regression} we have plotted the inferred time series for regions B-E (after training on region A) in comparison with the HPA flu rates.

%%% revisions %%%%%%%%%%%
Similarly to the previous section, we also report results for various smoothing window spans as well as for the lowess smoothing function (Table \ref{table_corrcoef2_dif_smoothing}). Interestingly, we see that, under both smoothing methods, correlations reach a maximum for a window of 5 points (\ie days) and then as the smoothing time span increases -- especially for the lowess function -- performance shows a decreasing behaviour. Overall, lowess does not perform better than moving average; in general, the two smoothing functions produce results of a very similar performance. Based on the facts that: (a) linear correlations for the spans of 5 and 7 do not differ much, (b) using a 7-point moving average produces a more interpretable result as it unveils a weekly tendency from the data, and (c) moving average performs slightly better than lowess, in the results presented in the following sections, we only perform smoothing using a 7-point moving average, leaving further investigation of smoothing spans and methodologies to future work.
%%%%%%%%%%%%%%%%%%%%%%%%%

\begin{table}[!t]
\renewcommand{\arraystretch}{1.3}
\caption{Linear regression using the markers of our choice - An element $(i,j)$ denotes the correlation coefficient between the weighted flu-scores time series and HPA's flu rates on region $j$, after training the weights on region $i$. The p-value for all the correlations is $<$ 10e-32.}
\label{table_lg_results}
\centering
\begin{tabular}{ccccccc}
\hline
Train/Test & \textbf{A} & \textbf{B} & \textbf{C} & \textbf{D} & \textbf{E}           & Avg.              \\\hline
\textbf{A} & -           & 0.8389      & 0.9605      & 0.9539      & 0.9723           & \textbf{0.9314}   \\%\hline
\textbf{B} & 0.7669      & -           & 0.8913      & 0.9487      & 0.8896           & 0.8741            \\%\hline
\textbf{C} & 0.8532      & 0.702       & -           & 0.8887      & 0.9445           & 0.8471            \\%\hline
\textbf{D} & 0.8929      & 0.9183      & 0.9388      & -           & 0.9749           & 0.9312            \\%\hline
\textbf{E} & 0.9274      & 0.8307      & 0.9204      & 0.9749      & -                & 0.9134            \\\hline
           &             &             &             & \multicolumn{2}{c}{Total Avg.} & \textbf{0.8994}   \\\cline{5-7}
\end{tabular}
\end{table}

%%%%% revisions %%%%%%%%%%%%
\begin{table}[!t]
\renewcommand{\arraystretch}{1.3}
\caption{Average linear correlation coefficients across all the considered regression scenarios between the inferred flu-scores derived from Twitter content and HPA's rates for two smoothing functions (n-point moving average and lowess) and various smoothing window sizes. The p-value for all the correlations is $<$ 1.0e-11.}
\label{table_corrcoef2_dif_smoothing}
\centering
\begin{tabular}{lcccccccc}
\hline
                            & --     & \textbf{3} & \textbf{5} & \textbf{7} & \textbf{9} & \textbf{11} & \textbf{13} & \textbf{15}\\\hline
\textbf{\emph{n}-point MA}  & 0.8849 & 0.9078     & 0.9101     & 0.8994     & 0.9039     & 0.9055      & 0.8618      & 0.8152\\
\textbf{lowess}             & 0.8849 & 0.8849     & 0.9072     & 0.9064     & 0.9002     & 0.8898      & 0.8627      & 0.83\\\hline
\end{tabular}
\end{table}
%%%%%%%%%%%%%%%%%%%%%%%%%%%%

%\begin{figure*}[!t]
%    \begin{center}
%    \subfigure[Region B (South England) - correlation: 83.69\%]{\includegraphics[width=3in]{Compare_HPA_Reg_SEng}
%    \label{fig_comp1}}
%    \hfil
%    \subfigure[Region C (North England) - correlation: 96.05\%]{\includegraphics[width=3in]{Compare_HPA_Reg_NEng}
%    \label{fig_comp2}}
%    \subfigure[Region D (England \& Wales) - correlation: 95.39\%]{\includegraphics[width=3in]{Compare_HPA_Reg_Eng_Wales}
%    \label{fig_comp3}}
%    \hfil
%    \subfigure[Region E (England, Wales \& N. Ireland) - correlation: 97.23\%] {\includegraphics[width=3in]{Compare_HPA_Reg_Eng_Wales_NIr}
%    \label{fig_comp4}}
%    \end{center}
%    \caption{Comparison plots for the inferred (using linear regression) versus the HPA's flu rates time series in regions B-E. Training has been performed using the time series of smoothed $\mathcal{F}$ (based on our selection of markers) and the HPA rates of region A (Central England \& Wales) for weeks 26-49, 2009.}
%    \label{fig_compare_regression}
%\end{figure*}

To assess the predictive power of the former result differently, we perform linear regression on the aggregated time series of the flu-scores and HPA's flu rates using the data from all the regions. The data belonging in weeks 28 and 41 (during the peak and the stabilised period of the epidemic respectively) form the test set; the remaining data is used for training the weights. This results to a linear correlation of 0.9234 with a p-value of 5.61e-30 on the test set. Additionally, we perform a 10-fold cross validation (1000 repeats, where the folds are randomly decided each time) using again linear regression for learning. On average, we obtain a linear correlation of 0.9412 with a standard deviation equal to 0.0154.

While in all cases the score was tested on unseen data, in the first set of experiments we trained on data gathered on one region, and then tested on the remaining regions, but on the same period of time; in the last two experiments, using an aggregation of our data sets, we carried out training and testing on different times. Together these two sets of experiments, provide strong support to the predictive power of the flu-score we developed.

%%%%%%%%%%%%%%%%%%%%%%%%
\subsection{What is missing?}
In the previous sections, we used a set of handpicked markers related to the target topic of illness to infer ILI scores in several regions in the UK. Obviously, this is not an optimal choice; how can one know that the optimal set of markers is being used, especially when she deals with a vague concept? In the following sections, we show ways to select those features automatically, enabling our approach to be general and independent of human effort and domain knowledge.

%%%%%%%%%%%%%%%%%%%%%%%%%%%%%%%%%%%%%%%%%%%%%%%%%%%%%%% LASSO paper 1 - correct notation to fit chapter 2
\section{Automatic extraction of ILI textual markers}
\label{section_lasso_paper_1}
In the previous sections, we made use of hand crafted ILI related textual markers. In this section, we present a method for extracting weighted markers (or features) automatically. The method selects a subset of keywords and their weights to maximise
the correlation with the HPA flu rates, while also minimising the size of the keyword set. It is formed by two parts: creating a set of candidate features, and then selecting the most informative ones.
%As long as the initial set is not too large, it can identify a useful subset.

At first, we create a pool of candidate markers from web articles related to influenza. We use an encyclopedic reference\footnote{ Influenza on Wikipedia, \url{http://en.wikipedia.org/wiki/Influenza}.} as well as a more informal reference where potential flu patients discuss their personal experiences.\footnote{ Swine Flu on NHS (with potential patients comments), \url{http://www.nhs.uk/Conditions/pandemic-flu/Pages/Symptoms.aspx}.} After preprocessing (tokenisation, stop-word removal), we extract a set of $K =$ 1560 stemmed candidate markers (1-grams). The latter is denoted by $\mathcal{M}_{C} = \{m_{ci}\}$, $i \in [1,K]$. $\mathcal{M}_C$ contains some words which form a very good description of the topic as well as many irrelevant ones. This formation choice for the set of candidate markers is discussed further and justified in Sections \ref{section_harry_potter_effect} and \ref{section_error_bounds_for_LASSO}.

After forming the candidate features, we compute their daily, regional, and unweighted flu-subscores $f(\mathcal{T}_r, m_{ci})$ given $\mathcal{T}_r$ which denotes the Twitter corpus for region $r$, $r \in \{\text{A-E}\}$. For a day $d$, the flu score on Twitter is represented as a vector $\mathcal{F}_{d,r} = [f(\mathcal{T}_r, m_{c1})\text{ ... }f(\mathcal{T}_r, m_{cK})]^{T}$. Consequently, for a region $r$ and a period of $\ell$ days, we can form an array with the time series of the flu-subscores for all the candidate features: $X_{r} = [\mathcal{F}_{1,r}\text{ ... }\mathcal{F}_{\ell,r}]^{T}$, where $\ell$ denotes the total number of days considered. The columns of $X_{r}$, \emph{i.e.} the time series of the flu-subscores of each candidate feature, are smoothed using a 7-point moving average (as in the previous cases); the resulting array is denoted as $X^{(s)}_{r}$.

The expanded and smoothed time series of the HPA's flu rates\index{flu rates} for region $r$ and for the same period of $\ell$ days are denoted by the vector $h^{(s)}_{r}$. At this point, one could use the correlation coefficient between each column of $X^{(s)}_{r}$ and $h^{(s)}_{r}$ or other linear regression methods (least squares, rigde regression, etc.) in order to rank or learn weights for the candidate features. For this purpose, the LASSO method has been chosen as it has the advantage of producing sparse solutions, \emph{i.e.} it will discard candidate features which are proven to be redundant in terms of predictability \cite{tibshirani1996regression}. LASSO\index{regression!LASSO} is an established method for estimating least squares parameters subject to an L1 penalty. It can be considered as a constrained optimisation task, which in our case is formulated as
\begin{equation}
\renewcommand{\arraystretch}{1.2}
\begin{array}{cc}
\displaystyle\min_w & \|X^{(s)}_{r}w - h^{(s)}_{r}\|_{2}^{2}\\
\text{s.t.} & \|w\|_{1} \leq t,
\end{array}
\end{equation}
where vector $w$ is the sparse solution, and $t$ is the shrinkage parameter. The shrinkage parameter can be expressed as
\begin{equation}
t = \alpha \times \|w^{(ls)}\|_{1},
\end{equation}
where $w^{(ls)}$ denotes the least squares estimates for our regression problem, and $\alpha\in(0,1)$ is the shrinkage percentage.

\begin{figure*}
    \begin{center}
    \subfigure[Region C - Correlation: 0.9349 (p-value: 1.39e-76)] {\includegraphics[width=4in]{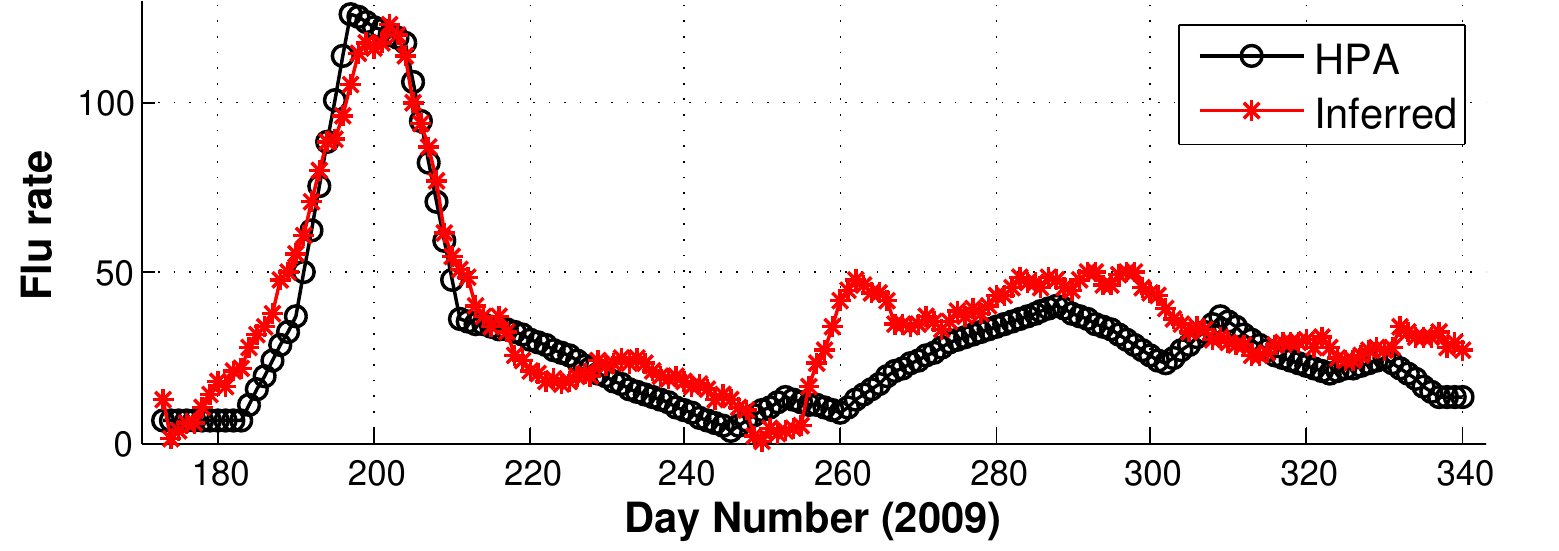}
    \label{fig_comp1}}
    \hfil
    \subfigure[Region D - Correlation: 0.9677 (p-value: 2.98e-101)]{\includegraphics[width=4in]{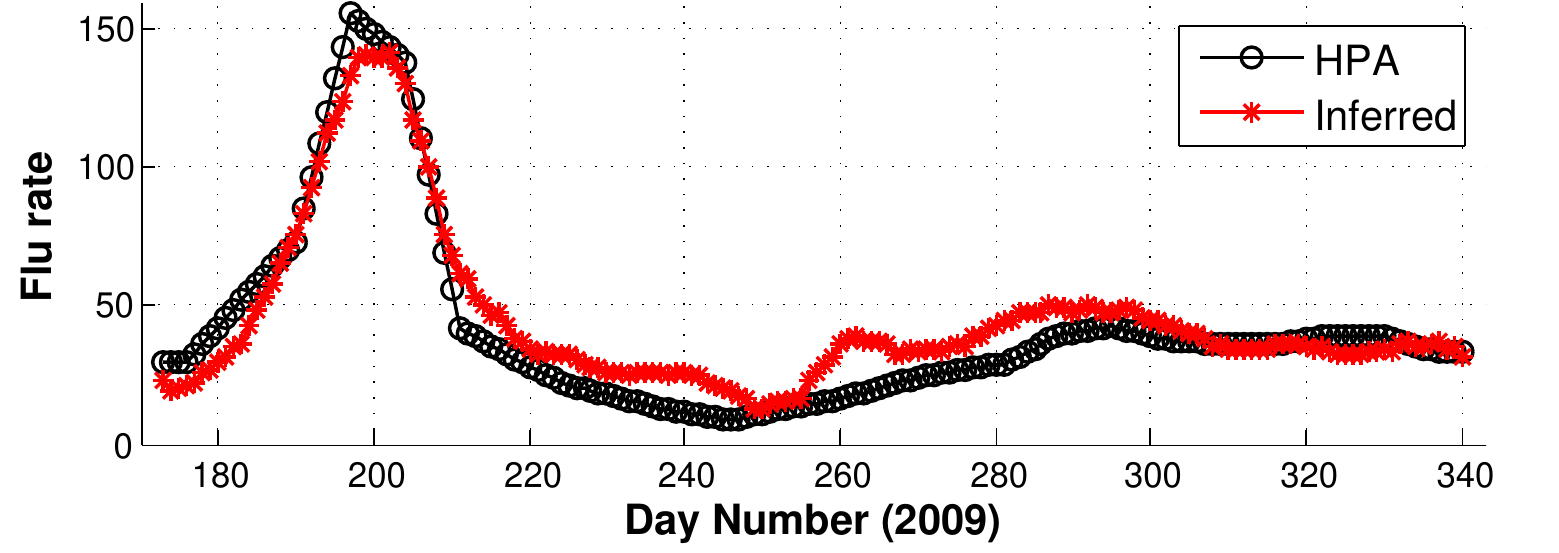}
    \label{fig_comp2}}
    \subfigure[Region E - Correlation: 0.9755 (p-value: 3.85e-111)]{\includegraphics[width=4in]{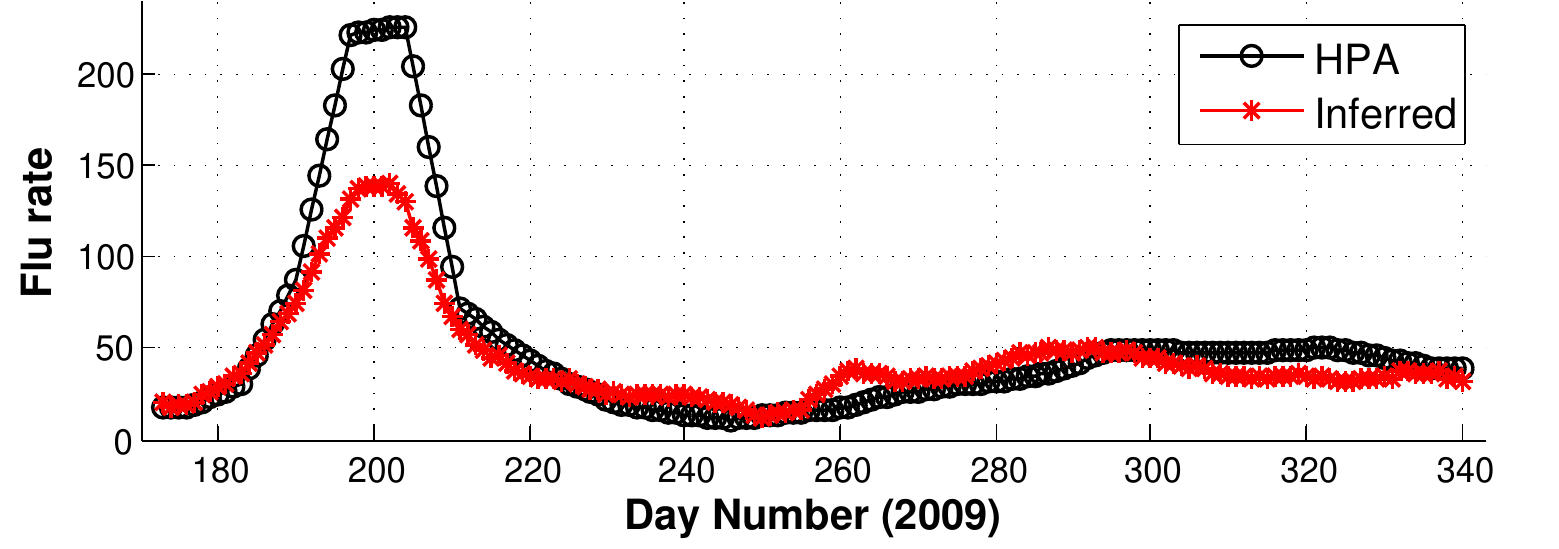}
    \label{fig_comp3}}
    \hfil
    \subfigure[Inference on the aggregated data set (per region) for weeks 28 and 41 - Correlation: 0.9713 (p-value: 3.96e-44)] {\includegraphics[width=4in]{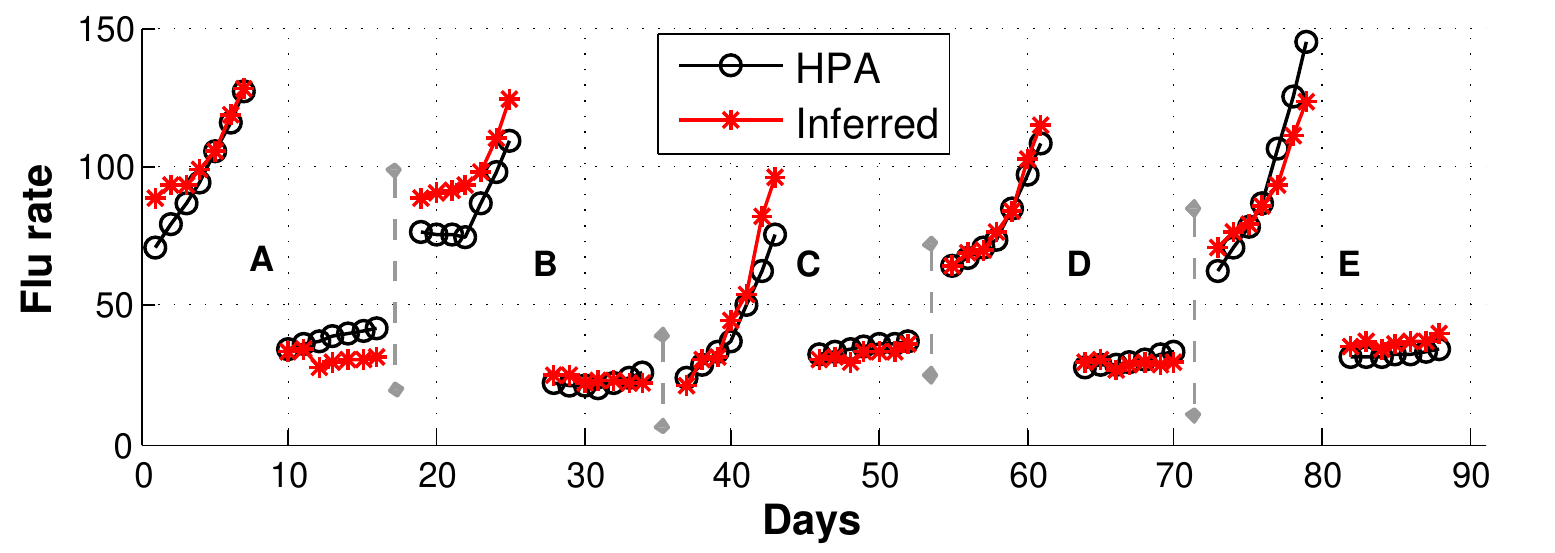}
    \label{fig_comp4}}
    \end{center}
    \caption{Comparison of the inferred versus the HPA's flu rates, after using LASSO method for learning. Figures \ref{fig_comp1}, \ref{fig_comp2}, and \ref{fig_comp3} present the results for the regional experiment (training on one region, validating on another, and testing on the remaining three), whereas \ref{fig_comp4} shows the results for the aggregated sets (5 partitions of a 14-day period each from left to right correspond to the results for regions A-E respectively). Note that for subfigure \ref{fig_comp4} the `Days' axis does not denote yearly day numbers.}
    \label{fig_compare_LASSO}
\end{figure*}

We use time series of a region $r_{i} \in \{\text{A-E}\}$ as the training set, the time series of a region $r_{j} \in \left\{\{\text{A-E}\} - r_{i}\right\}$ as the validation set for deciding the optimal shrinkage percentage $\alpha$, and we test on the data of the remaining three regions. We repeat this procedure for all possible five training set choices. LARS algorithm is applied to compute LASSO's estimates \cite{efron2004least}. The results of our method are captured in Table \ref{table_lasso_results}. Most of the possible training/validating choices lead to high linear correlations. The average linear correlation over all possible settings is 0.9256 indicating the robustness of our method. The experiments showed that the optimal scenario was to train on region A and use region B for validating $\alpha$, leading to an average correlation of 0.9594 on the remaining three regions (C-E) (for a shrinkage percentage $\alpha$ equal to 87\%). Figures \ref{fig_comp1}, \ref{fig_comp2}, and \ref{fig_comp3} show a comparison between the inferred and HPA's flu rates time series on regions C-E respectively (for the optimal choice). The weight vector $w$ had 97 non-zero values, \emph{i.e.} we were able to extract 97 markers (or features), which, in turn, are presented in Table \ref{table_features1}. The majority of the markers is pointing directly or indirectly to illness related vocabulary.

\begin{table}[!t]
\renewcommand{\arraystretch}{1.3}
\caption{97 stemmed markers extracted by applying LASSO regionally. The markers are sorted in a descending order based on their weights (read horizontally, starting from the top-left corner).}
\label{table_features1}
\centering
\fontsize{6}{6}
%\scriptsize
\begin{tabular}{cccccc}
\hline
lung     & unwel   & temperatur & like     & headach & season   \\%\hline
unusu    & chronic & child      & dai      & appetit & stai     \\%\hline
symptom  & spread  & diarrhoea  & start    & muscl   & weaken   \\%\hline
immun    & feel    & liver      & plenti   & antivir & follow   \\%\hline
sore     & peopl   & nation     & small    & pandem  & pregnant \\%\hline
thermomet& bed     & loss       & heart    & mention & condit   \\%\hline
high     & group   & tired      & import   & risk    & carefulli\\%\hline
work     & short   & stage      & page     & diseas  & recognis \\%\hline
servic   & wors    & case       & similar  & term    & home     \\%\hline
increas  & exist   & ill        & sens     & counter & better   \\%\hline
cough    & vomit   & earli      & neurolog & catch   & onlin    \\%\hline
fever    & concern & check      & drink    & long    & far      \\%\hline
consid   & ach     & breath     & flu      & member  & kidnei   \\%\hline
mild     & number  & sick       & throat   & famili  & water    \\%\hline
read     & includ  & swine      & confirm  & need    & nose     \\%\hline
medic    & phone   & cancer     & disord   & unsur   & suddenli \\%\hline
runni    &         &            &          &         &          \\\hline
\end{tabular}
\end{table}

\begin{table}[!t]
\renewcommand{\arraystretch}{1.3}
\caption{Linear correlations on the test sets after performing the LASSO - An element $(i,j)$ denotes the average correlation coefficient on the three remaining regions, after performing LASSO on region $i$ in order to learn the weights, and validating the shrinkage parameter $t$ on region $j$.}
\label{table_lasso_results}
\centering
\begin{tabular}{cccccc}
\hline
Train/Validate & \textbf{A} & \textbf{B}      & \textbf{C} & \textbf{D}       & \textbf{E}     \\\hline
\textbf{A}     & -          & \textbf{0.9594} & 0.9375     & 0.9348           & 0.9297         \\%\hline
\textbf{B}     & 0.9455     & -               & 0.9476     & 0.9267           & 0.9003         \\%\hline
\textbf{C}     & 0.9154     & 0.9513          & -          & 0.8188           & 0.908          \\%\hline
\textbf{D}     & 0.9463     & 0.9459          & 0.9424     & -                & 0.9337         \\%\hline
\textbf{E}     & 0.8798     & 0.9506          & 0.9455     & 0.8935           & -              \\\hline
&              &            &\multicolumn{2}{c}{Total Avg.}                   & \textbf{0.9256}\\\cline{4-6}
\end{tabular}
\end{table}

\begin{table}[!t]
\renewcommand{\arraystretch}{1.3}
\caption{73 stemmed markers extracted by applying LASSO on the aggregated data set of regions A-E. The markers are sorted in a descending order based on their weights (read horizontally, starting from the top-left corner). 72 of them have also occurred in Table \ref{table_features1} but here the order (\emph{i.e.} the corresponding weights) is different.}
\label{table_features2}
\centering
\fontsize{6}{6}
%\scriptsize
\begin{tabular}{cccccc}
\hline
muscl     & like     & appetit   & read      & unwel   & child      \\%\hline
work      & follow   & season    & page      & throat  & nose       \\%\hline
check     & suddenli & pleas     & immun     & phone   & swine      \\%\hline
sick      & dai      & symptom   & consid    & sens    & breath     \\%\hline
cough     & loss     & recognis  & peopl     & number  & mild       \\%\hline
home      & condit   & mention   & servic    & runni   & member     \\%\hline
wors      & diseas   & diarrhoea & high      & short   & onlin      \\%\hline
pregnant  & small    & exist     & headach   & unsur   & cancer     \\%\hline
stai      & concern  & fever     & earli     & tired   & carefulli  \\%\hline
import    & weaken   & nation    & famili    & similar & temperatur \\%\hline
feel      & ach      & flu       & case      & sore    & unusu      \\%\hline
spread    & vomit    & ill       & thermomet & pandem  & increas    \\%\hline
stage     & far      &           &           &         &            \\\hline
\end{tabular}
\end{table}

We also assess the performance of our method differently, following the same principle as in the previous section. We aggregate our regional data sets $X^{(s)}_{r}$ and $h^{(s)}_{r}$, and as before, we form a test set by using the data for weeks 28 and 41, a validation set (for deciding the optimal value of the shrinkage percentage $\alpha$) by using weeks 36 and 49, and a training set with the remaining data sets. The outcome of the experiment indicates that the optimal value for $\alpha$ is 0.0049; for this value we retrieve a linear correlation of 0.9713 (p-value is equal to 3.96e-44) on the test set. The corresponding vector of weights $w$ has 73 non-zero features which are shown in Table \ref{table_features2}. Only one of them (the stemmed word `pleas') was not included in the previously extracted set of features. Figure \ref{fig_comp4} presents a comparison of the inferred versus the HPA's flu rates for all the test points. Again, we have demonstrated how a list of markers can be automatically inferred from a large set of candidates by using a supervised learning algorithm and HPA's index as the target signal; this approach delivers a correlation greater than 0.97 with the target signal on unseen data.

%%%%%%%%%%%%%%%%%%%%%%%%%%%%%%%%%%%%%%%%%%%%%%%%%%%%%%%%%%%%%%%%%%%%%%%%%%
\subsection{What is missing from this approach?}
We have showcased how a sparse learning method can be applied in order to automatically select flu-related markers and then use them to track the levels of flu in several UK regions. In this experiment, we have chosen to smooth the input data (the time series of each marker's frequency) with a 7-point moving average to express a weekly tendency -- however, when applying this method on new unseen data, those data cannot or should not always be smoothed. Furthermore, the use of single words only (1-grams) might favour overfitting; single words unrelated with the target topic have a larger probability to correlate with it by chance than larger chunks of text \cite{Furnkranz1998,Tan2002}. Moreover, LASSO has known disadvantages (see Section \ref{section_LASSO}) and especially is inconsistent in selecting features. Finally, is this approach general? Will this methodology work when applied on other topics as well? We try to address those issues as well as others (see Section \ref{section_recap_previous_limitations}) in the next chapter.

%%%%%%%%%%%%%%%%%%%%%%%%%%%%%%%%%%%%%%%%%%%%%%%%%%%%%%%%%%%%%%%%%%%%%%%%%%%%%%%%%% harry potter effect
\section{The Harry Potter effect}
\label{section_harry_potter_effect}
A mainstream IR technique (for example the VSM, see Section \ref{section_vector_space_model}) proposes the formation of an index or vocabulary using all the n-grams in the target corpus \cite{Kowalski1997, Baeza-Yates1999,Manning2008}. In our approach, however, we form an index from several encyclopedic or other topic related web references, a choice that is partly justified in this as well as in the next section.

Following this basic principle of the VSM, for weeks 26--49 in 2009 (the same 168-day time period used in the previously presented experimental process), we index all 1-gram, after applying Porter stemming and removing stop words that appear more than 50 times in the Twitter corpus for England \& Wales. The retrieved index contains 47,193 terms. The time series of each 1-grams is smoothed by applying a 7-point moving average; the weekly ground truth is expanded and smoothed.

\begin{table}[t]
\renewcommand{\arraystretch}{1.3}
\caption{Top-20 linearly correlated terms with the flu rate in England \& Wales}
\label{table_harry_potter_corr_terms}
\centering
\begin{tabular}{ccc}
\toprule
\textbf{1-gram} & \textbf{Event} & \textbf{Corr. Coef.}\\\hline
latitud     & Latitude Festival         & 0.9367\\
flu         & Flu epidemic              & 0.9344\\
swine       & $\blacktriangle$       & 0.9212\\
harri       & Harry Potter Movie        & 0.9112\\
slytherin   & $\blacktriangle$       & 0.9094\\
potter      & $\blacktriangle$       & 0.8972\\
benicassim  & Benic\`{a}ssim Festival   & 0.8966\\
graduat     & Graduation (?)            & 0.8965\\
dumbledor   & Harry Potter Movie        & 0.8870\\
hogwart     & $\blacktriangle$       & 0.8852\\
quarantin   & Flu epidemic              & 0.8822\\
gryffindor  & Harry Potter Movie        & 0.8813\\
ravenclaw   & $\blacktriangle$       & 0.8738\\
princ       & $\blacktriangle$       & 0.8635\\
swineflu    & Flu epidemic              & 0.8633\\
ginni       & Harry Potter Movie        & 0.8620\\
weaslei     & $\blacktriangle$       & 0.8581\\
hermion     & $\blacktriangle$       & 0.8540\\
draco       & $\blacktriangle$       & 0.8533\\
snape       & $\blacktriangle$       & 0.8486\\
\bottomrule
\end{tabular}
\end{table}

At first, in order to get a quick characterisation of the data set, we compute the linear correlation of each 1-gram with the ground truth. Table \ref{table_harry_potter_corr_terms} shows the top-20 correlated markers together with their correlation coefficients (all correlations are statistically significant). There exist some flu related words such as `flu' or `swine' but the majority of the words is unrelated to the topic of flu and seems to derive from the well known Harry Potter fantasy novel/film series. The top correlated word is `latitud', something that at first sight might not make much sense.

\begin{figure}[t]
\centering
\includegraphics[width=5in]{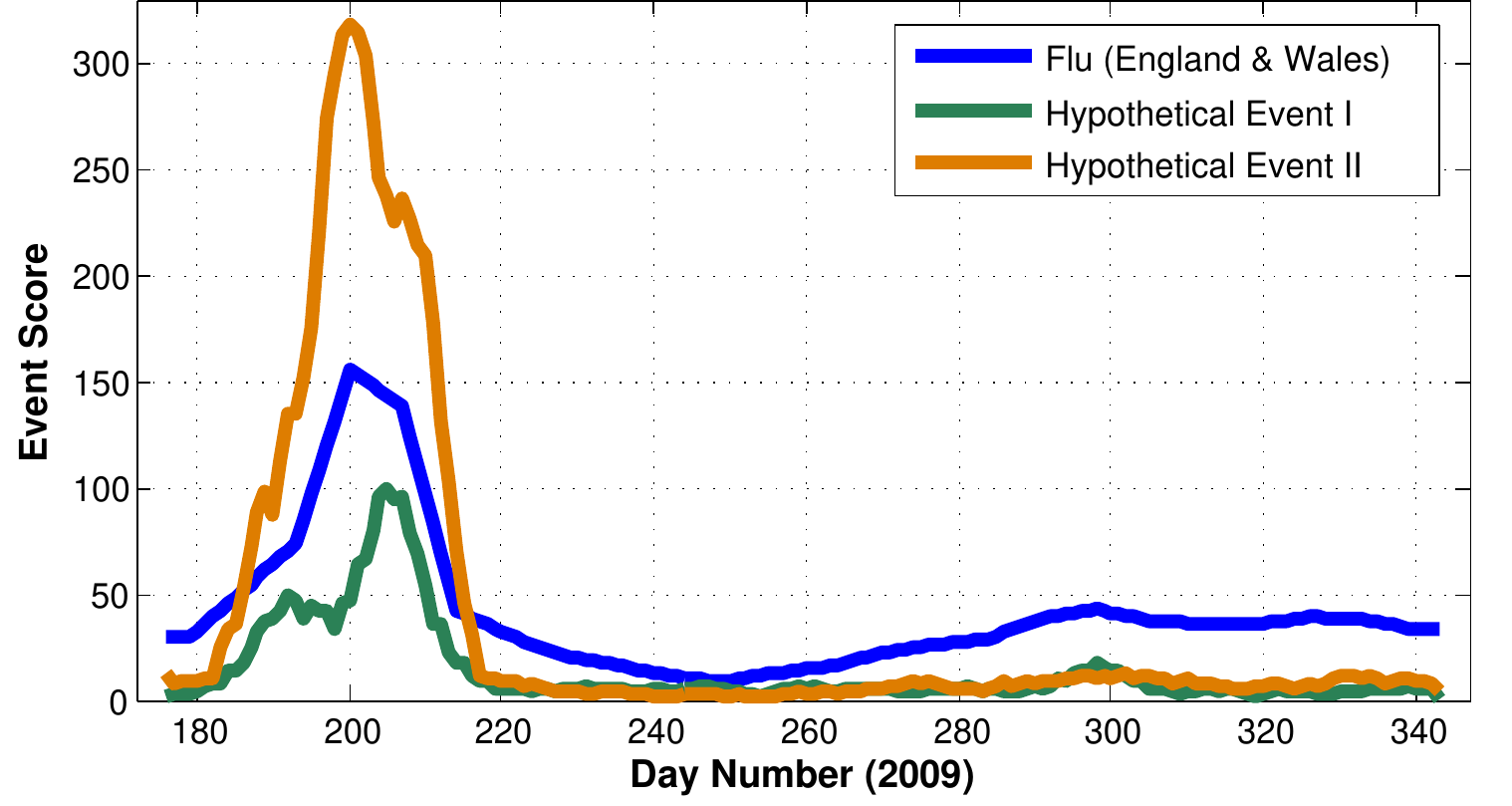}
\caption{The Harry Potter effect - There might exist other signals emerging from real-life events correlated with the target topic.}
\label{fig_harry_potter_effect}
\end{figure}

\begin{table}[!t]
\renewcommand{\arraystretch}{1.3}
\caption{Top-20 selected terms by LASSO (sorted in descending order based on their weight) in every iteration of the 5-fold cross validation.}
\label{table_harry_potter_lasso_terms}
\centering
\begin{tabular}{ccccc}
\toprule
it-\textbf{1} & it-\textbf{2} & it-\textbf{3} & it-\textbf{4} & it-\textbf{5}\\\hline
swine   & hetero    & flu       & graduat   & flu\\
thunder & khan      & khan      & flu       & graduat\\
graduat & flu       & swine     & hetero    & hetero\\
fern    & harri     & bastil    & harri     & media\\
harri   & lotteri   & fern      & swine     & harri\\
sale    & fern      & harri     & sundai    & swine\\
flu     & swine     & ladi      & manchest  & thunder\\
market  & meteor    & manchest  & school    & sundai\\
hetero  & chrome    & mobil     & weekend   & manchest\\
mondai  & memori    & sale      & meteor    & meteor\\
sundai  & player    & app       & god       & school\\
labour  & feel      & servic    & live      & record\\
gener   & sundai    & quiz      & app       & gener\\
bristol & saturdai  & market    & lotteri   & wood\\
prize   & help      & mondai    & moir      & death\\
public  & tori      & week      & player    & footbal\\
khan    & edward    & level     & memori    & canada\\
lotteri & ash       & danyl     & andi      & spammer\\
award   & xxx       & bank      & bed       & memori\\
clock   & wave      & graduat   & feel      & award\\
\bottomrule
\end{tabular}
\end{table}

Knowing that in the region of England \& Wales the flu epidemic peaked on the 16th of July and basing our search on the top correlated terms (Table \ref{table_harry_potter_corr_terms}), we tracked events that emerged at the same date and might have been a topic of conversation on Twitter. Indeed, a Harry Potter movie aired on the 15th of July in the UK.\footnote{ According to the Internet Movie Database -- \url{http://www.imdb.com/title/tt0417741/releaseinfo}.} Moreover, two 3-day music festivals kicked off on July 16th: the annual festival `Latitude'  featuring several popular artists\footnote{ According to the relevant article on Wikipedia, Latitude Festival -- \url{http://goo.gl/2oQmF}.} and an annual international festival in Spain (`Bennic\`{a}ssim') -- possibly popular among UK citizens -- which was interrupted by a large fire.\footnote{ According to the relevant article on Wikipedia, Festival Internacional de Benic\`{a}ssim -- \url{http://goo.gl/jfHKk}.}

This, conventionally named as the \textbf{Harry Potter effect}\index{Harry Potter effect}, is something expected to happen when other signals emerging from real-life events show similar behaviour compared to the target topic (see Figure \ref{fig_harry_potter_effect}). The possibility of having other events in correlation with the target one is increasing when the data are from a relatively small period of time and it is even higher when the target event emerged only once in this period.

We have also applied LASSO on the same data set, knowing that it might not select the top correlated terms but a different set of variables conforming with its constraints. After permuting the data set based on a random selection of the day-index, we perform learning by applying a 5-fold cross validation scheme. The performance of the selected features and their weights on the disjoint test set is excellent, \ie of the same level as in Section \ref{section_lasso_paper_1}. However, by observing the selected features and their weights in every iteration of the process, we have proof that this is a case of overfitting. Table \ref{table_harry_potter_lasso_terms} shows the top-20 selected terms (based on their weights) for all steps of the cross validation process. The selected features are slightly different in every iteration, something expected as they are based on training sets that are 20\% different from one another. Again, a few basic flu-related features have been selected, such as `flu' and `swine' as well as the word `harri'. However, most terms seem to describe other events (which we did not track), such as a lottery or a graduation ceremony.

With a minority of features related to the target topic of illness, it is safe to conclude that any significant performance on the test set is a result of overfitting. Bearing in mind this `Harry Potter effect', we chose to form the set of candidate markers in a more focused way as it has been described in the previous section.
%%%%%%%%%%%%%%%%%%%%%%%%%%%%%%%%%%%%%%%%%%%%%%%%%%%%%%%%%%%%% end of harry potter

%%%%%%%%%%%%%%%%%%%%%%%%%%%%%%%%%%%%%%%%%%%%%%%%%%%%%%%%%%%%%%%%%%%% error bounds
\section{How error bounds for L1-regularised regression affect our approach}
\label{section_error_bounds_for_LASSO}
Suppose that we have an input matrix $X$ with $N$ samples and $p$ variables (or dimensions), such that $x_{ij}\in X$ with $i\in\{1, ..., N\}$ and $j\in\{1, ..., p\}$, and an one dimensional target variable $y$ of size $N$. The empirical LASSO\index{regression!LASSO} estimate $\hat{w}$ is given by:
\begin{equation}
\label{eq_empirical_lasso_estimate}
\hat{w} = \argmin_{w}\left\{ \sum_{i=1}^{N}\left( y_i - w_0 - \sum_{j=1}^{p}x_{ij}w_j \right)^{2} + \lambda \sum_{j=1}^{p}|w_j| \right\},
\end{equation}
where $\lambda$ is the shrinkage parameter.

It is important to notice that statistical bounds exist linking LASSO's expected performance to the one derived on the training set (empirical), the number of dimensions, number of training samples and L1-norm or sparsity of $\hat{w}$.

A general error bound\index{regression!LASSO!error bound} drawn on the Euclidean distance between the optimal and the empirical LASSO coefficients suggests that
\begin{equation}
\label{eq:err_bound_1}
\|w - \hat{w}\|_{\ell_2} = \mathcal{O}(\sqrt{\frac{k\log{p}}{N}}),
\end{equation}
where $k$ is the number of nonzero elements in $\hat{w}$ \cite{Wainwright2009}. Equation \ref{eq:err_bound_1} makes clear that the error is proportional to the dimensionality of the problem, but disproportional to the number of samples or the achieved sparsity of $\hat{w}$.

In \cite{bartlett2009a} it is shown that LASSO's expected loss $\mathcal{L}(w)$ up to poly-logarithmic factors in $W_{1}$, $p$ and $N$ is bounded\index{regression!LASSO!error bound} by
\begin{equation}
\label{eq:err_bound_2}
\mathcal{L}(w) \leq \mathcal{L}(\hat{w}) + \mathcal{Q}\text{, with }\mathcal{Q} \sim \min{\left\{\frac{W_{1}^{2}}{N} + \frac{p}{N},\frac{W_{1}^{2}}{N} + \frac{W_{1}}{\sqrt{N}}\right\}},
\end{equation}
where $\mathcal{L}(\hat{w})$ denotes the empirical loss and $W_{1}$ is an upper bound for the L1-norm of $\hat{w}$, \emph{i.e.} $\|\hat{w}\|_{\ell_1} \leq W_{1}$; this error bound is sharp (not loose) and predictive. Therefore, to minimise the prediction error using a fixed set of training samples and given that the empirical error is relatively small, one should either reduce the dimensionality of the problem ($p$) or increase the shrinkage level of $\hat{w}$'s L1-norm (which intuitively might result in sparser solutions, \ie fewer selected features).

%high dimensional generalized models and the lasso\cite{VandeGeer2008}
%\begin{equation}
%\mathcal{L}(\hat{w}) \leq C_0 \cdot \min_
%\end{equation}

A more complex error bound\index{regression!LASSO!error bound} regarding the predictions ($Xw$), derived in \cite{Candes2009} and holding with probability at least $1 - 6p^{-2\log{2}} - p^{-1}(2\pi\log{p})^{-1/2}$ is the following:
\begin{equation}
\label{eq:err_bound_3}
\|Xw - X\hat{w}\|^{2}_{\ell_{2}} \leq C_0 \cdot (2\log{p}) \cdot k \cdot \sigma^{2},
\end{equation}
where $C_0 = 8(1 + \sqrt{2})^{2}$, $k$ is the number of nonzero elements in $\hat{w}$ and the prediction errors $\epsilon_{i}$, $j\in\{1, ..., N\}$ are assumed to be Gaussian with $\epsilon_{i} \sim \mathcal{N}(0,\sigma^{2})$. For this equation to hold, we also need an $X$ with no highly co-linear predictors.

In all the aforementioned error bounds, as well as in many others in literature \cite{Bickel2008, Greenshtein2007, Koltchinskii2009}, the main conclusion is similar: as the dimensionality of the problem increases, error increases as well. When $p/N$ (dimensionality -- sample-size ratio) is relatively small, \ie when the number of variables is far less than the number of samples, the error rate is reduced. Which, in turn, means that when the number of samples is relatively small, in order to achieve a satisfactory inference performance, one has to constrain the number of dimensions; this serves as an additional reasoning for our feature extraction modelling choice. Finally, we see that sparser solutions result to lower error rates -- which was one of the reasons for using LASSO in the first place.

%%%%%%%%% revision %%%%%%%%%%%%%%%%%
%%%%%%%%%%%%%%%%%%%%%%%%%%%%%%%%%%%%

%%%%%%%%%%%%%%%%%%%%%%%%%%%%%%%%%%%%%%%%%%%%%%%%%%%%%%%%%%%%%%%%%%%% end of error bounds

\section{Summary of the chapter}
\label{section_summary_of_first_steps_chapter}
In this chapter, we have presented a preliminary method for tracking the flu epidemic in the UK by using the contents of Twitter; our approach could give an early warning in various situations, but mostly can give timely and free information to health agencies to plan health care. This method is based on the textual analysis of micro-blog contents, is calibrated with the ground truth provided by the HPA, and could also be extended to the contents of text messages sent via mobile devices (besides privacy concerns), as the latter are of the same format with tweets (140 characters limitation).

At first, we formed a set of keywords or phrases which represented illness and some of the flu symptoms. After defining a basic function which combined the normalised frequencies of those manually selected terms to come up with a flu-score, we discovered that the time series of those flu scores were linearly correlated with the actual flu rates. By adding one more layer to this approach, we applied OLS regression to learn a weight for each term; as expected, the retrieved correlations between OLS predictions and the ground truth were greater than the original ones. The next step included the automatic selection of features from the textual stream. To achieve this we applied a sparse regressor, the LASSO, which was able to select a small, weighted subset of the initial feature space; the selected features were -- in their great majority -- drawn from the illness topic. This automatic method, improved the inference performance further. However, the original set of candidate features was not formed by using the entirety of the Twitter corpus, something that is usually the mainstream approach in IR. We have chosen to form the candidate features from text which was related with the target topic such as Wikipedia pages and discussions from health-oriented websites; nevertheless, the majority of the candidate features were irrelevant to the illness topic. This choice is justified firstly by presenting and explaining the `Harry Potter effect' in event detection and secondly by understanding how error bounds for LASSO are affected by the specific characteristics of a learning task.

The approach presented in this chapter can automatically learn the inputs (textual markers and their weights) of a scoring function that correlates very highly with the HPA flu score. The proposed method only requires access to the time series of geolocated blog contents and a ground truth. However, knowing that LASSO is an inconsistent learning function in terms of model selection, but also that the presented method was topic-specific, in the next chapter we focus on improving it in several aspects as well as making it -- to an extent -- generic. 

%% file: Chapters/Chapter5.tex
\chapter{Nowcasting Events from the Social Web with Statistical Learning}
\label{Chapter_Nowcasting_Events_From_The_Social_Web}
%\lhead{Chapter 5. Nowcasting Events from the Social Web with Statistical Learning}

\rule{\linewidth}{0.5mm}
We present a generic methodology -- which builds on the methodology presented in Chapter \ref{Chapter_first_steps} resolving some of its limitations -- for inferring the occurrence and magnitude of an event or phenomenon by exploring the rich amount of unstructured textual information on the social part of the web. Having geo-tagged user posts on the microblogging service of Twitter as our input data, we investigate two case studies. The first consists of a benchmark problem, where actual levels of rainfall in a given location and time are inferred from the content of tweets. The second one is a real-life task, where we infer regional ILI rates in the effort of detecting timely an emerging epidemic disease. Our analysis builds on a statistical learning framework, which performs sparse learning via the bootstrapped version of LASSO to select a consistent subset of textual features from a large amount of candidates. In both case studies, selected features indicate close semantic correlation with the target topics and inference, conducted by regression, has a significant performance; the performance of our method is also compared with another approach proposed to address a similar task, showing a significant improvement. We also apply a different, nonlinear and non-parametric learner -- an ensemble of CARTs\index{CART} -- in the core of our approach, and investigate possible differences in the feature selection and inference process. This chapter is an extended version of our paper ``Nowcasting Events from the Social Web with Statistical Learning'' \cite{Lampos2011a}.
\newline \rule{\linewidth}{0.5mm}
\newpage

%%%%%%%%%%%%%%%%%%%%%%%%%%%%%%%%%%%%%%%%%%%%%%
\section{Recap on previous method's limitations}
\label{section_recap_previous_limitations}
We have already described in Chapter \ref{Chapter_first_steps} a preliminary method which enabled us to detect flu-like illness in several UK regions by observing Twitter content. However, the core theoretical tool used in this approach, a sparse L1-regulariser known as the LASSO, has been proven to be a model-inconsistent learner as in many settings it selects more variables than necessary \cite{Lv2009, Zhao2006}. Taking a closer look at the proposed method, one could also argue that the use of 1-grams (unigrams) only could also lead to a bad choice of features and consequently to overfitting. Based on the simple notion that the entropy -- or the uncertainty -- of the English language decreases as the number of characters increases \cite{Shannon1951}, and considering the experimental proof for the learning improvements which may be achieved by the use of 2-grams (bigrams) \cite{Tan2002, Furnkranz1998}, in this chapter we present a hybrid scheme that combines unigrams and bigrams. Furthermore, the previous experimental process has been focused on a short time span (168 days), and there was no comparison with other methods proposed for similar tasks. One more critical point was the use of smoothing on the input data, something that improved the performance of our approach. However, in real-time applications, where inferences should be made about the running state of a hidden variable, this type of smoothing might cause the loss of important information. Lastly, our previous experiments were focused on a single specific event (Swine Flu epidemic in the UK, 2009) and therefore, no experimental proof existed allowing us to generalise our findings.

% abstract of the original paper
%Taking in account all the above, in this chapter we present a general methodology for inferring the occurrence and magnitude of an event or phenomenon by exploring the rich amount of unstructured textual information on the social part of the web. Having geo-tagged user posts on the microblogging service of Twitter as our input data, we investigate two case studies. The first consists of a benchmark problem, where actual levels of rainfall in a given location and time are inferred from the content of tweets. The second one is a real-life task, where we infer regional Influenza-like Illness rates in the effort of detecting timely an emerging epidemic disease. Our analysis builds on a statistical learning framework, which performs sparse learning via the bootstrapped version of LASSO to select a consistent subset of textual features from a large amount of candidates. In both case studies, selected features indicate close semantic correlation with the target topics and inference, conducted by regression, has a significant performance, especially given the short length -- approximately one year -- of Twitter's data time series. Our method is compared with another technique proposed to address a similar task, showing a significant improvement. We also apply a different, nonlinear learner -- an ensemble of CARTs -- in the core of our approach, and investigate possible differences in the feature selection and inference process.

\section{Nowcasting events}
% nowcasting
The term `\emph{nowcasting}'\index{nowcasting}, commonly used in economics \cite{Giannone2008, Banbura2011} as well as in meteorology \cite{Wilson1998, Menzel1998}, expresses the fact that we are making inferences regarding the current magnitude $\mathcal{M}(\varepsilon)$ of an event $\varepsilon$ -- in this special case `current' may refer to the very near future, but also the very recent past.

For a time interval $u = [t - \Delta t, t + \Delta t]$, where $t$ is a reference time and $(t + \Delta t)$ denotes the current time instance, consider $\mathcal{M}\left(\varepsilon^{(u)}\right)$ -- the magnitude of an event $\varepsilon$ within the time interval $u$ -- as a latent variable. For the same time instance $u$, we denote the entire web content as $\mathcal{W}^{(u)}$ and an observable proportion of it as $\mathcal{S}^{(u)}$, where $\mathcal{S}^{(u)} \subseteq \mathcal{W}^{(u)}$.

Making the hypothesis that an event emerging in real life will affect the content of the web, our aim is to use the observable web content in order to detect and also infer the magnitude of this event. Using the introduced notation, we want to infer $\mathcal{M}\left(\varepsilon^{(u)}\right)$ directly from $\mathcal{S}^{(u)}$. Given that $u$ is of short length, we are inferring the present value of the latent variable, \emph{i.e.} we are \textbf{nowcasting} $\mathcal{M}\left(\varepsilon^{(u)}\right)$.

\begin{figure*}[tp]
    \begin{center}
    \includegraphics[width=4in]{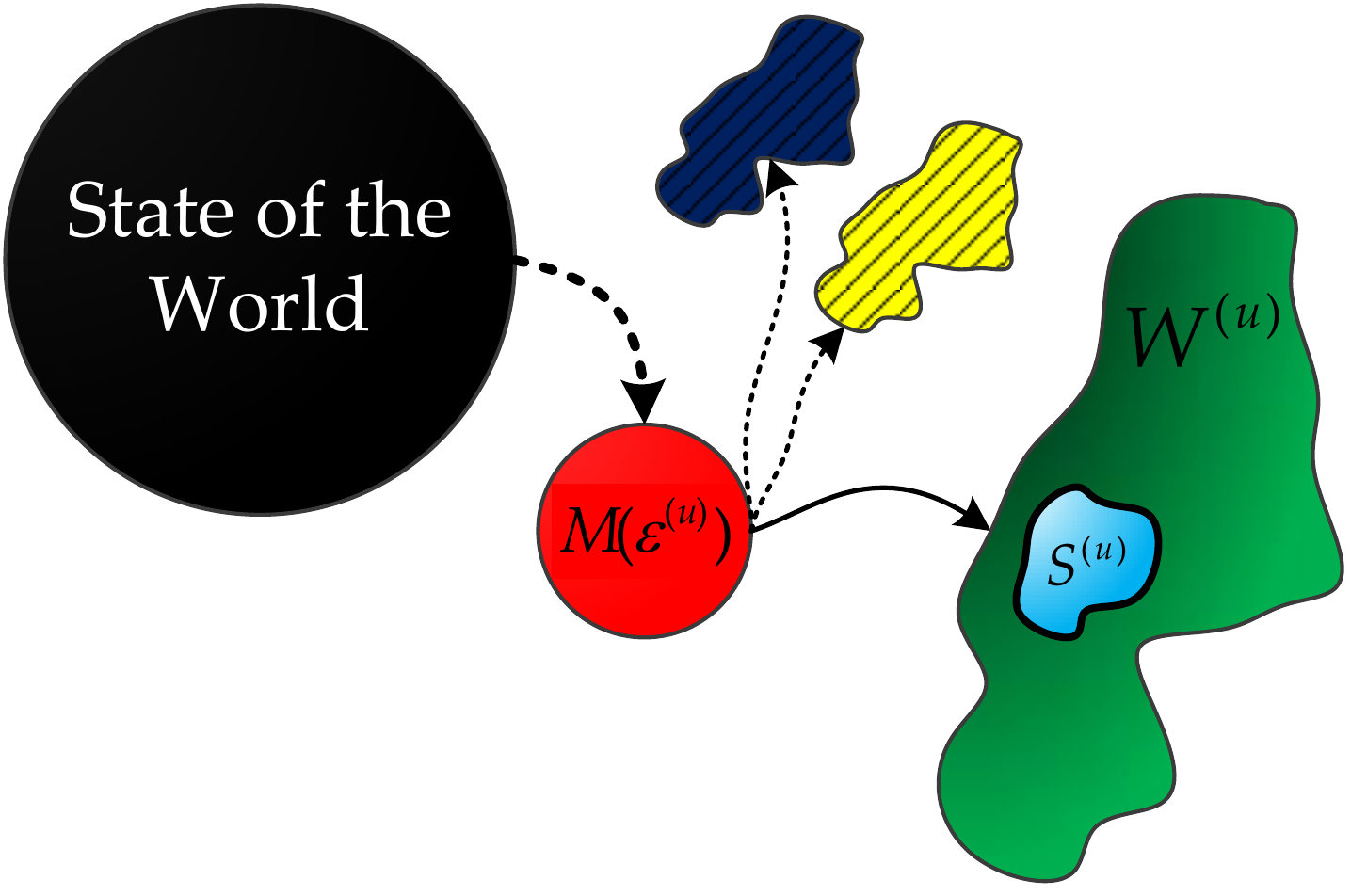}
    \end{center}
    \caption{An abstract graphical representation that describes the generic hypothesis behind our model. The State of the World during a time interval $u$ affects the occurrence and magnitude of an event $\varepsilon$, $\mathcal{M}\left(\varepsilon^{(u)}\right)$. In turn, this event might affect other entities including the content of the Web, $\mathcal{W}^{(u)}$. We are observing a subset of the Web content, $\mathcal{S}^{(u)}$.}
    \label{fig_nowcasting}
\end{figure*}

From a probabilistic point of view (see an abstract graphical representation in Figure \ref{fig_nowcasting}), we first have to learn the conditional probability distribution $\Pr\left(\mathcal{S}^{(u)}|\mathcal{M}(\varepsilon^{(u)})\right)$ from the observable information, and then -- following the Bayesian logic -- we use $\Pr\left(\mathcal{S}^{(u)}|\mathcal{M}(\varepsilon^{(u)})\right)$ to infer $\Pr\left(\mathcal{M}(\varepsilon^{(u)})|\mathcal{S}^{(u)}\right)$. An important assumption here is that during the time interval $u$, there was enough time for the web to depict this real-life situation in its content. When $u$ is of a relatively small value, this might not be true for the majority of the web sites, but has been shown to hold for social networks, such as Twitter \cite{Sakaki2010}. Obviously, an even more fundamental assumption made is that an identifiable -- possibly causal -- link between the shape of the web content and real-life events does exist; indeed, we have already presented results that support this assumption in the previous chapter.

\section{Related work}
\label{section:related_work_journal}
Recent work has been concentrated on exploiting user generated web content for conducting several types of inference. A significant subset of papers, examples of which are given in this paragraph, focuses on methodologies that are based either on manually selected textual features related to a latent event, \emph{e.g.} flu related keywords, or the application of sentiment or mood analysis, which in turn implies the use of predefined vocabularies, where words or phrases have been mapped to sentiment or mood scores \cite{Pang2008}. Corley \emph{et al.} reported a 76.7\% correlation between official ILI rates and the frequency of certain hand-picked influenza related words in blog posts \cite{corley2009monitoring}, whereas similar correlations were shown between user search queries which included illness related words and CDC\footnote{ Centers for Disease Control and Prevention (CDC)\index{Centers for Disease Control and Prevention|see{CDC}}\index{CDC}, \url{http://www.cdc.gov/}.} rates \cite{polgreen2008using}. Furthermore, sentiment analysis\index{sentiment analysis} has been applied in the effort of extracting voting intentions \cite{tumasjan2010predicting} or box-office revenues \cite{Asur2010a} from Twitter content. Similarly, mood analysis combined with a non linear regression model derived a 0.876 correlation with daily changes in Dow Jones Industrial Average closing values \cite{Bollen2011}. Finally, Sakaki \emph{et al.} presented a method which exploited the content, time stamp and location of a tweet to detect the existence of an earthquake \cite{Sakaki2010}.

However, in other approaches feature selection is performed automatically by applying statistical learning methods. Apart from the obvious advantage of reducing human involvement to a minimum, those methods tend to have an improved inference performance as they are enabled to explore the entire feature space or, in general, a greater amount of candidate features \cite{Guyon2003}. In \cite{ginsberg2008detecting}, Google researchers proposed a model able to automatically select flu related user search queries, which later on were used in the process of tracking ILI rates. Their method, a core component of Google Flu Trends, achieved an average correlation of 0.9 with CDC data, much higher than any other previously reported method. An extension of this approach has been applied on Twitter data achieving a 78\% correlation with CDC rates \cite{Culotta2010}. In both those works, features were selected based on their individual correlation with ILI rates; the subset of candidate features (user search queries or keywords) appearing to independently have the highest linear correlations with the target values formed the result of feature selection. Another technique -- part of our preliminary results presented in the previous chapter --, which applied sparse regression on Twitter content for automatic feature selection, resulted to a greater than 90\% correlation with HPA's flu rates for several UK regions \cite{Lampos2010f} -- an improved version of this methodology has been incorporated in Flu Detector \cite{Lampos2010}, an online tool for inferring flu rates based on tweets (see Chapter \ref{chapter:theory_in_practice}).

Besides the important differences regarding the information extraction and retrieval techniques or the data sets considered, the fundamental distinction between \cite{ginsberg2008detecting,Culotta2010} and \cite{Lampos2010f} lies on the feature selection principle -- a sparse regressor, such as LASSO, does not handle each candidate feature independently but searches for a subset of features which satisfies its constraints \cite{tibshirani1996regression} (see also Section \ref{section_LASSO}).

\section{Methodology}
\label{section_bolasso_methodology}
In this section, we describe the proposed methodology in detail. An abstract summary of it would include the following main operations:
\begin{enumerate}
  \item A \textbf{pipeline} that is used to collect and store the essential observations or input data in general; in our experiments this is Twitter content (and we have already described this process in Chapter \ref{chapter:data_characterisation_collection}).
  \item \textbf{Candidate Feature Extraction}\index{feature extraction}. A vocabulary of candidate features is formed by using n-grams, \emph{i.e.} phrases with \emph{n} tokens. We also refer to those n-grams as markers. Markers are extracted from text which is expected to contain topic-related words, \emph{e.g.} web encyclopedias as well as other more informal references. By construction the set of extracted candidates contains many features relevant with the target topic and much more with no direct semantic connection. We are considering two schemes of features: 1-grams and 2-grams. Then, we investigate ways to combine them.
  \item \textbf{Feature Selection}\index{feature selection}. A subset of the candidate features is selected by applying a sparse learning method.%In our experiments, we have applied Bolasso to select a consistent set of features. The weights of the selected features are then learnt via OLS regression on the reduced input space. % maybe try something better?
  \item \textbf{Inference and Validation}. Finally, we test our proposed method on unseen data and retrieve performance figures.
\end{enumerate}

\subsection{Forming vector space representations of text streams}
\label{section_forming_vsr_of_text_streams}
We define a text stream as a superset of documents. The word stream denotes that there is a continuous flow of documents given that the Web does not have a static behaviour. Documents can be tweets, status updates on Facebook, blog posts and so on as well as combinations of the aforementioned examples. As we have already mentioned, they are comprised by sets of words known as n-grams, where $n$ indicates the number of words in the set.

In our effort to extract useful signals from textual streams, some n-grams are more informative than others depending always on the task at hand. Our first step towards achieving our goal, which is to detect and quantify such signals, is to spot the most informative n-grams or at least a subset of them which guarantees a significant performance during inference.

All n-grams that take part in this `selection' process are denoted as candidate n-grams or features or markers. Suppose we have extracted a set of candidate features $\mathcal{C} = \{c_{i}\}$, $i \in \{1,...,|\mathcal{C}|\}$.

The user generated textual stream for a time interval $u$ is denoted by the set of posts $\mathcal{P}^{(u)} = \{p_{j}\}$, $j \in \{1,...,|\mathcal{P}^{(u)}|\}$.

A function $g$ indicates the frequency $\varphi$ (raw count) of a candidate marker $c_{i}$ in a user post $p_j$:
\begin{equation}
g(c_i, p_{j}) = \left\{
\begin{array}{l l}
\varphi & \quad \mbox{if $c_i \in p_{j}$,}\\
0 & \quad \mbox{otherwise.}\\
\end{array} \right.
\end{equation}
Given the limitation of a tweet's length (140 characters), when we deal with Twitter content, we choose to restrict function $g$ to its Boolean special case by setting the maximum value for $\varphi$ to 1.

We compute the score $s$ of a candidate marker $c_{i}$ in the set of user posts $\mathcal{P}^{(u)}$ as follows:
\begin{equation}
\displaystyle
s(c_i, \mathcal{P}^{(u)}) = \frac{\displaystyle \sum_{j=1}^{|\mathcal{P}^{(u)}|} g(c_i, p_{j})}{|\mathcal{P}^{(u)}|}.
\end{equation}
Therefore, the score of a candidate marker is a normalised frequency representing the ratio of its occurrences over the total number of posts in the text stream. When $g$ is a Boolean function this score can be also interpreted as the ratio of tweets containing the marker over the total number of tweets.

%(idea: what if we used a different normaliser?)
The scores of all candidate markers for the same time interval $u$ are kept in a vector $x$ given by:
\begin{equation}
x^{(u)} = [ s(c_1, \mathcal{P}^{(u)})\mbox{ ... }s(c_{|\mathcal{C}|}, \mathcal{P}^{(u)}) ]^{T}.
\end{equation}
The length of the time interval $u$ is a matter of choice and can vary depending on the inference task at hand. In our work, we frequently set it equal to the duration of a day.

For a set of time intervals $\mathcal{U} = \{u_{k}\}$, $k \in \{1,...,|\mathcal{U}|\}$ and given $\mathcal{P}^{(u_{k})}$ $\forall k$, we compute the scores of the candidate markers $\mathcal{C}$. Those are held in a $|\mathcal{U}| \times |\mathcal{C}|$ array $\mathcal{X}^{(\mathcal{U})}$:
\begin{equation}
\mathcal{X}^{(\mathcal{U})} = [ x^{(u_1)}\mbox{ ... }x^{(u_{|\mathcal{U}|})} ]^{T}.
\end{equation}
In most occasions time intervals $u_{k}$ are of the same size, but this is not a compulsory restriction.

\subsection{Feature selection}
\label{section_feature_selection_journal}
For the same set of $|\mathcal{U}|$ time intervals, we retrieve -- from authoritative sources -- the corresponding values of the target variable $y^{(\mathcal{U})}$:
\begin{equation}
y^{(\mathcal{U})} = [y_{1}\mbox{ ... }y_{|\mathcal{U}|}]^{T}.
\end{equation}

$\mathcal{X}^{(\mathcal{U})}$ and $y^{(\mathcal{U})}$ are used as an input in Bolasso\index{Bolasso}, the bootstrapped version of LASSO \cite{Bach2008}. In each bootstrap, LASSO selects a subset of the candidates and at the end Bolasso, by intersecting the bootstrap outcomes, attempts to make this selection consistent. LASSO is formulated as follows:
\begin{equation}
\begin{array}{cc}
\displaystyle\min_w & \|\mathcal{X}^{(\mathcal{U})}w - y^{(\mathcal{U})}\|_{2}^{2}\\
\mbox{s.t.}         & \|w\|_{1} \leq t,
%\mbox{s.t.}         & \|w\|_{1} \leq \alpha \cdot \|w_{ls}\|_{1},\mbox{ }\alpha\in(0,1]
\end{array}
\end{equation}
where $t$ is the regularisation parameter controlling the shrinkage of $w$'s L1-norm. In turn, $t$ can be expressed as
\begin{equation}
\label{eq:lasso_alpha}
t = \alpha \cdot \|w_{\mbox{\tiny OLS}}\|_{1},\mbox{ }\alpha\in(0,1],
\end{equation}
where $w_{\mbox{\tiny OLS}}$ is the OLS regression solution and $\alpha$ denotes the desired shrinkage percentage of $w_{\mbox{\tiny OLS}}$'s L1-norm. Bolasso's implementation applies LARS which is able to explore the entire regularisation path at the cost of one matrix inversion and decides the optimal value of the regularisation parameter ($t$ or $\alpha$) by either using the largest consistent region, \emph{i.e.} the largest continuous range on the regularisation path, where the set of selected variables remains the same, or by cross-validating \cite{efron2004least}. We explain Bolasso in detail in Section \ref{section:soft_bolasso_with_Ct_validation}, when looking at the entire algorithm (Algorithm \ref{algorithm_soft_bolasso_with_CT_validation}) that encapsulates our method.

\subsection{Inference and consensus threshold}
\label{section:inference_and_consensus_threshold}
After selecting a subset $\mathcal{F} = \{f_{i}\}$, $i \in \{1,...,|\mathcal{F}|\}$ of the feature space $\mathcal{C}$ ($\mathcal{F} \subseteq \mathcal{C}$), the VSR of the initial vocabulary $\mathcal{X}^{(\mathcal{U})}$ is reduced to an array $\mathcal{Z}^{(\mathcal{U})}$ of size $|\mathcal{U}| \times |\mathcal{F}|$. We learn the weights of the selected features by performing OLS regression:
\begin{equation}
\begin{array}{cc}
\displaystyle\min_{w_s} & \|(\mathcal{Z}^{(\mathcal{U})}w_s + \beta) - y^{(\mathcal{U})}\|_{2}^{2},\\
\end{array}
\end{equation}
where vector $w_s$ denotes the learnt weights for the selected features and scalar $\beta$ is regression's bias term.

A strict application of Bolasso implies that only features with a non zero weight in all bootstraps are going to be considered. In our methodology a soft version of Bolasso is applied (named as \emph{Bolasso-S} in \cite{Bach2008}), where features are considered if they acquire a non zero weight in a fraction of the bootstraps, which is referred to as Consensus Threshold\index{Consensus Threshold} (\textbf{CT}). CT ranges in $(0,1]$ and obviously is equal to 1 in the strict application of Bolasso. The value of CT, expressed by a percentage, is decided using a validation set. To constrain the computational complexity of the learning phase, we consider 21 discrete CTs from 50\% to 100\% with a step of 2.5\%.

Overall, the experimental process includes three steps: (\textbf{a}) training, where for each CT we retrieve a set of selected features from Bolasso and their weights from OLS regression, (\textbf{b}) validating CT, where we select the optimal CT value based on a validation set, and (\textbf{c}) testing, where the performance of our previous choices is computed. Training, validation and testing sets are by definition disjoint from each other.

%%%%%%%% revision %%%%%%%%%%%%%%%%%%
The MSE between inferred ($\mathcal{X}w$) and target values ($y$) defines the loss ($\mathcal{L}$) during all steps. For a sample of size $|\mathcal{D}|$ this is defined as:
\begin{equation}
\mathcal{L}(w) = \frac{1}{|\mathcal{D}|} \sum_{i=1}^{|\mathcal{D}|}{\ell(\langle x_{i}w \rangle,y_i)},
\end{equation}
where the loss function $\ell(\langle x_{i}w \rangle,y_i) = (\langle x_{i}w \rangle - y_i)^2$. This decision comes naturally from the definition of LASSO, which essentially tries to find the $w$'s that minimise the squared loss plus a regularisation factor. Although the squared loss tends to penalise outliers more than other loss functions (\eg the absolute loss), it is the one that has been investigated thoroughly in the relevant literature and research.\footnote{ For example the LARS algorithm that has been applied in our work uses the squared loss.} The Root Mean Square Error (\textbf{RMSE})\index{Root Mean Square Error|see{RMSE}}\index{RMSE} has been used as a more comprehensive metric (it has the same units with the target variables) for presenting results in the next sections.
%%%%%%%%%%%%%%%%%%%%%%%%%%%%%%%%%%%%

To summarise CT's validation, suppose that for all considered consensus thresholds $\mbox{CT}_i$, $i\in\{1,...,21\}$, training yields $\mathcal{F}_i$ sets of selected features respectively, whose losses on the validation set are denoted by $\mathcal{L}_{i}^{(\mbox{\footnotesize val})}$. Then, if $i^{(*)}$ denotes the index of the selected CT and set of features, then it is given by:
\begin{equation}
i^{(*)} = \argmin_i \mathcal{L}_{i}^{(\mbox{\footnotesize val})}.
\end{equation}
Therefore, $\mbox{CT}_{i^{(*)}}$ is the result of the validation process and $\mathcal{F}_{i^{(*)}}$ is used in the testing phase.

Taking into consideration that both target values (rainfall and flu rates) can only be zero or positive, we threshold the negative inferred values with zero during testing, \emph{i.e.} we set $x_i \leftarrow \mbox{max}\{x_i,0\}$. We perform this filtering only in the testing phase; during CT's validation, we want to keep track of deviations in the negative space as well.

\subsection{Soft-Bolasso with consensus threshold validation}
\label{section:soft_bolasso_with_Ct_validation}
In this section we go through Algorithm \ref{algorithm_soft_bolasso_with_CT_validation}\index{Bolasso!soft}, which describes the core of our approach. The algorithm takes as input the set of candidate features, training and validation sets with observations and ground truth, a set of consensus thresholds (CT) that are going to be explored, the number of bootstraps, an optional numeric parameter \emph{maxFeaturesNum} that causes an early stop in every bootstrap of LASSO (when this number of features is reached, LASSO terminates), and another numeric parameter \emph{percentLCR}, the use of which is explained later on. The observations are assumed to be standardised and the responses centred. The outputs of the algorithm are the set of selected candidate features and the optimal consensus threshold.

First, a set of $\lambda$'s is formed by performing LARS on the entire training set as a basis and then by re-applying LARS on random samples of the training set. In every bootstrap, the training set is sampled uniformly with replacement and then used as input to LARS. After executing all bootstraps, we end up with a set of weights for each bootstrap and value of $\lambda$. By identifying the nonzero weights over all bootstraps, we get a set of selected features for each pair of $\lambda_i$ and CT$_i$.

%%%%% revisions %%%%%%%%%%%%%%%
The next step is to decide the optimal value of $\lambda$ for each CT$_i$. We use two methods for this task: either the largest consistent region (LCR) on the regularisation path where the selected values do not change, or if less than \emph{percentLCR}\% of the specified $\lambda$'s lie in the LCR, 5-fold cross-validation. In our experiments, we set \emph{percentLCR} to 0.05 (5\%). LCR is in general a preferable choice as the creation of a consistent region for the selected features is the main aim of Bolasso. Hence, if such a region exists, there is evidence that Bolasso's solution has consistent characteristics and therefore it shall be used \cite{Bach2008}.
%%%%%%%%%%%%%%%%%%%%%%%%%%%%%%%

Finally, after computing the optimal value of $\lambda$ for each CT$_i$, and therefore retrieving the corresponding optimal set of selected features, a validation set is used for deciding the optimal CT, and therefore the set of selected features.

%%%%%%%%%%%%%%%% soft bolasso algorithm %%%%%%%%%%%%%%%%%%%
\IncMargin{1em}
\begin{algorithm}[tp]
\setstretch{1.3}
\caption{Soft-Bolasso with Consensus Threshold Validation}
\label{algorithm_soft_bolasso_with_CT_validation}
\KwIn{$\mathcal{C}_{1:n}$, $\mathcal{X}^{(train)}_{[1:m,1:n]}$, $y^{(train)}_{1:m}$, $\mathcal{X}^{(val)}_{[1:m',1:n]}$, $y^{(val)}_{1:m'}$, $\text{CT}_{1:v}$, $B$, maxFeaturesNum, percentLCR}
\KwOut{$\hat{\mathcal{C}}_{1:p}$, $\hat{\text{CT}}$}
$\lambda_{1:k} \leftarrow \text{computeLambdas}\left(\mathcal{X}^{(train)}_{[1:m,1:n]}\text{, }y^{(train)}_{1:m}\right)$\;
\For{$i\leftarrow 1$ \KwTo $B$}{
    $\mathcal{X}^{(samp_i)}_{[1:m,1:n]}$, $y^{(samp_i)}_{1:m} \leftarrow \text{sample}\left(\mathcal{X}^{(train)}_{[1:m,1:n]}\text{, }y^{(train)}_{1:m}\right)$\;
    $\mathcal{W}_{[1:k,1:n]}^{(i)} \leftarrow \text{LARS}\left(\mathcal{X}^{(samp_i)}_{[1:m,1:n]}\text{, }y^{(samp_i)}_{1:m}\text{, }\lambda_{1:k}\text{, maxFeaturesNum}\right)$\;
}
\For(\tcp*[f]{$\forall$ $\lambda_i$}){$i\leftarrow 1$ \KwTo $k$}{
    \For(\tcp*[f]{$\forall$ $\text{CT}_j$}){$j\leftarrow 1$ \KwTo $v$}{
        $\mathcal{\dot{C}}_{1:z_{i,j}}^{(i,j)} \leftarrow \text{getFeatures}\left(\mathcal{W}_{[i,1:n]}^{(1:B)},\text{CT}_j\right)$\tcp*[r]{$|z_{i,j}|$ varies per iteration}
    }
}
\tcp{Optimal value for the regularisation parameter}
\For{$j\leftarrow 1$ \KwTo $v$}{
    $\mathcal{\ddot{C}}_{1:s_{j}}^{(j)}\text{, }\lambda_{\text{LCR}_j} \leftarrow \text{LCR}\left(\lambda_{1:k},\mathcal{\dot{C}}_{1:z_{1:k,j}}^{(1:k,j)}\right)$\tcp*[r]{$|s_{j}|$ varies per iteration}
    \If{\normalfont$\left(|\lambda_{\text{LCR}_j}| < k * \text{percentLCR}\right)$}{
        $\mathcal{\ddot{C}}_{1:s_{j}}^{(j)} \leftarrow \text{CrossVal}\left(\mathcal{\dot{C}}_{1:z_{1:k,j}}^{(1:k,j)}\text{, }\mathcal{X}^{(train)}_{[1:m,1:n]}\text{, }y^{(train)}_{1:m}\right)$\;
    }
}
\tcp{Optimal value for the consensus threshold}
$\hat{\mathcal{C}}_{1:p}\text{, }\hat{\text{CT}} \leftarrow \text{validate}\left(\mathcal{\ddot{C}}_{1:s_{1:v}}^{(1:v)}\text{, }\mathcal{X}^{(val)}_{[1:m',1:n]}\text{, }y^{(val)}_{1:m'}\right)$\;
\Return $\hat{\mathcal{C}}_{1:p}$, $\hat{\text{CT}}$\;
\end{algorithm}\DecMargin{1em}
%%%%%%%%%%%%%%%%%%%%%%%%%%%%%%%%%%%%%%%%%%%%%%%%%%%%%%%%%%%%%%%%%%%%

\subsection{Extracting candidate features and forming feature classes}
\label{section:extracting_candidate_features_feature_classes}
Three classes of candidate features\index{feature class} have been investigated: unigrams or 1-grams (denoted by \textbf{\emph{U}}), bigrams or 2-grams (\textbf{\emph{B}}) and a hybrid combination (\textbf{\emph{H}}) of 1-grams and 2-grams. 1-grams being single words cannot be characterised by a consistent semantic interpretation in most of the topics. They take different meanings and express distinct outcomes based on the surrounding textual context. 2-grams on the other hand can be more focused semantically. However, their frequency in a corpus is expected to be lower than the one of 1-grams. Particularly, in the Twitter corpus which consists of very short pieces of text (tweets are at most 140 characters long), the daily frequency for some of them is sometimes close to zero.

%%%%%%%%%%%%%%%%%%%%%%% hybrid combination algorithm I %%%%%%%%%%%%%%%%%%%%%%%%
\IncMargin{1em}
\begin{algorithm}[tp]
\setstretch{1.3}
\caption{Hybrid Combination of 1-grams and 2-grams ($H$)}
\label{algorithm_hybrid_combination_1grams_2grams_I}
\KwIn{$\hat{\mathcal{C}}_{[1:v,1:p_{1:v}]}^{(\text{1g})}$, $\hat{\mathcal{C}}_{[1:v,1:q_{1:v}]}^{(\text{2g})}$, $\mathcal{X}^{(train\text{,1g})}_{[1:m,1:n_{\text{1g}}]}$, $\mathcal{X}^{(train\text{,2g})}_{[1:m,1:n_{\text{2g}}]}$, $y^{(train)}_{1:m}$, $\mathcal{X}^{(val\text{,1g})}_{[1:m',1:n_{\text{1g}}]}$, $\mathcal{X}^{(val\text{,2g})}_{[1:m',1:n_{\text{2g}}]}$, $y^{(val)}_{1:m'}$, $\text{CT}_{1:v}$}
\KwOut{$\hat{\text{CT}}^{(\text{h})}$, $\hat{\mathcal{C}}_{1:p_{\text{minL}} + q_{\text{minL}}}$}
\For{$i \leftarrow 1$ \KwTo $v$}{
    $\mathcal{X}^{(train_i)}_{[1:m,1:p_i + q_i]} \leftarrow \mathcal{X}^{(train\text{,1g})}_{[1:m,\hat{\mathcal{C}}_{[i,1:p_i]}^{(\text{1g})}]} \cup \mathcal{X}^{(train\text{,2g})}_{[1:m,\hat{\mathcal{C}}_{[i,1:q_i]}^{(\text{2g})}]}$\;
    $\mathcal{X}^{(val_i)}_{[1:m,1:p_i + q_i]} \leftarrow \mathcal{X}^{(val\text{,1g})}_{[1:m,\hat{\mathcal{C}}_{[i,1:p_i]}^{(\text{1g})}]} \cup \mathcal{X}^{(val\text{,2g})}_{[1:m,\hat{\mathcal{C}}_{[i,1:q_i]}^{(\text{2g})}]}$\;
    $\mathcal{W}_{[i,1:p_i + q_i]} \leftarrow \text{OLS}\left(\mathcal{X}^{(train_i)}_{[1:m,1:p_i + q_i]},y^{(train)}_{1:m}\right)$\;
    $\mathcal{L}_{i} = \text{MSE}\left(\mathcal{X}^{(val_i)}_{[1:m',1:p_i + q_i]}\mathcal{W}_{[i,1:p_i + q_i]},y^{(val)}_{1:m'}\right)$\;
}
\tcp{Retrieve CT's index where the minimum loss occurs}
$\text{minL} \leftarrow \displaystyle\argmin_i \mathcal{L}_{1:v} $\;
$\hat{\text{CT}}^{(\text{h})} \leftarrow \text{CT}_{\text{minL}}$\;
$\hat{\mathcal{C}}_{1:p_{\text{minL}} + q_{\text{minL}}} \leftarrow \hat{\mathcal{C}}_{[\text{minL},1:p_{\text{minL}}]}^{(\text{1g})} \cup \hat{\mathcal{C}}_{[\text{minL},1:q_{\text{minL}}]}^{(\text{2g})}$\;
\Return $\hat{\text{CT}}^{(\text{h})}$, $\hat{\mathcal{C}}_{1:p_{\text{minL}} + q_{\text{minL}}}$\;
\end{algorithm}\DecMargin{1em}

The hybrid class of features exploits the advantages of classes $U$ and $B$ and reduces the impact of their disadvantages. It is formed by combining the training results of $U$ and $B$ for all CTs. We have tried two approaches for forming a hybrid set of features, described in Algorithms \ref{algorithm_hybrid_combination_1grams_2grams_I} and \ref{algorithm_hybrid_combination_1grams_2grams_II}, and denoted by $H$ and $H_{\text{II}}$ respectively.

Suppose that for all considered consensus thresholds $\mbox{CT}_i$, $i\in\{1,...,|\text{CT}|\}$, 1-grams and 2-grams selected via Bolasso are denoted by $\mathcal{F}_{i}^{(U)}$ and $\mathcal{F}_{i}^{(B)}$ respectively. Then, in the first hybrid approach (\emph{H}) the pseudo-selected n-grams for all CTs for the hybrid class $\mathcal{F}_{i}^{(H)}$ are formed by their one-by-one union, $\mathcal{F}_{i}^{(H)} = \{ \mathcal{F}_{i}^{(U)}\cup\mathcal{F}_{i}^{(B)}\}\mbox{, }i\in\{1,...,|\text{CT}|\}$. Likewise, $\mathcal{Z}_{i}^{(H)} = \{ \mathcal{Z}_{i}^{(U)}\cup\mathcal{Z}_{i}^{(B)}\}\mbox{, }i\in\{1,...,|\text{CT}|\}$, where $\mathcal{Z}$ denotes the VSR of each feature class (using the notation of Section \ref{section:inference_and_consensus_threshold}). Validation and testing are performed on $\mathcal{Z}_{i}^{(H)}$ as it has already been described in Section \ref{section:inference_and_consensus_threshold}.

%%%%%%%%%%%%%%%%%%%%%%% hybrid combination algorithm II (all combinations) %%%%%%%%%%%%%%%%%%%%%%%%
\IncMargin{1em}
\begin{algorithm}[tp]
\setstretch{1.3}
\caption{Hybrid Combination of 1-grams and 2-grams ($H_{\text{II}}$)}
\label{algorithm_hybrid_combination_1grams_2grams_II}
\KwIn{$\hat{\mathcal{C}}_{[1:v,1:p_{1:v}]}^{(\text{1g})}$, $\hat{\mathcal{C}}_{[1:v,1:q_{1:v}]}^{(\text{2g})}$, $\mathcal{X}^{(train\text{,1g})}_{[1:m,1:n_{\text{1g}}]}$, $\mathcal{X}^{(train\text{,2g})}_{[1:m,1:n_{\text{2g}}]}$, $y^{(train)}_{1:m}$, $\mathcal{X}^{(val\text{,1g})}_{[1:m',1:n_{\text{1g}}]}$, $\mathcal{X}^{(val\text{,2g})}_{[1:m',1:n_{\text{2g}}]}$, $y^{(val)}_{1:m'}$, $\text{CT}_{1:v}$}
\KwOut{$\hat{\text{CT}}^{(\text{h,1g})}$, $\hat{\text{CT}}^{(\text{h,2g})}$, $\hat{\mathcal{C}}_{1:p_{\text{minL\_1g}} + q_{\text{minL\_2g}}}$}
\For{$i \leftarrow 1$ \KwTo $v$}{
    \For{$j \leftarrow 1$ \KwTo $v$}{
        $\mathcal{X}^{(train_{i,j})}_{[1:m,1:p_i + q_j]} \leftarrow \mathcal{X}^{(train\text{,1g})}_{[1:m,\hat{\mathcal{C}}_{[i,1:p_i]}^{(\text{1g})}]} \cup \mathcal{X}^{(train\text{,2g})}_{[1:m,\hat{\mathcal{C}}_{[j,1:q_j]}^{(\text{2g})}]}$\;
        $\mathcal{X}^{(val_{i,j})}_{[1:m,1:p_i + q_j]} \leftarrow \mathcal{X}^{(val\text{,1g})}_{[1:m,\hat{\mathcal{C}}_{[i,1:p_i]}^{(\text{1g})}]} \cup \mathcal{X}^{(val\text{,2g})}_{[1:m,\hat{\mathcal{C}}_{[j,1:q_j]}^{(\text{2g})}]}$\;
        $\mathcal{W}_{[1:p_i + q_j]}^{(i,j)} \leftarrow \text{OLS}\left(\mathcal{X}^{(train_{i,j})}_{[1:m,1:p_i + q_j]},y^{(train)}_{1:m}\right)$\;
        $\mathcal{L}_{i,j} = \text{MSE}\left(\mathcal{X}^{(val_{i,j})}_{[1:m',1:p_i + q_j]}\mathcal{W}_{[1:p_i + q_j]}^{(i,j)},y^{(val)}_{1:m'}\right)$\;
    }
}
\tcp{Retrieve CT's index where the minimum loss occurs}
$\text{minL\_1g},\text{minL\_2g} \leftarrow \displaystyle\argmin_{i,j} \mathcal{L}_{[1:v,1:v]} $\;
$\hat{\text{CT}}^{(\text{h,1g})} \leftarrow \text{CT}_{\text{minL\_1g}}$\;
$\hat{\text{CT}}^{(\text{h,2g})} \leftarrow \text{CT}_{\text{minL\_2g}}$\;
$\hat{\mathcal{C}}_{1:p_{\text{minL\_1g}} + q_{\text{minL\_2g}}} \leftarrow \hat{\mathcal{C}}_{[\text{minL\_1g},1:p_{\text{minL\_1g}}]}^{(\text{1g})} \cup \hat{\mathcal{C}}_{[\text{minL\_2g},1:q_{\text{minL\_2g}}]}^{(\text{2g})}$\;
\Return $\hat{\text{CT}}^{(\text{h,1g})}$, $\hat{\text{CT}}^{(\text{h,2g})}$, $\hat{\mathcal{C}}_{1:p_{\text{minL\_1g}} + q_{\text{minL\_2g}}}$\;
\end{algorithm}\DecMargin{1em}

In the second hybrid approach, the pseudo-selected n-grams are formed by exploring all possible $|\text{CT}|^2$ pairs of selected 1-grams and 2-grams. Thus, in that case $\mathcal{F}_{i}^{(H_{\text{II}})} = \{ \mathcal{F}_{i}^{(U)}\cup\mathcal{F}_{j}^{(B)}\}\mbox{, } \forall i,j\in\{1,...,|\text{CT}|\}$, and then this hybrid combination method proceeds in the same manner as the previous one. Note that compiling an optimal hybrid scheme is not the main focus here; our aim is to investigate whether a simple combination of 1-grams and 2-grams is able to deliver better results. The experimental results in the following sections do indeed indicate that feature class $H$ performs on average better than $U$ and $B$, something that does not always hold for $H_{\text{II}}$.

Finally, we create one more class of features by combining again all 1-grams and 2-grams, but this time we perform the entire training process -- \ie run the soft Bolasso algorithm --  on this unified data set. This method -- conventionally named as $UB$ -- is expected to perform worse than the hybrid combinations due to the fact that it increases the dimensionality of the problem at hand without providing more training points. Indeed, as we will see in the next sections, the first hybrid approach ($H$) performs better than all the other class features.

%%%%%%%%%%%%%%%%%%%%%%%%%%%%%%%%%%%%%%%%%%%%%%%%%%%%%%%%%%%%%
%%%%%%%%%%%%%%%%%%%%%%%%% end of methodology

\section{Comparing with other methodologies}
\label{section:comparing_with_other_methodologies}
As part of the evaluation process, we compare our results with a baseline approach which encapsulates the methodologies in \cite{ginsberg2008detecting} and \cite{Culotta2010}. Those approaches, as explained in Section \ref{section:related_work_journal}, mainly differ in the feature selection process which is performed via correlation analysis; Algorithm \ref{algorithm_corr_for_feature_selection} describes this process in detail.

Briefly, given a set $\mathcal{C}$ of $n$ candidate features, their computed VSRs for training and validation $\mathcal{X}^{(train)}$ and $\mathcal{X}^{(val)}$ and the corresponding response values $y^{(train)}$ and $y^{(val)}$, this feature selection process: \textbf{a)} computes the Pearson correlation coefficients ($\rho$) between each candidate feature and the response values in the training set, \textbf{b)} ranks the retrieved correlation coefficients in descending order, \textbf{c)} computes the OLS-fit loss ($\mathcal{L}$) of incremental subsets of the top-$k$ correlated terms on the validation set and \textbf{d)} selects the subset of candidate features with the minimum loss. The inference performance of the selected features is evaluated on a (disjoint) test set.

%%%%%%%%%%%% baseline method
\IncMargin{1em}
\begin{algorithm}[tp]
\setstretch{1.3}
\caption{Baseline Method: Feature Selection via Correlation Analysis}
\label{algorithm_corr_for_feature_selection}
\KwIn{$\mathcal{C}_{1:n}$, $\mathcal{X}^{(train)}_{[1:m,1:n]}$, $y^{(train)}_{1:m}$, $\mathcal{X}^{(val)}_{[1:m',1:n]}$, $y^{(val)}_{1:m'}$}
\KwOut{$\hat{\mathcal{C}}_{1:p}$}
$\rho_{1:n} \leftarrow \text{correlation}\left(\mathcal{X}^{(train)}_{[1:m,1:n]}\text{, }y^{(train)}_{1:m}\right)$\;
$\hat{\rho}_{1:n} \leftarrow \text{descendingCorrelationIndex}(\rho_{1:n})$\;
$\hat{\mathcal{C}}_{1:n} \leftarrow \mathcal{C}_{\hat{\rho}_{1:n}}$\;
$i \leftarrow 1$\;
\While{$i \leq k$}{
$\mathcal{L}_{i} \leftarrow \text{validate}\left(\mathcal{X}^{(train)}_{[1:m,\hat{\rho}_{1:i}]}\text{, }y^{(train)}_{1:m}\text{, }\mathcal{X}^{(val)}_{[1:m',\hat{\rho}_{1:i}]}\text{, }y^{(val)}_{1:m'}\right)$\;
$i \leftarrow i + 1$\;
}
$\displaystyle p \leftarrow \argmin_i \mathcal{L}_{i}$\;
\Return $\hat{\mathcal{C}}_{1:p}$\;
\end{algorithm}\DecMargin{1em}

\section{Case study I: nowcasting rainfall rates from Twitter}
\label{section_nowcasting_rainfall}
In the first case study, we exploit the content of Twitter to infer daily rainfall rates\index{rainfall rates} (measured in millimetres of precipitation) for five UK cities, namely \emph{Bristol}, \emph{London}, \emph{Middlesbrough}, \emph{Reading} and \emph{Stoke-on-Trent}. The choice of those locations has been based on the availability of ground truth, \emph{i.e.} daily rainfall measurements from weather stations installed in their vicinity.

We consider the inference of precipitation levels at a given time and place as a good benchmark problem, in that it has many of the properties of other more useful scenarios, while still allows us to verify the performance of the system since rainfall is a measurable variable. The event of rain is a piece of information available to the significant majority of Twitter users and affects various activities that could form a discussion topic in tweets. Furthermore, predictions about it are not always easy due to its non smooth and sometimes unpredictable behaviour \cite{jenkins2008climate}.

The candidate markers for this case study are extracted from weather related web references, such as Wikipedia's page on Rainfall, an English language course on weather vocabulary, a page with formal weather terminology and several others. The majority of the extracted candidate features is not directly related to the target topic, but there exists a subset of markers which could probably offer a good semantic interpretation. Markers with a count $\leq$ 10 in the Twitter corpus used for this case study are removed. Hence, from the extracted 2381 1-grams, 2159 have been kept as candidates; likewise the 7757 extracted 2-grams have been reduced to 930.\footnote{ Complete lists of the web references used and the extracted markers for the rainfall case study are available at \url{http://geopatterns.enm.bris.ac.uk/twitter/rainfall-cf.php}.}

\subsection{Experimental settings}
A year of Twitter data and rainfall observations (from the 1$^{st}$ of July, 2009 to the 30$^{th}$ of June, 2010) formed the input data for this experiment. For this time period and the considered locations, 8.5 million tweets have been collected. In each run of Bolasso the number of bootstraps is proportional to the size of the training sample (approximately 13\% using the same principle as in \cite{Bach2008}), and in every bootstrap we select at most 300 features by performing at most 900 iterations. A bootstrap is completed as soon as one of those two stopping criteria is met. This is an essential trade-off which guarantees a quicker execution of the learning phase, especially when dealing with large amounts of data.

The performance of each feature class ($U$, $B$, $H$, $H_{\text{II}}$, $UB$) is computed by applying a 6-fold cross validation. Each fold is based on 2 months of data starting from the month pair July-August (2009) and ending with May-June (2010). In every step of the cross validation, 5 folds are used for training, the first half (a month-long data) of the remaining fold for validating CT and the second half for testing the performance of the selected markers and their weights. Training is performed by using the VSRs of all five locations in a batch data set, CT's validation is carried out on the same principle (we learn the same markers-weights under the same CT for all locations), and finally testing is done both on the batch data set (to retrieve a total performance evaluation) as well as on each location separately. Finally, we also compute the inference performance of the baseline approach for feature selection (Algorithm \ref{algorithm_corr_for_feature_selection}) for the same training, validation and testing sets, considering the top $k =$ 300 correlated terms.

\subsection{Results}
\label{section:results_rainfall_bolasso}

The derived CTs as well as the numbers of selected features for all rounds of the 6-fold cross validation are presented on Table \ref{table_rainfall_ct}. In most rounds CT values are high and as a result the number of selected features is relatively small. In turn, this could mean that in most occasions a few markers were able to capture the rainfall rates signal. The last round, where the validation data set is based on July 2009, is an exception, where we derive lower CTs for most feature classes as well as larger numbers of selected features. This can be interpreted by the fact that July is not only the 2nd most rainy month in our data set, but also a summer month -- hence, tweets for rain could be followed or preceded by tweets discussing a sunny day, creating instabilities during the validation process. In addition, our data set is restricted to only one year of weather observations and therefore seasonal patterns like this one are not expected to be entirely captured.

\begin{table}
\caption{Nowcasting Rainfall Rates -- Derived Consensus Thresholds and numbers of selected features (in parentheses) for all Feature Classes (FC) in the rounds of 6-fold cross validation -- Fold $i$ denotes the validation/testing fold of round $7-i$}
\label{table_rainfall_ct}
\centering
\scriptsize
\(\begin{tabular}{c|cccccc}
\textbf{FC} & \textbf{Fold 6}  & \textbf{Fold 5}  & \textbf{Fold 4}  & \textbf{Fold 3}  & \textbf{Fold 2}  & \textbf{Fold 1} \\
\hline
$U$             & 100\% (\emph{4})  & 92.5\% (\emph{19}) & 90\% (\emph{17}) & 92.5\% (\emph{12}) & 90\% (\emph{17})   & 75\%   (\emph{28})  \\
$B$             & 90\% (\emph{21})  & 67.5\% (\emph{10}) & 95\% (\emph{10}) & 67.5\% (\emph{15}) & 90\% (\emph{9})    & 62.5\% (\emph{38}) \\
$H$             & 100\% (\emph{8})  & 92.5\% (\emph{27}) & 95\% (\emph{21}) & 92.5\% (\emph{27}) & 90\% (\emph{26})   & 52.5\% (\emph{131})\\
$H_{\text{II}}$ & 100/90\% (\emph{25}) & 92.5/67.5\% (\emph{29}) & 90/95\% (\emph{27}) & 92.5/57.5\% (\emph{24}) & 90/90\% (\emph{26}) & 82.5/100\% (\emph{23})\\
$UB$            & 100\% (\emph{5})  & 65\% (\emph{29})   & 95\% (\emph{8})  & 75\% (\emph{25})   & 87.5\% (\emph{12}) & 75\% (\emph{40})
\end{tabular}\)
\end{table}

Detailed performance evaluation results (total and per location for all feature classes) are presented on Table \ref{table_rainfall_results_bolasso}. For a better interpretation of the numerical values (in $mm$), consider that the average rainfall rate in our data set is equal to $1.8$ with a standard deviation of 3.9 and a range of [0,65]. Our method outperforms the baseline approach (see Algorithm \ref{algorithm_corr_for_feature_selection}) in all-but-one intermediate RMSE indications as well as in total for all feature classes, achieving an improvement of 10.74\% (derived by comparing the lowest total RMSEs for each method). The overall performance for our method indicates that feature class $H$ performs better than all others, and interestingly unigrams outperform bigrams as well as the remaining hybrid feature classes.

\begin{table}
\caption{Nowcasting Rainfall Rates -- RMSEs (in \emph{mm}) for all Feature Classes (FC) and locations in the rounds of 6-fold cross validation -- Fold $i$ denotes the validation/testing fold of round $7-i$. The last column holds the RMSEs of the baseline method.}
\label{table_rainfall_results_bolasso}
\renewcommand{\arraystretch}{1.2}
\setlength\tabcolsep{1mm}
\newcolumntype{C}{>{\centering\arraybackslash} m{1cm} }
\newcolumntype{V}{>{\centering\arraybackslash} m{1.40cm} }
\centering
\(\begin{tabular}{c|c|*{6}{C}|V||V}
\textbf{Location} & \textbf{FC} & \textbf{Fold 6}  & \textbf{Fold 5}  & \textbf{Fold 4}  & \textbf{Fold 3}  & \textbf{Fold 2}  & \textbf{Fold 1}  & \textbf{Mean RMSE} & \textbf{BS-Mean RMSE}\\\hline\hline
\emph{Bristol}         & $U$ & 1.164   & 1.723    & 1.836   & 2.911 & 1.607 & 2.348 & 1.931 & 2.173\\
                       & $B$ & 1.309   & 1.586    & 2.313   & 3.371 & 1.59  & 1.409 & 1.93  & 2.218\\
                       & $H$ & 1.038   & 1.631    & 2.334   & 2.918 & 1.579 & 2.068 & 1.928 & 2.094\\\hline
\emph{London}          & $U$ & 1.638   & 1.507    & 5.079   & 2.582 & 1.62  & 6.261 & 3.115 & 3.297\\
                       & $B$ & 1.508   & 5.787    & 4.887   & 3.403 & 1.478 & 6.568 & 3.939 & 4.305\\
                       & $H$ & 1.471   & 1.526    & 4.946   & 2.813 & 1.399 & 6.13  & 3.047 & 4.101\\\hline
\emph{Middlesbrough}   & $U$ & 4.665   & 1.319    & 3.102   & 2.618 & 2.949 & 2.536 & 2.865 & 2.951\\
                       & $B$ & 4.355   & 1.069    & 3.379   & 2.22  & 2.918 & 2.793 & 2.789 & 2.946\\
                       & $H$ & 4.47    & 1.098    & 3.016   & 2.504 & 2.785 & 2.353 & 2.704 & 6.193\\\hline
\emph{Reading}         & $U$ & 2.075   & 1.566    & 2.087   & 2.393 & 1.981 & 2.066 & 2.028 & 2.168\\
                       & $B$ & 0.748   & 2.74     & 1.443   & 3.016 & 1.572 & 3.429 & 2.158 & 2.159\\
                       & $H$ & 1.636   & 1.606    & 1.368   & 2.571 & 1.695 & 2.145 & 1.836 & 2.214\\\hline
\emph{Stoke-on-Trent}  & $U$ & 3.46    & 1.932    & 1.744   & 4.375 & 2.977 & 1.962 & 2.742 & 2.855\\
                       & $B$ & 3.762   & 1.493    & 1.433   & 2.977 & 2.447 & 2.668 & 2.463 & 2.443\\
                       & $H$ & 3.564   & 1.37     & 1.499   & 3.785 & 2.815 & 1.931 & 2.494 & 2.564\\\hline\hline
\textbf{Total RMSE}    & $U$ & 2.901   & 1.623    & 3.04    & 3.062 & 2.31  & 3.443 & \textbf{2.73}  & 2.915\\
                       & $B$ & 2.745   & 3.062    & 2.993   & 3.028 & 2.083 & 3.789 & \textbf{2.95}  & 3.096\\
                       & $H$                & 2.779   & 1.459    & 2.937   & 2.954 & 2.145 & 3.338 & \textbf{2.602} & 4.395\\
                       & $H_{\text{II}}$    & 2.68    & 3.009    & 2.851   & 2.962 & 2.145 & 3.461 & \textbf{2.851} &--\\
                       & $UB$               & 2.875   & 3.157    & 3.006   & 2.972 & 2.071 & 3.553 & \textbf{2.939} &--
%\multirow{3}*{\textbf{BS-Mean}}
\end{tabular}\)
\end{table}

\begin{figure*}[tp]
    \begin{center}
    \includegraphics[width=6in]{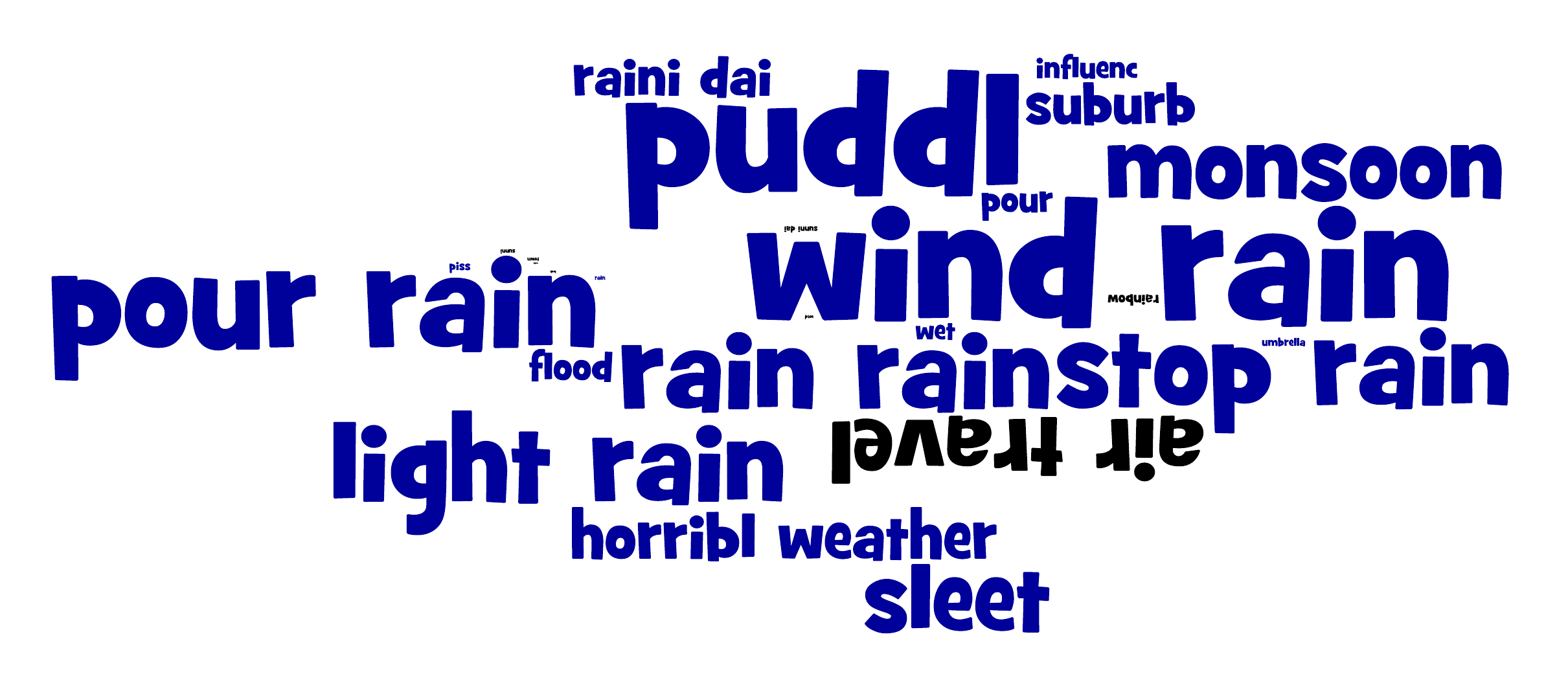}
    \end{center}
    \caption{Table \ref{table_features_rain_hybrid} in a word cloud, where font size is proportional to regression's weight and flipped words have negative weights}
    \label{fig_table_rain_hybrid_wordle}
\end{figure*}

Presenting all intermediate results for each round of the cross validation would have been intractable. In the remaining part of this Section, we present the results of learning and testing for cross-validation's round 5 only, where the month of testing is October, 2009. Tables \ref{table_features_rain_1grams}, \ref{table_features_rain_2grams} and \ref{table_features_rain_hybrid} list the selected features in alphabetical order together with their weights for feature classes $U$, $B$ and $H$ respectively.\footnote{ We do not show detailed results for the feature classes $H_{\text{II}}$ and $UB$ since feature class $H$ -- which is based on the same principle of combining 1-grams and 2-grams -- outperforms them.} For class $H$ we have also compiled a word cloud with the selected features as a more comprehensive representation of the selection outcome (Figure \ref{fig_table_rain_hybrid_wordle}). The majority of the selected 1-grams (Table \ref{table_features_rain_1grams}) has a very close semantic connection with the underlying topic -- stem `\emph{puddl}' holds the largest weight, whereas stem `\emph{sunni}' has taken a negative weight and interestingly, the word `\emph{rain}' has a relatively small weight. There also exist a few words without a direct semantic connection, but the majority of them has negative weights and in a way they might be acting as mitigators of non weather related uses of the remaining rainy weather oriented and positively weighted features. The selected 2-grams (Table \ref{table_features_rain_2grams}) have a clearer semantic connection with the topic; `\emph{pour rain}' acquires the highest weight. In this particular case, the features for class $H$ are formed by the exact union of the ones in classes $U$ and $B$, but take different weights (Table \ref{table_features_rain_hybrid} and Figure \ref{fig_table_rain_hybrid_wordle}).%In $H$ (Table \ref{table_features_rain_hybrid} and Figure \ref{fig_table_rain_hybrid_wordle}), stem `\emph{wind rain}' takes the highest weight.
%Stemmed 2-gram `\emph{pour rain}' has the biggest weight, whereas `\emph{sunni dai}' is weighted negatively (similarly to 1-grams case).

\begin{table}
\caption{Feature Class $U$ -- 1-grams selected by Bolasso for Rainfall case study (Round 5 of 6-fold cross validation) -- All weights (\textbf{w}) should be multiplied by $10^3$}
\label{table_features_rain_1grams}
\footnotesize
\renewcommand{\arraystretch}{1.1}
\setlength\tabcolsep{1mm}
\centering
\(\begin{tabular}{cc|cc|cc|cc|cc}
\textbf{1-gram} & \textbf{w} & \textbf{1-gram}  & \textbf{w} & \textbf{1-gram} & \textbf{w} & \textbf{1-gram} & \textbf{w} & \textbf{1-gram} & \textbf{w}\\\hline
flood       & 0.767  & piss     & 0.247                       & rainbow & -0.336 & todai     & -0.055 & wet & 0.781\\
influenc    & 0.579  & pour     & 1.109                       & sleet   & 1.766  & town      & -0.134 & &  \\
look        & -0.071 & \emph{\textbf{puddl}} & \textbf{3.152} & suburb  & 1.313  & umbrella  & 0.223  & &  \\
monsoon     & 2.45   & rain    & 0.19                         & sunni   & -0.193 & wed       & -0.14  & &
\end{tabular}\)
\end{table}

\begin{table}
\caption{Feature Class $B$ -- 2-grams selected by Bolasso for Rainfall case study (Round 5 of 6-fold cross validation) -- All weights (\textbf{w}) should be multiplied by $10^3$}
\label{table_features_rain_2grams}
\footnotesize
\renewcommand{\arraystretch}{1.1}
\setlength\tabcolsep{1mm}
\centering
\(\begin{tabular}{cc|cc|cc|cc|cc}
\textbf{2-gram} & \textbf{w} & \textbf{2-gram} & \textbf{w} & \textbf{2-gram} & \textbf{w} & \textbf{2-gram} & \textbf{w} & \textbf{2-gram} & \textbf{w}  \\\hline
air travel      & -2.167 & light rain  & 2.508  & raini dai & 2.046 & stop rain  & 3.843 & wind rain & 5.698 \\
horribl weather & 3.295  & \emph{\textbf{pour rain}} & \textbf{7.161} & rain rain & 4.490 & sunni dai  & -0.97 & &\\
\end{tabular}\)
\end{table}

\begin{table}
\caption{Feature Class $H$ -- Hybrid selection of 1-grams and 2-grams for Rainfall case study (Round 5 of 6-fold cross validation) -- All weights (\textbf{w}) should be multiplied by $10^3$}
\label{table_features_rain_hybrid}
\footnotesize
\renewcommand{\arraystretch}{1.1}
\setlength\tabcolsep{1mm}
\centering
\(\begin{tabular}{cc|cc|cc|cc|cc}
\textbf{\emph{n}-gram} & \textbf{w} & \textbf{\emph{n}-gram} & \textbf{w} & \textbf{\emph{n}-gram} & \textbf{w} & \textbf{\emph{n}-gram} & \textbf{w} & \textbf{\emph{n}-gram} & \textbf{w} \\\hline
air travel      & -1.841  & monsoon     & 2.042  & rain rain   & 2.272   & sunni       & -0.125 & wet         & 0.524\\
flood           & 0.781   & piss        & 0.24   & rainbow     & -0.294  & sunni dai   & -0.165 & \textbf{\emph{wind rain}}   & \textbf{3.399}\\
horribl weather & 1.282   & pour        & 0.729  & raini dai   & 1.083   & todai       & -0.041 & & \\
influenc        & 0.605   & pour rain   & 2.708  & sleet       & 1.891   & town        & -0.112 & & \\
light rain      & 2.258   & puddl       & 3.275  & stop rain   & 2.303   & umbrella    & 0.229  & & \\
look            & -0.067  & rain        & 0.122  & suburb      & 1.116   & wed         & -0.1   & &
\end{tabular}\)
\end{table}

\begin{figure*}[tp]
    \begin{center}
    \subfigure[\emph{Bristol} -- RMSE: 1.607] {\includegraphics[width=2.75in]{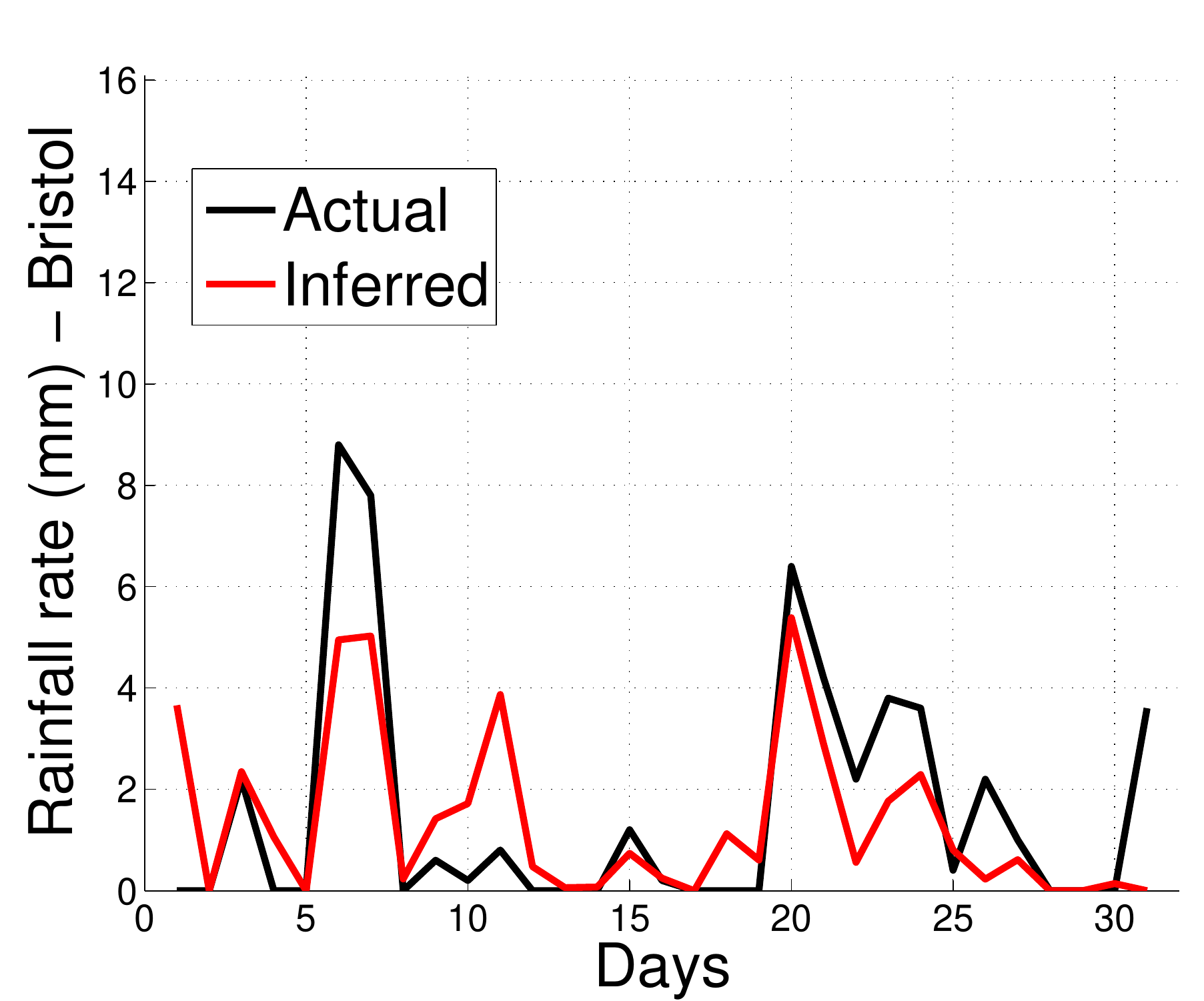}
    \label{fig_rain1gramsB}}
    \hfil
    \subfigure[\emph{London} -- RMSE: 1.62] {\includegraphics[width=2.75in]{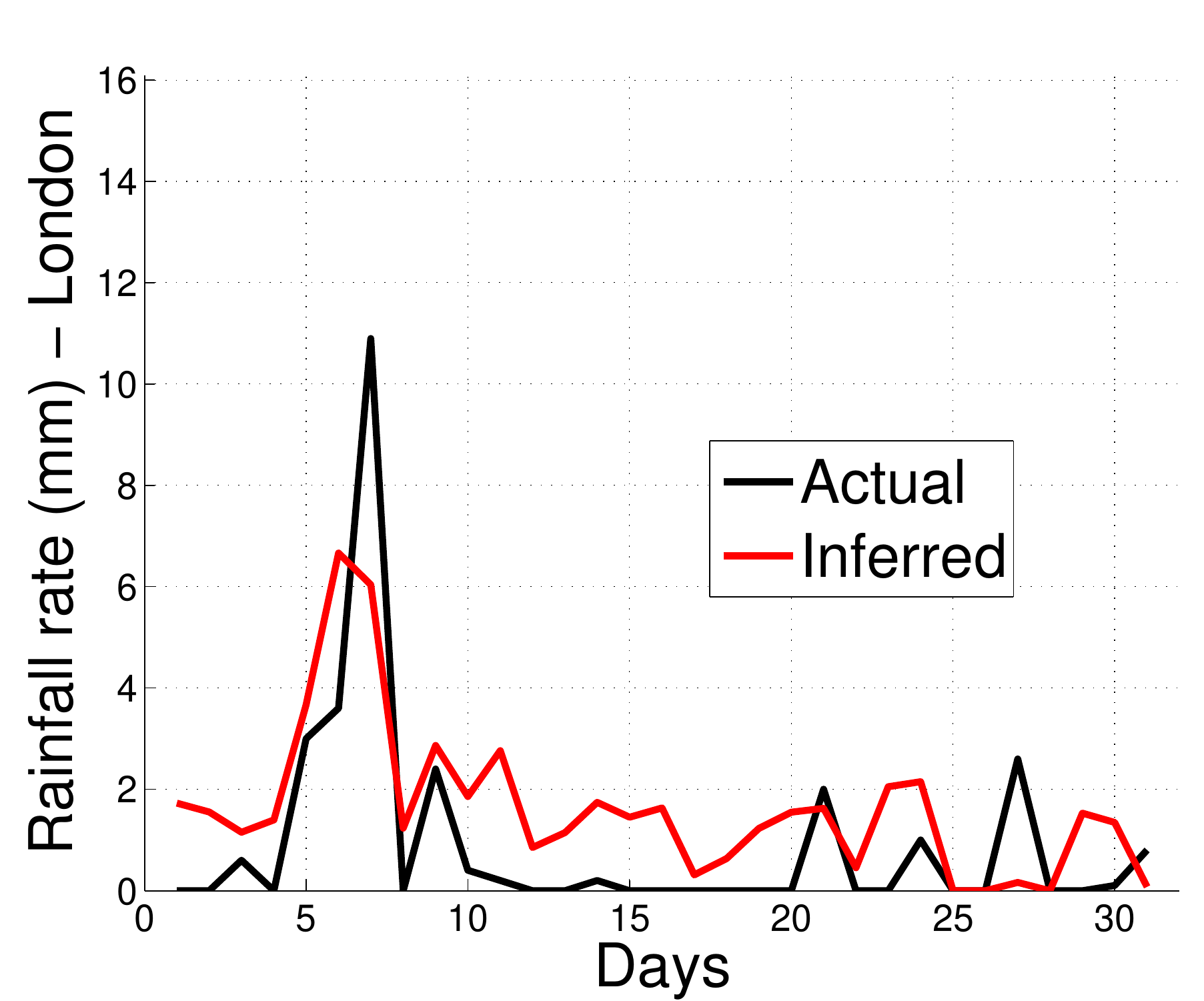}
    \label{fig_rain1gramsL}}
    \hfil
    \subfigure[\emph{Middlesbrough} -- RMSE: 2.949] {\includegraphics[width=2.75in]{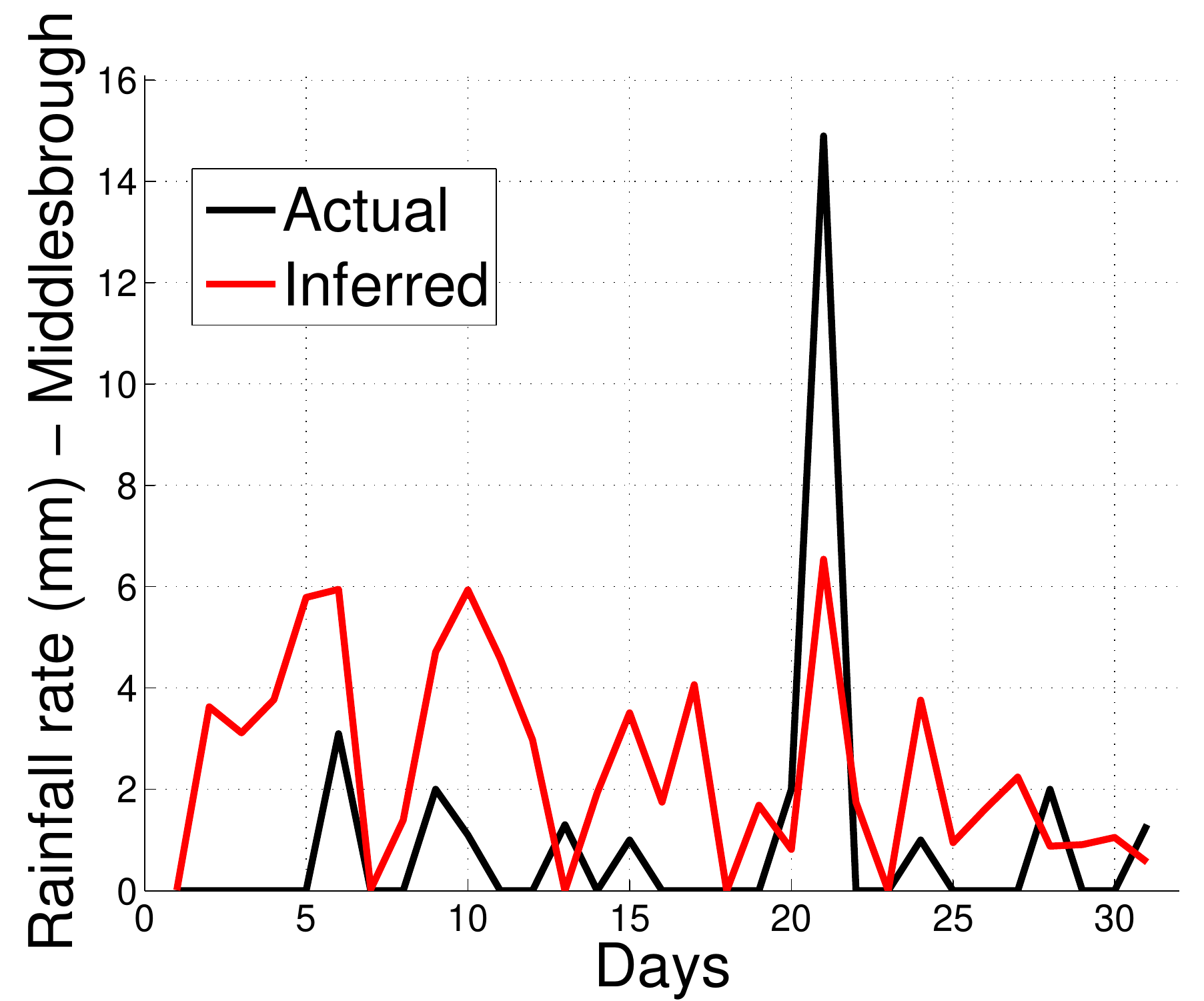}
    \label{fig_rain1gramsM}}
    \subfigure[\emph{Reading} -- RMSE: 1.981] {\includegraphics[width=2.75in]{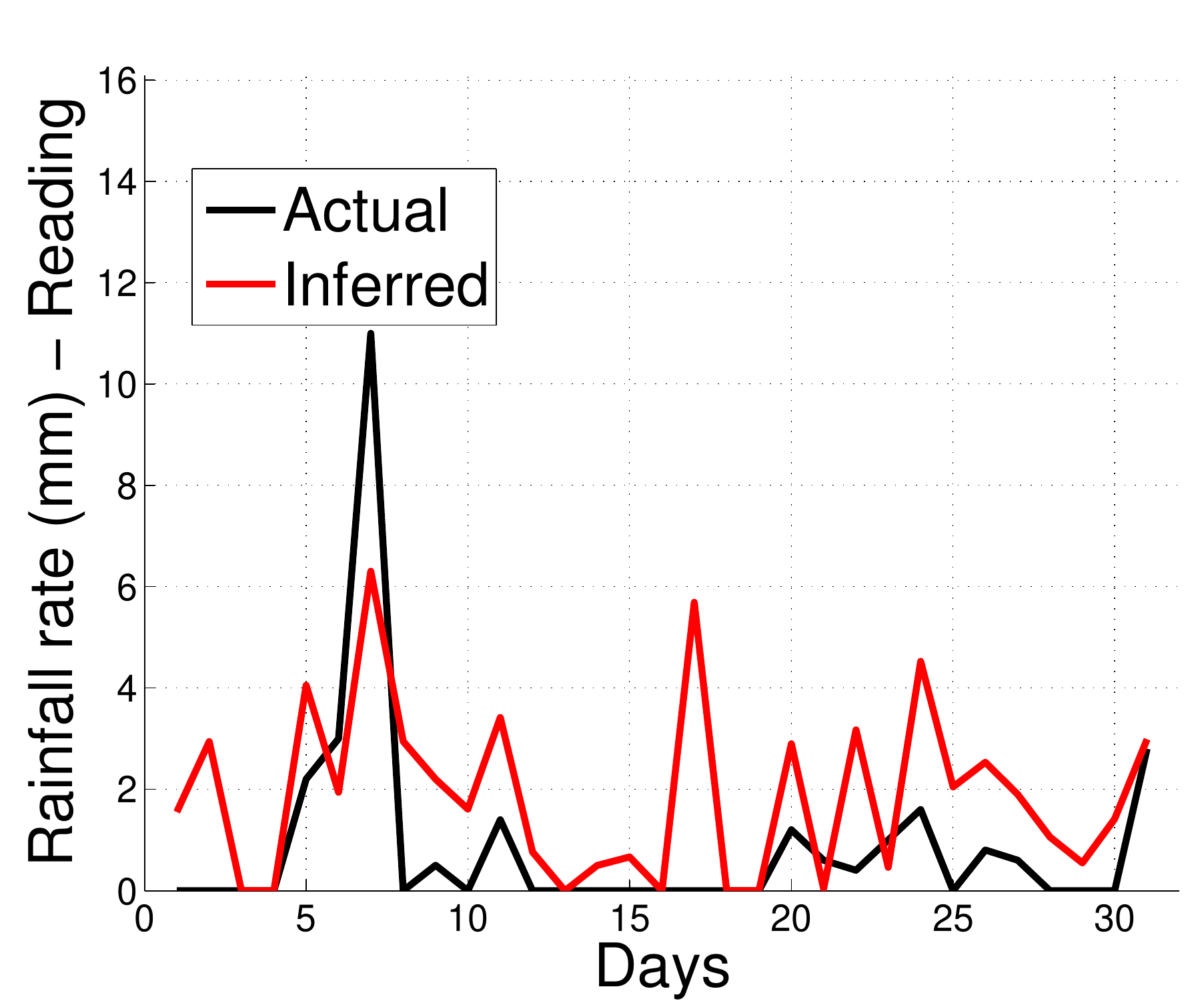}
    \label{fig_rain1gramsR}}
    \hfil
    \subfigure[\emph{Stoke-on-Trent} -- RMSE: 2.977] {\includegraphics[width=2.75in]{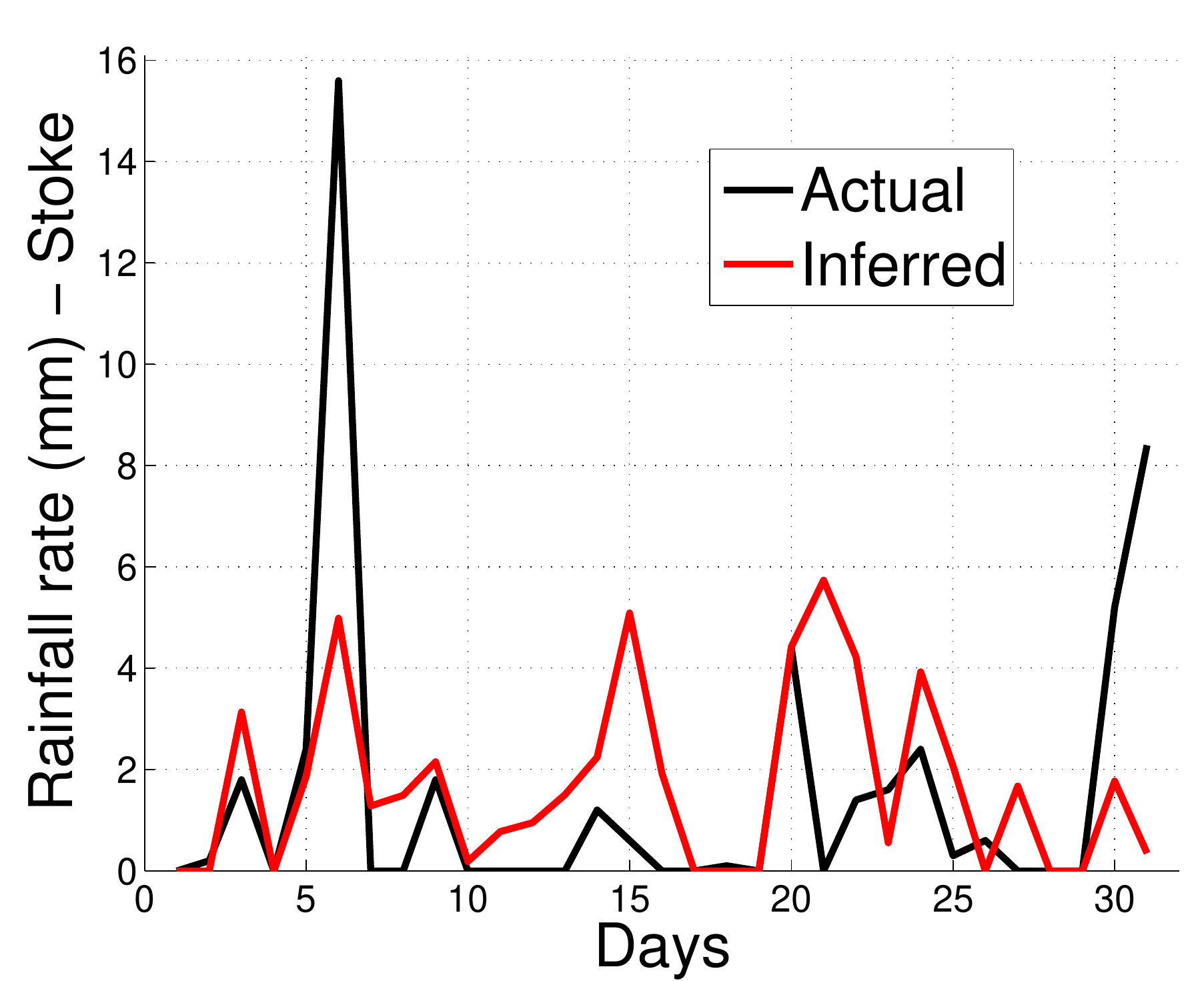}
    \label{fig_rain1gramsS}}
    \end{center}
    \caption{Feature Class $U$ -- Inference for Rainfall case study (Round 5 of 6-fold cross validation)}
    \label{fig_rain1grams}
\end{figure*}

\begin{figure*}[tp]
    \begin{center}
    \subfigure[\emph{Bristol} -- RMSE: 1.59] {\includegraphics[width=2.75in]{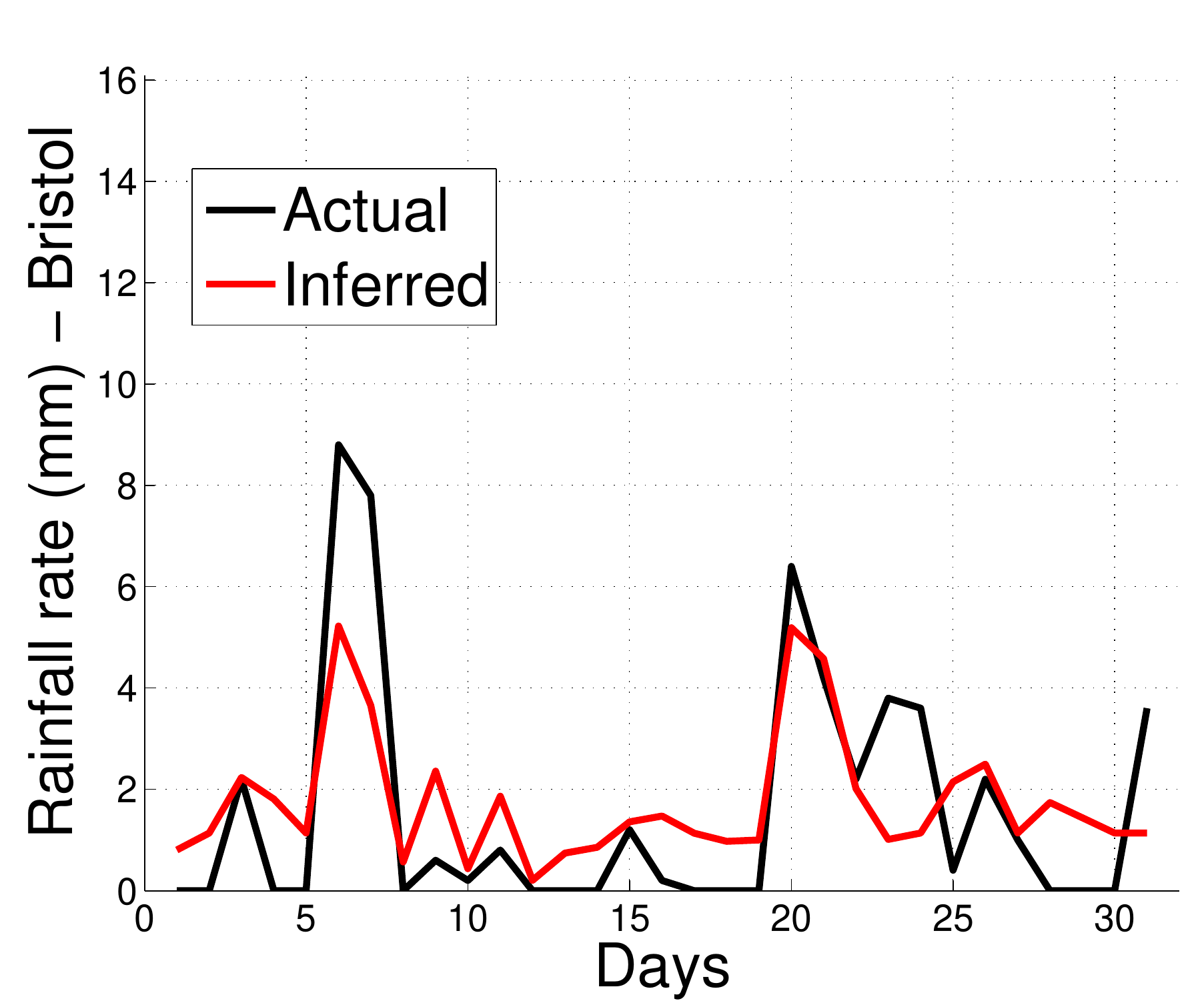}
    \label{fig_rain2gramsB}}
    \hfil
    \subfigure[\emph{London} -- RMSE: 1.478] {\includegraphics[width=2.75in]{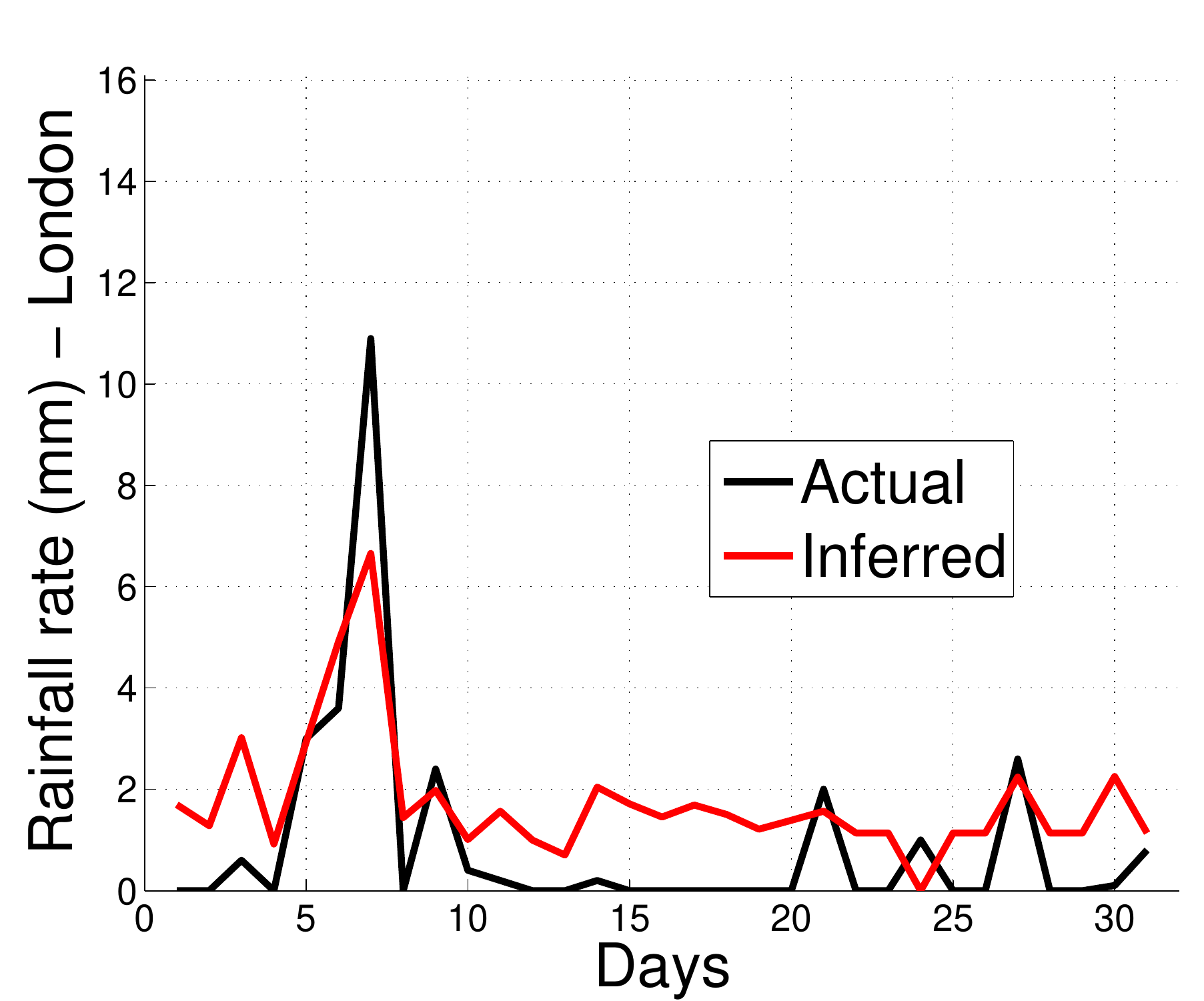}
    \label{fig_rain2gramsL}}
    \hfil
    \subfigure[\emph{Middlesbrough} -- RMSE: 2.918] {\includegraphics[width=2.75in]{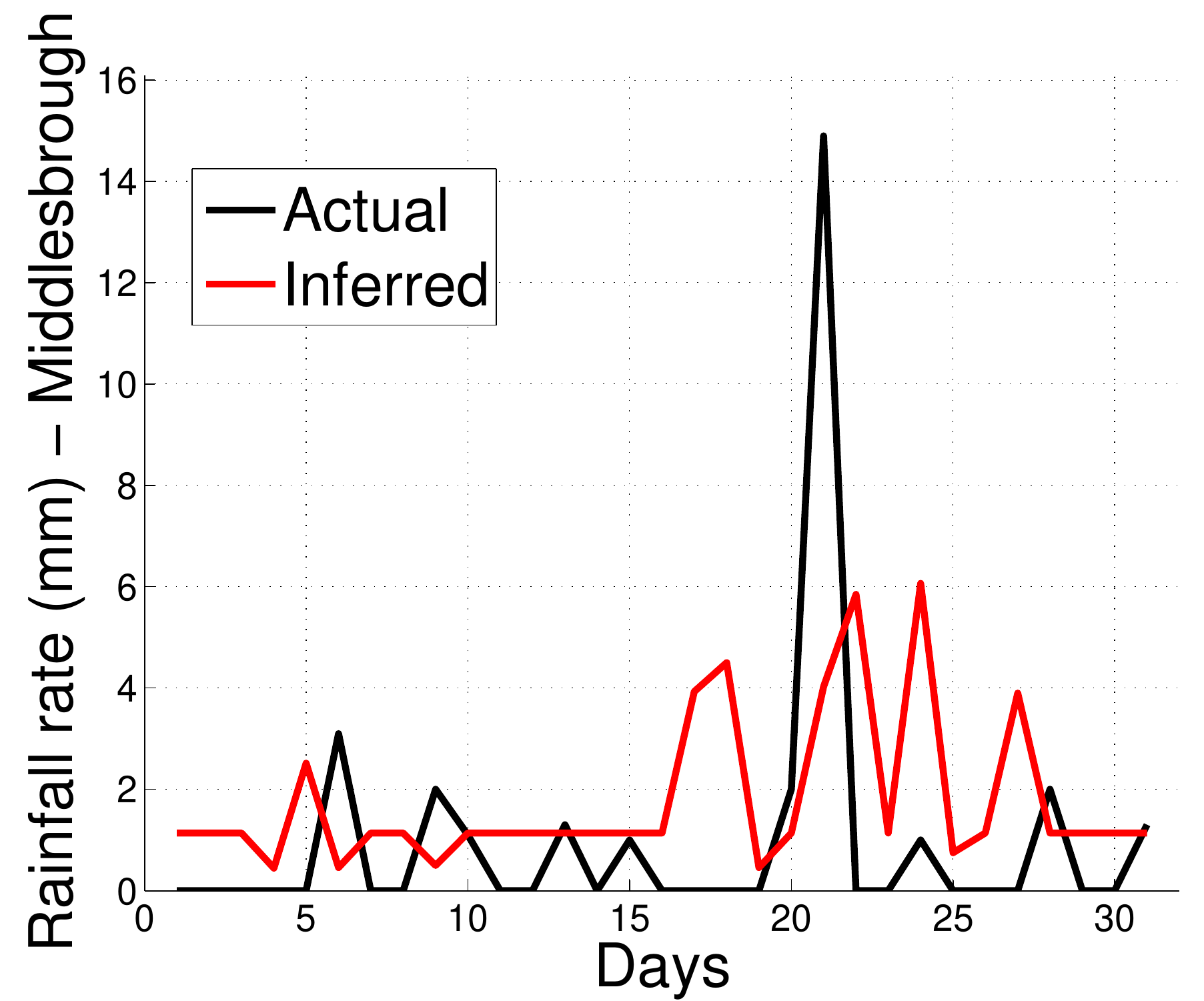}
    \label{fig_rain2gramsM}}
    \subfigure[\emph{Reading} -- RMSE: 1.572] {\includegraphics[width=2.75in]{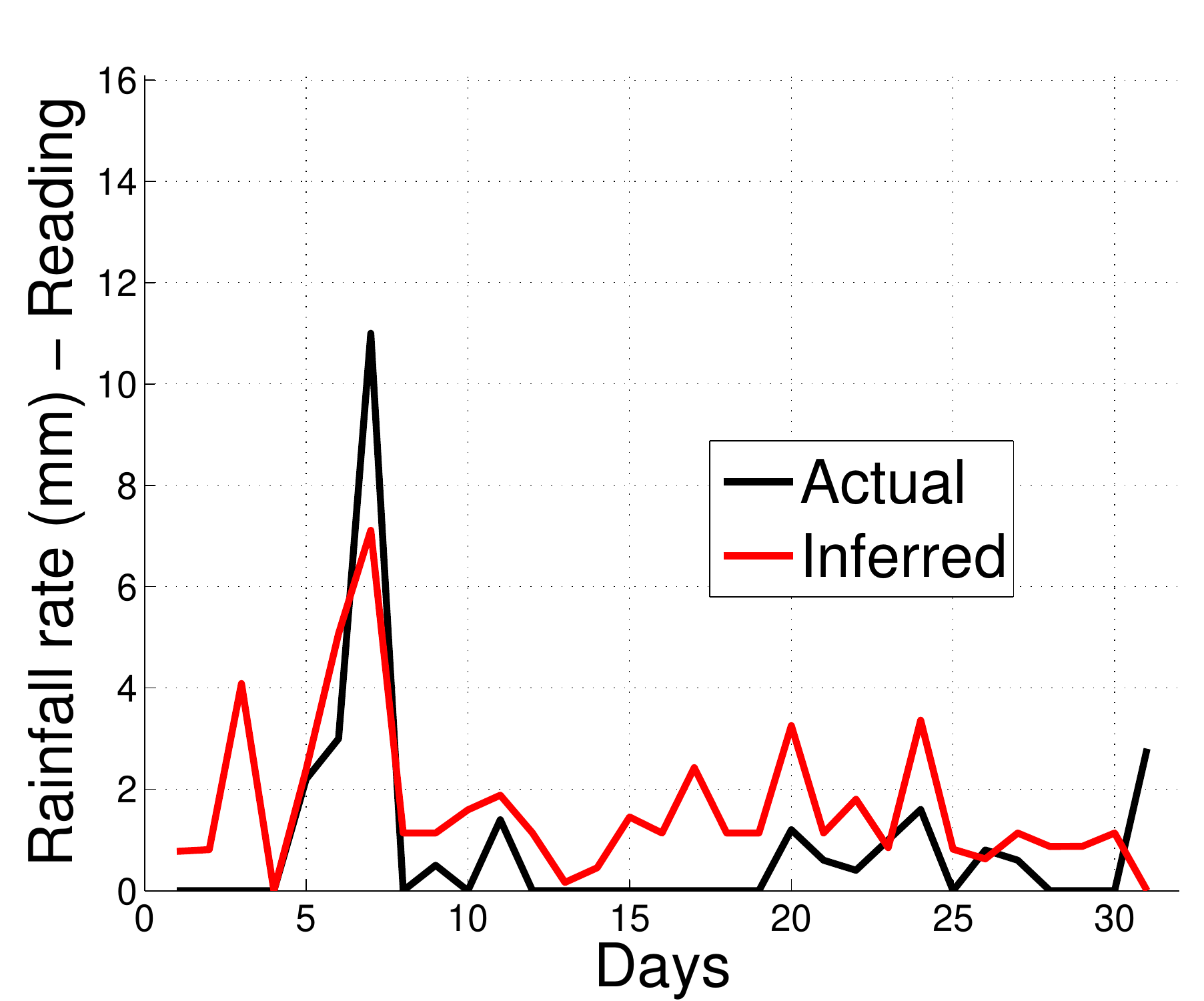}
    \label{fig_rain2gramsR}}
    \hfil
    \subfigure[\emph{Stoke-on-Trent} -- RMSE: 2.447] {\includegraphics[width=2.75in]{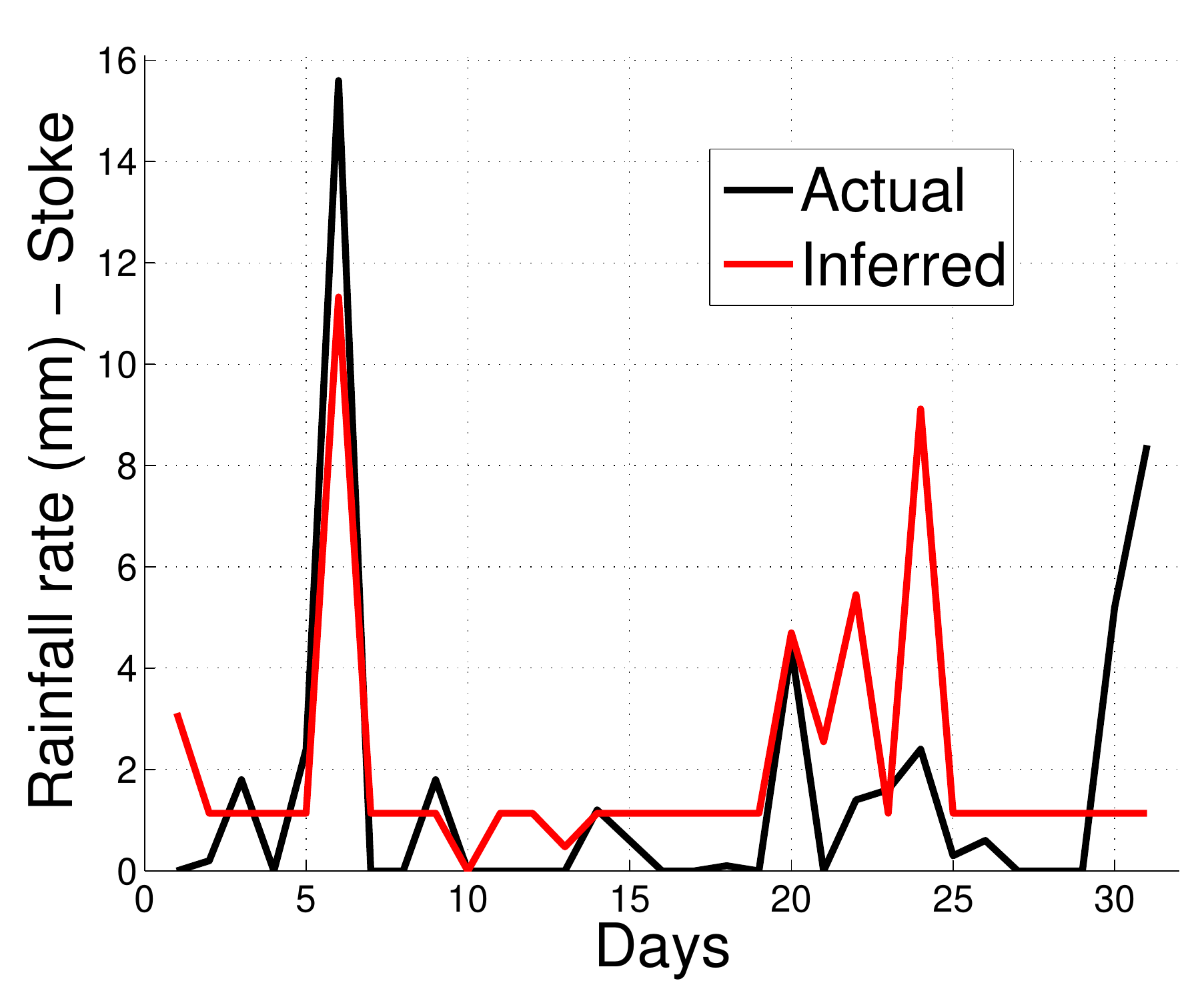}
    \label{fig_rain2gramsS}}
    \end{center}
    \caption{Feature Class $B$ -- Inference for Rainfall case study (Round 5 of 6-fold cross validation)}
    \label{fig_rain2grams}
\end{figure*}

\begin{figure*}[tp]
    \begin{center}
    \subfigure[\emph{Bristol} -- RMSE: 1.579] {\includegraphics[width=2.75in]{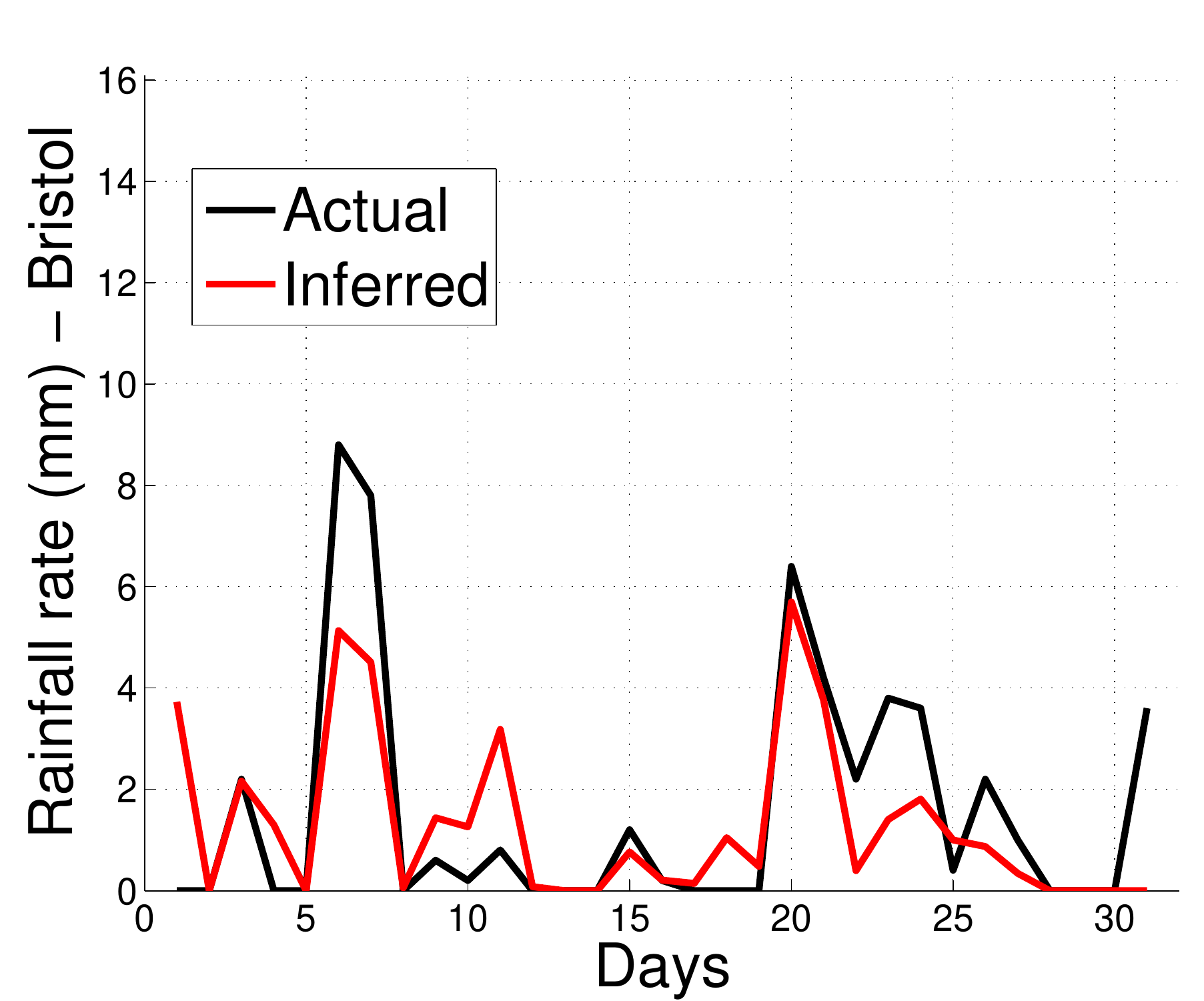}
    \label{fig_rainhybridB}}
    \hfil
    \subfigure[\emph{London} -- RMSE: 1.399] {\includegraphics[width=2.75in]{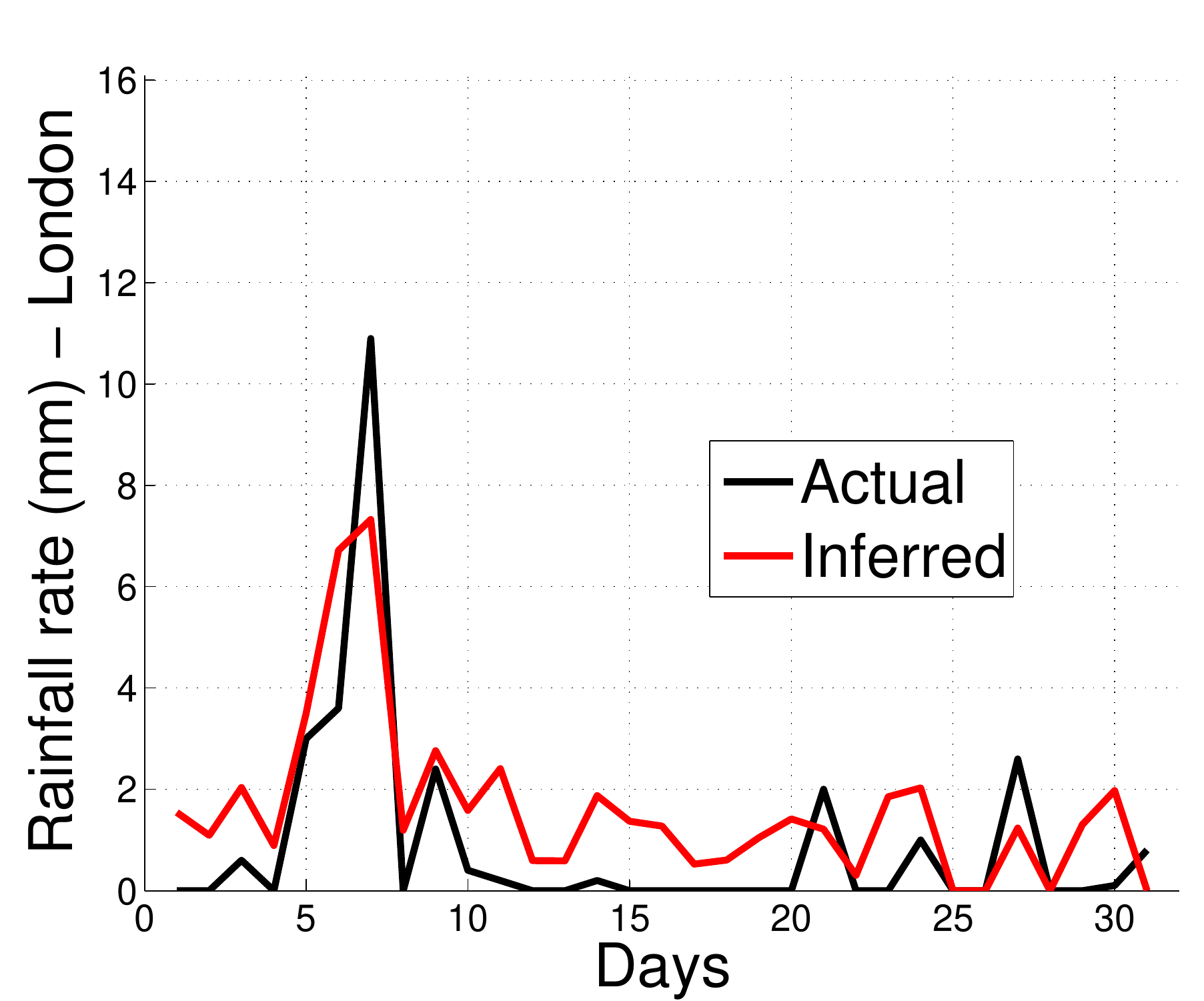}
    \label{fig_rainhybridL}}
    \hfil
    \subfigure[\emph{Middlesbrough} -- RMSE: 2.785] {\includegraphics[width=2.75in]{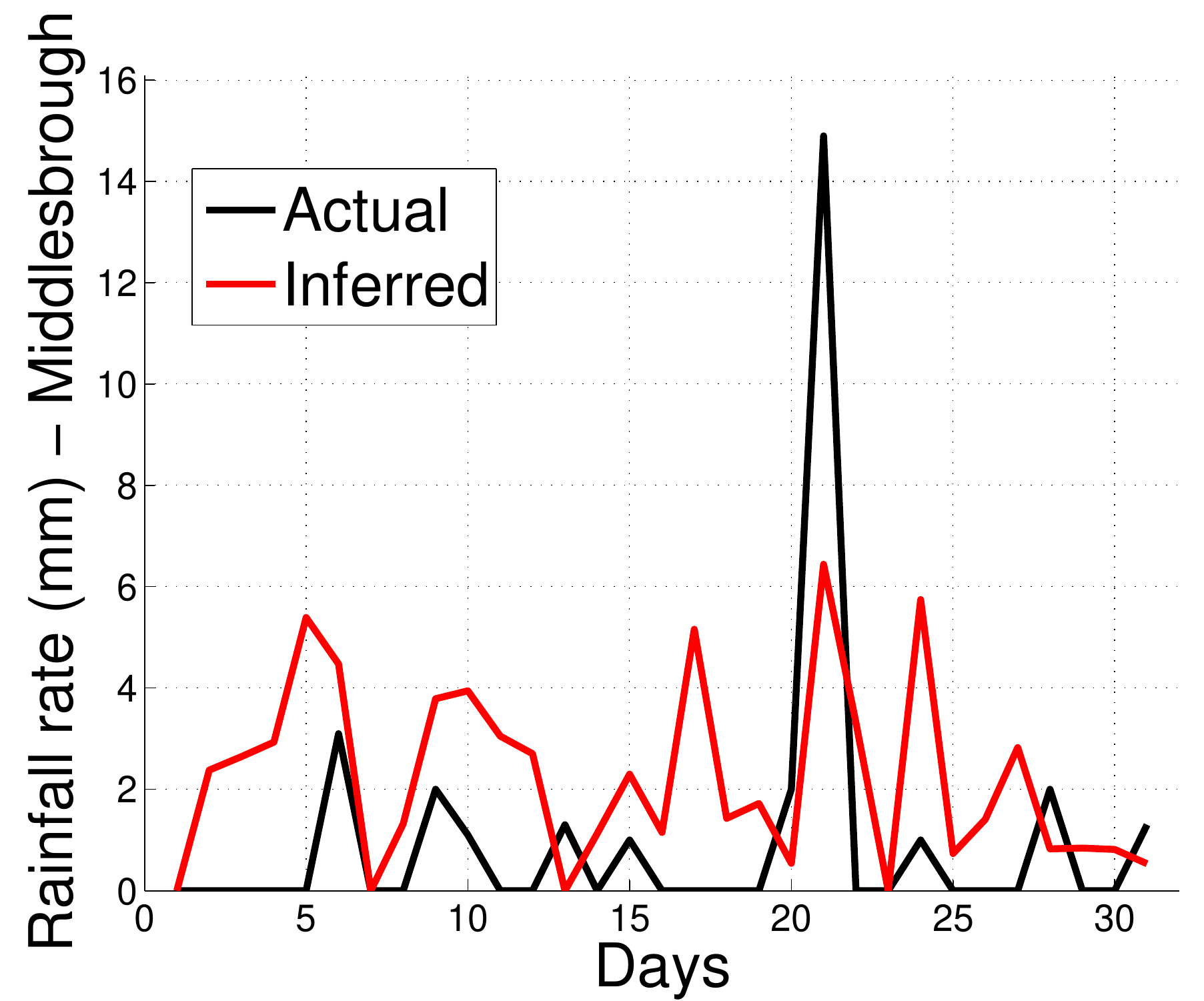}
    \label{fig_rainhybridM}}
    \subfigure[\emph{Reading} -- RMSE: 1.695] {\includegraphics[width=2.75in]{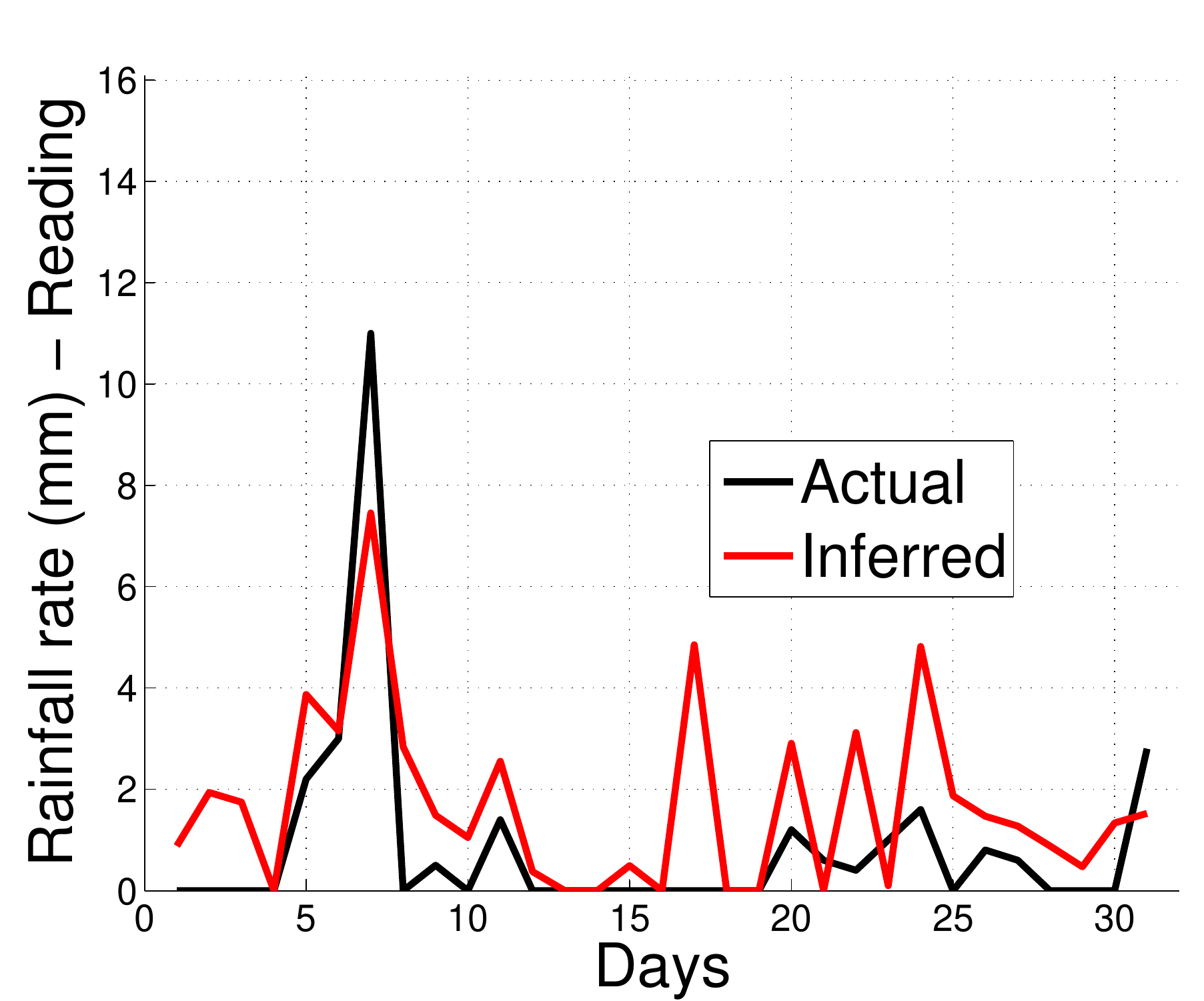}
    \label{fig_rainhybridR}}
    \hfil
    \subfigure[\emph{Stoke-on-Trent} -- RMSE: 2.815] {\includegraphics[width=2.75in]{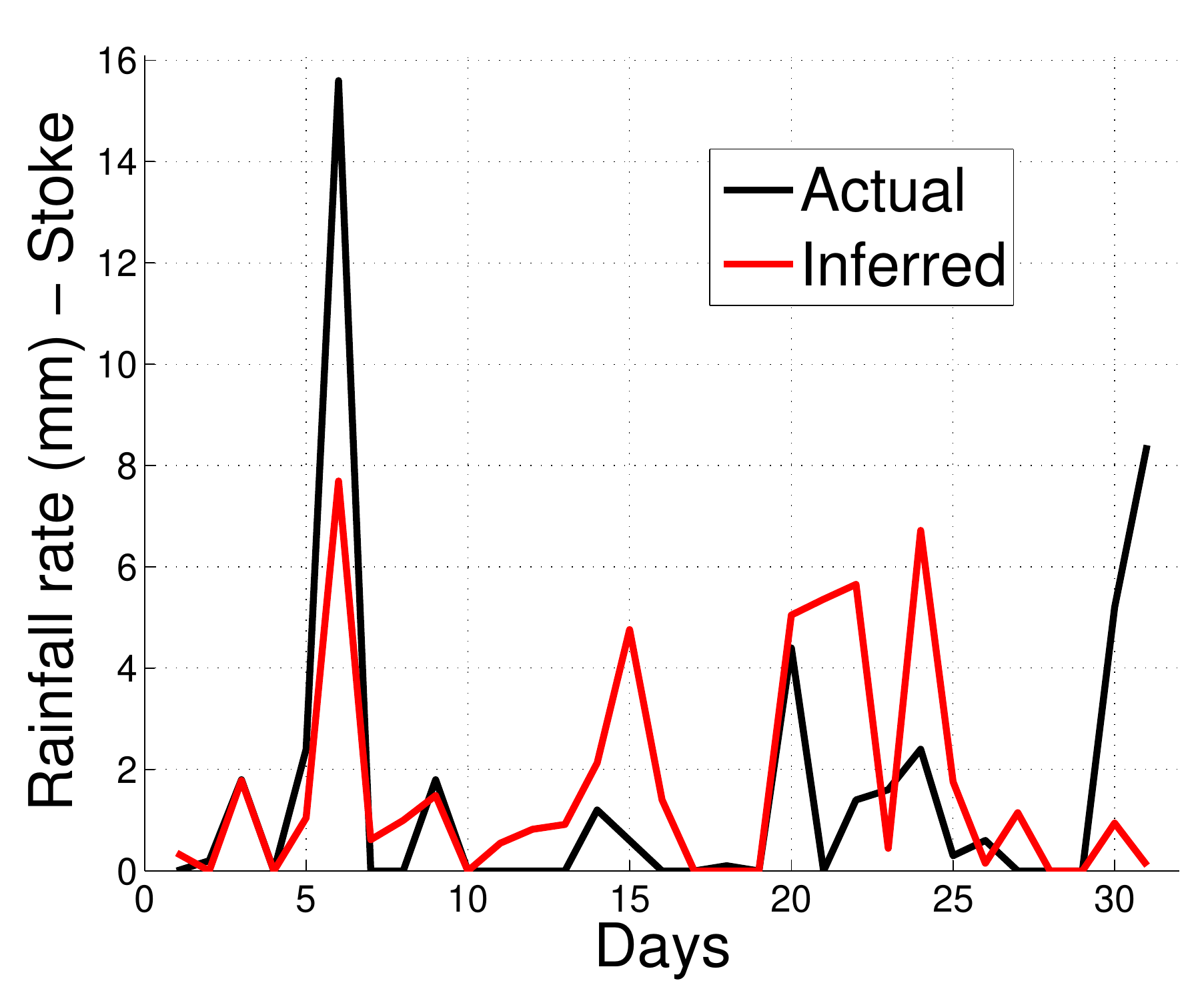}
    \label{fig_rainhybridS}}
    \end{center}
    \caption{Feature Class $H$ -- Inference for Rainfall case study (Round 5 of 6-fold cross validation)}
    \label{fig_rainhybrid}
\end{figure*}

Inference results per location for cross validation's round 5 are presented in Figures \ref{fig_rain1grams}, \ref{fig_rain2grams} and \ref{fig_rainhybrid} for $U$, $B$, and $H$ feature classes respectively. Overall, inferences follow the pattern of actual rain; for feature class $B$, we see that inferences in some occasions appear to have a positive lower bound (see Figure \ref{fig_rain2gramsS}), which is actually the positive bias term of OLS regression appearing when the selected markers have zero frequencies in the daily Twitter corpus of a location. As mentioned before, this problem is resolved in $H$ since it is very unlikely for 1-grams to also have a zero frequency. Results for class $H$, depicted in Figures \ref{fig_rainhybridB} (\emph{Bristol}), \ref{fig_rainhybridL} (\emph{London}) and \ref{fig_rainhybridR} (\emph{Reading}), demonstrate a good fit with the target signal.

%-- in the results per location, feature class $H$ has the best performance 11 times (out of 30), the same holds for $B$, and $U$ is better 8 times.
%MOVE THIS TO DISCUSSION ???
%However, the overall performance for class $U$ is better than $B$'s, something explained by the fact that smaller samples of tweets for non highly populated cities such as \emph{Stoke-on-Trent} or instabilities on Twitter service (which reduce the number of retrieved tweets for all locations) have resulted to zero daily counts of some 2-grams giving out an unstable (compared to 1-grams) behaviour in terms of inference. This is resolved in feature class $H$.

%%%%%%%%%%%%%%%%%%%%%%%%%%%%%%%%%%%%%%%%%%%%%%%%%%%%%%%%%%%%%%%%%%%%%%%%%%%%%%%%%
%%%%%%%%%%%%%%%%%%%%%%%%%%%%%%%%%%%%%%%%%%%%%%%%%%%%%%%%%%%%%%%%%%%%%%%%%%%%%%%%%
\section{Case study II: nowcasting flu rates from Twitter}
\label{section_nowcasting_flu}
In the second case study, we use the content of Twitter to infer regional flu rates\index{flu rates} in the UK. We base our inferences in three UK regions, namely \emph{Central England} \& \emph{Wales}, \emph{North England} and \emph{South England}. Ground truth, \emph{i.e.} official flu rate measurements, is derived from HPA\index{HPA}. HPA's weekly reports are based on information collected from the RCGP\index{RCGP} and express the number of GP\index{GP} consultations per 100,000 citizens, where the result of the diagnosis was ILI. According to HPA, a flu rate less than or equal to 30 is considered as baseline, flu rates below 100 are normal, between 100 and 200 are above average and over 200 are characterised as exceptional.\footnote{ ``Interpreting the HPA Weekly National Influenza Report'', July 2009 -- \url{http://goo.gl/GWZmB}.} To create a daily representation of HPA's weekly reports we use linear interpolation between the weekly rates. Given the flu rates $r_i$ and $r_{i+1}$ of two consecutive weeks, we compute a step factor $\delta$ from
\begin{equation}
\displaystyle\delta = \frac{r_{i+1} - r_i}{7},
\end{equation}
and then produce flu rates for the days in between using the equation
\begin{equation}
d_j = d_{j-1} + \delta\mbox{, }j\in\{2,...,7\},
\end{equation}
where $d_j$ denotes the flu rate of week's day $j$ and $d_1 = r_i$. The main assumption here is that a regional flu rate will be monotonically increasing or decreasing within the duration of a week.

Candidate features are extracted from several web references, such as the web sites of National Health Service, BBC, Wikipedia and so on, following the general principle, \emph{i.e.} including encyclopedic, scientific and more informal input. Similarly to the previous case study, extracted 1-grams are reduced from 2428 to 2044, and extracted 2-grams from 7589 to 1678. Here, we have removed n-grams with a count $\leq$ 50 since the number of tweets involved is approximately 5 times larger compared to the rainfall case study.\footnote{ Complete lists of the web references and the extracted markers for the flu case study are available at \url{http://geopatterns.enm.bris.ac.uk/twitter/influenza-cf.php}.}

In the flu case study, not enough peaks are present in the ground truth time series, as the collected Twitter data cover only one flu period with above average rates (Swine Flu epidemic in June-July 2009). During performance evaluation, this results in training mostly on non-flu periods where there is no strong flu signal -- hence, feature selection under those conditions is not optimal. To overcome this and assess properly the proposed methodology, we perform a random permutation of all data points based on their day index. The randomly permuted result for \emph{South England}'s flu rate is shown on Figure \ref{fig_gtFluSEngRand} -- we apply the same randomised index on all regions during performance evaluation.

\subsection{Experimental Settings}
For this experiment, we considered tweets and ground truth in the time period between the 21$^{st}$ of June, 2009 and 19$^{th}$ of April, 2010 (303 days). The total number of tweets used reaches approximately 50 million. Similarly to the previous case study, we are applying 5-fold cross validation using data of 60 or 61 days per fold. In each round of the cross-validation, 4 folds are used for training. From the remaining fold, 30 days of data are used for validating CT and the rest for testing. Bolasso settings are identical to the ones used in the rainfall case.

Notice that in the following experiments data points are not contiguous in terms of time since they have been permuted randomly based on their day index (as explained in the previous section). However, we have included an example at the end of the next section, where contiguous (time-wise) training, validating and testing data points have been used.

\subsection{Results}
The derived CTs and numbers of selected features for all rounds of the 5-fold cross validation are presented on Table \ref{table_flu_ct}. In the flu case study, most CTs (especially in $U$ and $H$ feature classes) get a value closer to the lower bound (50\%) after validation, and on average more features (compared to the rainfall case) are being selected. This is due to either the existence of only one significant flu period in the ground truth data or the general inadequacy of 1-grams to describe the underlying topic as effectively as in the previous case study.

\begin{table}[tp]
%\footnotesize
\caption{Nowcasting Flu Rates -- Derived Consensus Thresholds and numbers of selected features (in parentheses) for all Feature Classes (FC) in the rounds of 5-fold cross validation -- Fold $i$ denotes the validation/testing fold of round $6-i$}
\label{table_flu_ct}
\renewcommand{\arraystretch}{1.1}
\setlength\tabcolsep{1mm}
\scriptsize
\centering
\(\begin{tabular}{c|ccccc}
\textbf{FC} & \textbf{Fold 5}  & \textbf{Fold 4}  & \textbf{Fold 3}  & \textbf{Fold 2}  & \textbf{Fold 1}  \\\hline
$U$  & 52.5\% (\emph{90}) & 52.5\% (\emph{100}) & 52.5\% (\emph{108}) & 62.5\% (\emph{67}) & 50\% (\emph{62})  \\
$B$  & 55\% (\emph{42})   & 62.5\% (\emph{47})  & 92.5\% (\emph{14})  & 85\% (\emph{10})   & 52.5\% (\emph{36})\\
$H$  & 55\% (\emph{124})  & 62.5\% (\emph{131}) & 52.5\% (\emph{151}) & 60\% (\emph{103})  & 50\% (\emph{100})\\
$H_{\text{II}}$ & 82.5/55\% (\emph{82}) & 57.5/62.5\% (\emph{119}) & 72.5/92.5\% (\emph{50}) & 67.5/60\% (\emph{105}) & 50/80\% (\emph{81})\\
$UB$            & 90\% (\emph{26}) & 85\% (\emph{33}) & 60\% (\emph{29}) & 100\% (\emph{11}) & 57.5\% (\emph{128})
\end{tabular}\)
\end{table}

\begin{table}[tp]
\caption{Nowcasting Flu Rates -- RMSEs for all Feature Classes (FC) and locations in the rounds of 5-fold cross validation -- Fold $i$ denotes the validation/testing fold of round $6-i$. The last column holds the RMSEs of the baseline method.}
\label{table_flu_results_bolasso}
%\footnotesize
\renewcommand{\arraystretch}{1.1}
\setlength\tabcolsep{1mm}
\newcolumntype{C}{>{\centering\arraybackslash} m{1.2cm} }
\newcolumntype{V}{>{\centering\arraybackslash} m{1.5cm} }
\centering
\(\begin{tabular}{c|c|*{5}{C}|V||V}
\textbf{Region} & \textbf{FC} & \textbf{Fold 5}  & \textbf{Fold 4}  & \textbf{Fold 3}  & \textbf{Fold 2}  & \textbf{Fold 1}  & \textbf{Mean RMSE} & \textbf{BS-Mean RMSE} \\\hline\hline
\emph{Central England} & $U$ & 11.781   & 9.005  & 16.147    & 13.252    & 10.912    & 12.219 & 12.677\\
\& \emph{Wales}        & $B$ & 11.901   & 12.2   & 21.977    & 12.426    & 14.615    & 14.624 & 15.665\\
                       & $H$ & 8.36     & 8.826  & 14.618    & 12.312    & 12.62     & 11.347 & 11.691\\\hline
\emph{North England}   & $U$ & 9.757    & 6.708  & 9.092     & 13.117    & 8.489     & 9.432  & 10.511\\
                       & $B$ & 9.659    & 9.969  & 10.716    & 12.057    & 8.699     & 10.22  & 12.299\\
                       & $H$ & 9.782    & 7.112  & 6.65      & 13.694    & 7.607     & 8.969  & 9.752\\\hline
\emph{South England}   & $U$ & 9.599    & 8.285  & 13.656    & 14.673    & 11.061    & 11.455 & 13.617\\
                       & $B$ & 13.536   & 9.209  & 16.188    & 14.279    & 8.531     & 12.348 & 12.977\\
                       & $H$ & 9.86     & 7.881  & 13.448    & 14.34     & 8.872     & 10.88  & 12.768\\\hline\hline
\textbf{Total RMSE}    & $U$ & 10.426   & 8.056  & 13.29     & 13.699    & 10.222    & \textbf{11.139} & 12.438\\
                       & $B$ & 11.806   & 10.536 & 16.93     & 12.958    & 10.986    & \textbf{12.643} & 13.815\\
                       & $H$                & 9.359    & 7.971  & 12.094    & 13.475    & 9.93      & \textbf{10.566} & 11.617\\
                       & $H_{\text{II}}$    & 10.237   & 7.699  & 12.918    & 13.417    & 9.963     & \textbf{10.847} &--\\
                       & $UB$               & 9.121    & 8.176  & 14.371    & 14.729    & 31.168    & \textbf{15.513} &--
\end{tabular}\)
\end{table}

Table \ref{table_flu_results_bolasso} holds the performance results for all rounds of the cross validation. For a more comprehensive interpretation of the numerical values consider that the average ILI rate across the regions used in our experiments is equal to 26.659 with a standard deviation of 29.270 and ranges in [2,172]. Again, feature class $U$ performs better than $B$, whereas $H$ outperforms all others. However, in this case study, feature class $H_{\text{II}}$ is the second best in terms of performance. Similarly to the previous case study, our method improves on the performance of the baseline approach by a factor of 9.05\%.
%In the regional results per fold, class $H$ has the best performance 8 times, $B$ 5 times and $U$ only 2 times (out of 15 subcases in total).

% selected features for fold 1
\begin{table}
\caption{Feature Class $U$ -- 1-grams selected by Bolasso for Flu case study (Round 1 of 5-fold cross validation) -- All weights (\textbf{w}) should be multiplied by $10^4$}
\label{table_flu_1grams}
\footnotesize
\renewcommand{\arraystretch}{1.1}
\setlength\tabcolsep{1mm}
\centering
\(\begin{tabular}{cc|cc|cc|cc|cc}
\textbf{1-gram} & \textbf{w} & \textbf{1-gram}  & \textbf{w} & \textbf{1-gram} & \textbf{w} & \textbf{1-gram} & \textbf{w} & \textbf{1-gram} & \textbf{w}\\\hline
acut        & -1.034 & cleav   & 0.735  & hippocr           & -6.249 & properti   & -0.66  & speed    & -0.286\\
afford      & -0.181 & complex & -0.499 & holidai           & -0.017 & psycholog  & -1.103 & spike    & 0.145 \\
allergi     & -2.569 & cough   & 0.216  & huge              & -0.33  & public     & 0.212  & stage    & 0.109 \\
approv      & -0.672 & cruis   & -1.105 & \textbf{\emph{irrig}}    & \textbf{10.116} & radar      & 0.284  & strength & 0.873 \\
artifici    & 2.036  & daughter& 0.187  & item     & -0.337 & reach      & 0.247  & strong   & 0.336 \\
assembl     & 0.589  & dilut   & 4.165  & knock    & 0.261  & reliev     & -0.254 & swine    & 1.262 \\
asthmat     & 4.526  & drag    & 0.098  & lethal   & -0.73  & remain     & -0.755 & tast     & 0.13  \\
attempt     & 0.375  & erad    & 0.201  & major    & -0.367 & rough      & 0.068  & team     & -0.031\\
behavior    & -1.747 & face    & -0.008 & medic    & 1.06   & run        & 0.242  & throat   & 0.07  \\
better      & 0.066  & fellow  & 0.542  & member   & 0.354  & rush       & -0.159 & tissu    & 0.533 \\
bind        & 0.675  & fluid   & 2.002  & mercuri  & -0.588 & scari      & 0.198  & transmit & 1.352 \\
blood       & 0.059  & fuss    & 0.575  & metro    & -0.397 & seal       & -0.161 & troop    & 0.532 \\
boni        & 1.308  & germ    & 0.211  & mile     & -0.081 & season     & -0.103 & typic    & 0.585 \\
bulg        & -0.966 & guilti  & -0.608 & miss     & 0.071  & seizur     & 2.448  & underli  & 0.774 \\
caution     & 2.578  & habit   & 0.619  & nurs     & 0.223  & self       & 0.127  & unquot   & 8.901 \\
cellular    & -2.125 & halt    & 1.472  & perform  & 0.084  & sik        & -0.634 & upcom    & 0.642 \\
checklist   & -1.494 & harbour & -0.472 & personnel& -1.451 & site       & 0.042  & wave     & 0.042 \\
chicken     & 0.317  & health  & -0.241 & pictur   & -0.134 & soak       & 0.413  & wikipedia& 0.824
\end{tabular}\)
\end{table}

% selected features for fold 1
\begin{table}[tp]
\caption{Feature Class $B$ -- 2-grams selected by Bolasso for Flu case study (Round 1 of 5-fold cross validation) -- All weights (\textbf{w}) should be multiplied by $10^4$}
\label{table_flu_2grams}
\footnotesize
\renewcommand{\arraystretch}{1.1}
\setlength\tabcolsep{0.5mm}
\centering
\(\begin{tabular}{cc|cc|cc|cc}
\textbf{2-gram} & \textbf{w} & \textbf{2-gram}  & \textbf{w} & \textbf{2-gram} & \textbf{w} & \textbf{2-gram} & \textbf{w}\\\hline
case swine      & 12.783 & flu bad         & 6.641   & need take       & 0.887   & talk friend     & -4.9         \\
check code      & 6.27   & flu jab         & 4.66    & pain night      & 14.149  & time knock      & 10.002           \\
check site      & 0.568  & flu relat       & 10.948  & physic emotion  & 7.95    & total cost      & -11.582           \\
\textbf{\emph{confirm swine}}   & \textbf{31.509} & flu symptom     & 7.693   & sleep well      & 1.319   & underli health  & 25.535            \\
cough fit       & 7.381  & flu web         & -8.017  & sore head       & 4.297   & virus epidem    & -28.204          \\
cough lung      & 7.974  & ground take     & -15.208 & spread viru     & 20.871  & visit doctor    & -12.327            \\
cough night     & 16.73  & health care     & -0.636  & stai indoor     & 5.482   & weight loss     & -0.447         \\
die swine       & 9.722  & healthcar worker& 3.876   & suspect swine   & 3.863   & woke sweat      & -33.133          \\
effect swine    & 27.675 & home wors       & 22.167  & swine flu       & 1.153   & wonder swine  & 11.5085           \\
feel better     & 0.655  & ion channel     & 9.755   & symptom swine   & 5.895   &  &         \\
feel slightli   & 1.712  & kick ass        & -0.335  & take care       & 0.382   &  &
\end{tabular}\)
\end{table}

% selected features for fold 1 - second column has a wrong value (mistake spreads)
\begin{table}[tp]
\caption{Feature Class $H$ -- Hybrid selection of 1-grams and 2-grams for Flu case study (Round 1 of 5-fold cross validation) -- All weights (\textbf{w}) should be multiplied by $10^4$}
\label{table_flu_hybrid}
\footnotesize
\renewcommand{\arraystretch}{1.1}
\setlength\tabcolsep{0.5mm}
\centering
\(\begin{tabular}{cc|cc|cc|cc}
\textbf{\emph{n}-gram} & \textbf{w} & \textbf{\emph{n}-gram}  & \textbf{w} & \textbf{\emph{n}-gram} & \textbf{w} & \textbf{\emph{n}-gram} & \textbf{w}\\\hline
acut            & -0.796 & effect swine        & 19.835 & medic           & 0.48   & spike          & 0.032  \\
afford          & -0.106 & erad                & 0.27   & member          & 0.169  & spread viru    & 12.918 \\
allergi         & -2.332 & face                & 0.012  & mercuri         & -0.414 & stage          & 0.101  \\
approv          & -0.516 & feel better         & 0.15   & metro           & -0.365 & stai indoor    & 1.969  \\
artifici        & 1.319  & feel slightli       & 0.775  & mile            & -0.092 & strength       & 0.739 \\
assembl         & 0.231  & fellow              & 0.319  & miss            & 0.073  & strong         & 0.018 \\
asthmat         & 2.607  & flu bad             & 4.953  & need take       & 0.759  & suspect swine  & 2.503 \\
attempt         & 0.322  & flu jab             & -0.11  & nurs            & 0.118  & swine          & -0.203\\
behavior        & -1.349 & flu relat           & 3.183  & pain night      & 9.823  & swine flu      & 1.577\\
bind            & 0.437  & flu symptom         & 1.471  & perform         & 0.083  & symptom swine  & 1.626\\
blood           & 0.05   & flu web             & -5.463 & personnel       & -1.359 & take care      & 0.21\\
boni            & 0.984  & fluid               & 1.87   & physic emotion  & 6.192  & talk friend    & -2.518\\
bulg            & -0.733 & fuss                & 0.234  & pictur          & -0.124 & tast           & 0.08\\
case swine      & 4.282  & germ                & 0.111  & properti        & -0.372 & team           & -0.044\\
caution         & 1.174  & ground take         & -3.022 & radar           & 0.287  & throat         & 0.251\\
cellular        & -2.072 & guilti              & -0.394 & reach           & 0.201  & time knock     & 6.523\\
check code      & 4.495  & habit               & 0.381  & remain          & -0.666 & tissu          & -0.012\\
check site      & 0.149  & halt                & 0.819  & rough           & 0.075  & total cost     & -4.794\\
checklist       & -1.595 & health              & -0.04  & run             & 0.143  & transmit       & 1.535\\
chicken         & 0.286  & health care         & -0.393 & rush            & -0.07  & troop          & 0.767\\
cleav           & 0.991  & healthcar worker    & 1.339  & scari           & 0.109  & underli        & -0.221\\
\emph{\textbf{confirm swine}}   & \textbf{21.874} & hippocr             & -6.038 & seal            & -0.091 & underli health & 11.707\\
cough           & 0.234  & holidai             & -0.021 & season          & -0.064 & unquot         & 8.753\\
cough fit       & 2.395  & home wors           & 6.302  & seizur          & 2.987  & upcom          & 0.071\\
cough lung      & 2.406  & huge                & -0.199 & self            & 0.059  & viru epidem    & -8.805\\
cough night     & 6.748  & ion channel         & 4.974  & sik             & -0.542 & visit doctor   & -3.456\\
cruis           & -1.186 & irrig               & 8.721  & site            & 0.06   & wave           & 0.033\\
daughter        & 0.048  & item                & -0.219 & sleep well      & 0.753  & weight loss    & -0.296\\
die swine       & 0.196  & kick ass            & -0.15  & soak            & 0.41   & wikipedia      & 0.66\\
dilut           & 2.708  & knock               & 0.24   & sore head       & 2.023  & woke sweat     & -19.912\\
drag            & 0.147  & major               & -0.376 & speed           & -0.198 & wonder swine   & 7.266
\end{tabular}\)
\end{table}

\begin{figure*}[tp]
    \begin{center}
    \includegraphics[width=6in]{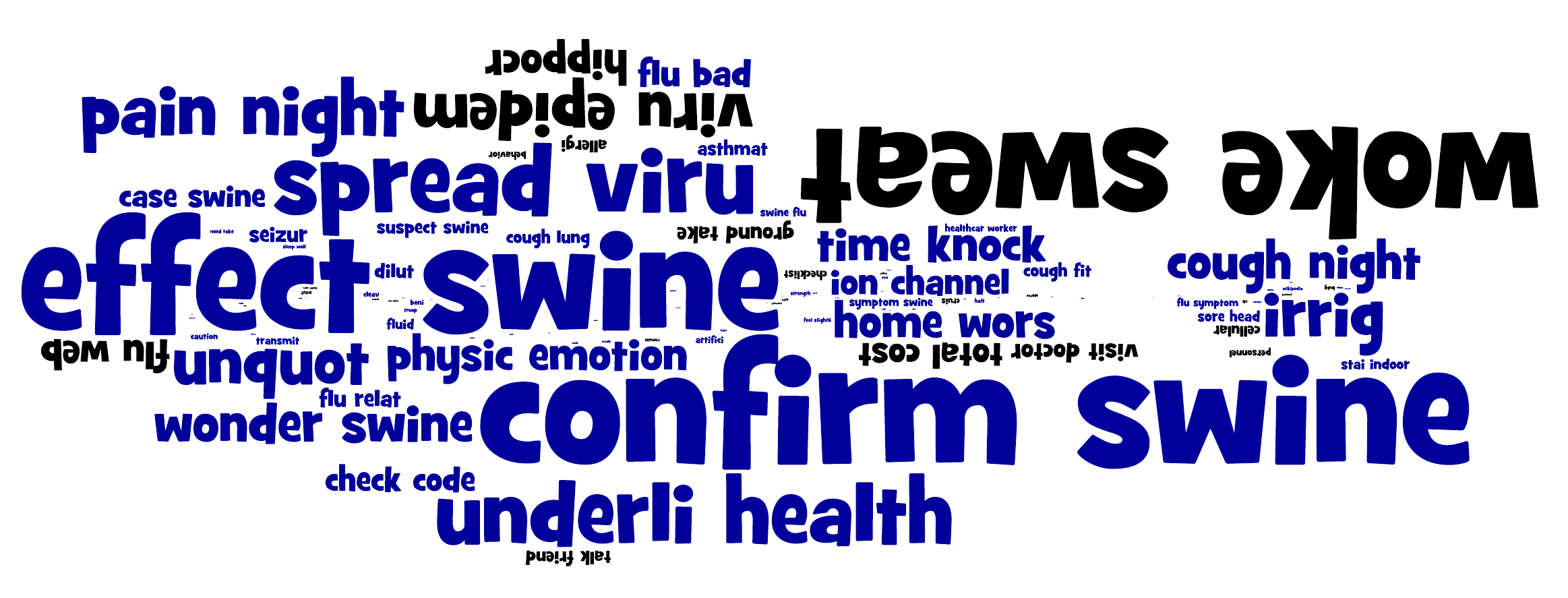}
    \end{center}
    \caption{Table \ref{table_flu_hybrid} in a word cloud, where font size is proportional to regression's weight and flipped words have negative weights}
    \label{fig_table_flu_hybrid_wordle}
\end{figure*}

For this case study, we present all intermediate results for cross validation's round 1. Tables \ref{table_flu_1grams}, \ref{table_flu_2grams} and \ref{table_flu_hybrid} show the selected features for $U$, $B$ and $H$ feature classes respectively. From the selected 1-grams (Table \ref{table_flu_1grams}), stem `\emph{irrig}'\footnote{\emph{Irrigation} describes the procedure of cleaning a wound or body organ by flushing or washing out with water or a medicated solution (WordNet).} has the largest weight. Many illness related markers have been selected such as `\emph{cough}', `\emph{health}', `\emph{medic}', `\emph{nurs}', `\emph{throat}' and so on, but there exist also words with no clear semantic relation. Surprisingly, stem `\emph{flu}' has not been selected as a feature in this round (and has only been selected in round 5). On the contrary, almost all selected 2-grams (Table \ref{table_flu_2grams}) can be considered as flu related -- `\emph{confirm swine}' has the largest weight for both feature classes $B$ and $H$ (Table \ref{table_flu_hybrid} and Figure \ref{fig_table_flu_hybrid_wordle}). As a general remark, keeping in mind that those features have been selected using data containing one significant flu period, they cannot be considered as very generic ones.
%The existence of at least one more flu peak in the ground truth signal might have removed \emph{swine flu} oriented markers, concentrating on more general descriptions for a flu-like illness.

% 1grams - iter 1 - randomised
\begin{figure*}[tp]
    \begin{center}
    \subfigure[\emph{C. England} \& \emph{Wales} -- RMSE: 11.781] {\includegraphics[width=2.75in]{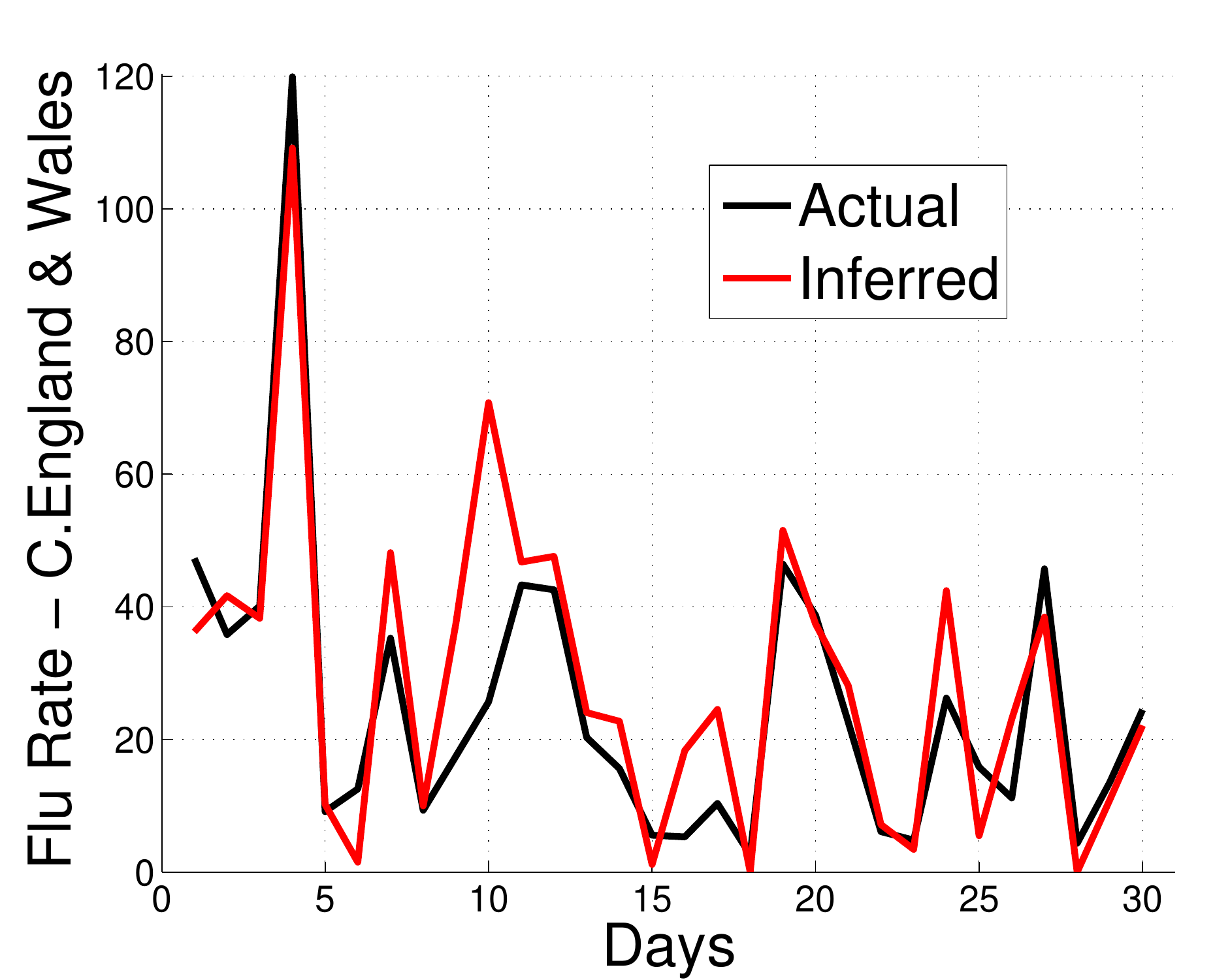}
    \label{fig_flu1gramsCengW}}
    \hfil
    \subfigure[\emph{N. England} -- RMSE: 9.757] {\includegraphics[width=2.75in]{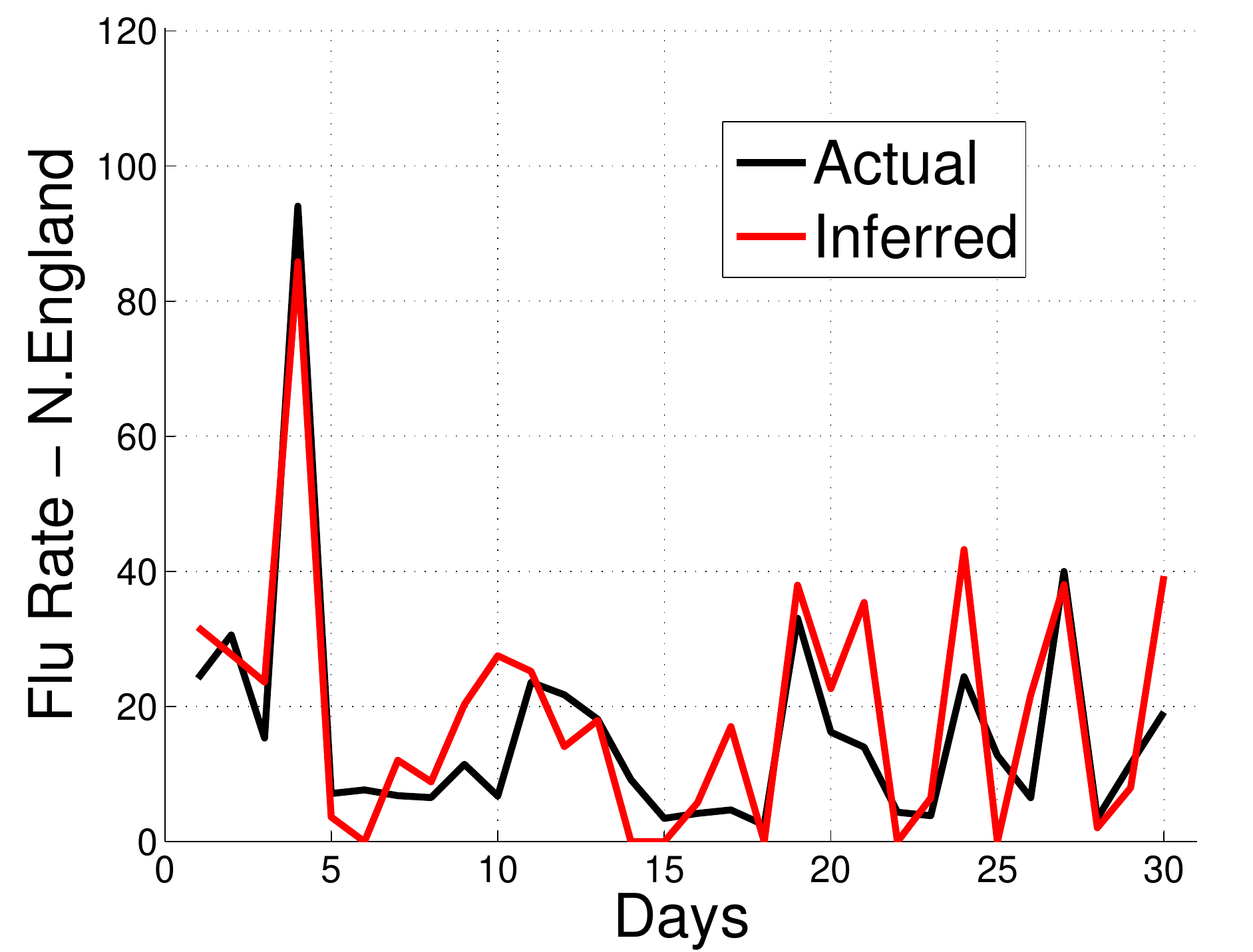}
    \label{fig_flu1gramsNEng}}
    \hfil
    \subfigure[\emph{S. England} -- RMSE: 9.599] {\includegraphics[width=2.75in]{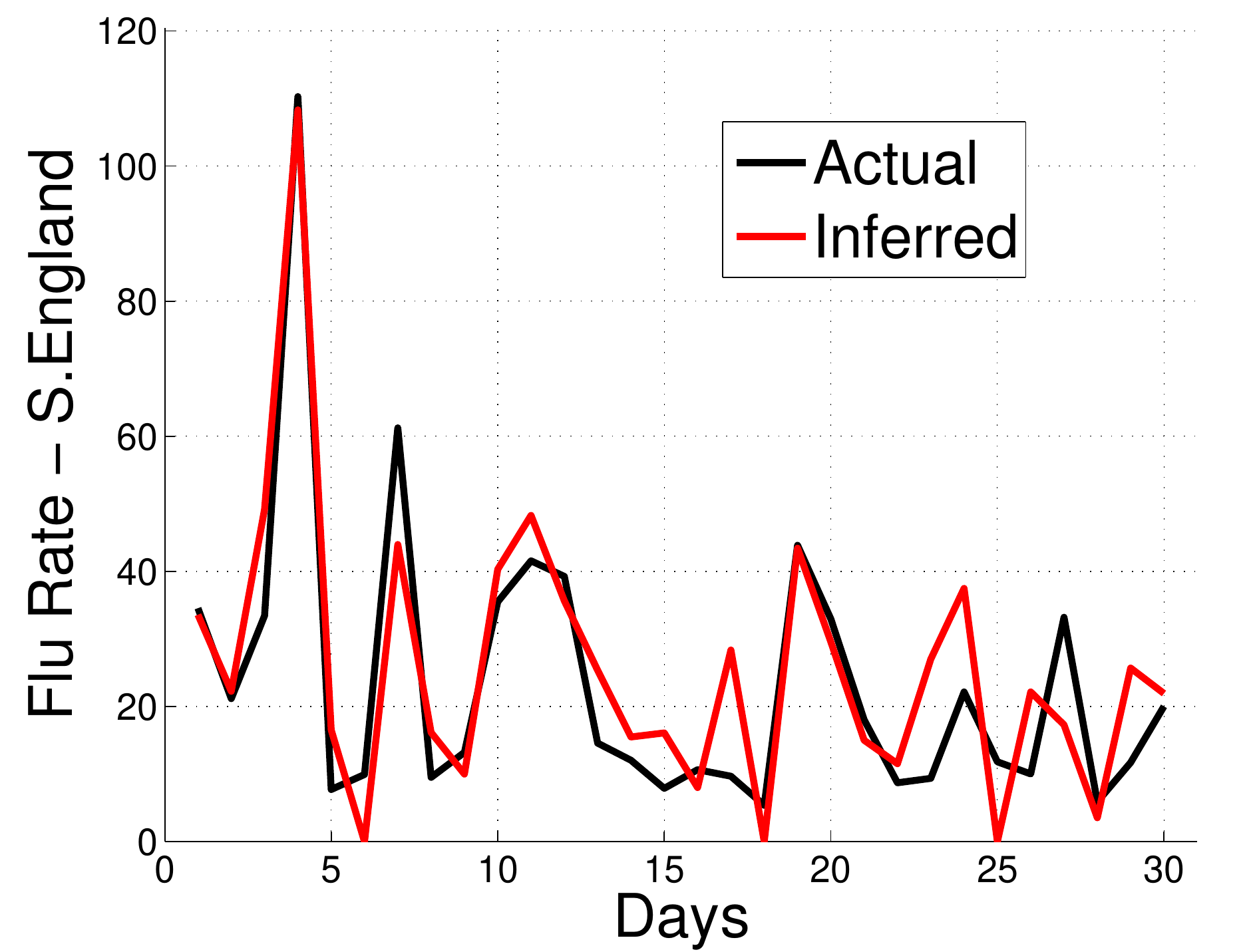}
    \label{fig_flu1gramsSEng}}
    \end{center}
    \caption{Feature Class $U$ -- Inference for Flu case study (Round 1 of 5-fold cross validation)}
    \label{fig_flu1grams}
\end{figure*}

% 2grams - iter 1 - randomised
\begin{figure*}[tp]
    \begin{center}
    \subfigure[\emph{C. England} \& \emph{Wales} -- RMSE: 11.901] {\includegraphics[width=2.75in]{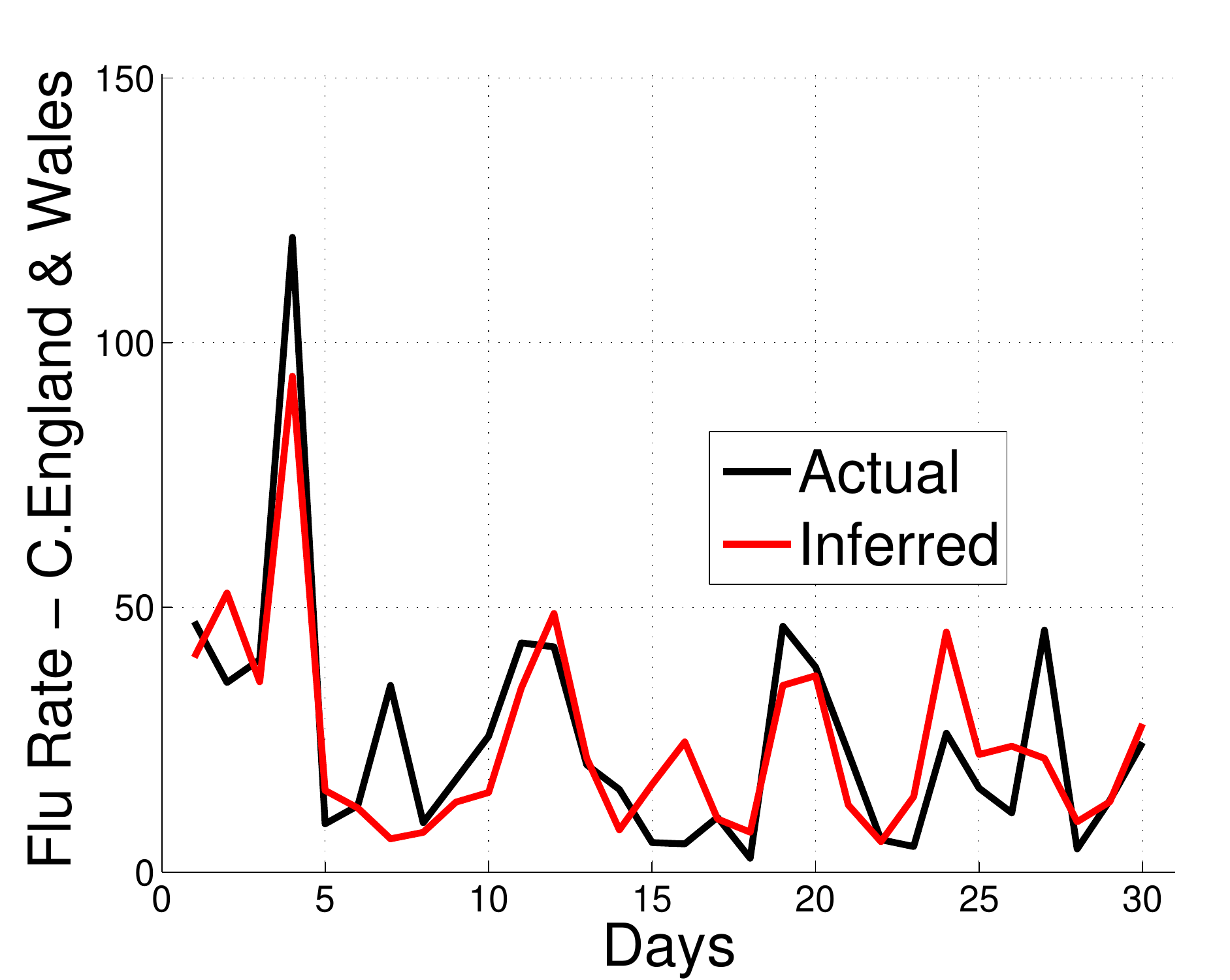}
    \label{fig_flu2gramsCengW}}
    \hfil
    \subfigure[\emph{N. England} -- RMSE: 9.659] {\includegraphics[width=2.75in]{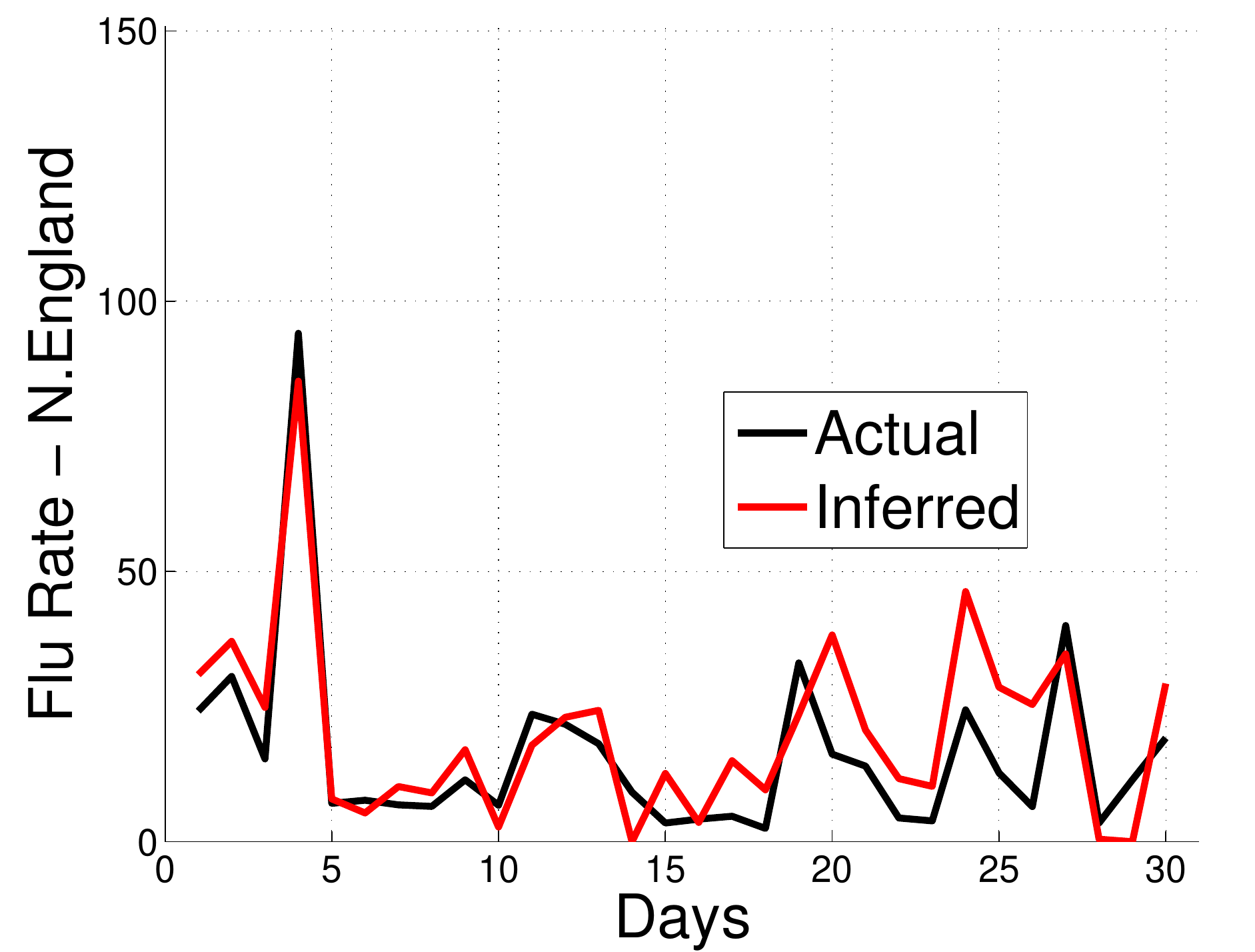}
    \label{fig_flu2gramsNEng}}
    \hfil
    \subfigure[\emph{S. England} -- RMSE: 13.536] {\includegraphics[width=2.75in]{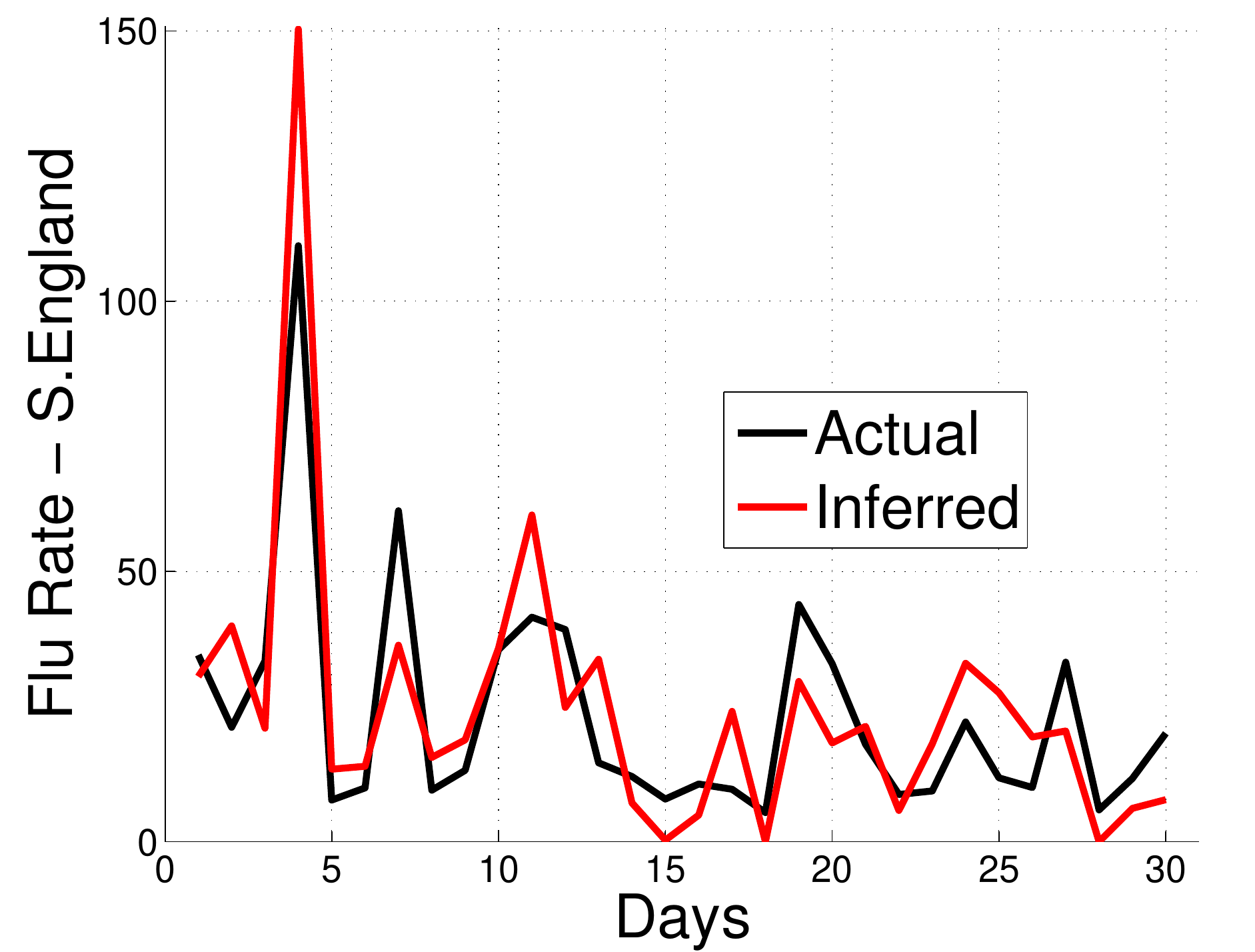}
    \label{fig_flu2gramsSEng}}
    \end{center}
    \caption{Feature Class $B$ -- Inference for Flu case study (Round 1 of 5-fold cross validation)}
    \label{fig_flu2grams}
\end{figure*}

% hybrid - iter 1 - randomised
\begin{figure*}[tp]
    \begin{center}
    \subfigure[\emph{C. England} \& \emph{Wales} -- RMSE: 8.36] {\includegraphics[width=2.75in]{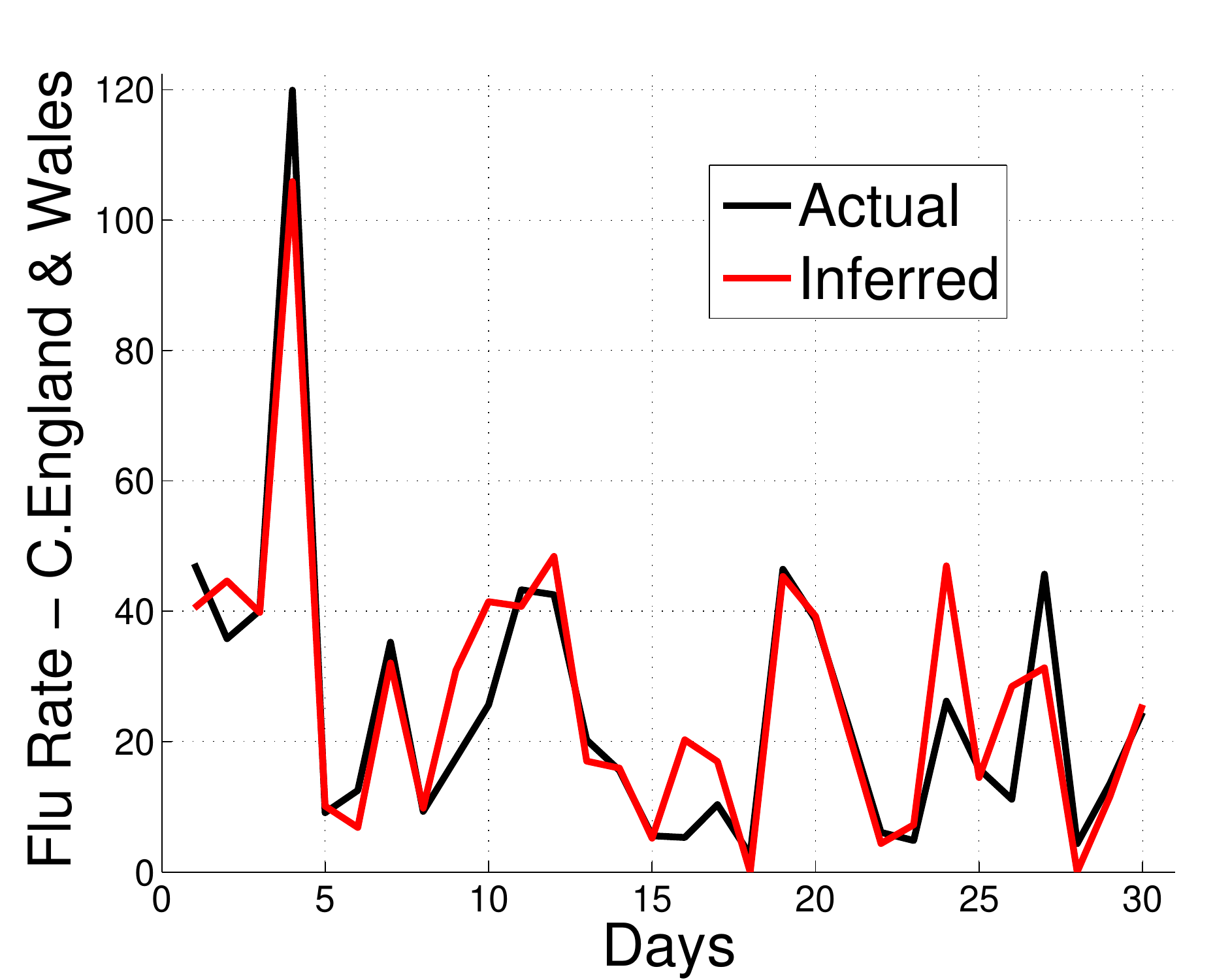}
    \label{fig_fluhybridCengW}}
    \hfil
    \subfigure[\emph{N. England} -- RMSE: 9.782] {\includegraphics[width=2.75in]{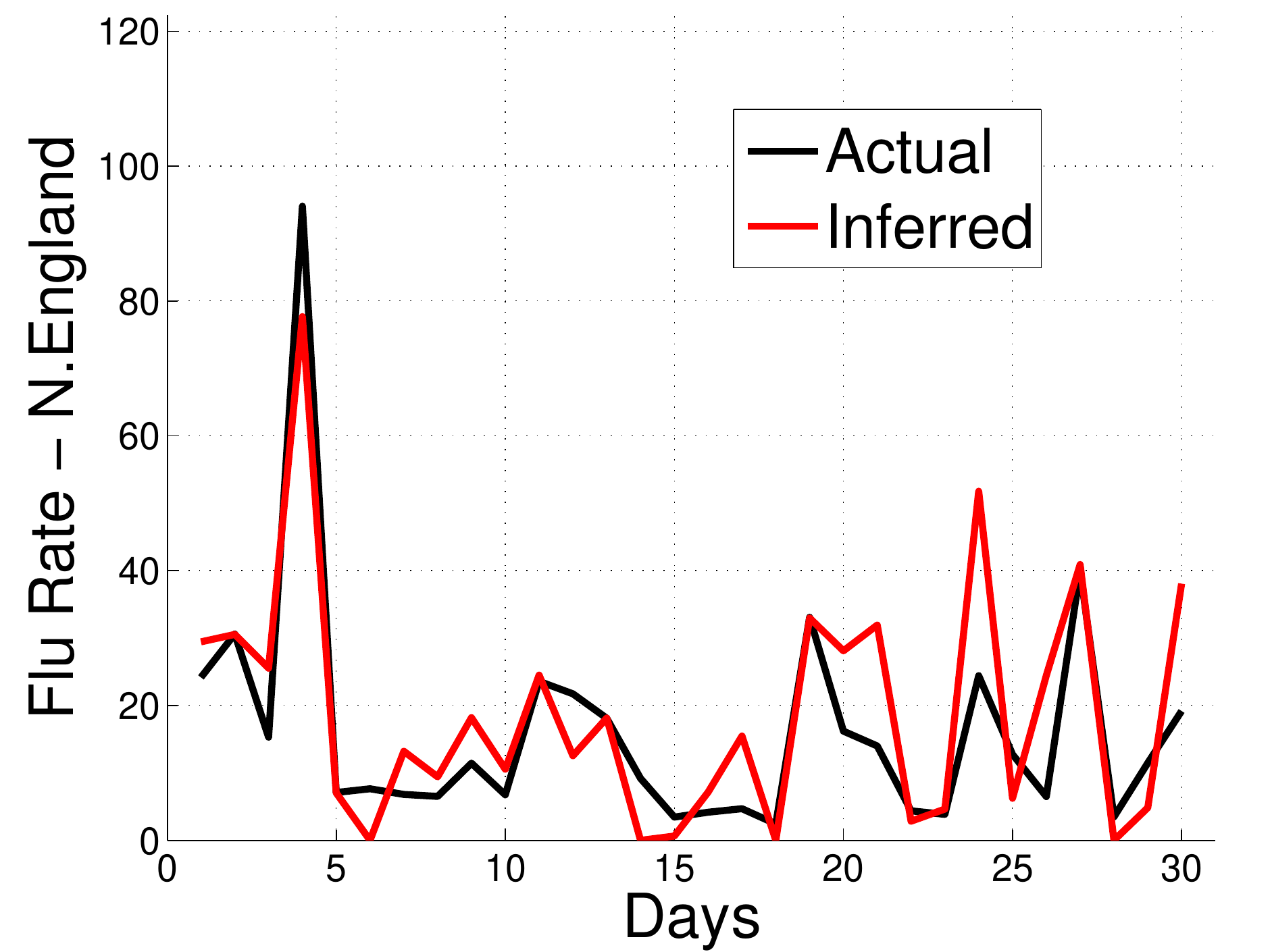}
    \label{fig_fluhybridNEng}}
    \hfil
    \subfigure[\emph{S. England} -- RMSE: 9.86] {\includegraphics[width=2.75in]{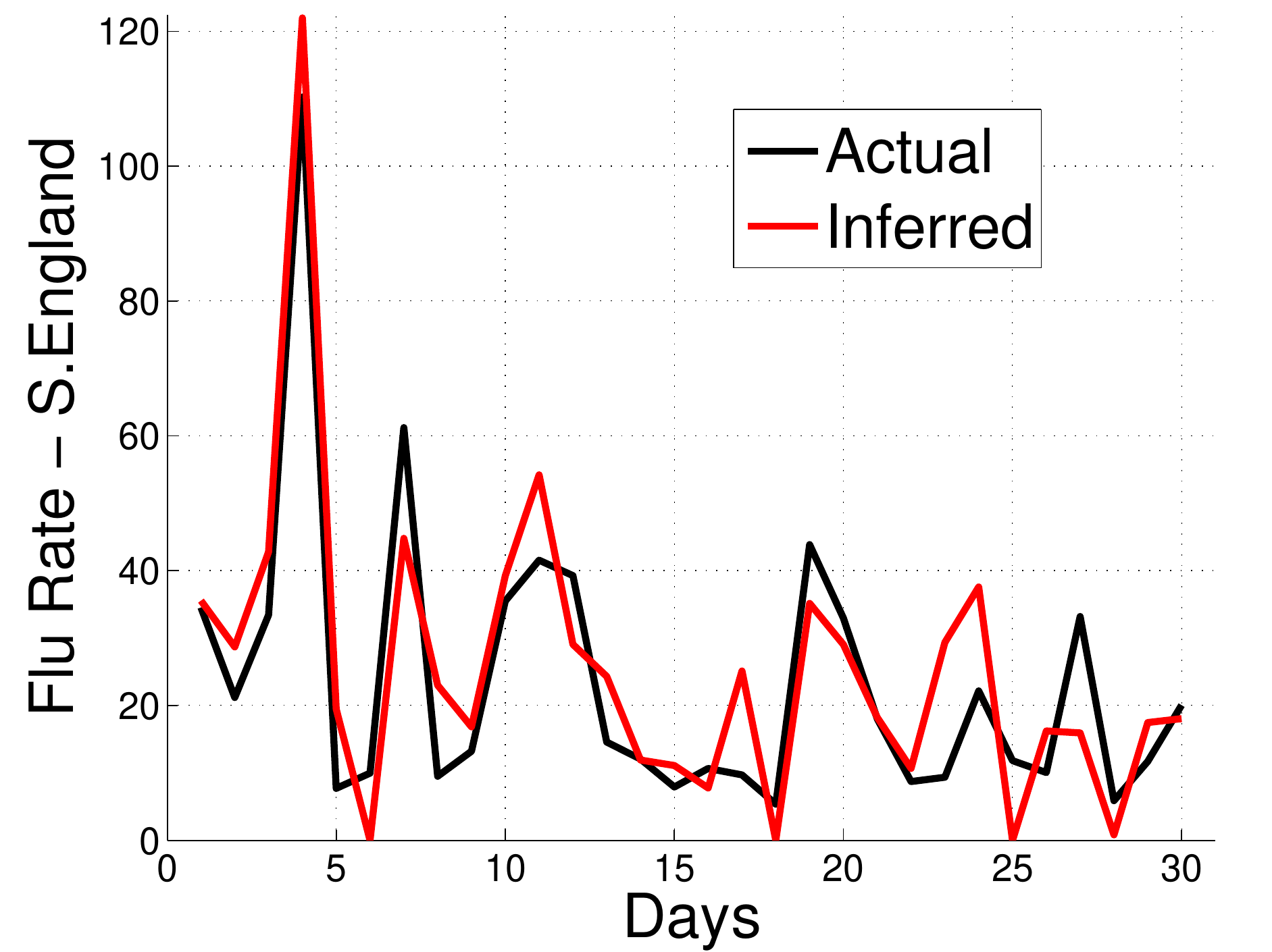}
    \label{fig_fluhybridSEng}}
    \end{center}
    \caption{Feature Class $H$ -- Inference for Flu case study (Round 1 of 5-fold cross validation)}
    \label{fig_fluhybrid}
\end{figure*}

Regional inference results are presented on Figures \ref{fig_flu1grams}, \ref{fig_flu2grams} and \ref{fig_fluhybrid} for classes $U$, $B$ and $H$ respectively. There is a clear indication that the inferred signal has a strong correlation with the actual one; \emph{e.g.} for feature class $H$ (Figure \ref{fig_fluhybrid}) the linear correlation coefficients between the inferred and the actual flu rate for \emph{Central England} \& \emph{Wales}, \emph{North England} and \emph{South England} are equal to 0.933, 0.855 and 0.905 respectively. Using all folds of the cross validation, the average linear correlation for classes $U$, $B$ and $H$ is equal to 0.905, 0.868 and 0.911 respectively, providing additional evidence for the significance of the inference performance.\footnote{ All p-values for the correlation coefficients listed are $\ll 0.05$ indicating statistical significance.}

% non randomised - south england - fold 4
\begin{figure*}[tp]
    \begin{center}
    \subfigure[Feature Class $U$ -- RMSE: 10.332 -- RMSE-smoothed: 7.982] %MSE: 106.754 -- MSE-smoothed: 63.71]
    {\includegraphics[width=2.75in]{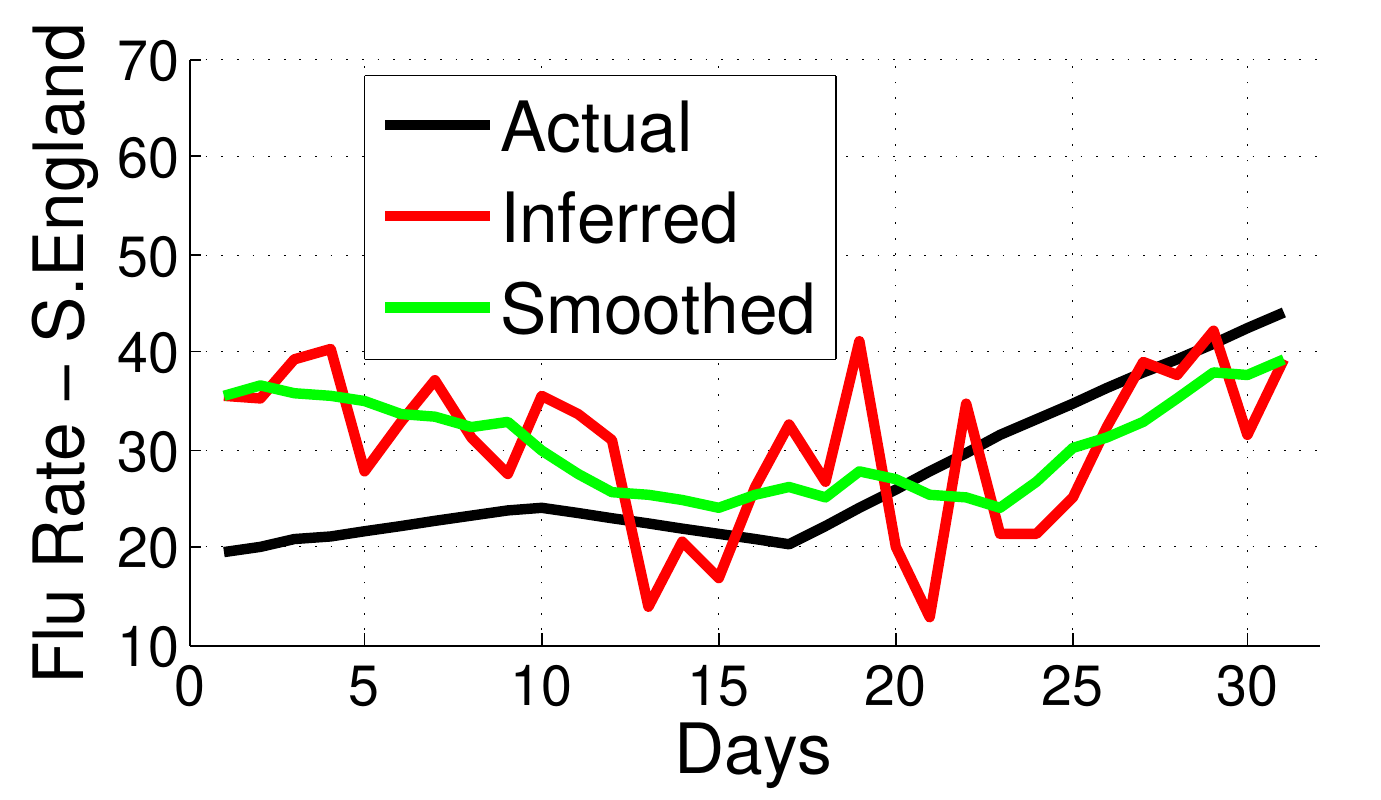}
    \label{fig_fluCont1grams}}
    \hfil
    \subfigure[Feature Class $B$ -- RMSE: 8.292 -- RMSE-smoothed: 5.995] %MSE: 68.752 -- MSE-smoothed: 35.938]
    {\includegraphics[width=2.75in]{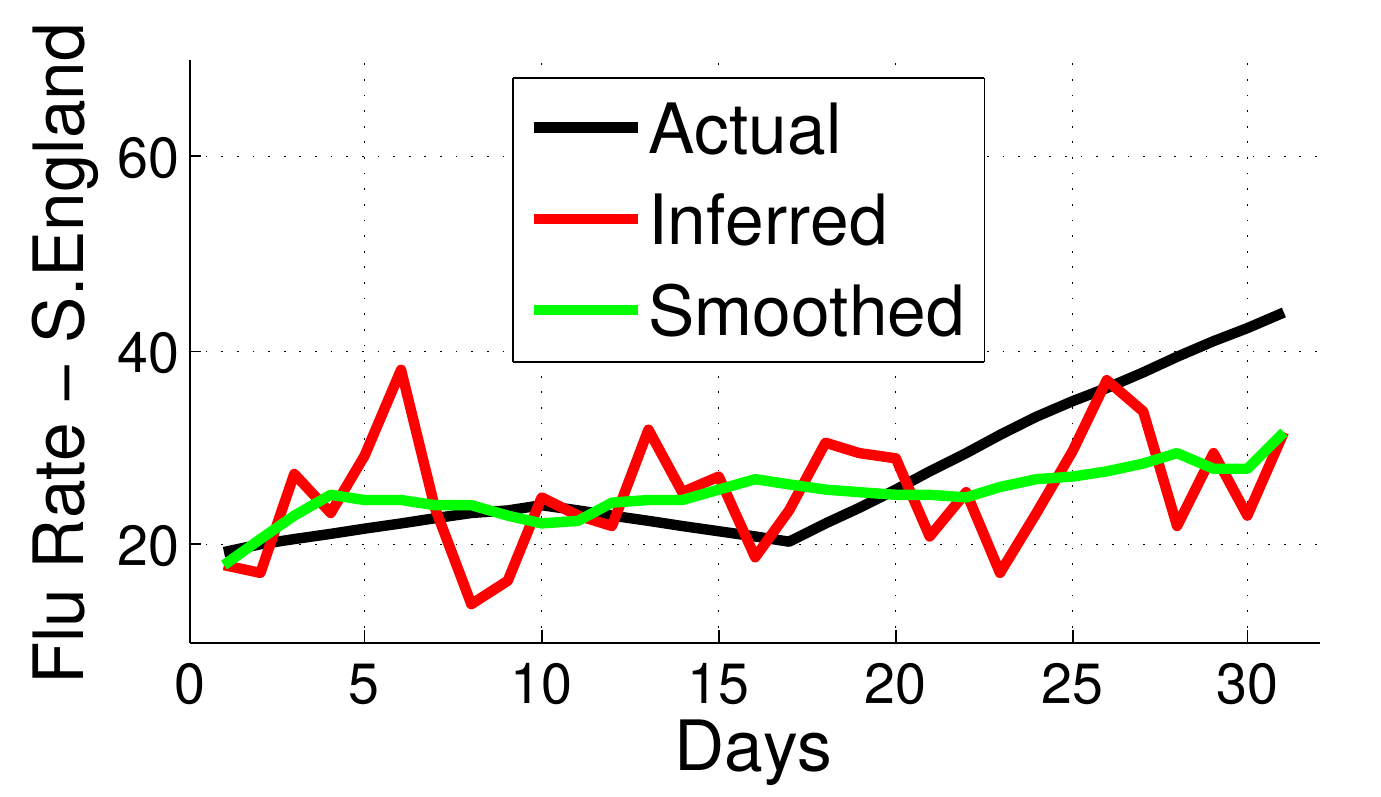}
    \label{fig_fluCont2grams}}
    \hfil
    \subfigure[Feature Class $H$ -- RMSE: 7.468 -- RMSE-smoothed: 5.537] % MSE: 55.778 -- MSE-smoothed: 30.664]
    {\includegraphics[width=2.75in]{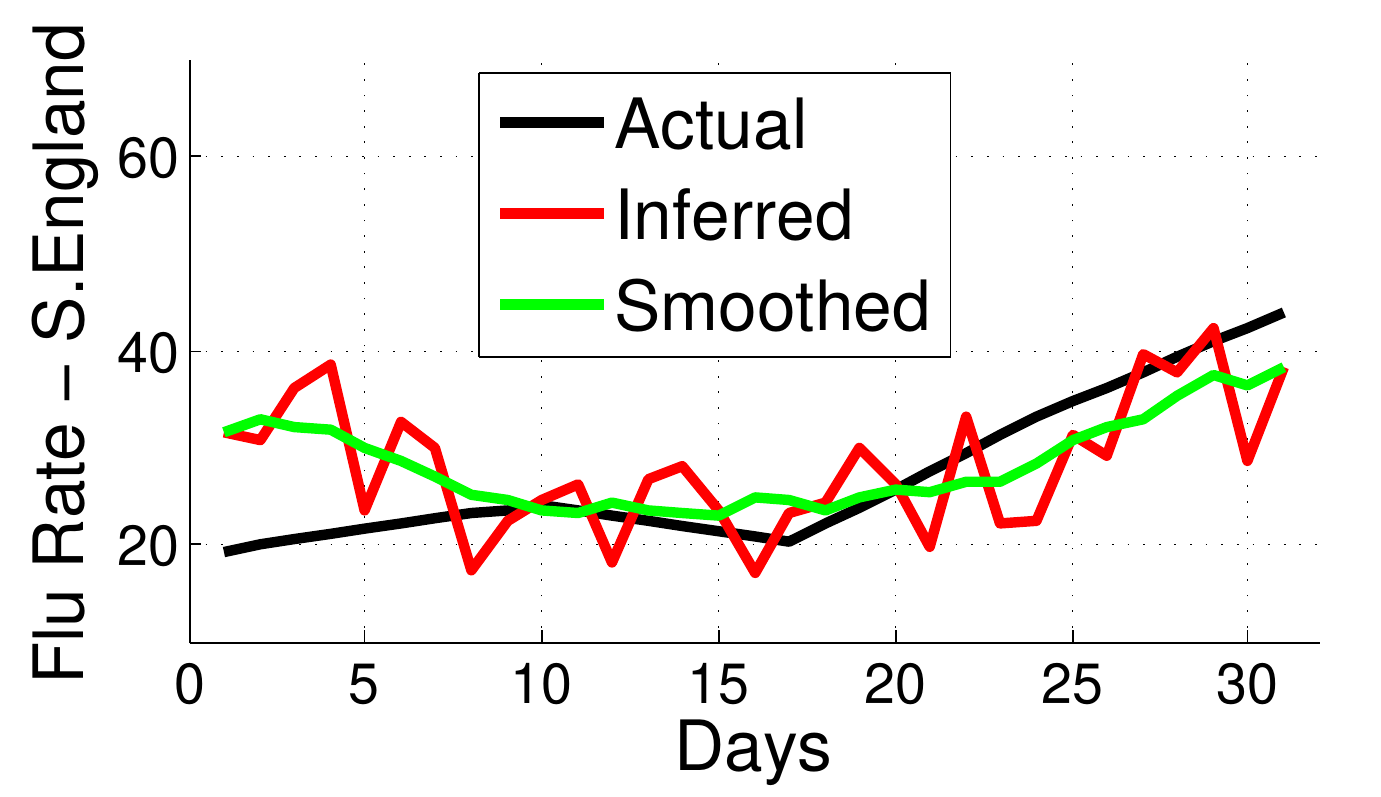}
    \label{fig_fluContHybrid}}
    \end{center}
    \caption{Flu inference results for continuous training, validating and testing sets for \emph{South England} -- Testing is performed on data from the $19^{th}$ of September to the $19^{th}$ October, 2009}
    \label{fig_fluCont}
\end{figure*}

Finally, we present some additional experimental results where training, validating and testing have been carried out in a contiguous time-wise manner. From the 303 days of data, we used days 61-90 for validating CT, 91-121 for testing (from 19th of September to 19th of October, 2009) and the remaining days have been used for training. In this formation setting, we train on data from the Swine Flu epidemic period and then test on a period where influenza existed but its rate was within a normal range. In Figure \ref{fig_fluCont}, we show the inference outcome for \emph{South England} for three distinctive feature classes ($U$, $B$ and $H$).\footnote{ Results for the other 2 regions were similar. South England region was chosen because it is the one with the highest population (as it includes the city of London).} We have also included a smoothed representation of the inferences (using a 7-point moving average) to induce a weekly trend. Class $H$ has the best performance; in this example, class $B$ performs better than $U$.

\section{Discussion}
\label{section_discussion_nowcasting_journal_paper}
\begin{figure*}[tp]
    \begin{center}
    \subfigure[Rainfall rate for \emph{London} (July, 2009 - June, 2010)] {\includegraphics[width=2.75in]{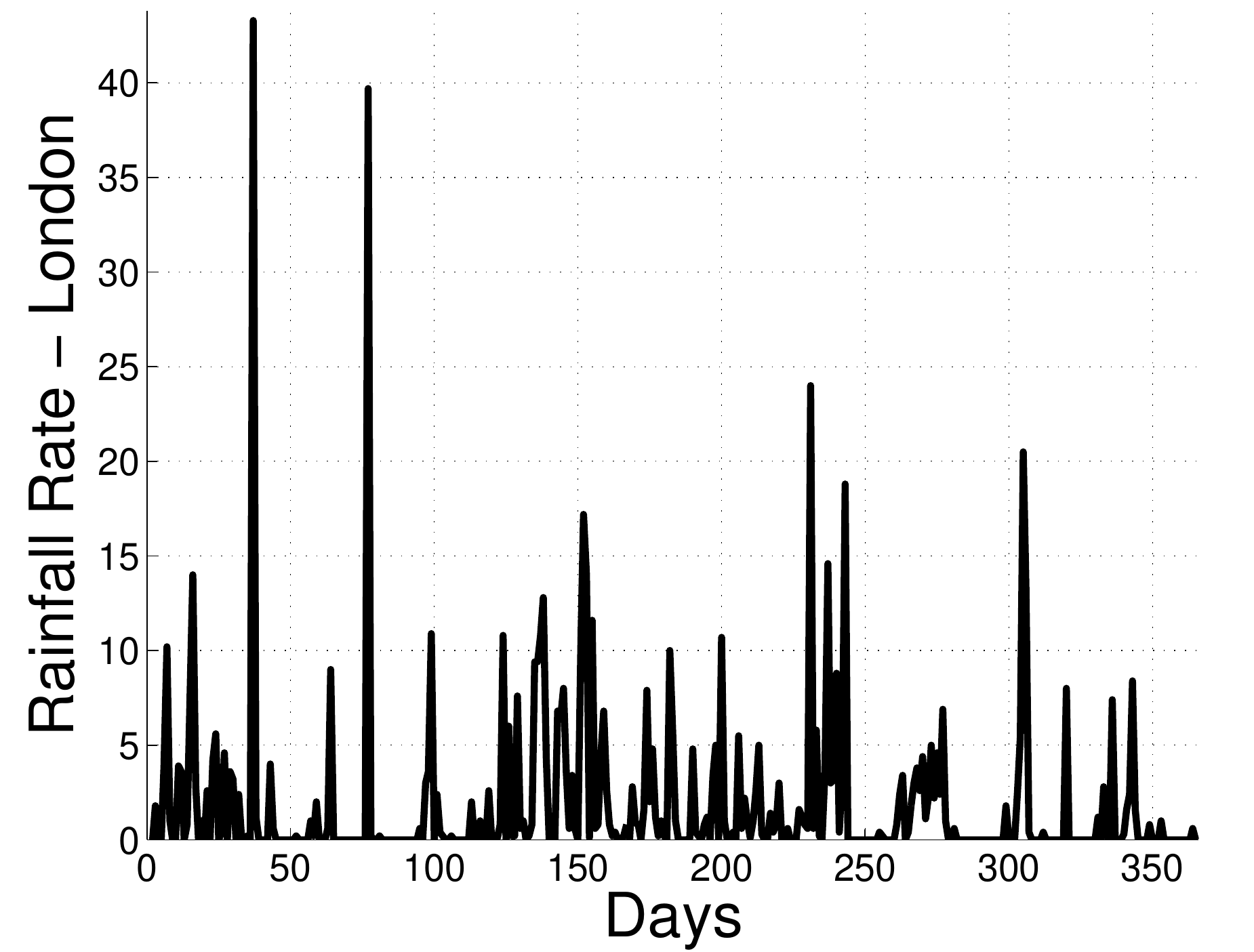}
    \label{fig_gtRainLondon}}
    \hfil
    \subfigure[Flu Rate for \emph{South England} (June, 2009 - April, 2010)] {\includegraphics[width=2.75in]{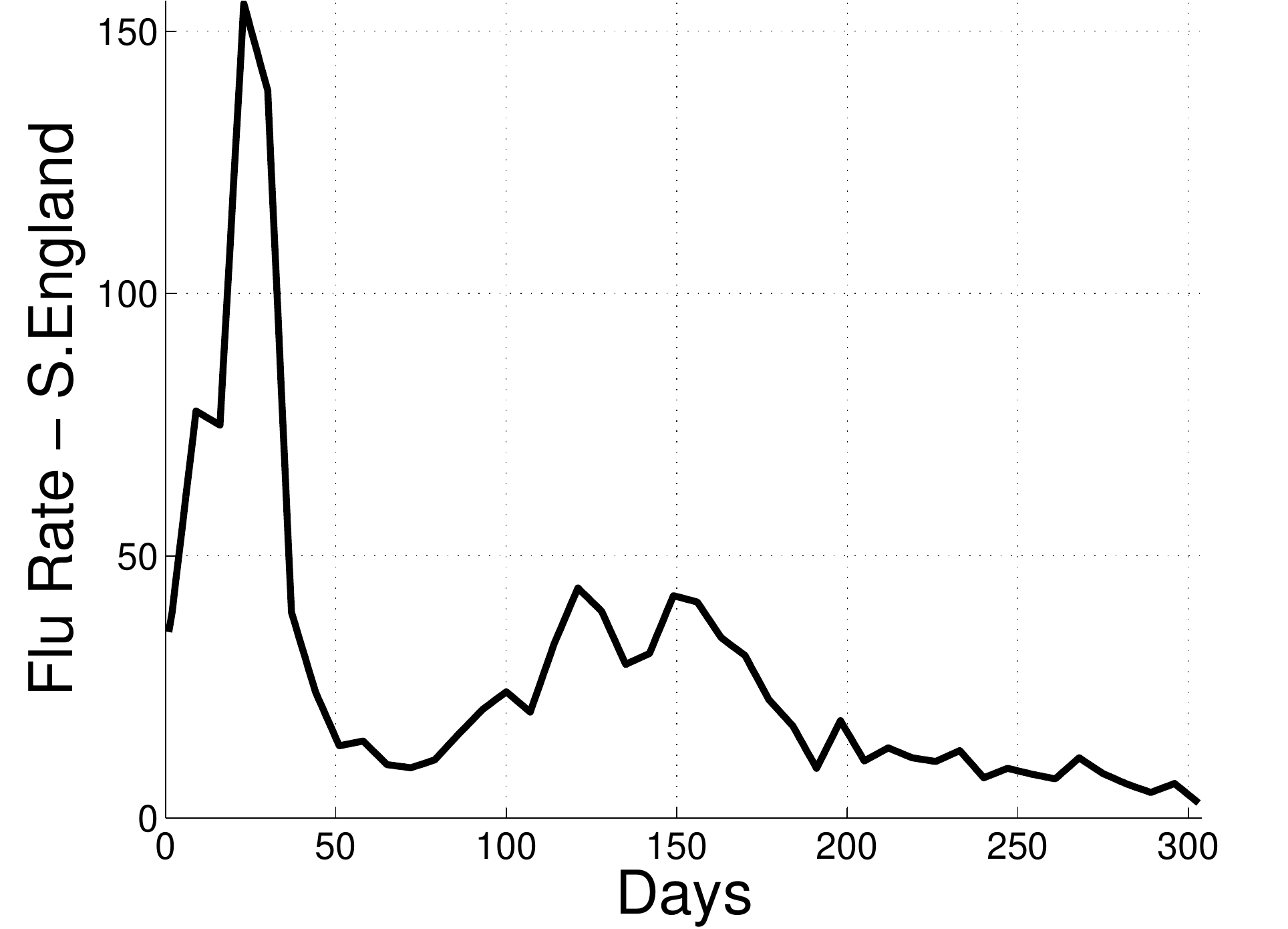}
    \label{fig_gtFluSEng}}
    \hfil
    \subfigure[Flu Rate for \emph{South England} randomly permuted] {\includegraphics[width=2.75in]{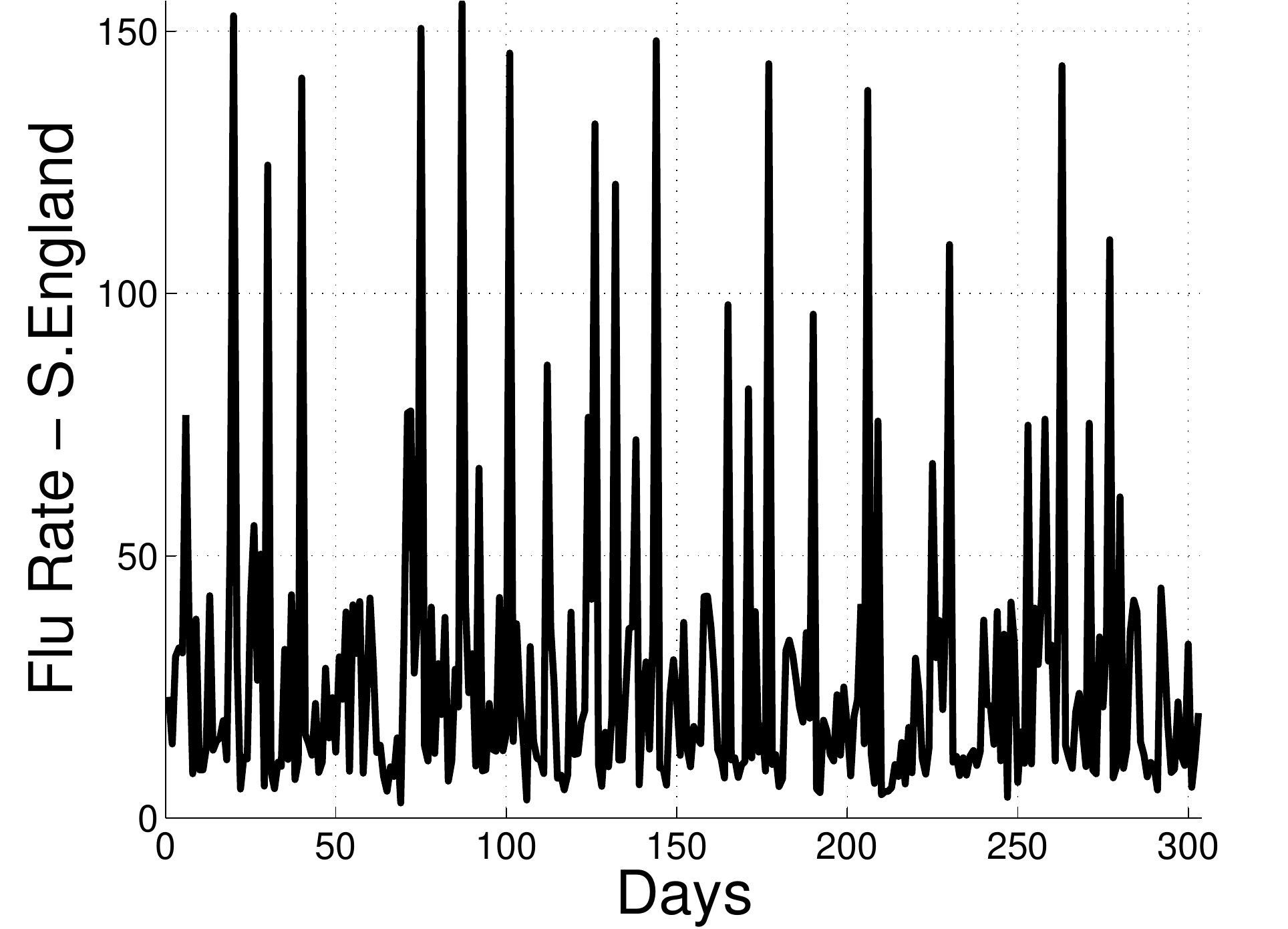}
    \label{fig_gtFluSEngRand}}
    \end{center}
    \caption{Comparing smoothness of ground truth between the two case studies}
    \label{fig_gt}
\end{figure*}

The experimental results provided practical proof for the effectiveness of our method in two case studies: rainfall and flu rates inference. Rain and flu are observable pieces of information available to the general public and therefore are expected to be parts of discussions in the social web media. While samples from both rainfall and ILI rates can be described by exponentially distributed random variables (see Section \ref{section:properties_of_target_events}), those two phenomena have a distinctive property. Precipitation, especially in the UK, is rather unstable, \emph{i.e.} prone to daily changes, whereas a flu rate evolves much more smoothly. Figures \ref{fig_gtRainLondon} (rainfall in \emph{London}) and \ref{fig_gtFluSEng} (flu rate in \emph{South England}) provide a clear picture for this. Consequently, rainfall rates\index{rainfall rates} inference is a much harder problem; users discussing the weather of a preceding day or the forecast for the next one, especially when the current weather conditions contradict, affect not only the inference process but learning as well. For example, in the 6$^{th}$ round of the cross validation, where we derived the worst inference performance, we see that in the VSRs\index{VSR} of the test set (which includes 67 rainy out of 155 days in total for all 5 locations), 1-gram `\emph{flood}' has the exact same average frequency during rainy and non rainy days -- furthermore, the average frequency of stem `\emph{rain}' in days with no rain was equal to 68\% of the one in rainy days. Similar statistics are also observed in the training set or for 2-grams -- \emph{e.g.} the average frequencies of  `\emph{rain hard}' and `\emph{pour rain}' in the training set (716/1515 rainy days) for non rainy days are equal to 42\% and 13\% of the ones in rainy days respectively.

The proposed method is able to overcome those tendencies by selecting features with a more stable behaviour to the extent possible. However, the figures in the two previous sections make clear that inferences have a higher correlation with the ground truth in the flu case study -- even when deploying a randomly permuted version of the data set, which in turn encapsulates only one major flu period, and therefore is of worse quality compared to the rainfall data. Based on those experimental results and the properties of the target events which reach several extremes, we argue that the proposed method is applicable to other events as well, which are at least drawn from similar distribution functions with flu and rainfall rates (see Section \ref{section:properties_of_target_events}).

Another important point in this methodology regards the feature extraction approach. A mainstream IR technique implies the formation of a vocabulary index from the entire corpus \cite{Manning2008}. Instead, we have chosen to form a more focused and restricted in numbers set of candidate features from online references related with the target event, a choice justified by LASSO's risk bound and the `Harry Potter effect' (see Sections \ref{section_error_bounds_for_LASSO} and \ref{section_harry_potter_effect} respectively). The short time span of our data limits the amount of training samples, and therefore directs us in the choice of reducing the number of the candidate features to minimise the risk error and avoid overfitting. By having fewer and slightly more focused on the target event's domain candidates, we constrain the dimensionality over training samples ratio and those issues are resolved. Nevertheless, the size of 1-gram vocabularies in both case studies was not small (approx. 2400 words) and 99\% of the daily tweets for each location or region contained at least one candidate feature. However, for 2-grams this proportion was reduced to 1.5\% and 3\% for rainfall and flu rates case studies respectively, meaning that this class of features required a much higher number of tweets in order to properly contribute.

The experimental process made also clear that a manual selection of very obvious keywords which logically describe a topic, such as `\emph{flu}' or `\emph{rain}', might not be optimal especially when using 1-grams; more rare words (`\emph{puddl}' or `\emph{irrig}') exhibited more stable indications about the target events' magnitude. Finally, it is important to note how CT operates as an additional layer in the feature selection process facilitating the adaptation on the special characteristics of each data set. CT's validation showed that a blind application of strict bolasso (CT = 1) would not have performed as good as the relaxed version we applied; only 3 times in the 33 validation sets used in our experiments the optimal value for CT was set equal to 1.

%%%%%%%%%%%%%%%%%%%%%%%%%%%%%%%%%%%%%%%%%%%%%%%%%%%%%%%%%%%%%%%
\section{Applying different learning methods}
\label{section:applying_different_learning_methods}
In the previous sections, we presented a methodology that used soft-Bolasso with CT validation to nowcast events emerging in real life by using Twitter content. This approach is linear since we end up with a set of predictors and their corresponding weights, and in order to make an inference we just have to compute the inner product between their observed values and their weights. In other words the model makes the initial assumption that the relationship between the observations and the response targets is linear in the parameters of the functional form chosen \cite{Poole1971}. Moreover, LASSO -- as other linear regression techniques -- is based on the assumption that either the observations are independent or the responses are conditionally independent given the observations \cite{tibshirani1996regression}.

In this section, we apply an other class of learning functions on the same problem. The motivation behind that is two-fold: firstly, we want to acquire a comparison metric based on different theoretical tools and secondly, we aim to investigate whether our problem is affected by nonlinearities in the data. Thus, for this task we have chosen CART\index{CART}, a decision tree method which is nonlinear and nonparametric, \ie it does not make any assumption on the probability distribution of the data. In total, we have applied three CART methods: the basic CART, its pruned version and finally an ensemble of `bagged' CARTs. Our attention is drawn mostly on the latter one as it is expected to achieve a better performance; the former are used as intermediate results which assist for a more comprehensive interpretation.

%%%%%%%%%%%%%%%%%%%%%%%%%%%%%%%%%%%%%%%%%%%%%%%%%%%%%%%%%%%%%%
\subsection{Nowcasting rainfall rates with CART}
\label{section:nowcasting_rainfall_rates_with_CART}
In this series of experiments we use the same data as in Section \ref{section_nowcasting_rainfall} -- \ie tweets, candidate features, locations and rainfall observations\index{rainfall rates} from 01/07/2009 to 30/06/2010 -- in order to be able to compare the performance of the different learners. The performance of each method is computed by applying a 6-fold cross validation; the process is identical to the one used for measuring the performance of soft-Bolasso with CT validation. Each fold is based on 2 months of data starting from the month pair July-August (2009) and ending with May-June (2010). In every step of the cross validation, 5 folds are used for training, the first half (a month-long data) of the remaining fold for either deciding the optimal pruning level (for pruned CART) or the optimal number of trees (for the ensemble of CARTs), and the second half for testing the performance of the derived learning function.

\begin{figure}[tp]
\centering
\includegraphics[width=6in]{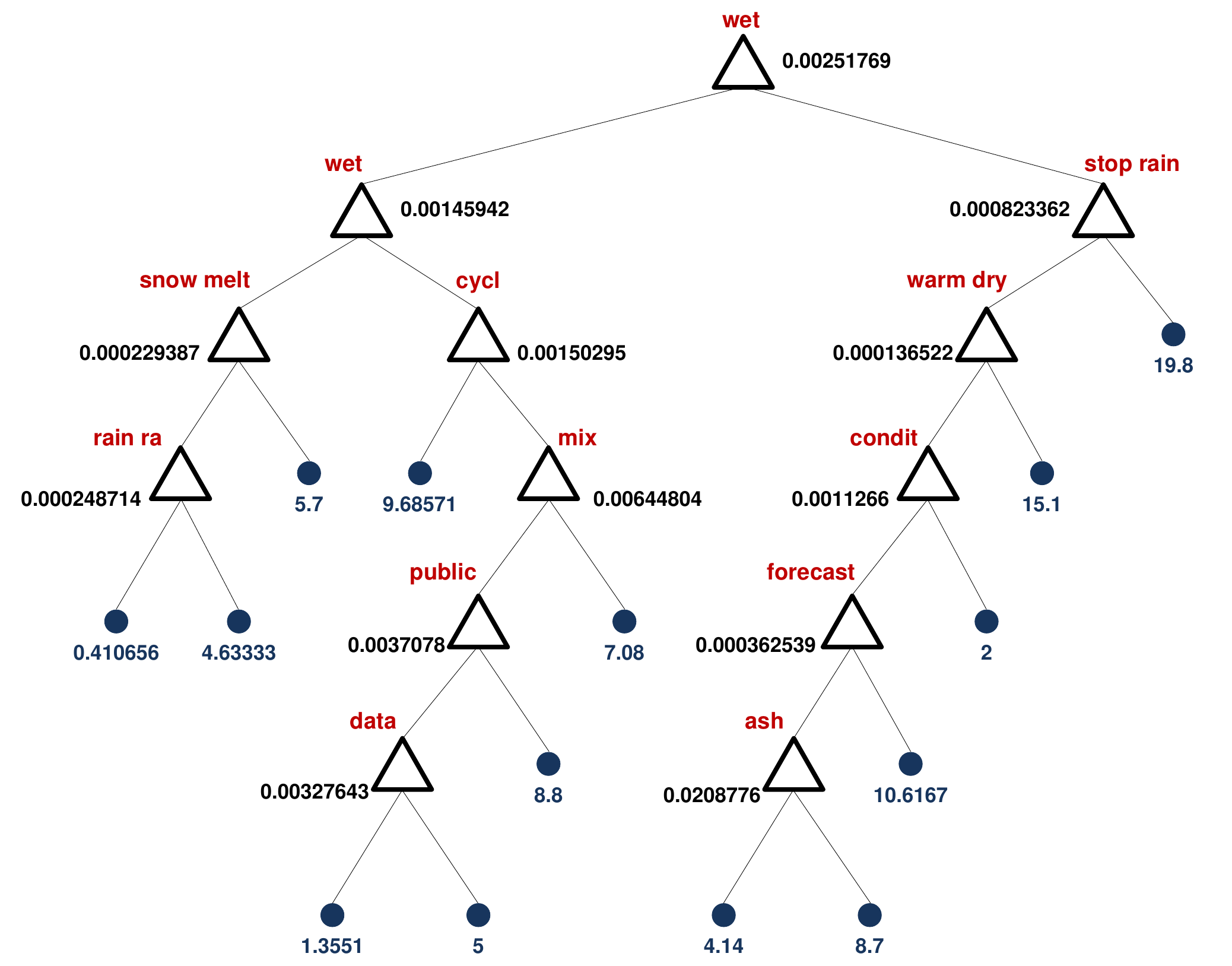}
\caption{CART based on data geolocated in Bristol using the entire data set (365 days) for inferring rainfall rates. This figure only shows a very small fraction of the tree derived by pruning. Numbers ($x_i$) on each node denote the comparison target for the node's (word's) frequency. If the frequency of the word is larger than $x_i$, then the next node to visit lies on the right, else we follow the edge on the left side.}
\label{fig_cart_bristol_rain}
\end{figure}

In total, three feature classes are considered: 1-grams, 2-grams and a concatenation of 1-grams and 2-grams. In our experiments, we have set a lower threshold on the percentage of tree-levels that should be present after pruning. This is equal to 25\%, \ie the pruned CART must retain at least the 25\% top levels of the full tree. To provide the reader with an insight on how an inferred CART looks like, Figure \ref{fig_cart_bristol_rain} shows a fraction of a tree that has been produced by training on all data (1-grams and 2-grams combined) regarding Bristol. We can see that most top words reflect on the topic of weather.

Combining the derivations in \cite{Sutton2005} and \cite{hastie2005elements} we use 150 bootstraps to form the ensemble of CARTs. In each bootstrap the training set is sampled uniformly with replacement; the number of trees that is going to be used in the ensemble is decided by using the validation set. The ensemble makes an inference by averaging across the predictions of all trees participating in it.

The inference performance of all methods for each feature class in terms of RMSE is presented in Tables \ref{table_rain_Bolasso_vs_CART_vs_TreeBagger_1grams}, \ref{table_rain_Bolasso_vs_CART_vs_TreeBagger_2grams} and \ref{table_rain_Bolasso_vs_CART_vs_TreeBagger_hybrid}. Pruning improves the performance of CART in all cases, but as expected, CART -- full or pruned -- cannot compete with the more elaborate bootstrap methods. The ensemble of CARTs has a lower RMSE than Bolasso for 1-grams and 2-grams, but when the feature classes are combined, it cannot improve on Bolasso; in that case the error is equal to 2.637, very close to Bolasso's 2.602. In most folds the optimal number of trees used in the ensemble is kept below 50; we also see that the performance of the ensemble of CARTs improves as we add more features.

\begin{table}[tp]
\caption{Rainfall Rates Inference using CARTs (1-grams) -- RMSEs for all methods in the rounds of 6-fold cross validation -- Fold $i$ denotes the validation/testing fold of round $7-i$.}
\label{table_rain_Bolasso_vs_CART_vs_TreeBagger_1grams}
\footnotesize
\renewcommand{\arraystretch}{1.1}
\setlength\tabcolsep{1mm}
\newcolumntype{C}{>{\centering\arraybackslash} m{1.1cm} }
\newcolumntype{V}{>{\centering\arraybackslash} m{1.5cm} }
\centering
\(\begin{tabular}{l|*{6}{C}|V}
\textbf{Method} & \textbf{Fold 6} & \textbf{Fold 5}  & \textbf{Fold 4}  & \textbf{Fold 3}  & \textbf{Fold 2}  & \textbf{Fold 1}  & \textbf{Mean RMSE}\\\hline
\textbf{Bolasso} ($U$)            & 2.901   & 1.623    & 3.04    & 3.062   & 2.31    & 3.443   & \textbf{2.73}\\\hline
\textbf{CART-full}                & 3.784   & 2.743    & 5.633   & 4.65    & 4.318   & 4.849   & \textbf{4.33}\\\hline
\textbf{CART-pruned}              & 3.367   & 2.478    & 5.549   & 4.315   & 4.152   & 4.705   & \textbf{4.094}\\
\% of tree-levels (min 25\%)      & 31.4\%  & 25.2\%   & 27.35\% & 25.79\% & 36.55\% & 25.22\% & --\\\hline
\textbf{Ensemble of CARTs}        & 2.853   & 1.713    & 3.395   & 2.667   & 2.164   & 3.487   &\textbf{2.713}\\
\# of trees (max 150)             & 43      & 50       & 13      & 93      & 60      & 15      & --\\
\end{tabular}\)
\end{table}

\begin{table}[tp]
\caption{Rainfall Rates Inference using CARTs (2-grams) -- RMSEs for all methods in the rounds of 6-fold cross validation -- Fold $i$ denotes the validation/testing fold of round $7-i$.}
\label{table_rain_Bolasso_vs_CART_vs_TreeBagger_2grams}
\footnotesize
\renewcommand{\arraystretch}{1.1}
\setlength\tabcolsep{1mm}
\newcolumntype{C}{>{\centering\arraybackslash} m{1.1cm} }
\newcolumntype{V}{>{\centering\arraybackslash} m{1.5cm} }
\centering
\(\begin{tabular}{c|*{6}{C}|V}
\textbf{Method} & \textbf{Fold 6} & \textbf{Fold 5}  & \textbf{Fold 4}  & \textbf{Fold 3}  & \textbf{Fold 2}  & \textbf{Fold 1}  & \textbf{Mean RMSE}\\\hline
\textbf{Bolasso} ($B$)            & 2.745   & 3.062    & 2.993   & 3.028   & 2.083   & 3.789   & \textbf{2.95}\\\hline
\textbf{CART-full}                & 4.692   & 5.066    & 3.8     & 3.897   & 4.14    & 4.064   & \textbf{4.276}\\\hline
\textbf{CART-pruned}              & 2.98    & 4.69     & 3.45    & 3.148   & 3.401   & 3.993   & \textbf{3.61}\\
\% of tree-levels (min 25\%)      & 26.18\% & 25.14\%  & 25.24\% & 29.04\% & 25.15\% & 25.62\% & --\\\hline
\textbf{Ensemble of CARTs}        & 2.865   & 1.609    & 3.487   & 2.816   & 2.165   & 3.361   & \textbf{2.717}\\
\# of trees (max 150)             & 42      & 14       & 6       & 50      & 21      & 30      & --\\
\end{tabular}\)
\end{table}

\begin{table}[tp]
\caption{Rainfall Rates Inference using CARTs (1-grams \& 2-grams) -- RMSEs for all methods in the rounds of 6-fold cross validation -- Fold $i$ denotes the validation/testing fold of round $7-i$.}
\label{table_rain_Bolasso_vs_CART_vs_TreeBagger_hybrid}
\footnotesize
\renewcommand{\arraystretch}{1.1}
\setlength\tabcolsep{1mm}
\newcolumntype{C}{>{\centering\arraybackslash} m{1.1cm} }
\newcolumntype{V}{>{\centering\arraybackslash} m{1.5cm} }
\centering
\(\begin{tabular}{c|*{6}{C}|V}
\textbf{Method} & \textbf{Fold 6} & \textbf{Fold 5}  & \textbf{Fold 4}  & \textbf{Fold 3}  & \textbf{Fold 2}  & \textbf{Fold 1}  & \textbf{Mean RMSE}\\\hline
\textbf{Bolasso} ($H$)            & 2.779   & 1.459    & 2.937   & 2.954    & 2.145     & 3.338     & \textbf{2.602}\\\hline
\textbf{CART-full}                & 3.79    & 2.698    & 3.513   & 3.822    & 4.69      & 5.042     & \textbf{3.926}\\\hline
\textbf{CART-pruned}              & 3.494   & 2.473    & 3.604   & 3.717    & 3.946     & 4.859     & \textbf{3.682}\\
\% of tree-levels (min 25\%)      & 25.21\% & 27.57\%  & 25.56\% & 29.11\%  & 25.2\%    & 25.22\%   & --\\\hline
\textbf{Ensemble of CARTs}        & 2.758   & 1.589    & 3.097   & 2.732    & 2.289     & 3.356     & \textbf{2.637}\\
\# of trees (max 150)             & 39      & 24       & 149     & 40       & 19        & 31        & --\\
\end{tabular}\)
\end{table}

Feature selection in an ensemble of $N$ trees $\mathcal{T}$ can be performed by first computing an importance factor $\delta$ for each variable $x_i$ in every subtree $\mathcal{T}_j$, and then averaging over all the trees in the ensemble \cite{Tuv2009}. The importance of variable $x_i$ in a tree $\mathcal{T}_j$ is given by
\begin{equation}
\delta(x_i,\mathcal{T}_j) = \sum_{t\in\mathcal{T}_j}\left[\Delta{\sum_{i \in t}\frac{(y_i - \bar{y})^2}{N(t)}}\right],
\end{equation}
where $y_i$'s denote the observations of node $t$, $\bar{y}$ denotes their mean, $N(t)$ is the number of observations for node $t$ and $\Delta$ expresses the decrease of the enclosed summation as we move down the tree. To compute the overall importance of a variable, we just have to average the above equation over all trees in the ensemble
\begin{equation}
\delta(x_i,\mathcal{T}_{1,...,N}) = \frac{1}{N} \sum_{n=1}^{N} \delta(x^{(u_i)},\mathcal{T}_i).
\end{equation}

Tables \ref{table_rain_TreeBagger_features_1grams}, \ref{table_rain_TreeBagger_features_2grams} and \ref{table_rain_TreeBagger_features_hybrid} show the most important features -- based on their $\delta$ -- for round 5 of the 6-fold cross validation in all the investigated feature classes based on the ensemble of CARTs. Those results are directly comparable with the ones in Tables \ref{table_features_rain_1grams}, \ref{table_features_rain_2grams} and \ref{table_features_rain_hybrid} respectively; the features that have been selected by Bolasso are typed in bold. We see that in all feature classes the distinctively important features are related to the topic of weather; the n-grams `\emph{rain}' and `\emph{rain rain}' are shown to have the largest $\delta$. However, those features are not identical to the ones selected by Bolasso and in some occasions we see that very irrelevant ones have a relatively high importance such as the 1-grams `\emph{sentenc}' and `\emph{archiv}' in Table \ref{table_rain_TreeBagger_features_hybrid}.

\begin{table}
\caption{Rainfall Rates Inference using CARTs (1-grams) -- Most important features in the ensemble of CARTs (Round 5 of 6-fold cross validation) -- $\delta$ should be multiplied by $10^{-3}$ and features in bold have been also selected by Bolasso ($U$).}
\label{table_rain_TreeBagger_features_1grams}
\footnotesize
\renewcommand{\arraystretch}{1.1}
\setlength\tabcolsep{1mm}
\centering
\(\begin{tabular}{cc|cc|cc|cc|cc}
\textbf{1-gram} & \textbf{$\delta$} & \textbf{1-gram}  & \textbf{$\delta$} & \textbf{1-gram} & \textbf{$\delta$} & \textbf{1-gram} & \textbf{$\delta$} & \textbf{1-gram} & \textbf{$\delta$}\\\hline
\textbf{rain}        & 5.308     & \textbf{umbrella}   & 0.82     & raini     & 0.336  & thunder  & 0.198  & displac  & 0.166\\
\textbf{wet}         & 4.037     & \textbf{puddl}      & 0.75     & wettest   & 0.244  & lai      & 0.179  & sleep    & 0.152 \\
rai                  & 1.871     & \textbf{monsoon}    & 0.617    & \textbf{sunni}     & 0.207  & act      & 0.168  & &  \\
\textbf{pour}        & 0.963     & sentenc    & 0.344    & manag     & 0.201  & factor   & 0.167  & &
\end{tabular}\)
\end{table}

\begin{table}
\caption{Rainfall Rates Inference using CARTs (2-grams) -- Most important features in the ensemble of CARTs (Round 5 of 6-fold cross validation) -- $\delta$ should be multiplied by $10^{-3}$ and features in bold have been also selected by Bolasso ($B$).}
\label{table_rain_TreeBagger_features_2grams}
\footnotesize
\renewcommand{\arraystretch}{1.1}
\setlength\tabcolsep{1mm}
\centering
\(\begin{tabular}{cc|cc|cc|cc|cc}
\textbf{2-gram} & $\delta$ & \textbf{2-gram} & $\delta$ & \textbf{2-gram} & $\delta$ & \textbf{2-gram} & $\delta$ & \textbf{2-gram} & $\delta$  \\\hline
\textbf{rain rain}  & 3.593 & unit state         & 0.493 & long time    & 0.367 & dai week                & 0.255 & \textbf{wind rain}  & 0.195 \\
rain ra             & 3.358 & like rain          & 0.455 & rain lot     & 0.358 & wind direct             & 0.255 & \textbf{light rain} & 0.192 \\
\textbf{stop rain}  & 2.484 & flash flood        & 0.429 & dai todai    & 0.318 & \textbf{horribl weather}& 0.249 & dai dai             & 0.178\\
\textbf{pour rain}  & 2.346 & look like          & 0.405 & start rain   & 0.316 & act like                & 0.234 & oh love             & 0.172\\
\textbf{raini dai}  & 1.062 & ic cap             & 0.393 & love weather & 0.282 & met offic               & 0.215 & rain dai            & 0.169\\
love watch          & 0.513 & \textbf{sunni dai} & 0.384 & al al        & 0.281 & climat chang            & 0.205 & rain start          & 0.165\\
\end{tabular}\)
\end{table}

\begin{table}
\caption{Rainfall Rates Inference using CARTs (1-grams \& 2-grams) -- Most important features in the ensemble of CARTs (Round 5 of 6-fold cross validation) -- $\delta$ should be multiplied by $10^{-3}$ and features in bold have been also selected by Bolasso ($H$).}
\label{table_rain_TreeBagger_features_hybrid}
\footnotesize
\renewcommand{\arraystretch}{1.1}
\setlength\tabcolsep{1mm}
\centering
\(\begin{tabular}{cc|cc|cc|cc|cc}
\textbf{n-gram} & $\delta$ & \textbf{n-gram} & $\delta$ & \textbf{n-gram} & $\delta$ & \textbf{n-gram} & $\delta$ & \textbf{n-gram} & $\delta$  \\\hline
\textbf{rain}      & 4.164 & rai                & 0.72  & \textbf{pour}      & 0.39  & manag    & 0.21  & white     & 0.184 \\
\textbf{rain rain} & 2.873 & \textbf{stop rain} & 0.627 & archiv             & 0.343 & \textbf{flood}    & 0.199 & raini     & 0.174 \\
\textbf{wet}       & 2.686 & sentenc            & 0.54  & \textbf{raini dai} & 0.26  & \textbf{puddl}    & 0.195 & wettest   & 0.172 \\
rain ra            & 2.542 & \textbf{umbrella}  & 0.495 & blind              & 0.258 & christma & 0.191 & japan     & 0.166 \\
\textbf{pour rain} & 1.299 & \textbf{monsoon}   & 0.4   & solar              & 0.252 & displac  & 0.187 &           &\\
\end{tabular}\)
\end{table}

%%%%%%%%%%%%%%%%%%%%%%%%%%%%%%%%%%%%%%%%%%%%%%%%%%%%%%%5
\subsection{Nowcasting flu rates with CART}
\label{section:nowcasting_flu_rates_with_cart}
In this series of experiments we have used the same data as in Section \ref{section_nowcasting_flu} -- \ie tweets, candidate features, UK regions and official flu rates\index{flu rates} from 21/06/2009 to 19/04/2010 -- in order to be able to compare the performance of the different learners. The performance of each method is computed by applying a 5-fold cross validation; 4 folds using data of 60 or 61 days each are used for training, and from the remaining fold, 30 days of data are used for validation (in order to decide the optimal prune level or number of trees in the ensemble) and the rest for testing. The same experimental process and settings as in the previous section are applied with the minor difference that we have set the maximum pruning to the 85\% (not 75\%) of CART's levels.

\begin{table}[tp]
\caption{Flu Rates Inference using CARTs (1-grams) -- RMSEs for all methods in the rounds of 6-fold cross validation -- Fold $i$ denotes the validation/testing fold of round $6-i$.}
\label{table_flu_Bolasso_vs_CART_vs_TreeBagger_1grams}
\footnotesize
\renewcommand{\arraystretch}{1.1}
\setlength\tabcolsep{1mm}
\newcolumntype{C}{>{\centering\arraybackslash} m{1.1cm} }
\newcolumntype{V}{>{\centering\arraybackslash} m{1.5cm} }
\centering
\(\begin{tabular}{l|*{5}{C}|V}
\textbf{Method} & \textbf{Fold 5} & \textbf{Fold 4}   & \textbf{Fold 3}  & \textbf{Fold 2}  & \textbf{Fold 1}  & \textbf{Mean RMSE}\\\hline
\textbf{Bolasso} ($U$)            & 10.426  & 8.056   & 13.29    & 13.699    & 10.222    & \textbf{11.139}\\\hline
\textbf{CART-full}                & 13.749  & 17.924  & 13.42    & 15.027    & 10.368    & \textbf{14.098}\\\hline
\textbf{CART-pruned}              & 13.706  & 17.832  & 13.381   & 13.409    & 10.281    & \textbf{13.722}\\
\% of tree levels (min 15\%)      & 48.84\% & 21.97\% & 15.79\%  & 17.04\%   & 16.15\%   & --\\\hline
\textbf{Ensemble of CARTs}        & 5.758   & 6.42    & 16.659   & 13.783    & 5.542     & \textbf{9.632}\\
\# of trees (max 150)             & 58      & 144     & 44       & 85        & 17        & --\\
\end{tabular}\)
\end{table}

\begin{table}[tp]
\caption{Flu Rates Inference using CARTs (2-grams) -- RMSEs for all methods in the rounds of 6-fold cross validation -- Fold $i$ denotes the validation/testing fold of round $6-i$.}
\label{table_flu_Bolasso_vs_CART_vs_TreeBagger_2grams}
\footnotesize
\renewcommand{\arraystretch}{1.1}
\setlength\tabcolsep{1mm}
\newcolumntype{C}{>{\centering\arraybackslash} m{1.1cm} }
\newcolumntype{V}{>{\centering\arraybackslash} m{1.5cm} }
\centering
\(\begin{tabular}{c|*{5}{C}|V}
\textbf{Method} & \textbf{Fold 5}  & \textbf{Fold 4}  & \textbf{Fold 3}  & \textbf{Fold 2}  & \textbf{Fold 1}  & \textbf{Mean RMSE}\\\hline
\textbf{Bolasso} ($B$)            & 11.806   & 10.536  & 16.93     & 12.958    & 10.986    & \textbf{12.643}\\\hline
\textbf{CART-full}                & 14.567   & 14.646  & 24.852    & 19.682    & 13.243    & \textbf{17.398}\\\hline
\textbf{CART-pruned}              & 13.798   & 12.266  & 23.451    & 19.208    & 13.021    & \textbf{16.349}\\
\% of tree levels (min 15\%)      & 15.6\%   & 15.17\% & 18.12\%   & 18.8\%    & 21.74\%   & --\\\hline
\textbf{Ensemble of CARTs}        & 10.152   & 8.747   & 20.754    & 14.241    & 11.763    & \textbf{13.131}\\
\# of trees (max 150)             & 7        & 26      & 30        & 42        & 126       & --\\
\end{tabular}\)
\end{table}

\begin{table}[tp]
\caption{Flu Rates Inference using CARTs (1-grams \& 2-grams) -- RMSEs for all methods in the rounds of 6-fold cross validation -- Fold $i$ denotes the validation/testing fold of round $6-i$.}
\label{table_flu_Bolasso_vs_CART_vs_TreeBagger_hybrid}
\footnotesize
\renewcommand{\arraystretch}{1.1}
\setlength\tabcolsep{1mm}
\newcolumntype{C}{>{\centering\arraybackslash} m{1.1cm} }
\newcolumntype{V}{>{\centering\arraybackslash} m{1.5cm} }
\centering
\(\begin{tabular}{c|*{5}{C}|V}
\textbf{Method} & \textbf{Fold 5} & \textbf{Fold 4}  & \textbf{Fold 3}  & \textbf{Fold 2}  & \textbf{Fold 1}  & \textbf{Mean RMSE}\\\hline
\textbf{Bolasso} ($H$)            & 9.359    & 7.971   & 12.094    & 13.475    & 9.93      & \textbf{10.566}\\\hline
\textbf{CART-full}                & 14.409   & 9.882   & 17.399    & 16.001    & 10.775    & \textbf{13.693}\\\hline
\textbf{CART-pruned}              & 14.445   & 10.274  & 17.677    & 14.822    & 10.586    & \textbf{13.561}\\
\% of tree levels (min 15\%)      & 53.03\%  & 16.15\% & 16.67\%   & 15.79\%   & 17.16\%   & --\\\hline
\textbf{Ensemble of CARTs}        & 6.684    & 6.132   & 14.203    & 13.803    & 6.191     & \textbf{9.402}\\
\# of trees (max 150)             & 19       & 104     & 47        & 24        & 20        & --\\
\end{tabular}\)
\end{table}

The inference performance of all methods for each feature class in terms of RMSE is presented in Tables \ref{table_flu_Bolasso_vs_CART_vs_TreeBagger_1grams}, \ref{table_flu_Bolasso_vs_CART_vs_TreeBagger_2grams} and \ref{table_flu_Bolasso_vs_CART_vs_TreeBagger_hybrid}. Again, pruning may have improved the performance of CART in all cases, but this was not enough for competing with the ensemble of CARTs or Bolasso. Bolasso performs better than the ensemble of CARTs in the case of 2-grams, but the best overall performance is derived by the ensemble when combining 1-grams and 2-grams; the RMSE drops to 9.402 and compared to Bolasso's RMSE (10.556), this is an improvement of 11.02\%.

\begin{table}
\caption{Flu Rates Inference using CARTs (1-grams) -- Most important features in the ensemble of CARTs (Round 1 of 5-fold cross validation) -- $\delta$ should be multiplied by $10^{-2}$ and features in bold have been also selected by Bolasso ($U$).}
\label{table_flu_TreeBagger_features_1grams}
\footnotesize
\renewcommand{\arraystretch}{1.1}
\setlength\tabcolsep{1mm}
\centering
\(\begin{tabular}{cc|cc|cc|cc|cc}
\textbf{1-gram} & \textbf{$\delta$} & \textbf{1-gram}  & \textbf{$\delta$} & \textbf{1-gram} & \textbf{$\delta$} & \textbf{1-gram} & \textbf{$\delta$} & \textbf{1-gram} & \textbf{$\delta$}\\\hline
\textbf{swine}  & 116.504  & swineflu       & 13.743 & lower        & 4.895 & symptom           & 1.846 & hot     & 1.385\\
flu             & 69.77    & school         & 9.84   & \textbf{team}& 3.641 & \textbf{season}   & 1.7   & \textbf{erad}    & 1.114\\
\textbf{site}   & 44.269   & \textbf{wave}  & 9.438  & tamiflu      & 3.015 & autumn            & 1.641 & warm    & 1.109\\
\textbf{member} & 24.629   & summer         & 8.609  & tour         & 2.666 & fact              & 1.599 & live    & 1.096\\
\textbf{strong} & 14.701   & \textbf{run}   & 5.346  & quarantin    & 2.379 & factor            & 1.386 & featur  & 1.071\\
\end{tabular}\)
\end{table}

\begin{table}
\caption{Flu Rates Inference using CARTs (2-grams) -- Most important features in the ensemble of CARTs (Round 1 of 5-fold cross validation) -- $\delta$ should be multiplied by $10^{-2}$ and features in bold have been also selected by Bolasso ($B$).}
\label{table_flu_TreeBagger_features_2grams}
\footnotesize
\renewcommand{\arraystretch}{1.1}
\setlength\tabcolsep{1mm}
\centering
\(\begin{tabular}{cc|cc|cc|cc}
\textbf{2-gram} & \textbf{$\delta$} & \textbf{2-gram}  & \textbf{$\delta$} & \textbf{2-gram} & \textbf{$\delta$} & \textbf{2-gram} & \textbf{$\delta$}\\\hline
\textbf{swine flu}     & 134.501 & confirm case           & 6.982 & \textbf{wonder swine}  & 4.494 & \textbf{healthcar worker} & 1.995 \\
\textbf{case swine}    & 62.402  & need check             & 6.648 & pig flu                & 3.812 & flu kill                  & 1.861 \\
\textbf{suspect swine} & 26.794  & \textbf{confirm swine} & 6.608 & swine flue             & 2.942 & flu vaccin                & 1.572 \\
\textbf{flu symptom}   & 23.463  & \textbf{underli health}& 5.362 & \textbf{symptom swine} & 2.914 & light switch              & 1.333 \\
\textbf{die swine}     & 12.209  & \textbf{feel better}   & 5.09  & death toll             & 2.447 & \textbf{kick ass}         & 1.124 \\
\textbf{check site}    & 12.019  & \textbf{flu jab}       & 5.088 & unit kingdom           & 2.242 & skin care                 & 1.042 \\
contract swine         & 9.568   & \textbf{weight loss}   & 4.698 & \textbf{flu bad}       & 2.072 & flu work                  & 1.034 \\
\end{tabular}\)
\end{table}

\begin{table}
\caption{Flu Rates Inference using CARTs (1-grams \& 2-grams) -- Most important features in the ensemble of CARTs (Round 1 of 5-fold cross validation) -- $\delta$ should be multiplied by $10^{-2}$ and features in bold have been also selected by Bolasso ($H$).}
\label{table_flu_TreeBagger_features_hybrid}
\footnotesize
\renewcommand{\arraystretch}{1.1}
\setlength\tabcolsep{1mm}
\centering
\(\begin{tabular}{cc|cc|cc|cc}
\textbf{n-gram} & \textbf{$\delta$} & \textbf{n-gram}  & \textbf{$\delta$} & \textbf{n-gram} & \textbf{$\delta$} & \textbf{n-gram} & \textbf{$\delta$}\\\hline
\textbf{swine flu}  & 121.821 & \textbf{wave}           & 10.002 & tamiflu              & 1.85  & \textbf{boni}               & 1.346 \\
\textbf{swine}      & 87.641  & \textbf{flu symptom}    & 6.77   & \textbf{check site}  & 1.79  & \textbf{healthcar worker}   & 1.139 \\
flu                 & 34.451  & \textbf{run}            & 4.994  & mass                 & 1.765 & live               & 1.133 \\
\textbf{member}     & 23.675  & swineflu                & 3.961  & \textbf{season}      & 1.719 & spur               & 1.096 \\
\textbf{case swine} & 15.701  & \textbf{strong}         & 3.913  & lower                & 1.618 & hayfev             & 1.091 \\
school              & 15.463  & summer                  & 2.653  & hot                  & 1.443 &                    &       \\
\textbf{site}       & 13.971  & \textbf{suspect swine}  & 2.52   & fact                 & 1.413 &                    &       \\
\end{tabular}\)
\end{table}

Tables \ref{table_flu_TreeBagger_features_1grams}, \ref{table_flu_TreeBagger_features_2grams} and \ref{table_flu_TreeBagger_features_hybrid} show the most important features -- based on their $\delta$ -- for round 1 of the 5-fold cross validation in all the investigated feature classes, when the ensemble of CARTs is applied. Those results are directly comparable with the ones in Tables \ref{table_flu_1grams}, \ref{table_flu_2grams} and \ref{table_flu_hybrid} respectively; similarly to the previous section, the features that have also been selected by Bolasso are typed in bold. Again, the set of markers selected by the ensemble of CARTs has differences compared to one selected by Bolasso; still, the selected features are mostly related to the topic of flu or illness. The most important words are now directly related to the swine flu epidemic (`\emph{swine}' and `\emph{swine flu}'). In addition, 1-gram `\emph{irrig}' which had the highest weight when Bolasso was applied, is not within the most important features of the ensemble.

\subsection{Discussion}
\label{section_ensemble_of_carts_vs_Bolasso}
Overall, we have seen that in most subcases -- 4 out of 6 -- the ensemble of CARTs outperforms Bolasso. Still, for the rainfall rates inference task, the performance of both learners is quite similar and the best one is achieved from Bolasso's hybrid class; in the flu rates inference problem, we see that the ensemble of CARTs achieves a significantly better performance than Bolasso.

In terms of feature selection, we may conclude that the two learning functions do not select the same sets of features; for features selected by both learners weighting and/or ranking might also be different with high probability. As far as the semantic correlation of the selected features with the target topic is concerned, we see that Bolasso tends to select more `reliable' features. Particularly for the task of rainfall rates inference, even the hybrid class of the ensemble of CARTs has selected very irrelevant words, such as `\emph{sentenc}', `\emph{archiv}', `\emph{blind}', `\emph{manag}' and so on; that might have been the reason why Bolasso has outperformed it in that case. Nevertheless, the selected 2-grams by both learning methods seem to have a closer semantic correlation with the topics, reconfirming our initial assumptions (see Section \ref{section_recap_previous_limitations}).

Based on the fact that the the ensemble of CARTs gives out an importance ranking of a large amount of features and hence, a feature importance threshold must be defined manually, one might argue that Bolasso offers a more preferable (or natural) way to perform feature selection. On the other hand, as the ensemble seems to outperform Bolasso during the inference process, we could perhaps assume that this problem has nonlinear characteristics which cannot be incorporated in a linear learning function.

Another statistical fact that might also affect the inference performance of the schemes is the underlying distribution of the ground truth data; in the next section, we are examining this aspect, and show that rainfall and flu rates might be generated from different probability density functions (\textbf{PDF}s)\index{PDF}.

\begin{figure*}[tp]
    \begin{center}
    \subfigure[\emph{C. England} \& \emph{Wales} -- RMSE: 4.72] {\includegraphics[width=2.75in]{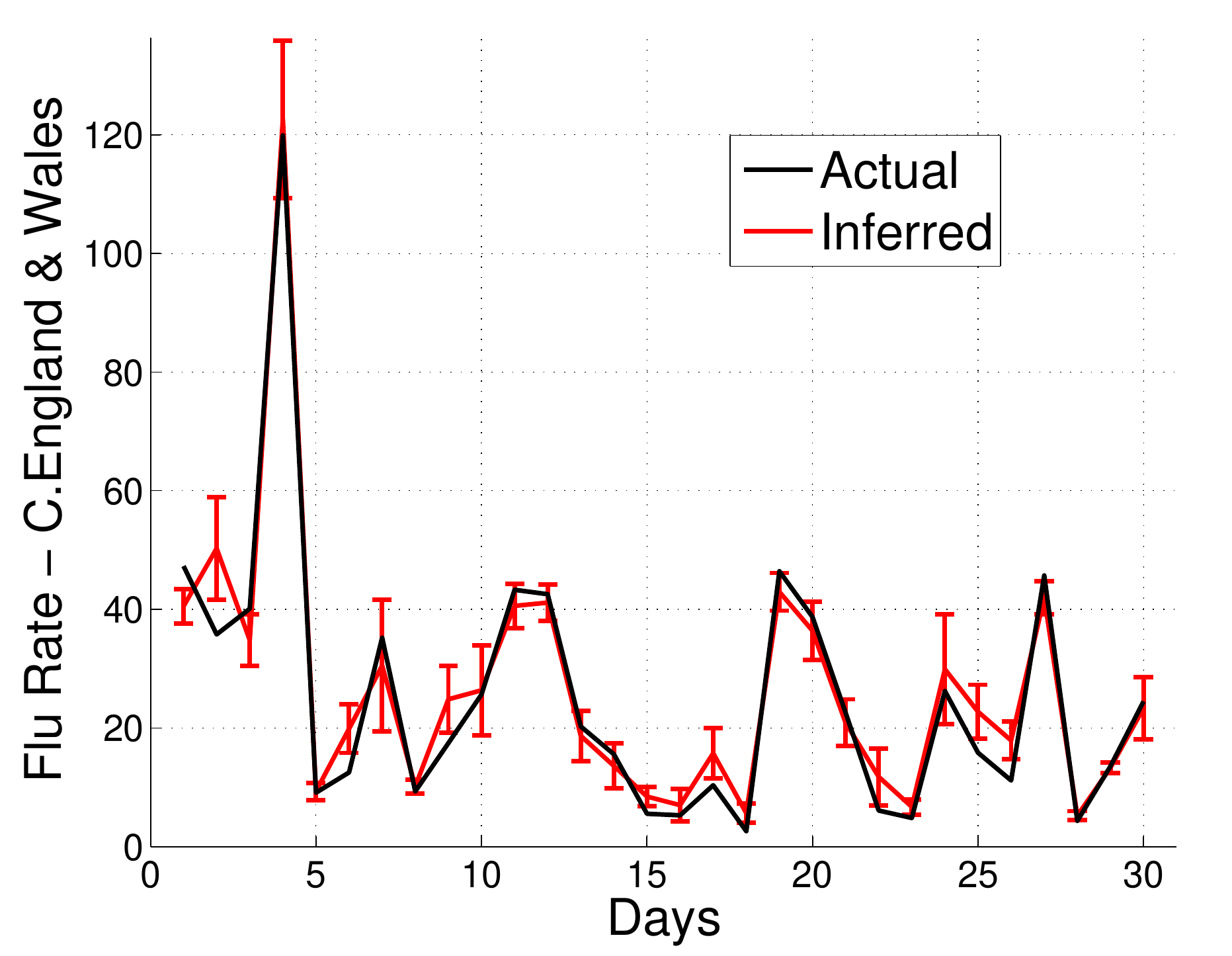}
    \label{fig_fluhybridCengW_treebagger_errorbars}}
    \hfil
    \subfigure[\emph{N. England} -- RMSE: 8.01] {\includegraphics[width=2.75in]{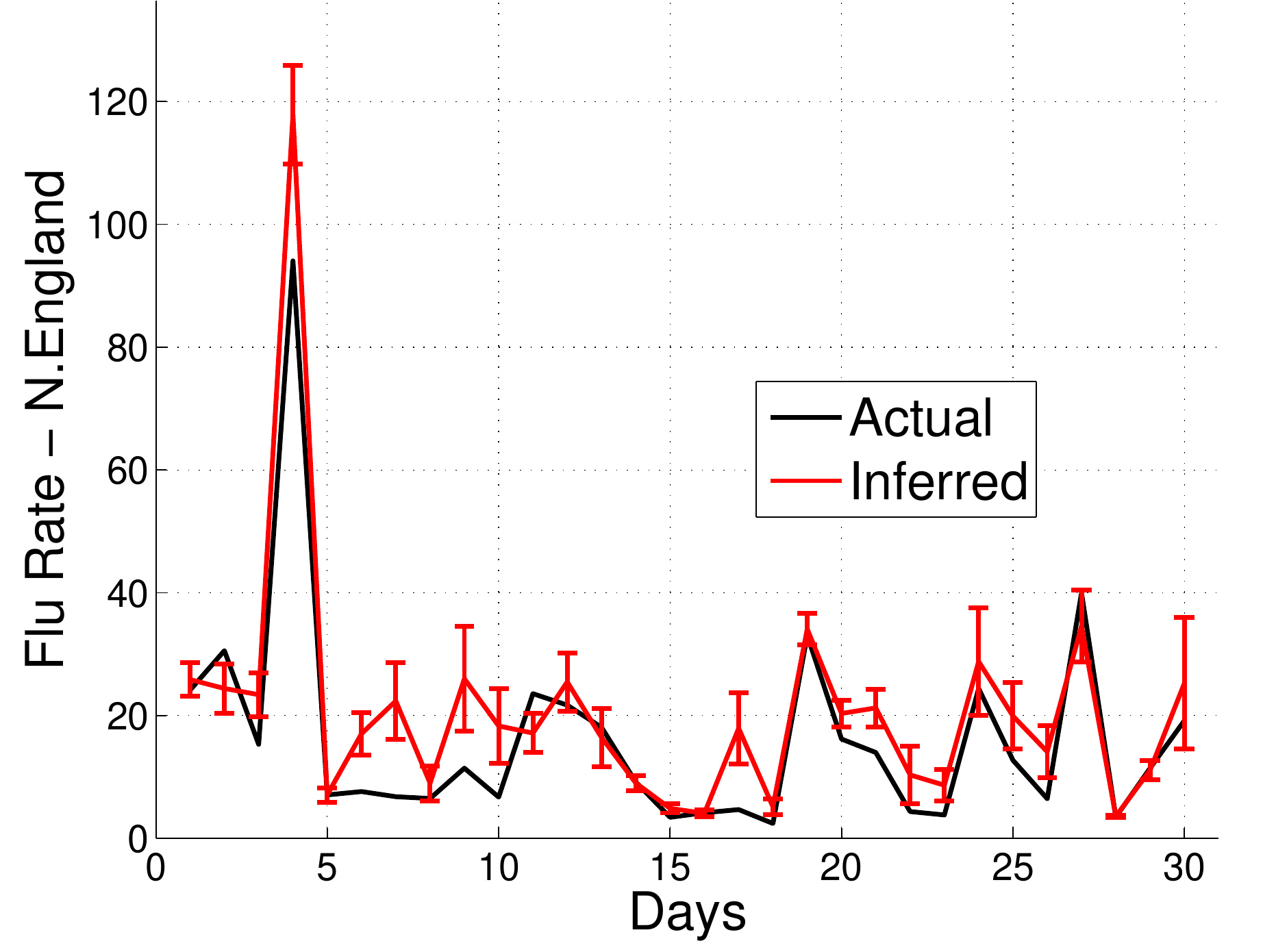}
    \label{fig_fluhybridNEng_treebagger_errorbars}}
    \hfil
    \subfigure[\emph{S. England} -- RMSE: 6.89] {\includegraphics[width=2.75in]{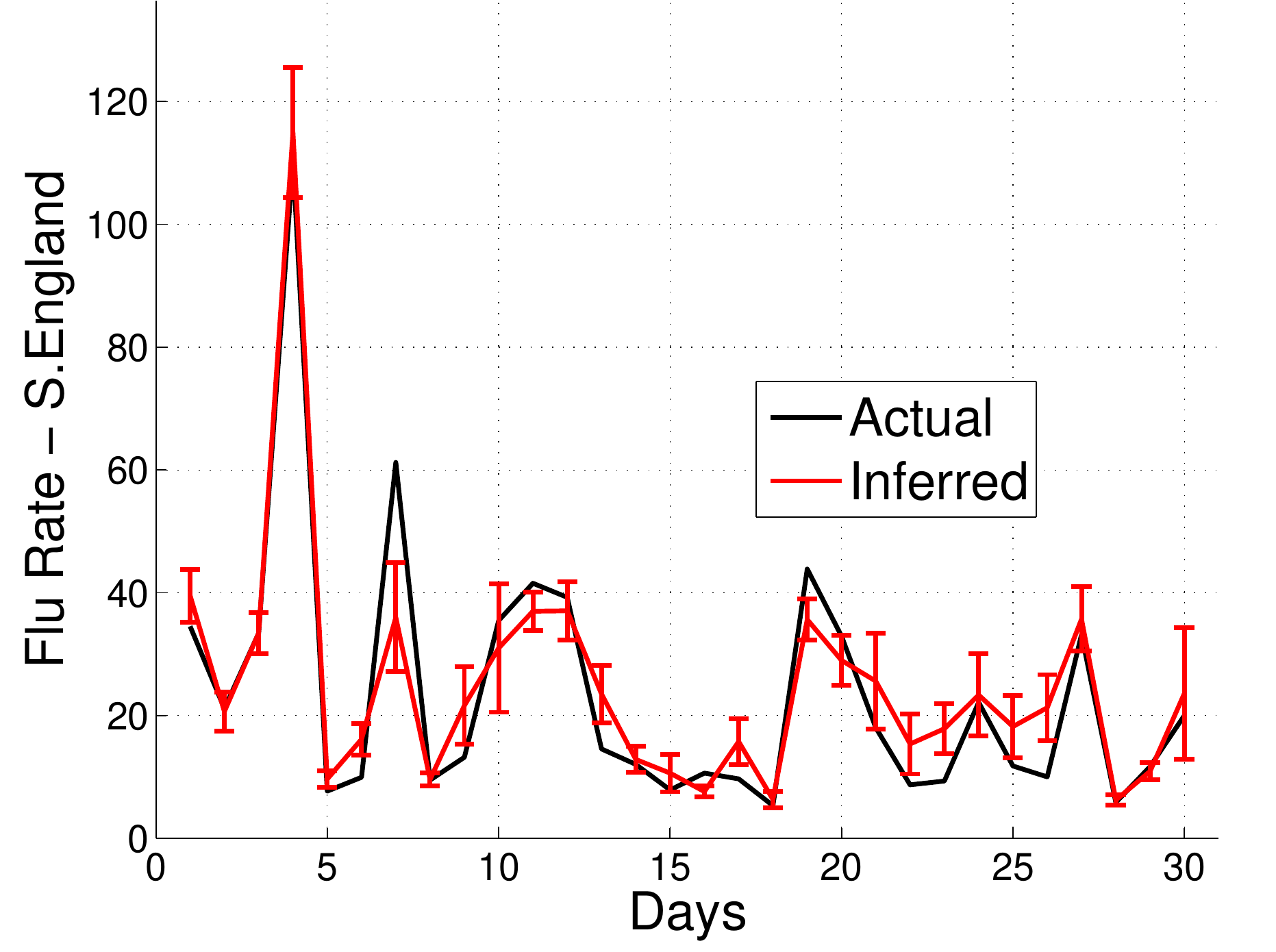}
    \label{fig_fluhybridSEng_treebagger_errorbars}}
    \end{center}
    \caption{Inference plots for the flu case study (1-grams \& 2-grams) using an ensemble of CARTs (Round 1 of 5-fold cross validation). Error bars depict 95\% CIs for the inferred values. This figure is comparable to Figure \ref{fig_fluhybrid}.}
    \label{fig_fluhybrid_treebagger_errorbars}
\end{figure*}

%%%%%%% revisions %%%%%%%%%%%%%%%%%%%%%%%
Finally, a general limitation of the inference methods used in our proposed models is that they only produce single inferred values without an associated spread. Spreads on the inferences -- usually indicated via confidence intervals (\textbf{CI}s)\index{confidence interval} -- may assist in identifying locations or regions where our models operate with high confidence and vice versa. One approach for tackling that task could be to approximate the SE of the inferences via the bootstrap samples in the ensemble of CARTs (see Algorithm \ref{algorithm_Bootstrap}). In this case, SE is equal to the standard deviation of the inferred values of each CART divided by the square root of the number of CARTs in the ensemble. A 95\% CI can then be computed by multiplying SE with the .975 quantile of the normal distribution. Figure \ref{fig_fluhybrid_treebagger_errorbars} shows inferences with error bars (95\% CIs) for Round 1 of the 5-fold cross validation for the flu case study. For the majority of samples CIs tend to be quite tight; we can also observe that CIs across the three considered regions behave similarly for the same daily time intervals (their average pairwise linear correlation is equal to 0.7369).

However, a possibly better way to address this task could be a model based on Gaussian processes. A Gaussian process\index{Gaussian process} can be seen as a collection of random variables, where it is hypothesised that any finite number those random variables can be described by a joint Gaussian distribution; it is completely defined by its mean and covariance function \cite{rasmussen2006gp}. Within the advantages of this framework lies not only the natural way of acquiring spreads for inferences in regression problems, but also the ability of blending various other parameters and information inputs in the model in a quite structured manner (by configuring the covariance function of the Gaussian process). Therefore, temporal as well as spatial characteristics of event detection process could also be explored resulting in probably better and definitely more general solutions.
%%%%%%%%%%%%%%%%%%%%%%%%%%%%%%%%%%%%%%%%%

%%%%%%% revisions %%%%%%%%%%%%%%%%%%%%%%%
%%%%%%%%%%%%%%%%%%%%%%%%%%%%%%%%%%%%%%%%%

\section{Properties of the target events}
\label{section:properties_of_target_events}
In this chapter, we have proposed a methodology which is capable of using Twitter content to nowcast rainfall\index{rainfall rates} and flu rates\index{flu rates} in the UK. The purpose of this section is to identify the key properties of those two target events. We assume that the proposed theoretical framework will be -- at least -- also applicable for inferring the magnitude of other events with similar characteristics. Section \ref{section_discussion_nowcasting_journal_paper} has already given an initial insight on the main similarities and differences between the random variables of rainfall and flu rates; it has been shown that both are quantities that oscillate, but rainfall tends to do so with a higher frequency, whereas flu rates -- in our data set -- show a much more smooth behaviour.

\begin{figure*}
    \begin{center}
    \subfigure[Rainfall rates] {\includegraphics[width=2.85in]{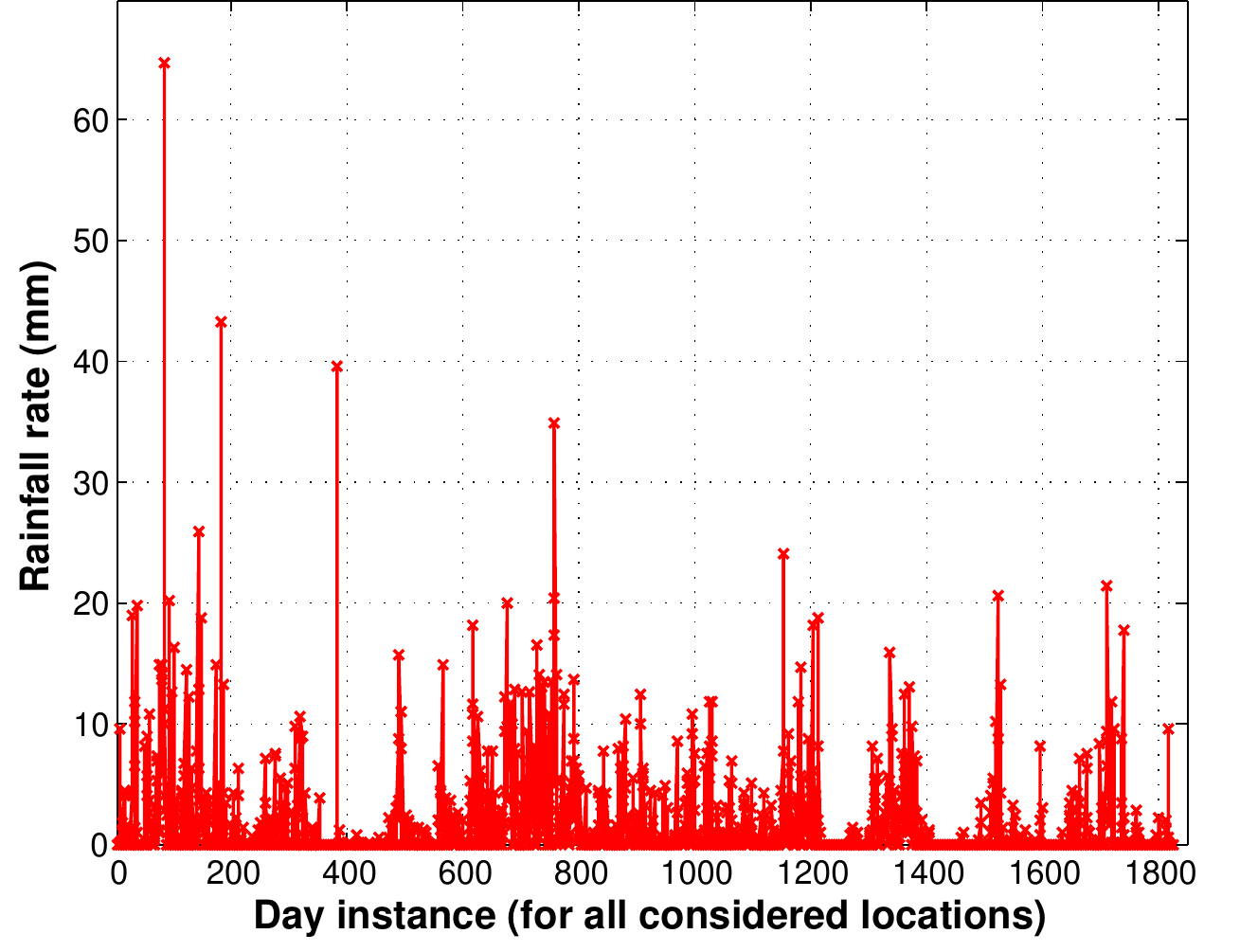}
    \label{fig_rain_symmetric}}
    \hfil
    \subfigure[Flu rates] {\includegraphics[width=2.85in]{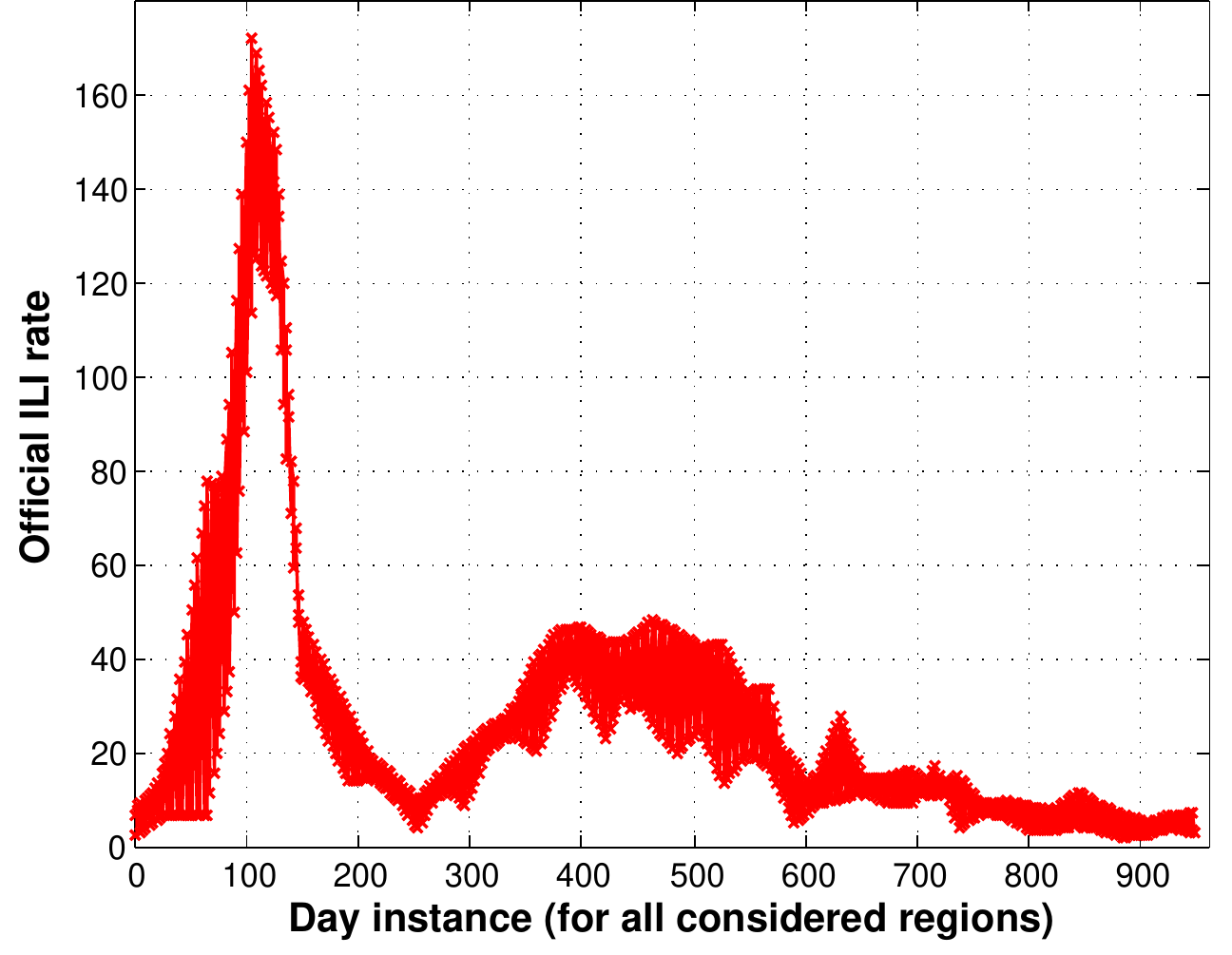}
    \label{fig_flu_symmetric}}
    \end{center}
    \caption{Ground truth figures by concatenating the data for all locations (rainfall) or all regions (flu) in a chronological order.}
    \label{fig_all_ground_truth}
\end{figure*}

Figure \ref{fig_all_ground_truth} holds the plots for the entire sets of ground truth points for rainfall (\ref{fig_rain_symmetric}) and flu (\ref{fig_flu_symmetric}) rates. In each data set we have concatenated the scores of different locations or regions (respectively for rainfall and flu) in a chronological order. Rainfall's mean is equal to 1.7573mm, standard deviation to 3.912mm, the minimum and maximum values are 0mm and 64.8mm respectively and the median is equal to 0mm. Likewise the mean flu rate is 26.0334 (ILI-diagnosed patients per 100,000 citizens), its standard deviation 28.8542, minimum and maximum 2 and 171.9 respectively and the median is equal to 15.8571. Figure \ref{fig_all_ground_truth} reconfirms the argument of the previous paragraph; rainfall rates are -- in general -- much more unstable than flu rates and many times are equal to 0 (with also a zero median), whereas flu rates are never become 0 in our data set.

\begin{figure*}
    \begin{center}
    \subfigure[CDF for rainfall rates] {\includegraphics[width=2.85in]{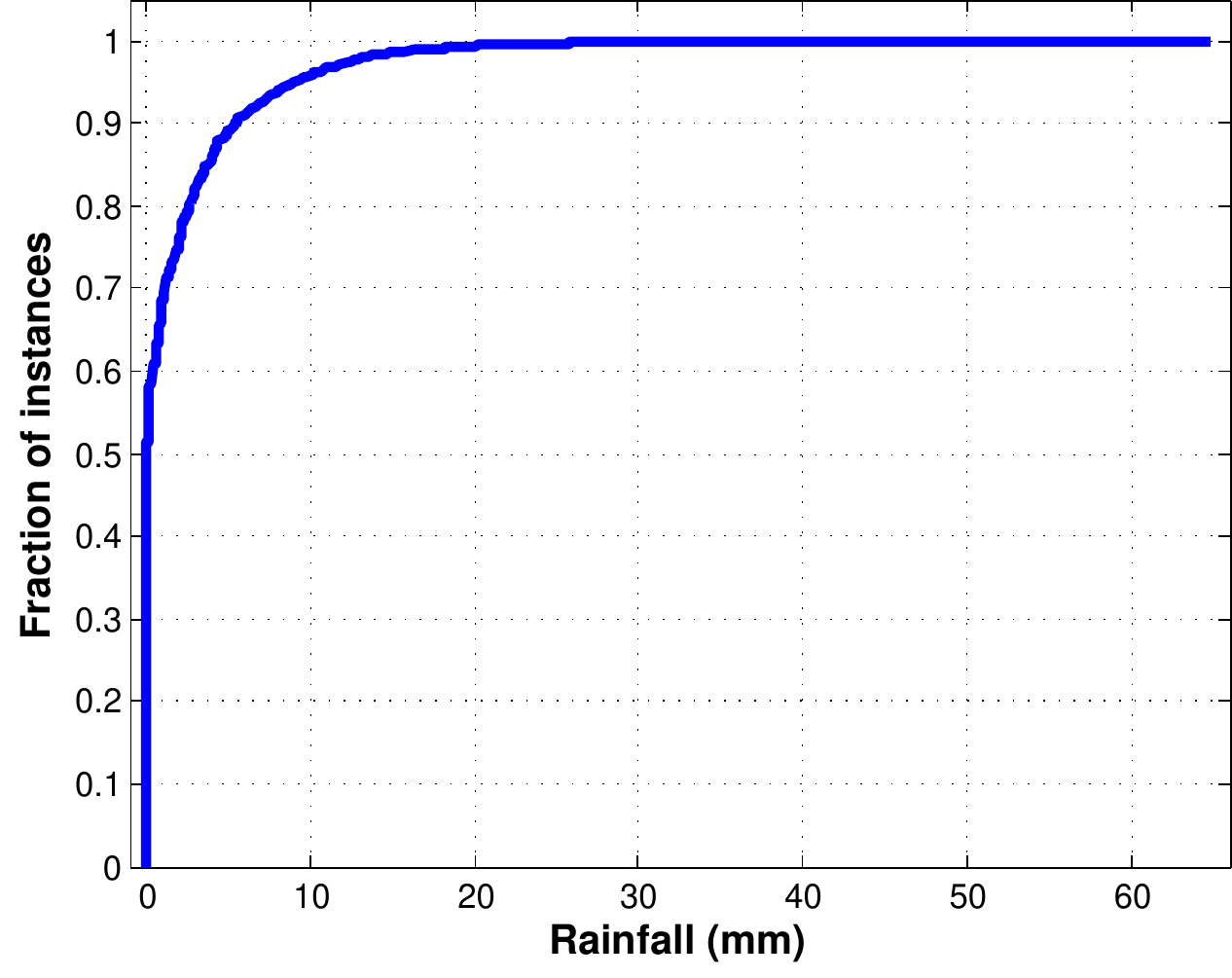}
    \label{fig_CDF_rain}}
    \hfil
    \subfigure[CDF for flu rates] {\includegraphics[width=2.85in]{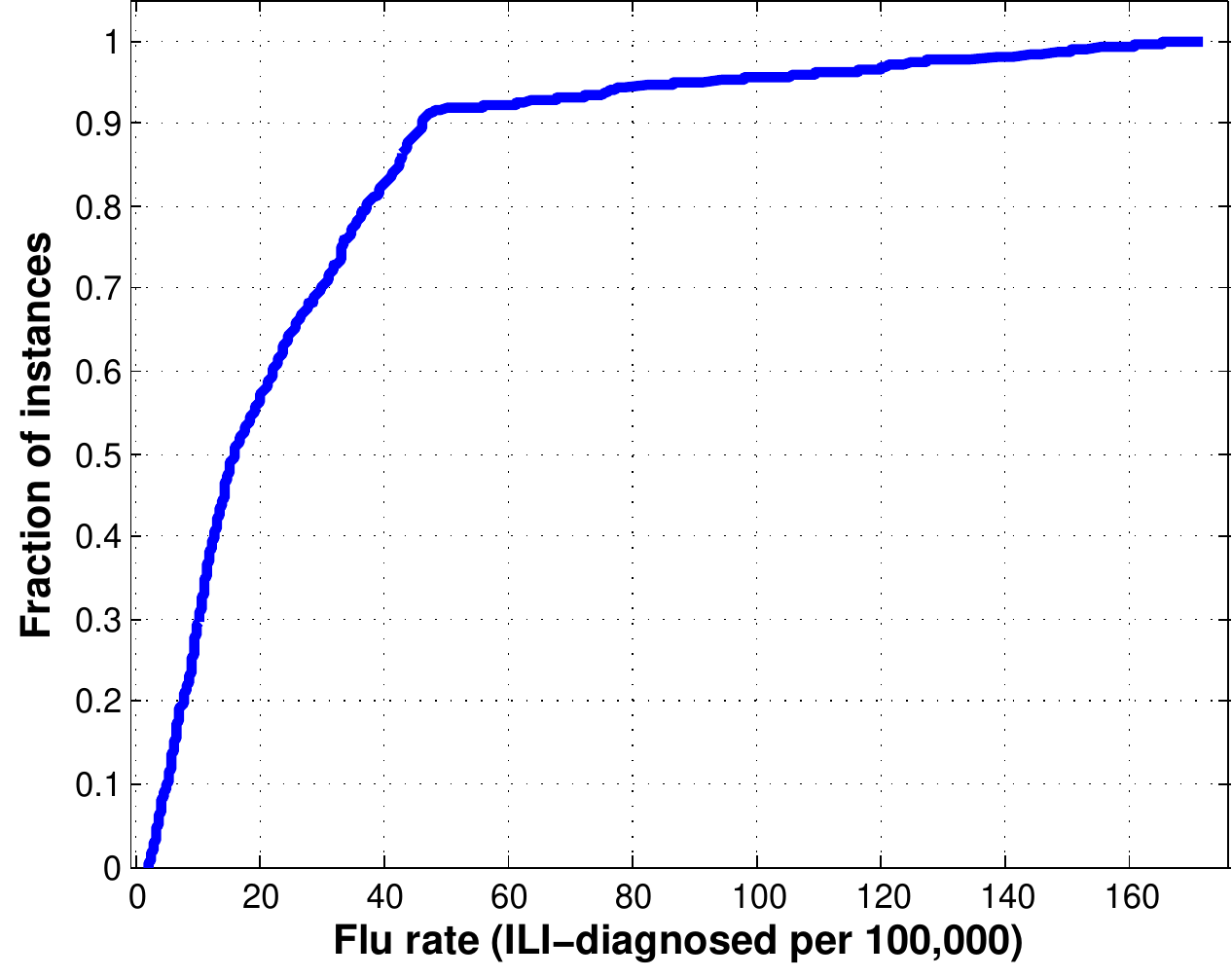}
    \label{fig_CDF_flu}}
    \end{center}
    \caption{Empirical CDFs for rainfall and flu rates}
    \label{fig_CDF_rain_flu}
\end{figure*}

Figure \ref{fig_CDF_rain_flu} shows the empirical cumulative density functions (\textbf{CDF})\index{CDF} for the two random variables. For a point `x' (on the x-axis) the corresponding value of the plot on the y-axis denotes the ratio of the observations which are smaller than `x' over the total number of observations. From Figure \ref{fig_CDF_rain} it is made quite obvious that approx. 60\% of the rainfall rates are equal to 0 and 95\% is below 10mm. Likewise from Figure \ref{fig_CDF_flu} we can see that only 10\% of the sample lies between 45 and 172 (ILI-diagnosed patients per 100,000 citizens). Therefore, both events seem to be inactive for the majority of time, and tend to acquire notable values for short periods only.

\begin{figure*}
    \begin{center}
    \subfigure[Rainfall rates -- Histogram vs. Exponential distribution]{\includegraphics[width=2.85in]{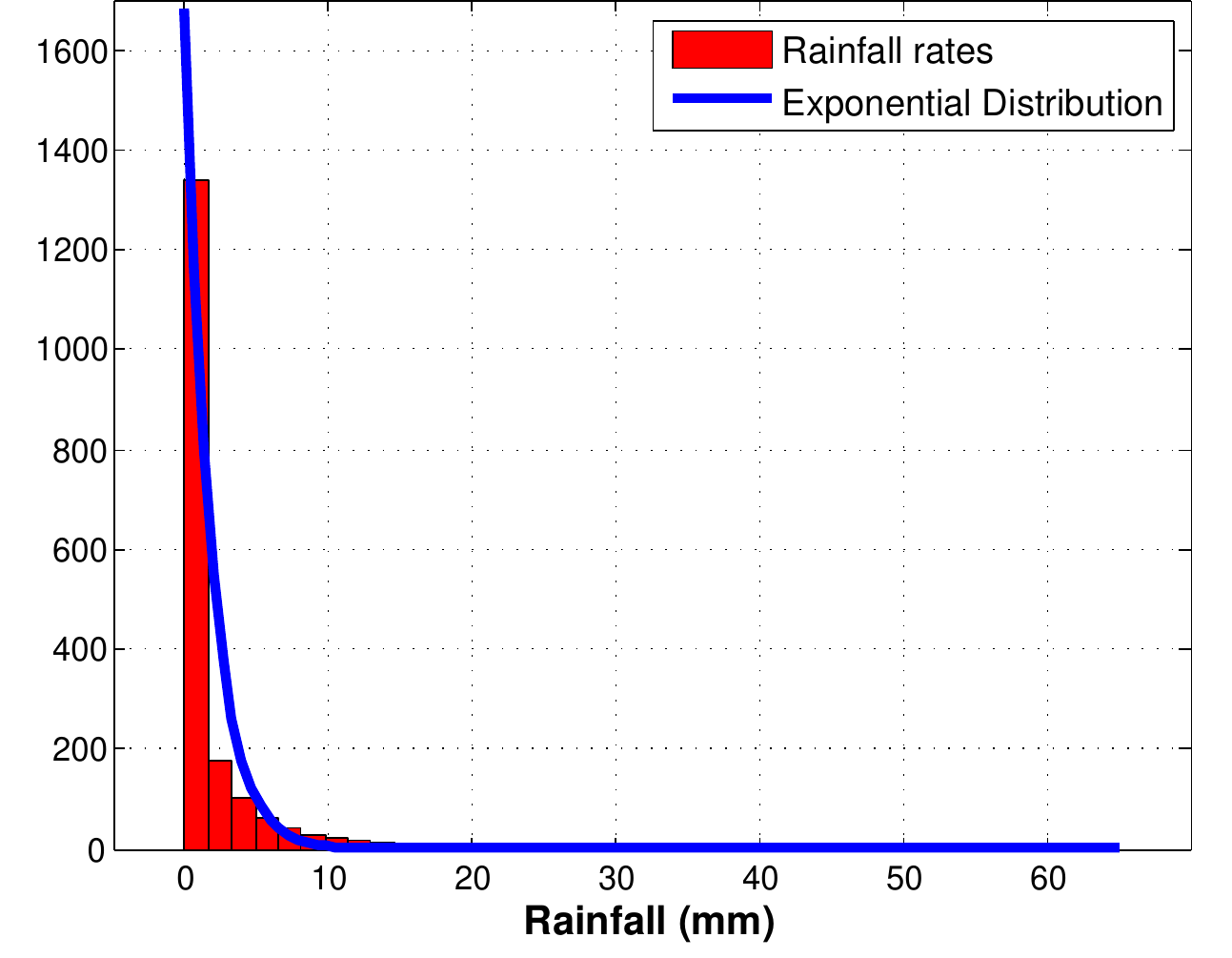}
    \label{fig_PDF_rain}}
    \hfil
    \subfigure[Flu rates -- Histogram vs. Exponential vs. Log-normal vs. Gamma distribution] {\includegraphics[width=2.85in]{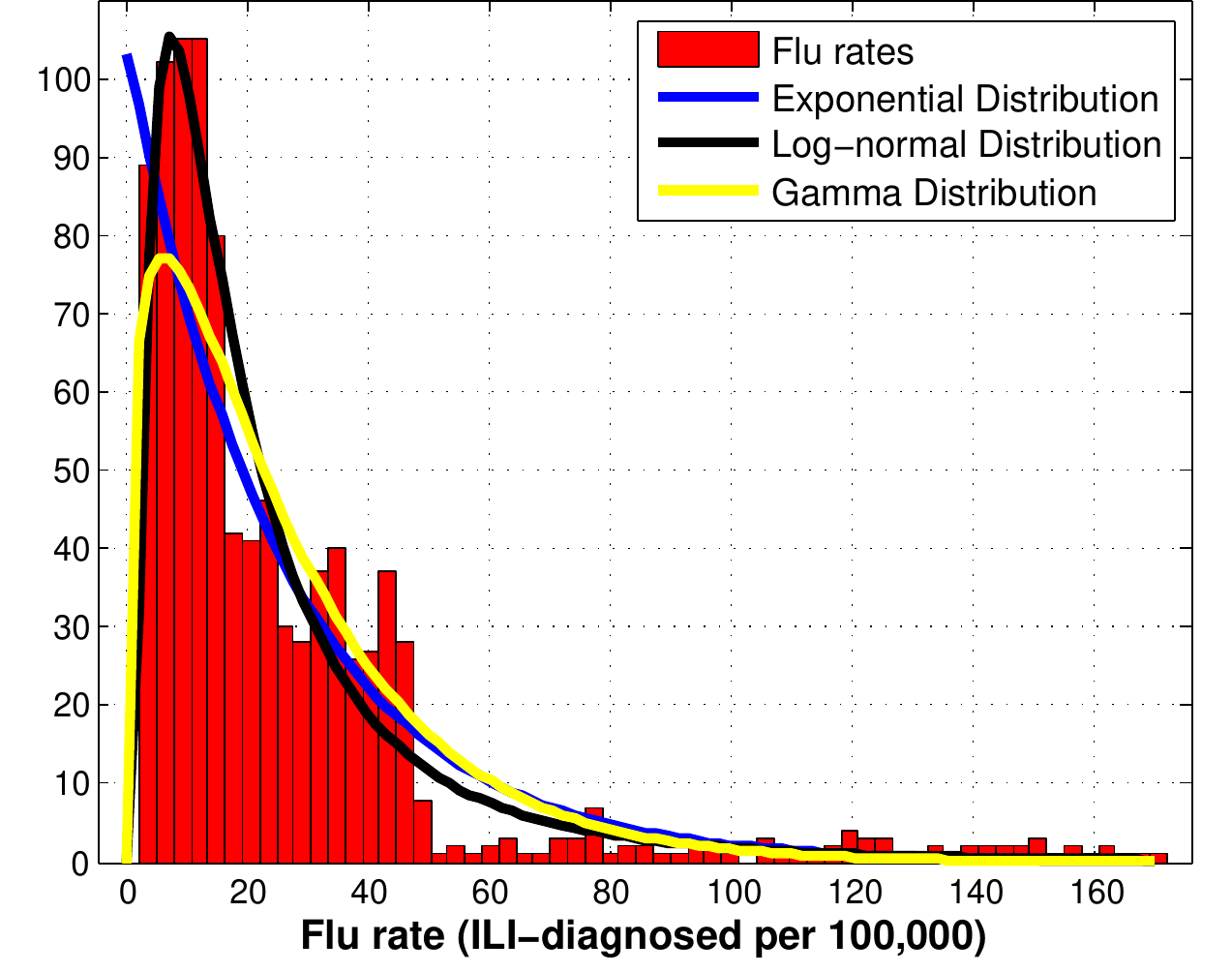}
    \label{fig_PDF_flu}}
    \end{center}
    \caption{Histograms and PDF approximations for rainfall and flu rates}
    \label{fig_PDF_rain_flu}
\end{figure*}

Finally, we plot the histograms for both variables and try to fit some commonly applied PDFs\index{PDF} to them (Figure \ref{fig_PDF_rain_flu}). The histogram derived by the rainfall rates points out that the data is most likely to follow an exponential distribution; the best fit occurred for an exponential PDF with $\lambda =$ 0.569 (Figure \ref{fig_PDF_rain}). However, flu rates -- based on the fact that zero or very small values were not the most dominant class -- did not show the best fit with an exponential density function. Figure \ref{fig_PDF_flu} shows their histogram as well as the fits for three PDFs, an exponential ($\lambda =$ 0.0384), a log-normal ($\mu =$ 2.82451 and $\sigma =$ 0.9254) and a gamma ($\kappa =$ 1.29098 and $\theta =$ 20.1657). We see that the log-normal PDF offers the best fit with the data, meaning that the logarithm of the samples can be approximated by a normal (Gaussian) distribution.

Summarising the results above, we have seen that our method has been able to track two events that can be approximated by an exponential (rainfall) and a log-normal (flu) distribution. The common ground for both events is that they tend to oscillate -- with very different frequencies though --, and that in most occasions and under their metric schemes, they tend to be assigned with small values. Therefore, those targets events are inactive or can be considered as inactive for the majority of time, and become active -- or, more precisely, take significant values -- for shorter periods of time. As the number of distinguishable peaks increases, we expect the event-tracking process to be more feasible (recall the `Harry Potter effect' in Section \ref{section_harry_potter_effect}); by also comparing the semantic correlation with the target topic between the selected features for rainfall (stronger), where more peaks are present, and the ones for flu (less strong), we could argue in favour of such a hypothesis.

\section{Summary of the chapter}
\label{section_summary_chapter_nowcasting}
In this chapter, we presented a general methodology for inferring the occurrence and magnitude of an event or phenomenon by exploring the rich amount of unstructured textual information on the social part of the web -- in particular, Twitter; our aim was to improve on several aspects of the methodology in \cite{Lampos2010f} and Chapter \ref{Chapter_first_steps}. The specific procedures that we followed, namely soft-Bolasso with consensus threshold validation for feature selection from a large set of candidates or an ensemble of CARTs, have been proven to work significantly better compared to another relevant state-of-the-art approach \cite{ginsberg2008detecting}, but the general claim is that statistical learning techniques can be deployed for the selection of features and, at the same time, for the inference of a useful statistical estimator. Several feature classes have been applied, 1-grams, 2-grams and hybrid combinations of them; the combination of 1-grams and 2-grams gave out the best inference performance, whereas the selected 2-grams formed a better semantic representation of the target event.

The overall inference performance of the ensemble of CARTs, which is a nonlinear learning function, was better than the one derived from Bolasso; however feature selection alone was more naturally performed under Bolasso. Comparisons with even more variants of Machine Learning methods are surely of interest, though they would not change the main message: that one can learn the estimator from data, by means of supervised learning. In the case of ILI, other methods (\emph{e.g.} \cite{corley2009monitoring,polgreen2008using}) propose to simply count the frequency of the disease name. This can work well when people can diagnose their own disease (maybe easier in some cases than others) and no other confounding factors exist. However, from our experimental results, one can conclude that this is not an optimal choice.

Furthermore, it is not obvious that a function of Twitter content should correlate with the actual health state of a population or rainfall rates. There are various possible sampling biases that may prevent this signal from emerging. An important result of this study is that we find that it is possible to make up for any such bias by calibrating the estimator on a large data set of Twitter posts and actual HPA readings or precipitation measurements. While it is true that Twitter users do not represent the general population and Twitter content might not represent any particular state of theirs, we find that actual states of the general population (health or weather oriented) can be inferred as a linear or a nonlinear function of the signal in Twitter.

Lastly, some insight on the characteristics of the target events was given. Flu rates are better described by a log-normal PDF, whereas rainfall rates show an exponential behaviour. We hypothesise that our methods are generic and therefore able to nowcast events with similar characteristics.

%% file: Chapters/Chapter6.tex
\chapter{Detecting Temporal Mood Patterns by Analysis of Social Web Content}
\label{chapter:detecting_temporal_mood_patterns}

\rule{\linewidth}{0.5mm}
In this chapter, we look into methods that are able to extract affective norms\index{affective norm} from the content of Twitter. In particular, we are focusing on four mood types, namely anger, fear, joy and sadness and compute seasonal as well as average circadian and daily mood scores. Circadian mood patterns are used to explain behavioural patterns throughout the day; seasonal characteristics are also identified by comparing the patterns of winter and summer. We also show that daily mood patterns can be correlated with significant events or annual celebrations emerging in real-life. This chapter has been partially based on content or ideas from our papers ``Detecting Temporal Mood Patterns in the UK by Analysis of Twitter Content'' \cite{Lampos2012p}\footnote{ This is a joint work (names of the co-authors are listed in Section \ref{section:intro_publications}); V. Lampos contributed in all aspects of the research and also wrote the paper.} and ``Effects of the Recession on Public Mood in the UK'' \cite{Lansdall-Welfare2012}.\footnote{ Based on the fact that V. Lampos was not the first author of \cite{Lansdall-Welfare2012}, we do not cover the exact results and the main topic of this publication, but focus on (as well as extend) the methods and outcomes, where he had a major contribution.}
\newline \rule{\linewidth}{0.5mm}
\newpage

\section{Detecting circadian mood patterns from Twitter content}
\label{section:detecting_circadian_patterns}
%Patterns of circadian and seasonal changes of mood in the UK general population were estimated based on the content of twitter messages. We analysed two seasonal sets of messages (120 million in total) captured throughout the day during 24 weeks of data collection divided equally between winter and summer. Our aim was to estimate the hourly and seasonal variations in the use of emotionally loaded words contained in twitter messages, possibly reflecting underlying circadian and seasonal rhythms.
%
%\noindent\textbf{--- SUMMARY WILL BE BASED ON THE DISCUSSION SECTION ---}\\
%TO BE COMPLETED\\
%\textbf{--- END OF --}
%
Social media, particularly the micro-blogging website `Twitter', provide a novel way to gather real time data in large quantities directly from the users. This data, which is also time-stamped and geolocated, can be analysed in various ways to examine patterns in a wide range of subjects. Several methods have been already proposed for exploiting this rich information in order to detect events emerging in the population \cite{Lampos2011a}, track a possible epidemic disease \cite{Lampos2010,Signorini2011} or even infer the result of an election \cite{tumasjan2010predicting}.

Mood seems to change throughout the day, spontaneously or in response to life's vicissitudes. Nonetheless exploring these variations in real time and with large samples is difficult. The use of information gathered via Twitter messaging permits overcoming some of these obstacles. A better knowledge of mood variations is important to confirm widely accepted notions linking certain psychiatric phenomena to diurnal \cite{Kronfeld-Schor2012} or even seasonal patterns \cite{Grimaldi2009}. Recent studies \cite{Golder2011,Lansdall-Welfare2012} using information captured from Twitter messaging found circadian and seasonal patterns in the content of emotionally loaded wording. As for circadian variation, the volume of words representing negative affect reached its lowest point in the mornings and improved steadily throughout the day. This finding is contrary to the clinical concept of diurnal variation of mood among depressed patients, with the highest load of depressive symptoms early in the mornings. Seasonal patterns, in particular an increase of depressive symptoms in winter months, were not confirmed either \cite{Golder2011}. However, this study showed seasonal changes in Positive Affectivity\index{Positive Affectivity|see{PA}}\index{PA} (\textbf{PA}) and speculated that ``winter blues'' may be associated with diminished PA rather than increased Negative Affectivity\index{Negative Affectivity|see{NA}}\index{NA} (\textbf{NA}).

We used a similar methodology to gather daily data during two different seasons in our effort to explore circadian and seasonal patterns of mood variation among Twitter users in the UK.

\subsection{Data and methods}
\label{section:methods}
We assessed approximately 120 million tweets, 50 and 70 million for winter and summer periods respectively. For each of the 54 most populated urban centres in the UK, we periodically retrieved the 100 most recent tweets, geolocated to within a 10km range of an urban centre. We attempted to reduce any potential sampling bias by applying the same daily sampling frequency per urban centre. Data was collected for 12 weeks during each season (06/12/2010 to 28/02/2011 for winter and 06/06/2011 to 28/08/2011 for summer). We used a stemmed version of the text in tweets by applying Porter's Algorithm \cite{porter1980}. The tweets were divided into 24 bins, one for each hour of the day, and the emotional valence of each hour was assessed by a text analysis tool. We estimated the level of activity for four emotions -- namely fear, sadness, joy and anger -- based on the contents of Twitter messages and on emotion-related words in the text.

The emotion-related word-lists were retrieved -- a priori -- from WordNet Affect \cite{Strapparava2004}, a tool that builds on WordNet \cite{Miller1995} by selecting and labelling synsets, which represent affective concepts. The word-lists were further processed via stemming and filtering in a way that only single words are kept in the end. After this preprocessing, they were formed by 146 (anger), 91 (fear),\footnote{ The word `alarm' was removed from the word list for fear, since it showed a strong correlation (summer: 0.8774; winter: 0.8662) with the occurrence of the word `clock', resulting in a peak when people wake up.} 224 (joy) and 115 (sadness) stemmed words respectively (see Appendix \ref{Ap:detecting_temporal_mood_patterns}). WorNet Affect as well as stemming have been regularly applied in other emotion-detection and text mining applications and thus, are considered as standard ways for this type of information retrieval \cite{Strapparava2008,Calvo2010}.

\subsubsection{Mean Frequency Mood Scoring}
\label{section_MFMS_circadian}
To extract a circadian (24-hour) pattern for the considered mood types, two approaches -- each one based on different assumptions -- have been applied. Suppose that $n$ terms are used to track an emotion over $M = D \times 24$ hourly time intervals, where $D$ is the total number of days considered and each time interval is denoted by $t_{d,h}$, where $d\in\{$1, ..., $D\}$ and $h\in\{$1, ..., 24$\}$. The collected number of tweets for a time interval is denoted by $N(t_{d,h})$. For every $t_{d,h}$ we count the appearances of each term in the Twitter corpus, $c_i^{(t_{d,h})}$, where $i\in\{$1, ..., $n\}$. Those `hourly' term counts are normalised by being divided with the respective number of tweets collected in that time interval and therefore, instead of working with raw counts, we now derive an hourly frequency $f_i^{(t_{d,h})}$ for each term, \ie $\displaystyle f_i^{(t_{d,h})} = \frac{c_i^{(t_{d,h})}}{N(t_{d,h})}$. Next, we compute the mean frequency $F(t_{d,h})$ of the $n$ terms for each time interval $t_{d,h}$:
\begin{equation}
F(t_{d,h}) = \frac{1}{n}\sum_{i=1}^{n} f_i^{(t_{d,h})}.
\end{equation}
Finally, the mood score $\mathcal{M}(h)$ for an hourly interval $h$ is given by averaging across all days in the data set:
\begin{equation}
\mathcal{M}(h) = \frac{1}{D} \sum_{j=1}^{D} F(t_{d,h}) = \frac{1}{D} \sum_{j=1}^{D}\left(\frac{1}{n} \sum_{i=1}^{n} \frac{c_i^{(t_{j,h})}}{N(t_{j,h})} \right).
\end{equation}
This method is based on the assumption that the frequency of a word indicates its importance. Words with higher frequencies will have a larger impact on the computed means and therefore, in the final mood score. We refer to this approach as Mean Frequency Mood Scoring\index{Mean Frequency Mood Scoring|see{MFMS}}\index{MFMS} (\textbf{MFMS}).

\subsubsection{Mean Standardised Frequency Mood Scoring}
In the second mood scoring approach, after computing the hourly term frequencies $f_i^{(t_{d,h})}$ ($i\in\{$1, ..., $n\}$, $d\in\{$1, ..., $D\}$ and $h\in\{$1, ..., 24$\}$), we form $n$ vectors $W_i$ of size $M$ which hold the frequencies of each word in all time intervals, \ie $W_i = \{f_i^{(t_{1,1})},...,f_i^{(t_{1,24})},...,f_i^{(t_{D,24})}\}$, $i\in\{$1, ..., $n\}$. Next, we standardise $W_i$'s by first subtracting their mean ($\bar{f}_i$) from each entry and then by dividing with their standard deviation ($\sigma_{f_i}$):
\begin{equation}
sf_{i}^{(t_{d,h})} = \frac{f_i^{(t_{d,h})} - \bar{f}_i}{\sigma_{f_i}}\mathbf{, \emph{i} }\in\{\text{1}, ..., n\}.
\end{equation}
In that way we `replace' all frequencies with their standardised versions. The following steps of the method remain the same as before, and hence the `standardised' mood score $\mathcal{M}_{s}(h)$ for an hourly interval $h$ is given by:
\begin{equation}
\mathcal{M}_{s}(h) = \frac{1}{D} \sum_{j=1}^{D}\left(\frac{1}{n}\sum_{i=1}^{n}sf_{i}^{(t_{j,h})}\right).
\end{equation}
We refer to this approach as Mean Standardised Frequency Mood Scoring\index{Mean Standardised Frequency Mood Scoring|see{MSFMS}}\index{MSFMS} (\textbf{MSFMS}). MSFMS is now based on the assumption that each mood-word has an equally-weighted contribution to the final mood score.

\subsection{Experimental results}
\label{section:results}
Figures \ref{fig_MFMS_circadian} and \ref{fig_MSFMS_circadian} show the 24-hour circadian pattern for the mood types of anger, fear, joy and sadness extracted by applying MFMS and MSFMS respectively. For each emotion type, we have computed the circadian pattern for winter and summer as well as the one of the aggregated data set. In all figures, we have also included 95\% CIs\index{confidence interval} using a faded colour. The CIs have been approximated by multiplying the SE\index{SE} of the sample mean with the .975 quantile of the normal distribution, which is assumed to represent the mood scores of each time interval across all considered dates.

Table \ref{table_winter_summer_mood_corr} shows the linear correlation coefficients for the inferred circadian patterns between winter and summer under both scoring schemes. Joy is the most correlated emotion in both metrics with correlations higher than 0.92; there is high positive correlation for the remaining emotions as well, pointing out that the overall mood trend does not change significantly between seasons, but might have specific deviations of a smaller impact, which are pointed out and discussed in the following sections.

\subsubsection{Testing the stability of the extracted patterns}
A statistical significance test has been performed on all the inferred average circadian rhythms testing their stability across the daily samples. Consider an emotion type and let its average circadian mood pattern -- based on all relevant data -- be denoted by a vector $v$. For every day $i\in\{$1, ..., $D\}$ in the data set, we compute its circadian mood pattern $d_i$ and then, compute $d_i$'s linear correlation with $v$, $\rho(v,d_i)$. We form $k =$ 1,000 random permutations of $d_i$, compute their linear correlation with $v$ and count the number of times ($N_i$) that the pseudo-random correlation is higher or equal to $\rho(v,d_i)$. The p-value of this statistical test is, therefore, given by:
\begin{equation}
\mathbf{p-value} = \frac{1}{D} \sum_{i = 1}^{D} \frac{N_i}{k}.
\end{equation}
We consider a mood pattern as stable, when its p-value is below 0.05.

%% discuss here on the stat significance
Based on this statistical significance test, the circadian patterns (on the aggregated data sets) derived by applying MFMS for anger, fear and sadness are slightly above the 0.05 threshold, and hence they might not be very stable representations; patterns derived by applying MSFMS (on the aggregated data sets) are all statistically significant. As the proof for instability is not very strong and considering the fact that the scoring schemes are making different assumptions on the weighting of mood terms, we present the extracted signals for both scoring schemes, but investigate further (in terms of peak stability, periodicity and negative versus positive affectivity) only the ones derived by MSFMS.

\subsubsection{Circadian patterns by applying MFMS}
\begin{figure}[tp]
    \begin{center}
    \includegraphics[width=6in]{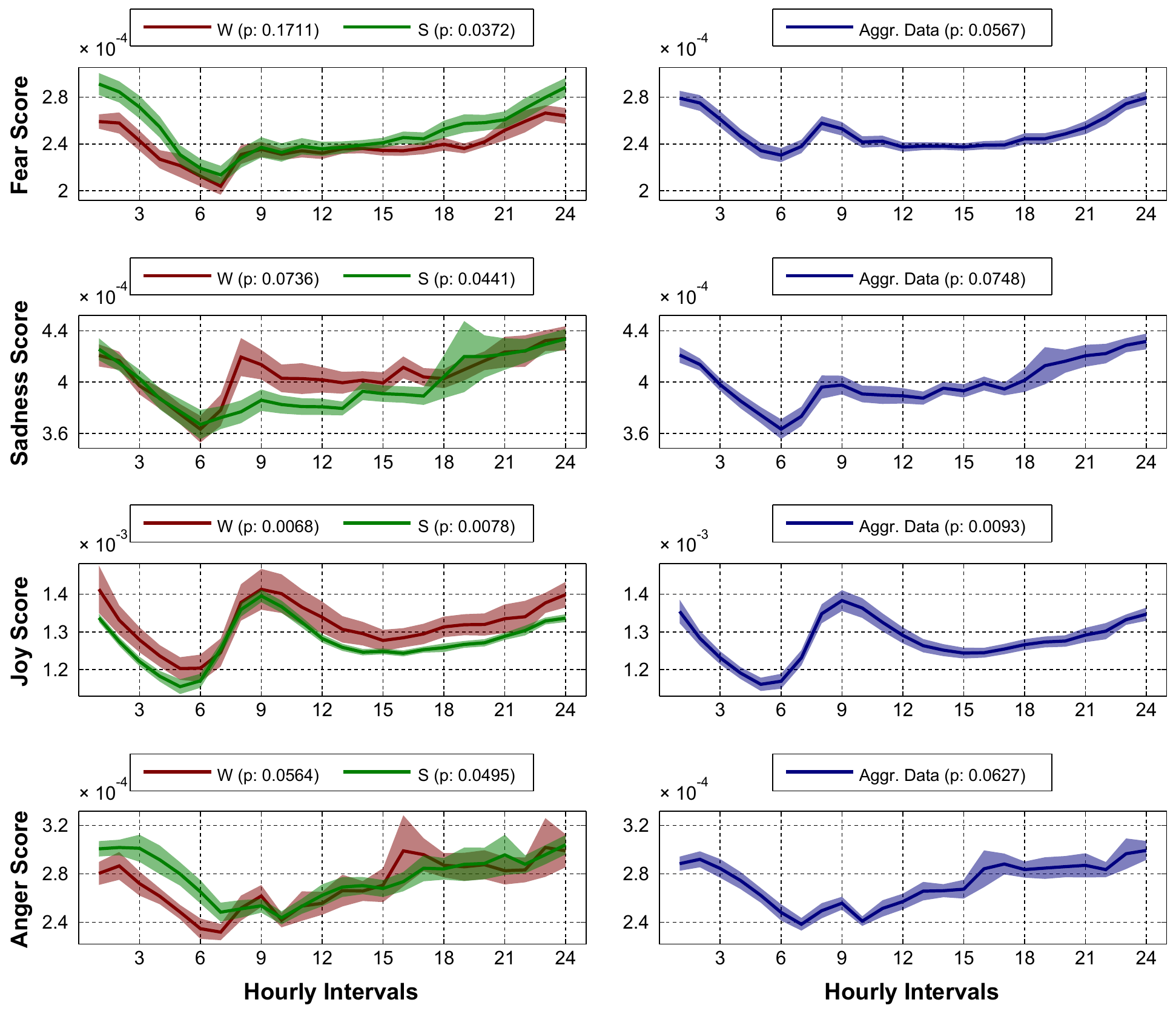}
    \end{center}
    \caption{These plots represent the variation over a 24-hour period of the emotional valence for fear, sadness, joy and anger obtained by applying MFMS. The red line represents days in the winter, while the green one represents days in the summer. The average circadian pattern was extracted by aggregating the two seasonal data sets. Faded colourings represent 95\% CIs on the sample mean. For all displayed patterns the corresponding stability p-values are being reported.}
    \label{fig_MFMS_circadian}
\end{figure}

In this section, we describe the circadian rhythms based on MFMS\index{MFMS} depicted in Figure \ref{fig_MFMS_circadian}. Wee see that fear has a first peak in the morning (7--8a.m.), then drops and from 3p.m. starts increasing again reaching a maximum during 11--12p.m.; the seasonal patterns are highly correlated with the summer rates being slightly higher in all hourly intervals (apart from 7--8a.m.). Sadness has a minimum at 5--6a.m., an intermediate peak at 8--9a.m. and then steadily increases reaching a maximum before midnight; a noticeable difference between the seasons is that winter's morning peak happens one hour earlier than the summer's one and is much stronger. Joy peaks in the morning (8--9a.m.), is decreasing until afternoon (4p.m.), then increases again until 12--1a.m., where it starts to drop reaching a minimum at 4--5a.m.; the circadian pattern is similar among the two seasons, with winter having slightly increased levels of joy. Anger reaches a minimum during the interval 6--7a.m. and then increases until midnight where it peaks; in the summer, we observe increased rates during the late night, while in the winter anger rates are higher in the mornings (7--10a.m.) and during afternoon (2--6p.m.).

\subsubsection{Circadian patterns by applying MSFMS}
\begin{figure}[tp]
    \begin{center}
    \includegraphics[width=6in]{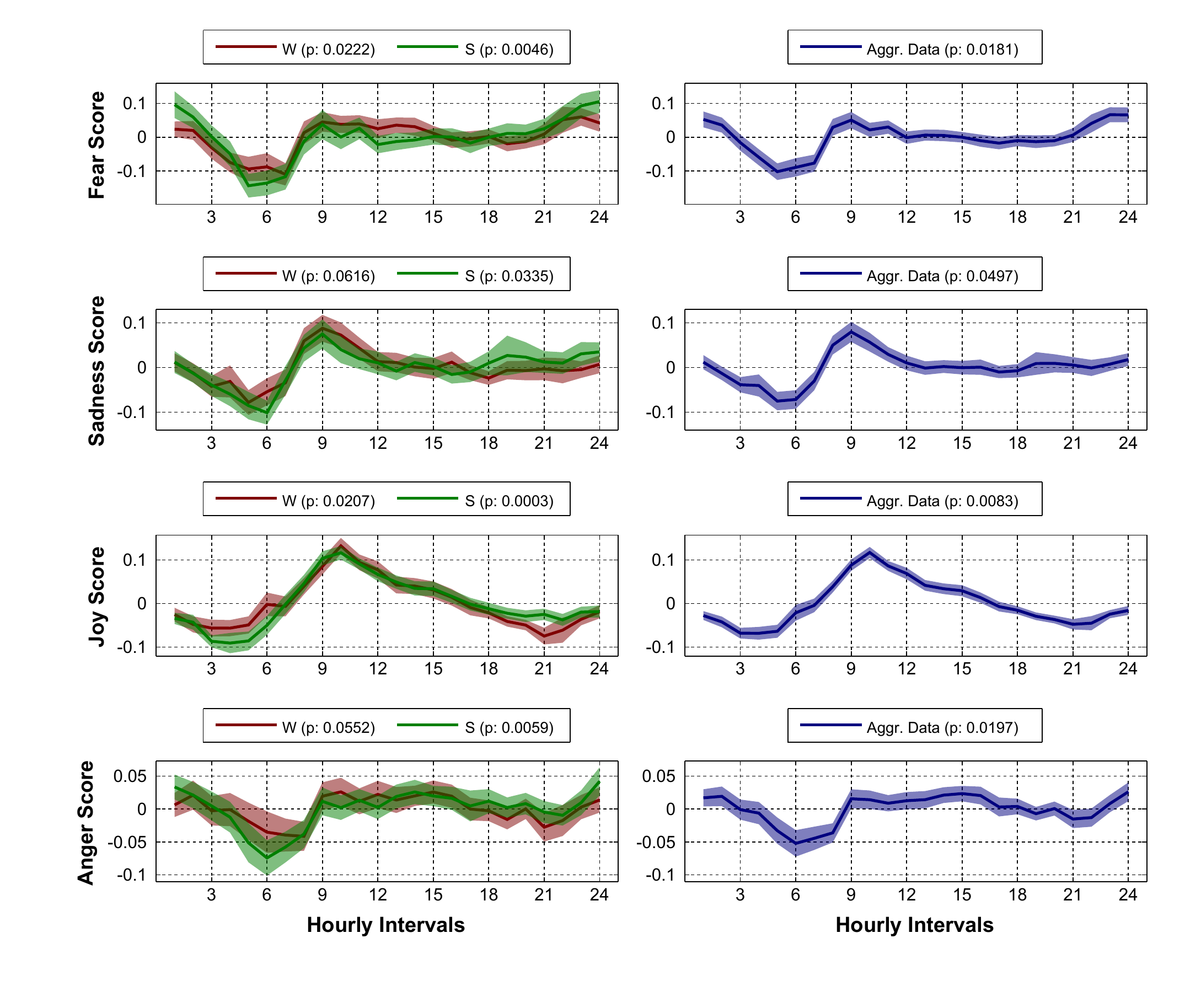}
    \end{center}
    \caption{These plots represent the variation over a 24-hour period of the emotional valence for fear, sadness, joy and anger obtained by applying MSFMS. The red line represents days in the winter, while the green one represents days in the summer. The average circadian pattern was extracted by aggregating the two seasonal data sets. Faded colourings represent 95\% CIs on the sample mean. For all displayed patterns the corresponding stability p-values are being reported.}
    \label{fig_MSFMS_circadian}
\end{figure}

MSFMS\index{MSFMS} produces circadian patterns with some noticeable differences compared to the previously presented ones. In Figure \ref{fig_MSFMS_circadian}, we see that fear has a first peak at 8--9a.m., then drops until 8p.m. when it starts increasing again reaching a maximum before midnight; we observe higher levels of fear very early in the mornings and during most of the daytime in winter (4a.m.--3p.m.), something that is reversed in the summer between 6p.m. and 4a.m. Sadness peaks in the mornings (8--9a.m.) and then decreases until 5p.m., where it becomes more steady. After midnight, the sadness score is decreasing reaching a minimum at 5--6a.m.; during the winter it is higher between the early morning hours and noon (3a.m.--1p.m.), whereas between 6p.m. and midnight it has higher rates in the summer. Joy also peaks in the morning (9--10a.m.) and then decreases until 9p.m., where it slightly starts to increase again, reaching a significantly smaller peak before midnight; the minimum scores for joy happen between 3 and 4a.m. The levels of joy during the night hours but before midnight tend to be higher in the summer, however, after midnight `winter tweets' seem to be more joyful. Anger has a minimum at 5--6a.m., a first peak during 8--9a.m., a second peak at 2--4p.m., then decreases until 9p.m., where it starts increasing again until it reaches a maximum before midnight; in winter during the core daytime we see higher levels of anger, whereas in the summer anger is higher from 4p.m. until the late night hours.

\subsubsection{Correlations across scoring schemes and mood types}
Table \ref{table_mood_scoring_corr} shows the linear correlations between the circadian patterns derived under MFMS and MSFMS in winter, summer and on the aggregated data set. All circadian patterns are positively correlated across the two different scorings schemes; some results drawn on seasonal patterns, however, are not statistically significant. Fear and joy show the highest correlations (0.7091 and 0.5185 respectively) and thus, we expect to derive similar patterns independently of the applied scoring approach, something, in fact, displayed in Figures \ref{fig_MFMS_circadian} and \ref{fig_MSFMS_circadian}. Table \ref{table_mood_types_corr} shows the linear correlation coefficients for the circadian patterns of all mood pairs under both scoring schemes. Mood pairs containing the emotion of joy have on average the lowest correlations. Sadness and fear are the most correlated mood types; there also exists some correlation between anger and fear or sadness. The only clearly uncorrelated emotions are joy and anger.

%%%%%%%%%%%%%%%%%%%%
\begin{table}[tp]
\small
\centering
\renewcommand{\arraystretch}{1.2}
\newcolumntype{C}{>{\centering\arraybackslash} m{1.8cm} }
\newcolumntype{V}{>{\arraybackslash} m{1.8cm} }
\caption{Winter vs. Summer circadian pattern linear correlations across all the considered mood types for MFMS and MSFMS. All correlations are statistically significant.}{
\(\begin{tabular}{V|C|C|C|C|}
\cline{2-5}
                                         &   \multicolumn{4}{c|}{\textbf{Winter vs. Summer Linear Correlation Coefficients}}\\\cline{2-5}
                                         &   \textbf{Fear}       & \textbf{Sadness}     & \textbf{Joy}    & \textbf{Anger} \\\hlinewd{2pt}
\multicolumn{1}{|l|}{\textbf{MFMS}}      &   0.9108              & 0.7779               & 0.9598          & 0.7402   \\\hline
\multicolumn{1}{|l|}{\textbf{MSFMS}}     &   0.8381              & 0.8309               & 0.9274          & 0.7925   \\\hline
\end{tabular}\)
}
\label{table_winter_summer_mood_corr}
\end{table}
%%%%%%%%%%%%%%%%%%%%%
%%%%%%%%%%%%%%%%%%%%%

%%%%%%%%%%%%%%%%%%%%
%%%%%%%%%%%%%%%%%%%%
\begin{table}[tp]
\small
\centering
\renewcommand{\arraystretch}{1.2}
\newcolumntype{C}{>{\centering\arraybackslash} m{1.8cm} }
\newcolumntype{V}{>{\arraybackslash} m{1.8cm} }
\caption{MFMS vs. MSFMS circadian pattern linear correlations across all the considered mood types in winter, summer and on the aggregated data set. Correlations with an asterisk ($\ast$) are not statistically significant.}{
\(\begin{tabular}{V|C|C|C|C|}
\cline{2-5}
                                            & \multicolumn{4}{c|}{\textbf{MFMS vs. MSFMS Linear Correlation Coefficients}}\\\cline{2-5}
                                            & \textbf{Fear} & \textbf{Sadness}   & \textbf{Joy}   & \textbf{Anger}  \\\hlinewd{2pt}
\multicolumn{1}{|l|}{\textbf{Winter}}       & 0.722         & 0.5028             & 0.3588$^{\ast}$& 0.2968$^{\ast}$\\\hline
\multicolumn{1}{|l|}{\textbf{Summer}}       & 0.7993        & 0.3916$^{\ast}$    & 0.6473         & 0.3813$^{\ast}$\\\hline
\multicolumn{1}{|l|}{\textbf{Average}}      & 0.7091        & 0.4122             & 0.5185         & 0.4211\\\hline
\end{tabular}\)
}
\label{table_mood_scoring_corr}
\end{table}
%%%%%%%%%%%%%%%%%%%%%
%%%%%%%%%%%%%%%%%%%%%

%%%%%%%%%%%%%%%%%%%%
%%%%%%%%%%%%%%%%%%%%
\begin{table}[tp]
\small
\centering
\renewcommand{\arraystretch}{1.2}
\newcolumntype{C}{>{\centering\arraybackslash} m{1.8cm} }
\newcolumntype{V}{>{\arraybackslash} m{1.8cm} }
\caption{Linear correlations amongst the considered mood types for MFMS and MSFMS (below and above table's main diagonal respectively). Correlations with an asterisk ($\ast$) are not statistically significant.}
\(\begin{tabular}{|V|C|C|C|C|}
\hline
\textbf{MSFMS}/ \textbf{MFMS}  & \textbf{Fear}   & \textbf{Sadness}   & \textbf{Joy}   & \textbf{Anger}\\\hlinewd{2pt}
\textbf{Fear}                  & --              & 0.7727             & 0.3274$^{\ast}$& 0.6549         \\\hline
\textbf{Sadness}               & 0.8069          & --                 & 0.6976         & 0.4858         \\\hline
\textbf{Joy}                   & 0.5702          & 0.5959             & --             & 0.3339$^{\ast}$  \\\hline
\textbf{Anger}                 & 0.6154          & 0.7912             & 0.0751$^{\ast}$& --\\\hline
\end{tabular}\)
\label{table_mood_types_corr}
\end{table}
%%%%%%%%%%%%%%%%%%%%%
%%%%%%%%%%%%%%%%%%%%%

\subsubsection{Peak moments in the mood signals (MSFMS)}
\begin{figure*}[tp]
    \begin{center}
    \subfigure[\textbf{Fear}] {\includegraphics[width=2.8in]{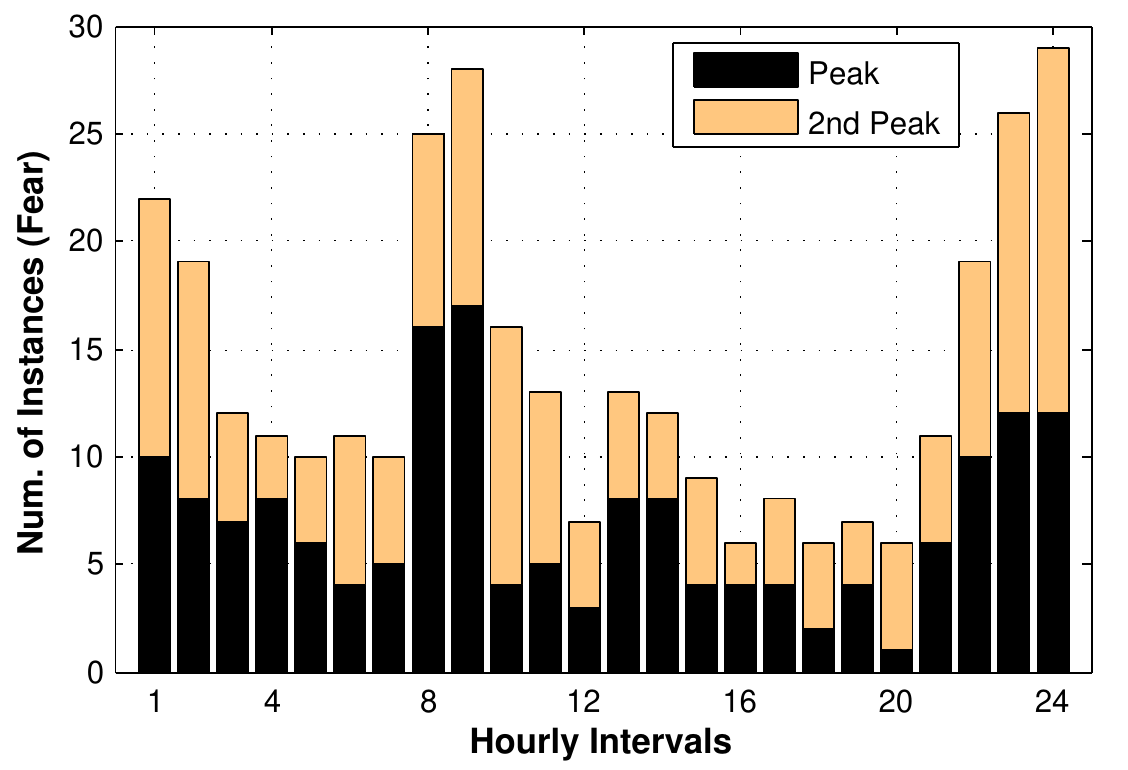}
    \label{fig_peaks_MSFMS_fear}}
    \hfil
    \subfigure[\textbf{Sadness}] {\includegraphics[width=2.8in]{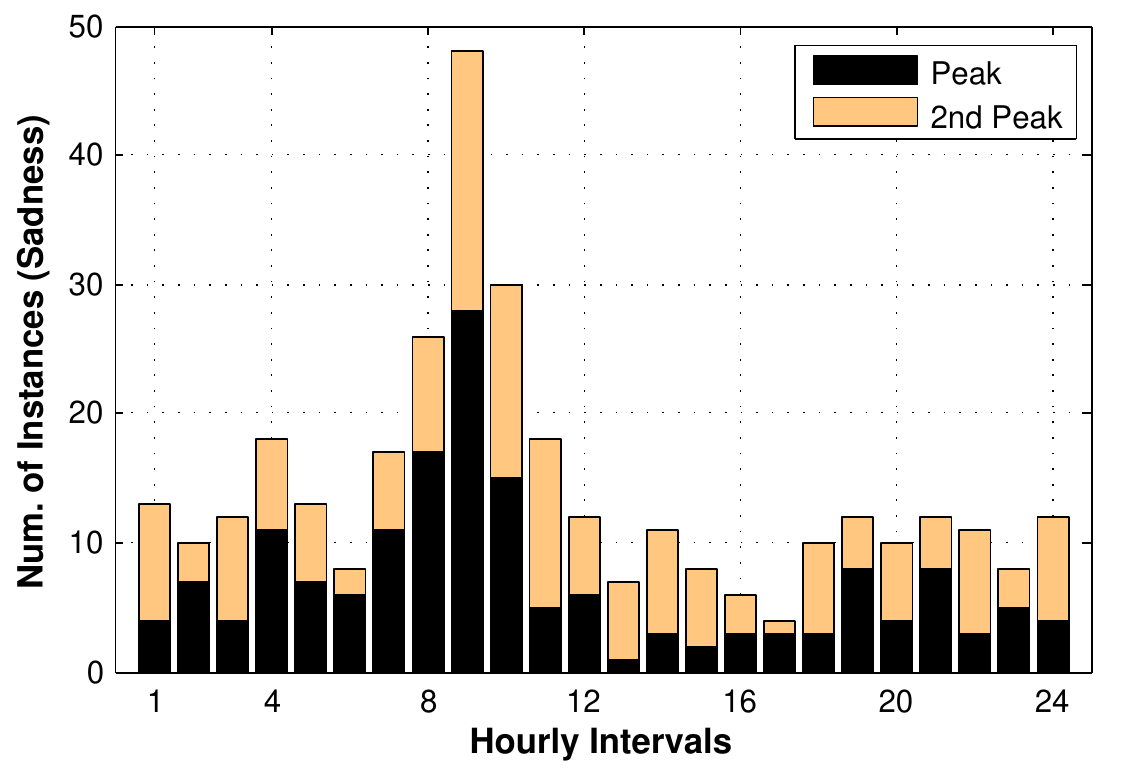}
    \label{fig_peaks_MSFMS_sadness}}
    \hfil
    \subfigure[\textbf{Joy}] {\includegraphics[width=2.8in]{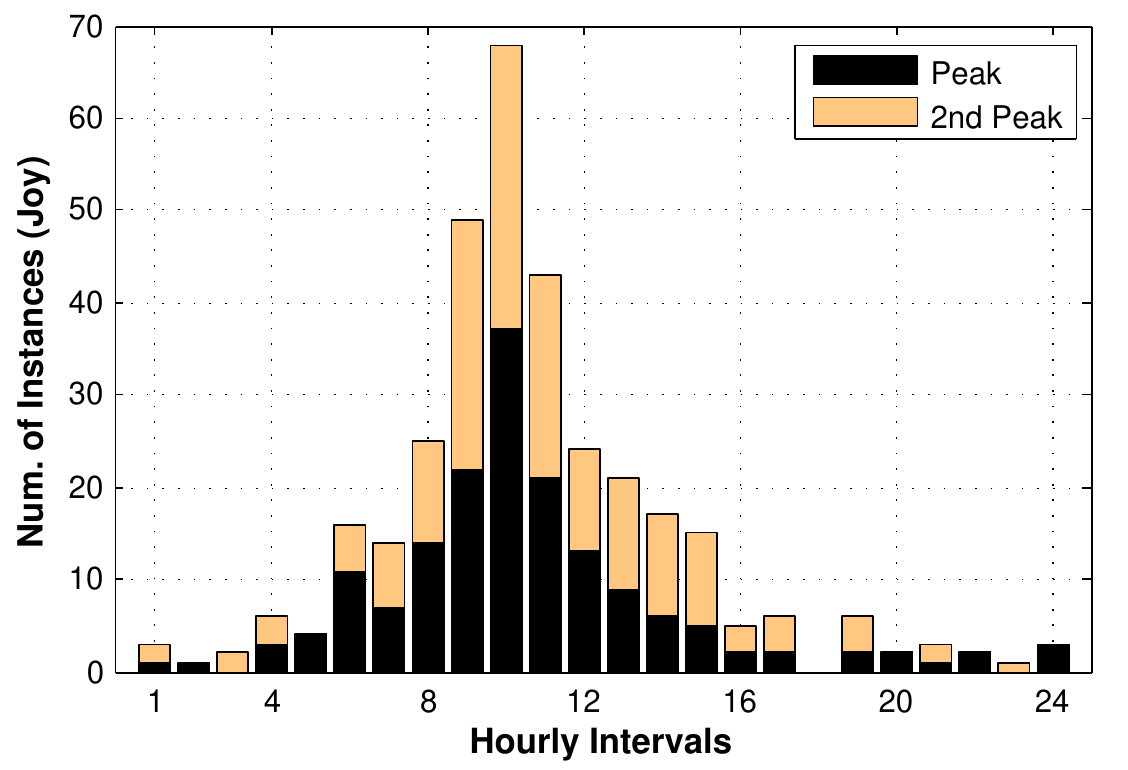}
    \label{fig_peaks_MSFMS_joy}}
    \hfil
    \subfigure[\textbf{Anger}] {\includegraphics[width=2.8in]{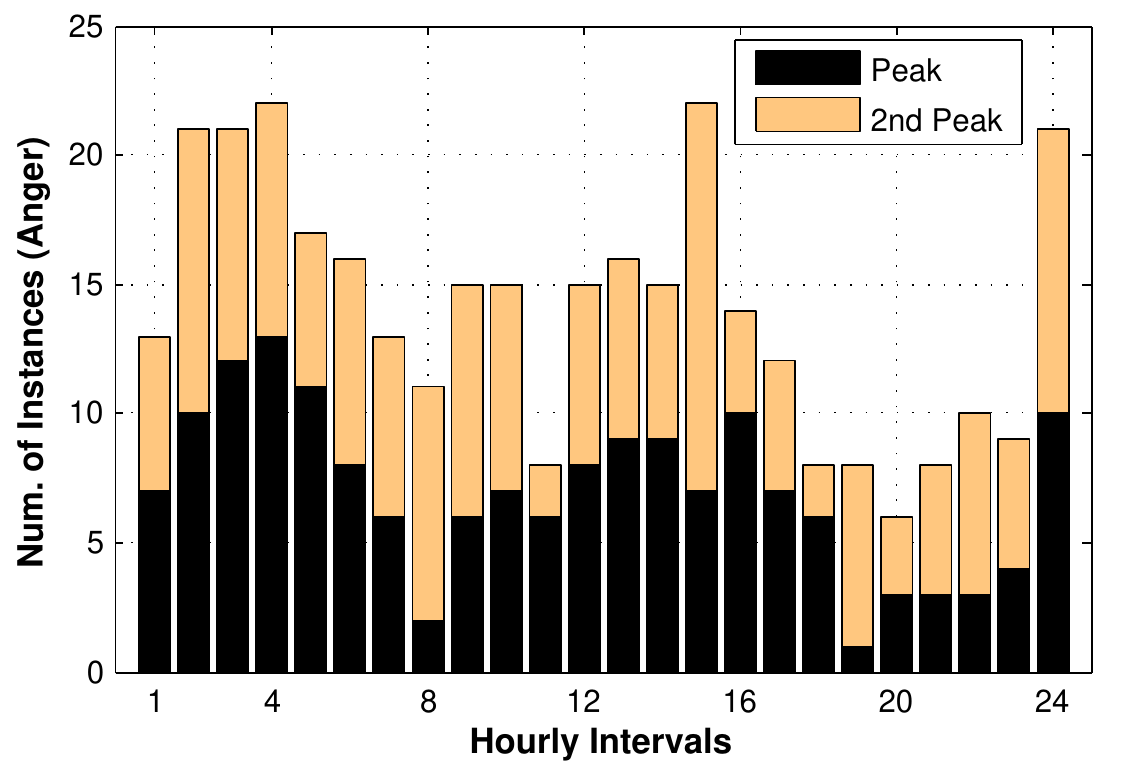}
    \label{fig_peaks_MSFMS_anger}}
    \end{center}
    \caption{Bar plots showing the number of samples which reach a maximum (peak) or second-highest (2nd peak) mood score for each hourly interval and for all mood types when MSFMS is applied.}
    \label{fig_peaks_MSFMS}
\end{figure*}

In this section, we investigate further the peak moments of each diurnal pattern extracted by applying the MSFMS scheme. The patterns presented in the previous sections show the average behaviour of each mood type. Some emotions, such as fear and anger, do not have one clearly distinctive peak in those average patterns and as a result, the stability or importance of the several occurring peaks is unclear. By extracting the peak moments from the daily mood signals in our data set (168 days in total) and forming their histogram, we can observe the distribution of peak moments across all days; this is depicted in Figure \ref{fig_peaks_MSFMS} with the black coloured rectangles for each hourly interval. Since the difference between the highest mood score and the second-highest value might be small in some samples, we have also visualised the frequency of the second-highest mood scores on top of the peak frequency.

We can see that joy (Figure \ref{fig_peaks_MSFMS_joy}) retrieves its highest values in the mornings (8--11a.m.) and especially between 9 and 10a.m. Sadness (Figure \ref{fig_peaks_MSFMS_sadness}) has a similar peaking behaviour; it usually peaks between 7 and 9a.m. The emotion of fear (Figure \ref{fig_peaks_MSFMS_fear}) has two peaking time periods; one occurs during late evenings and midnights (10p.m.--1a.m.) and the other one in the mornings (7--9a.m.). Anger (Figure \ref{fig_peaks_MSFMS_anger}) has the most unstable behaviour amongst the investigated emotional types. Still, we can derive that it rarely peaks in the evenings, but reaches high levels quite often during midnight, late night hours (1--4a.m.) or right after lunch time (2-3a.m.).

\subsubsection{Periodicity in mood scores (MSFMS)}
\label{section:circadian_mood_periodicity}
\begin{figure*}[tp]
    \begin{center}
    \subfigure[\textbf{Fear}] {\includegraphics[width=2.8in]{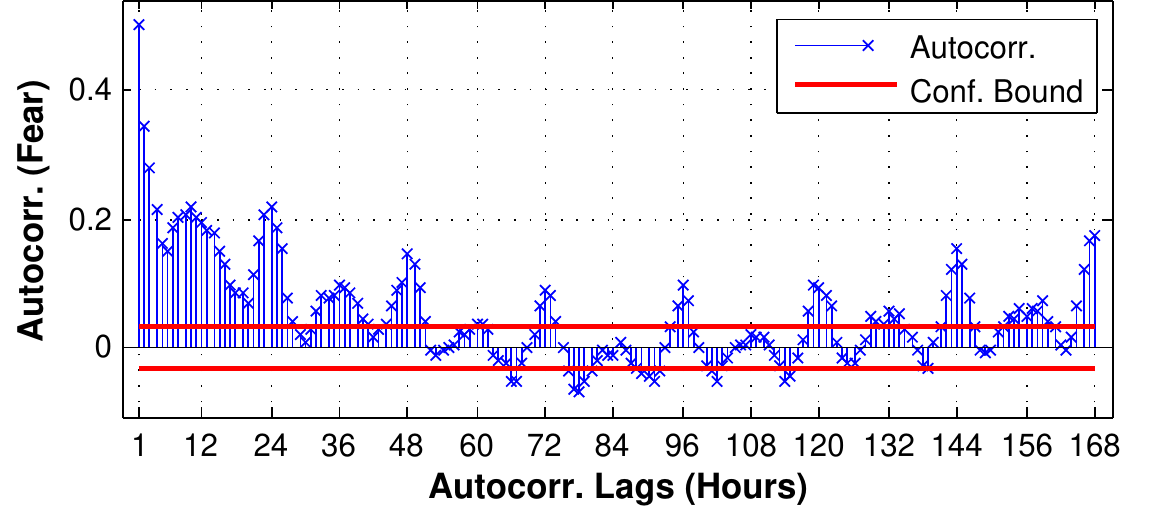}
    \label{fig_autocorr_MSFMS_fear}}
    \hfil
    \subfigure[\textbf{Sadness}] {\includegraphics[width=2.8in]{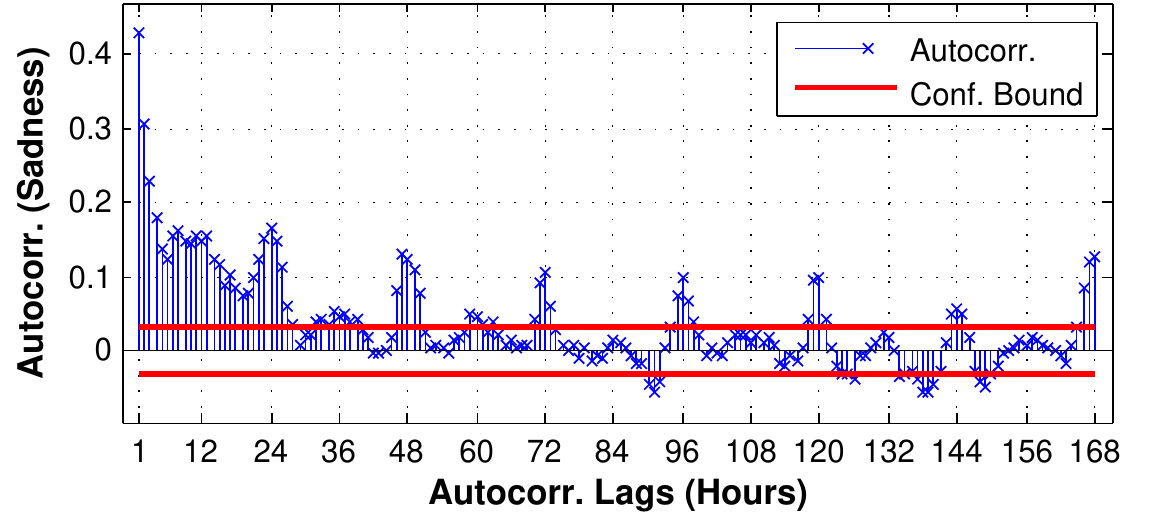}
    \label{fig_autocorr_MSFMS_sadness}}
    \hfil
    \subfigure[\textbf{Joy}] {\includegraphics[width=2.8in]{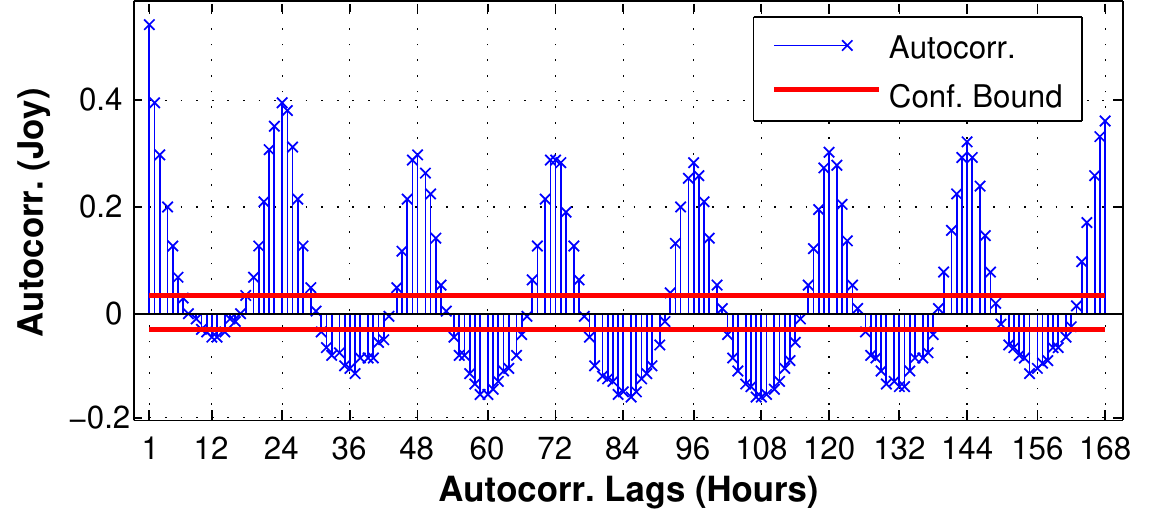}
    \label{fig_autocorr_MSFMS_joy}}
    \hfil
    \subfigure[\textbf{Anger}] {\includegraphics[width=2.8in]{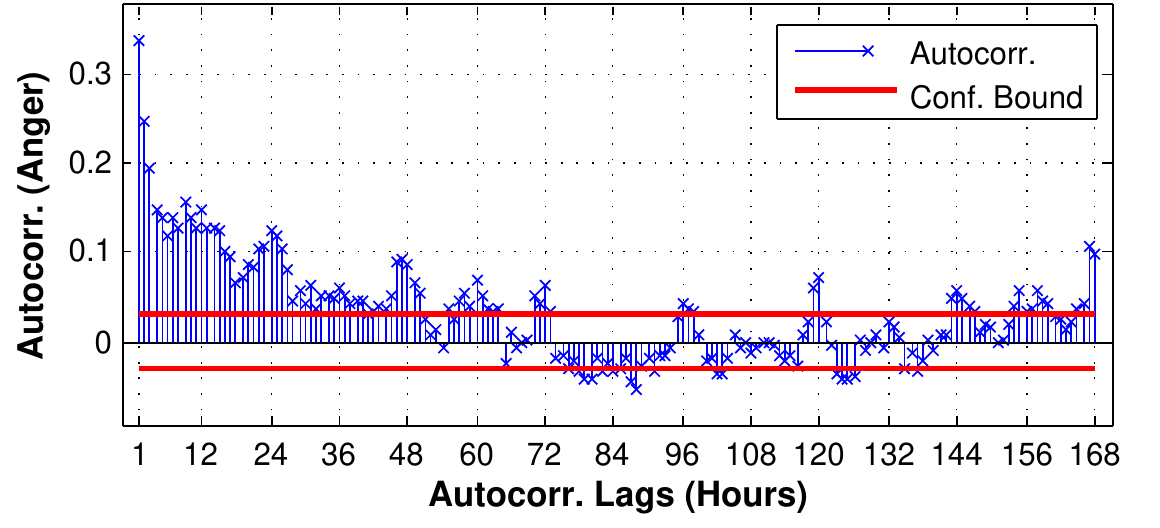}
    \label{fig_autocorr_MSFMS_anger}}
    \end{center}
    \caption{Autocorrelation figures for all emotions under MSFMS. The considered lags range from 1 to 168 hours (7-days). Correlations outside the confidence bounds are statistically significant.}
    \label{fig_autocorr_MSFMS}
\end{figure*}

As an additional step in our experimental process, we compute autocorrelation\index{autocorrelation} figures (see Figure \ref{fig_autocorr_MSFMS}) for the four mood types based on the MSFMS scheme\footnote{ The autocorrelations derived by the MFMS scheme were similar, hence, pointing to the same conclusion.} using lags ($\ell$) ranging from 1 to 168 hours, \ie the total number of hours in a week. Consecutive time intervals (hours) are highly correlated for all emotional types; by common logic this is something expected as there is high probability that the mood in the population at a time instance $t$ is affected by the mood in the previous time instance $t -$ 1. From the autocorrelation plots, it also becomes evident that for each emotional type there exists some level of daily ($\ell =$ 24) as well as weekly periodicity ($\ell =$ 168). Autocorrelations for $\ell \in \{$25, ..., 167$\}$ are lower than the one observed for $\ell =$ 168 indicating that the weekly pattern (or period) in the mood signal is stronger than the intermediate ones.

We test the statistical significance of the observed levels of periodicity using autocorrelation as the test statistic. Suppose that a vector $v$ of length 4032 ($=$ 168 days $\times$ 24 hourly intervals) holds the scores for a mood type. We compute $v$'s autocorrelations for $\ell =$ 1, 24 and 168; then we compute the same autocorrelations for 1,000 randomly permuted versions of $v$ and count how many times (say $k$) the levels of autocorrelation for the randomised signal were greater or equal to the ones computed for $v$. The p-value for this test is therefore equal to $k/1,000$. In our tests for all mood types and considered lags, the computed p-values were equal to 0, indicating that the observed levels of periodicity are statistically significant.

Overall, the emotion of joy has the highest levels of autocorrelation and shows a strong periodic behaviour, whereas periodicity seems to be less strong for the mood type of anger (which also had displayed the most unstable behaviour in terms of peaking times). Those periodicities in affect apart from a generally interesting result, also justify our initial assumption that the extracted mood scores are able to overcome several biases, which might be created, for example, when specific -- possibly unfortunate -- significant events emerge in real-life and temporarily influence the content of the tweets.

\subsubsection{Comparison between Positive and Negative Affectivity (MSFMS)}
\begin{figure}[tp]
    \begin{center}
    \includegraphics[width=5in]{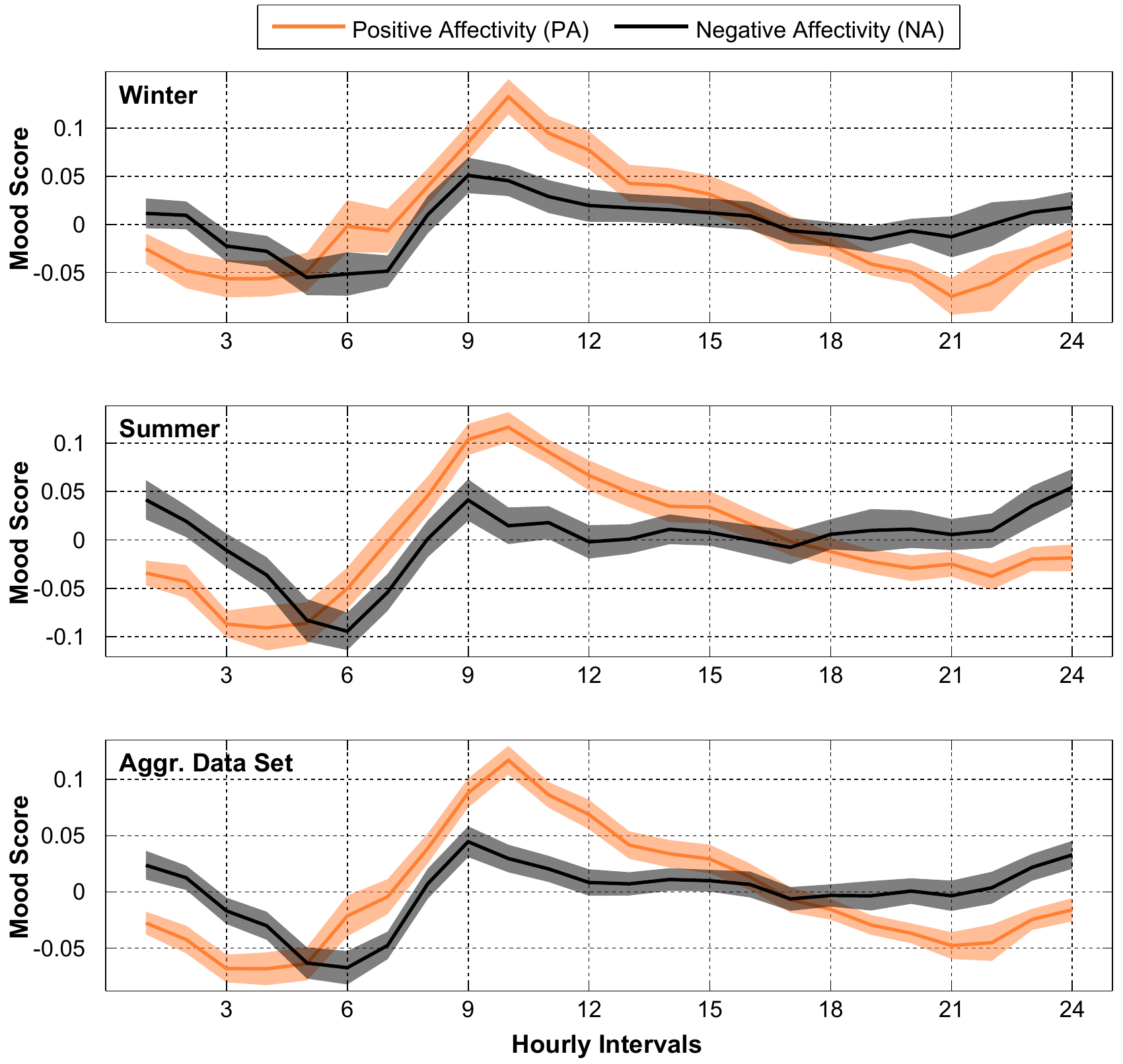}
    \end{center}
    \caption{These plots represent the variation over a 24-hour period of the emotional valence for Positive versus Negative Affectivity during winter, summer and on the aggregated data set by applying MSFMS scheme. Faded colourings represent 95\% CIs. The stability p-values of all extracted patterns are $\ll$ 0.05.}
    \label{fig_MSFMS_circadian_PA_vs_NA}
\end{figure}

Finally, we merged all mood types expressing NA\index{NA} (anger, fear and sadness) and compared them to the one expressing PA\index{PA} (joy) using MSFMS.\footnote{ In this particular case, MFMS scheme does not, nonetheless, output comparable results due to its dependence on the frequency range of each specific emotion word-list.} Figure \ref{fig_MSFMS_circadian_PA_vs_NA} shows the extracted PA versus NA patterns for winter, summer as well as the entire data set. From all the plots we derive that PA is higher during the daytime, while NA is stronger during the evenings and nights. By comparing winter to summer plots and counting in the seasonal differences between the hours of day and night, we can also see that in winter NA begins to dominate over PA slightly earlier in the afternoons. Winter's most negatively affected hour is 8--9a.m.; in the summer we observe increased levels of negativity during midnight as well.

\subsection{Discussion}
\label{section:discussion}
In the previous sections, we performed mood analysis\index{mood analysis} on Twitter content geolocated in the UK trying to extract diurnal -- seasonal and average -- patterns for the emotions of anger, fear, joy and sadness. As far as we are aware this is the first study of real-time mood variation at a population level using social media information in the UK. We developed two methods for computing mood scores, MFMS and MSFMS; their main difference is based on the fact that the second standardises the frequencies of each mood term over the considered amount of time and thus, reduces any impacts that highly frequent terms might have. Statistical analysis indicates that MSFMS scheme produces more stable results; hence, we show more elaborate findings considering only the latter. Still, the extracted patterns of both scorings schemes are positively correlated with a minimum linear correlation (on the average signals) equal to 0.4122.

Under MSFMS, the mood types of fear, joy and sadness have clear peaks in the mornings. Fear also peaks during midnight; on the flip side, anger has a more ambiguous behaviour with peaks emerging during afternoons, midnights and late night/very early morning hours. The combined NA circadian rhythm shows two peaks, one in the mornings and another one during midnights, whereas PA, which also peaks in the mornings, does so, an hour after NA's peak; the comparison between NA and PA indicates that PA takes higher values in the mornings, but NA increases as the day evolves and takes over PA in the evenings until the early morning hours. The mood signals are highly correlated across the two different seasons (winter and summer); still there exist some noticeable fluctuations, for example the emotion levels are higher during late night/early morning hours in the winter or there exist stronger emotional indicators during evenings and midnights in the summer.

The comparison of our findings with results from widely reported psychiatric studies reveals some similarities but also several differences. According to \cite{Rusting1998}, NA increases as the day progresses with a peak in the evenings; something reported also in our results for the summer's NA rhythm. Similarly, \cite{Murray2002} shows that from afternoon until the late night/early morning hours NA is -- on average -- stronger than PA. In \cite{Hasler2008}, PA -- in terms of `laughing' or `socialising' -- has a symmetric peak in the mornings, however the circadian patterns for anger -- expressed by the action of `arguing' -- and sadness -- expressed by `sighing' -- are not very correlated with our results since they show differences in peak moments as well as in monotonicity. Furthermore, in \cite{Stone2006} we observe that the emotional state of `angry' is reaching a maximum at the same time period as the emotion of anger in our findings; the same holds for the emotional states of `depressed/blue', `worry' and the emotion of sadness. The contradicting point in this study is that PA oriented emotional states (such as `happy' or `enjoy') are reported to have an increasing behaviour through the day.

A recent similar study presented circadian patterns for PA and NA across the different days of the week and their daily averages among different cultures \cite{Golder2011}. Their findings for the regions of UK and Australia\footnote{ They do not present diurnal rhythms for the UK alone.} are in contrast with the ones we derived for the UK alone. PA is shown to retrieve a peak value during midnight; in our work and under both scoring schemes this is happening in the mornings. Moreover, NA peaks during the late night hours, but we have reported reduced levels of NA during this time period of the day (again under both scoring schemes). Importantly, their results vary -- significantly -- across different cultures; for example, the circadian rhythms for India or Africa indicate a much better fit with the ones we extracted.

There are limitations when conducting these studies. Among these, it is obvious that people cannot send messages whilst asleep and therefore there is a drop in signalling over night. However, we adjusted our results by normalising (MFMS) or both normalising and standardising (MSFMS) to reduce the impact of this effect. In this work, we did not consider the influence that significant events or natural phenomena emerging in real-life might have on the extracted signals and our method for collecting Twitter messages excluded content geolocated in rural UK areas. However, we argue that such biases are usually resolved to a significant level, when working with large-scale samples of data; the periodical characteristics and the statistical significance (especially under MSFMS) provide further proof for the stability of the extracted signals.

On the other hand, limitations arising due to attributes of the general population that might not be present in Twitter users, such as behavioural motives of the elderly, do create unresolved biases; therefore, in the present script, we can only make claims about the population of UK Twitter users and not about the general population. Finally, further validation of the affective words using, for example, labelled data and supervised statistical learning techniques might improve the accuracy of the proposed mood scoring schemes. %In spite of all this, Twitter does represent a great opportunity to capture real-time data for massive numbers of people in the general population and use it to understand several aspects of human behaviour and life.

%%%%%%%%%%%%%%%%%%%%%%%%%%%%%%%%%%%%%%%%%%%%%%%%%%%%%%%%%%%%%%%%%%%%
%%%% yearly %%%%%%%%%%%%%%%%%%%%%%%%%%%%%%%%%%%%%%%%%%%%%%%%%%%%%%%%

\section{Detecting daily mood patterns from Twitter content}
\label{section:detecting_daily_mood_patterns}
In this section, we are investigating daily mood patterns on Twitter. Similarly to the previous section, we base our affective analysis on WordNet affect \cite{Strapparava2004}, and focus on three negative affective norms (anger, fear and sadness) and one positive (joy).

\subsection{Data and methods}
\label{section:data_methods_mood_daily}
The methods for data collection and processing have already been described in Section \ref{section:coll_store_process_Twitter_data}. All collected tweets are geolocated in the UK, and therefore results are partly affected by this. The considered time period for this study is the year 2011; the total number of tweets used in our experimental process is approx. 320 million.

MFMS\index{MFMS} and MSFMS\index{MSFMS} are applied again, with the only difference that hourly time intervals are replaced with daily time intervals, \ie we are extracting mood patterns on a daily basis. Therefore, using the same notation as in Section \ref{section:methods}, the mood score $\mathcal{M}(d)$ for a day $d\in\{$1, ..., 365$\}$ is given by:
\begin{equation}
\mathcal{M}(d) = \frac{1}{n} \sum_{i=1}^{n} \frac{c_i^{(t_{d})}}{N(t_{d})},
\end{equation}
where $c_i^{(t_{d})}$ is the count of term $i$ in the Twitter corpus of day $d$ and $N(t_{d})$ is the number of tweets collected in that day. Likewise MSFMS is defined as:
\begin{equation}
\mathcal{M}_{s}(d) = \frac{1}{n}\sum_{i=1}^{n}sf_{i}^{(t_{d})},
\end{equation}
where $sf_{i}^{(t_{d})}$ is the standardised frequency of term $i$ for day $d$.

Recall that MFMS is based on the assumption that the frequency of a word defines its importance (or weight) in the extracted mood signal, whereas MSFMS `treats' all terms equally.

\subsection{Experimental results}
\label{section:results_mood_analysis_daily}
\begin{figure*}
    \begin{center}
    \subfigure[Anger (MFMS)]{\includegraphics[width=4.2in]{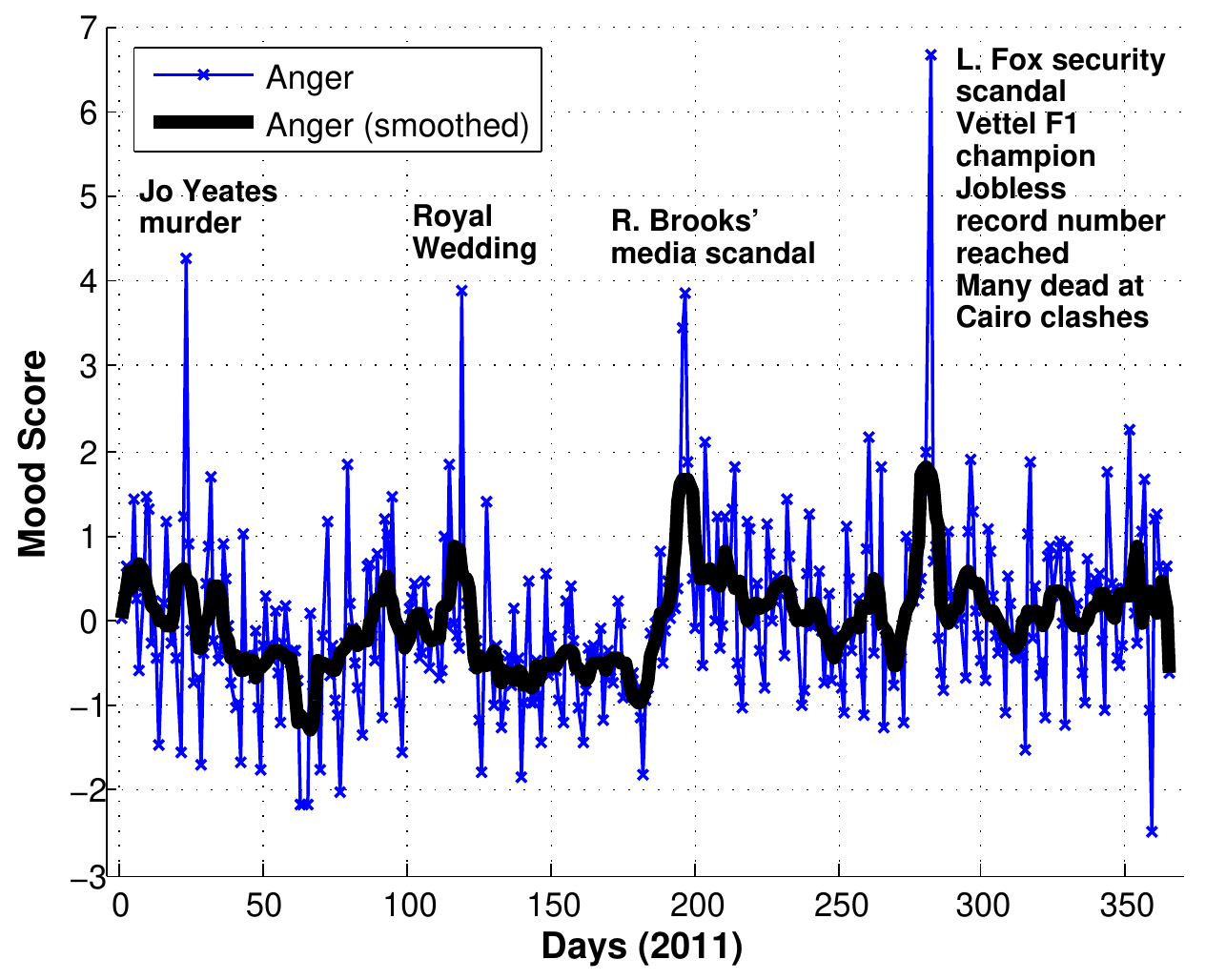}
    \label{fig_mood_anger_2011_MFMS}}
    \hfil
    \subfigure[Anger (MSFMS)]{\includegraphics[width=4.2in]{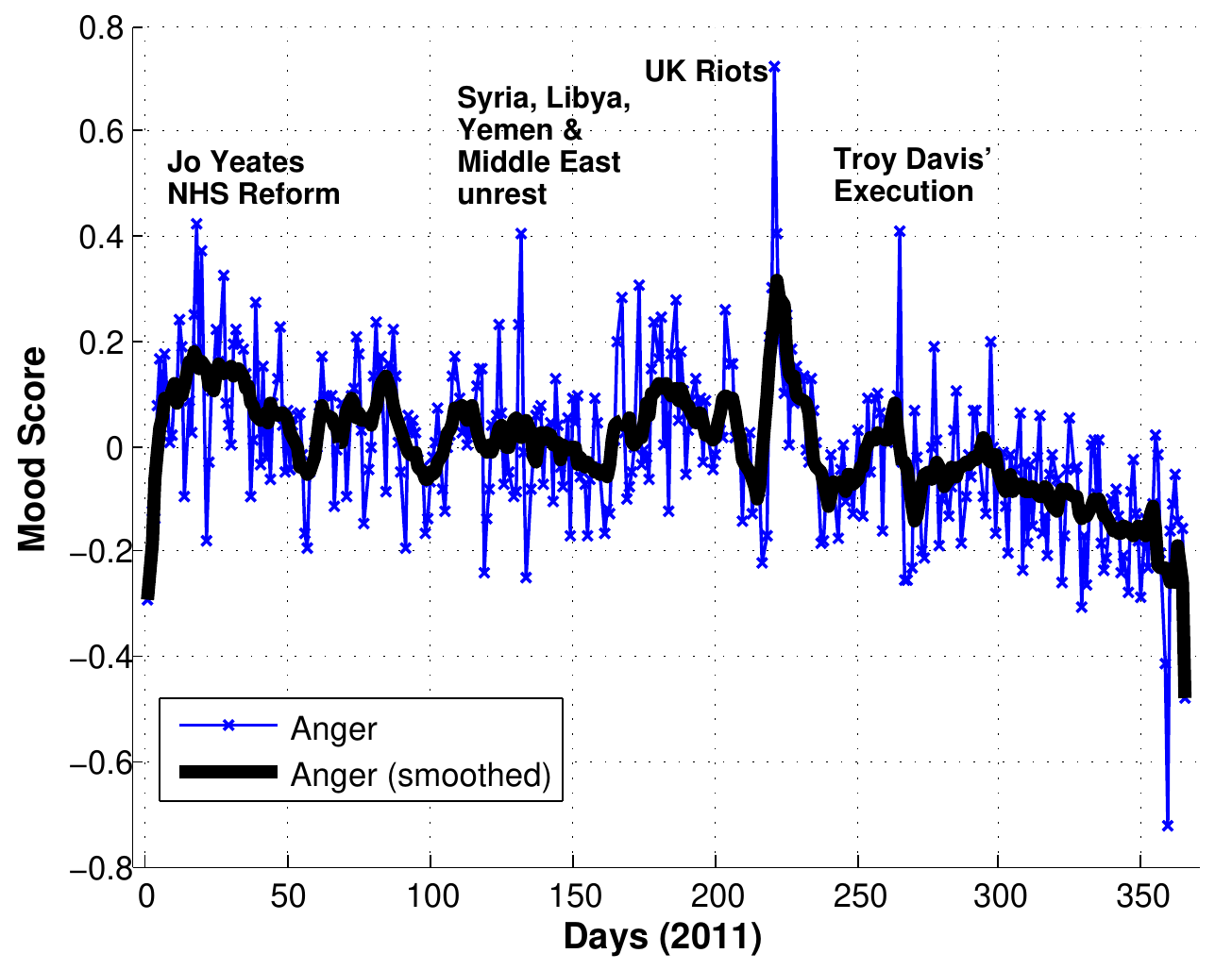}
    \label{fig_mood_anger_2011_MSFMS}}
    \end{center}
    \caption{Daily Scores for Anger (exact and smoothed) in 2011 based on Twitter content geolocated in the UK using MFMS and MSFMS.}
    \label{fig_mood_anger_2011}
\end{figure*}

\begin{figure*}
    \begin{center}
    \subfigure[Fear (MFMS)]{\includegraphics[width=4.2in]{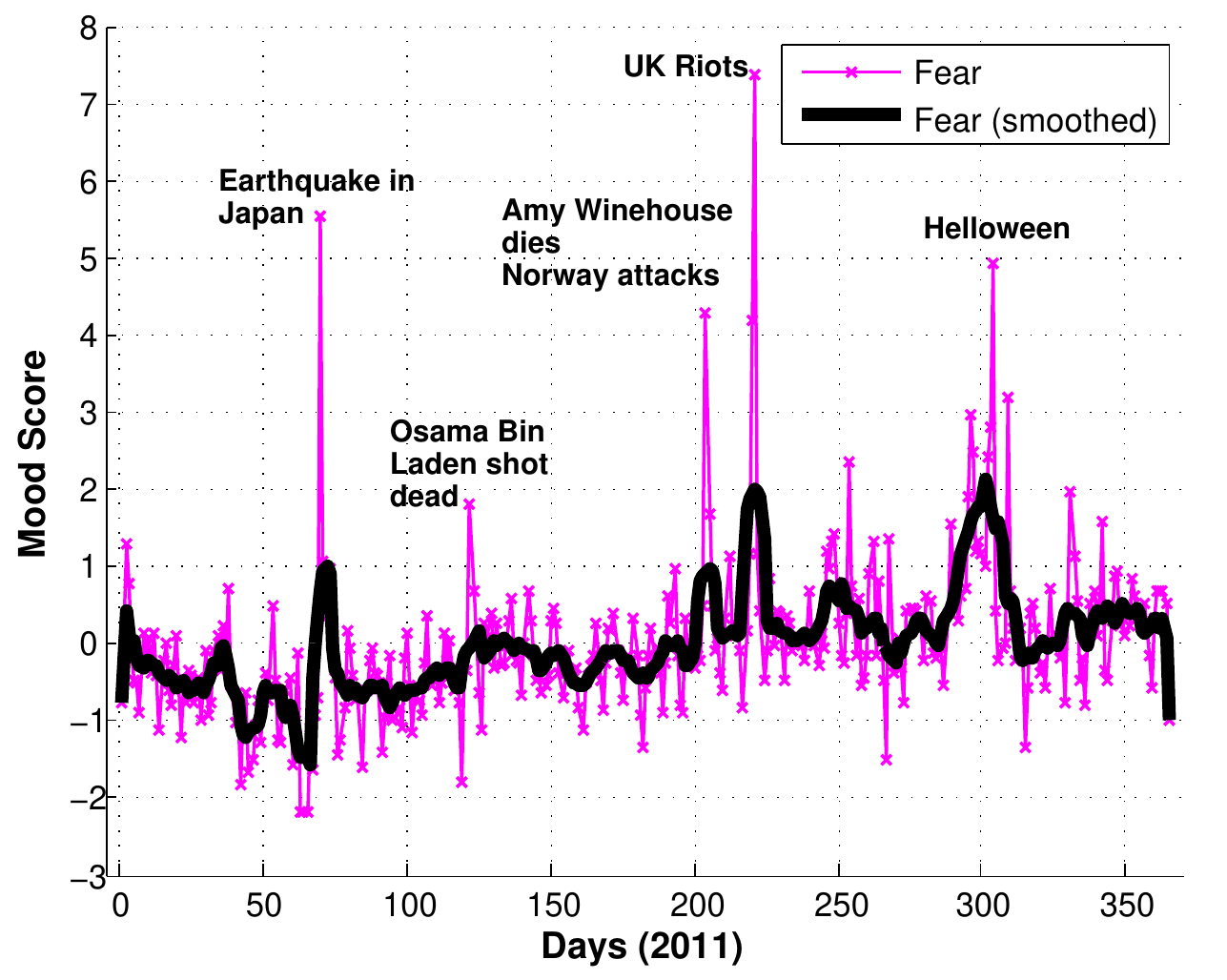}
    \label{fig_mood_fear_2011_MFMS}}
    \hfil
    \subfigure[Fear (MSFMS)]{\includegraphics[width=4.2in]{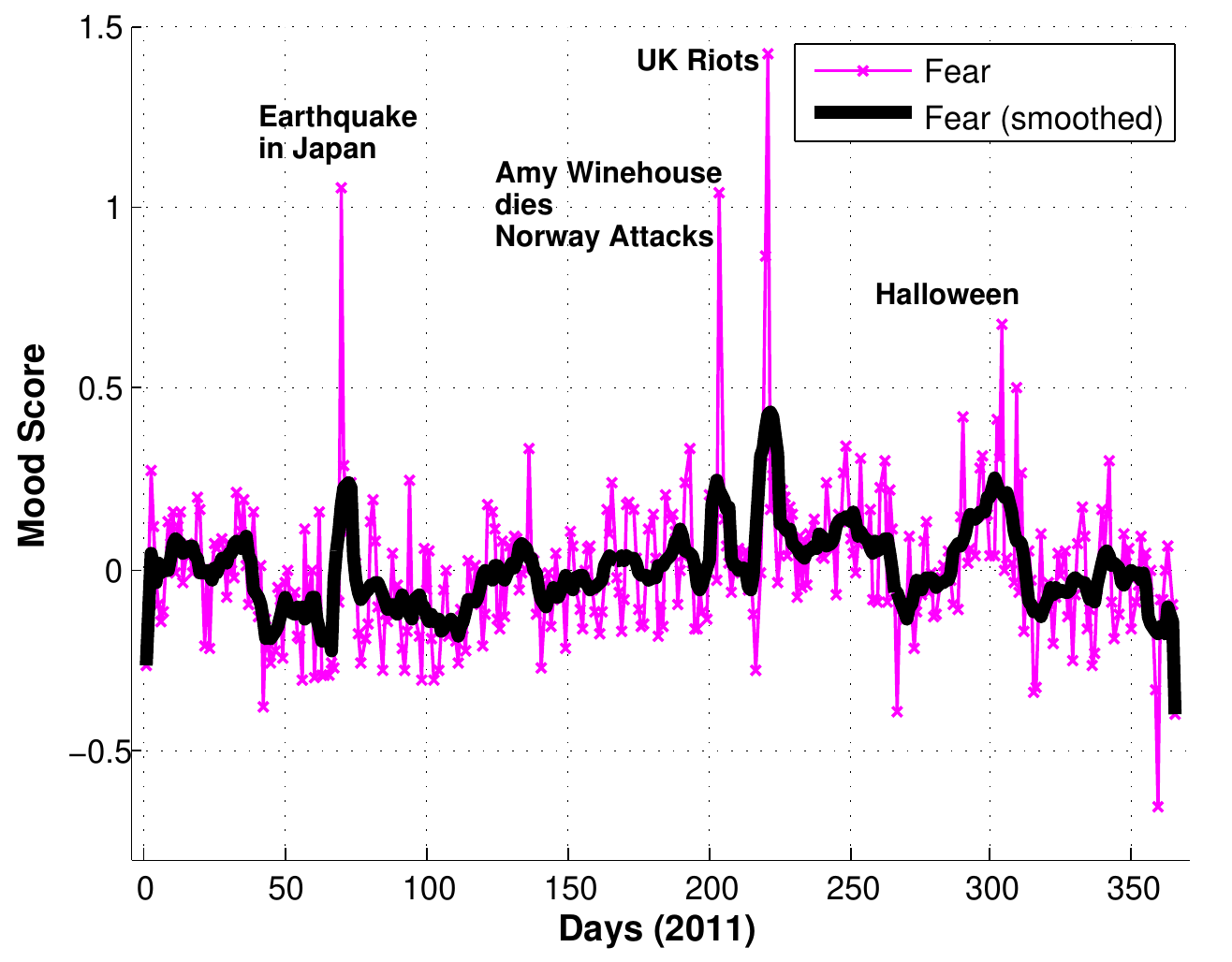}
    \label{fig_mood_fear_2011_MSFMS}}
    \end{center}
    \caption{Daily Scores for Fear (exact and smoothed) in 2011 based on Twitter content geolocated in the UK using MFMS and MSFMS.}
    \label{fig_mood_fear_2011}
\end{figure*}

\begin{figure*}
    \begin{center}
    \subfigure[Joy (MFMS)]{\includegraphics[width=4.2in]{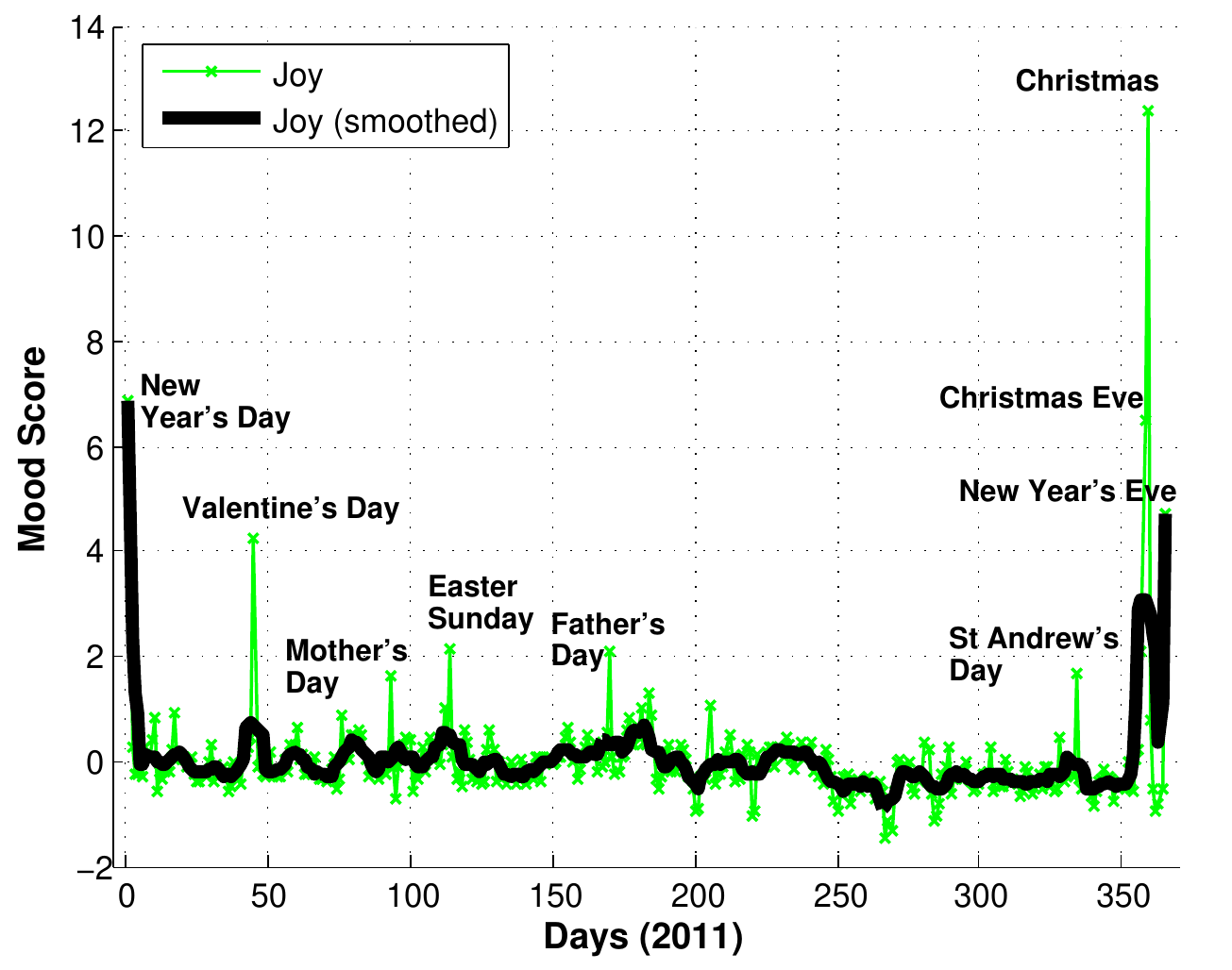}
    \label{fig_mood_joy_2011_MFMS}}
    \hfil
    \subfigure[Joy (MSFMS)]{\includegraphics[width=4.2in]{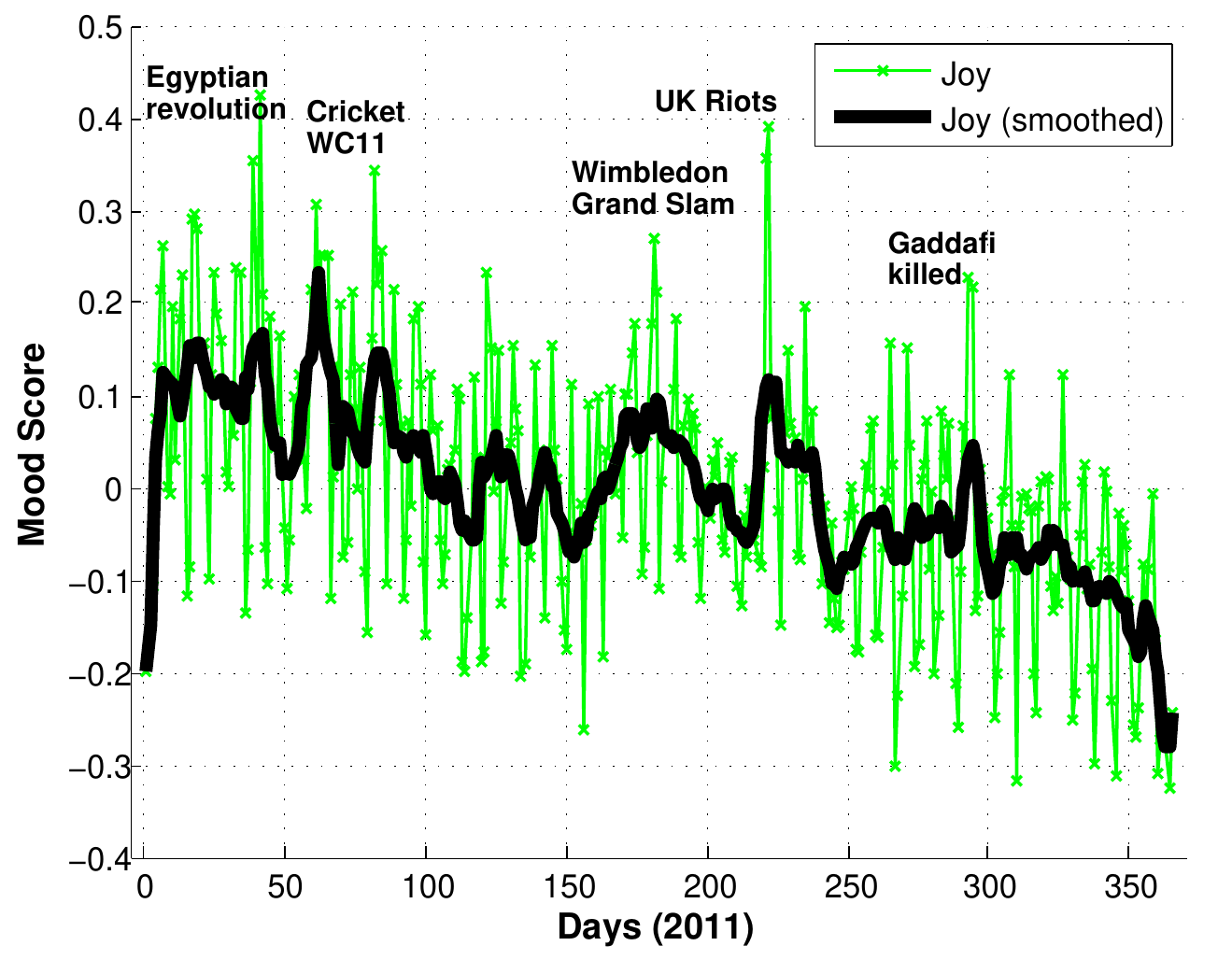}
    \label{fig_mood_joy_2011_MSFMS}}
    \end{center}
    \caption{Daily Scores for Joy (exact and smoothed) in 2011 based on Twitter content geolocated in the UK using MFMS and MSFMS.}
    \label{fig_mood_joy_2011}
\end{figure*}

\begin{figure*}
    \begin{center}
    \subfigure[Sadness (MFMS)]{\includegraphics[width=4.2in]{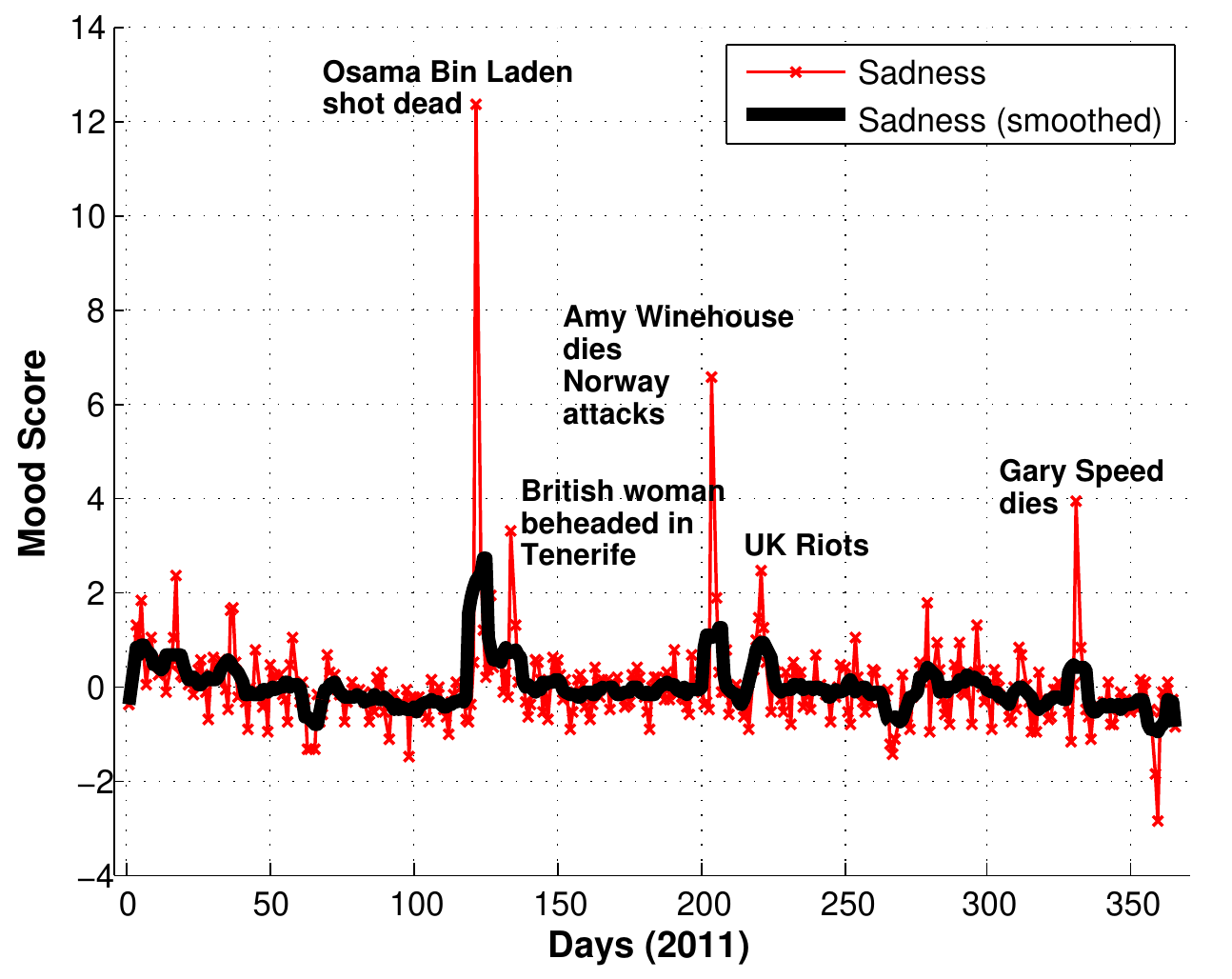}
    \label{fig_mood_sadness_2011_MFMS}}
    \hfil
    \subfigure[Sadness (MSFMS)]{\includegraphics[width=4.2in]{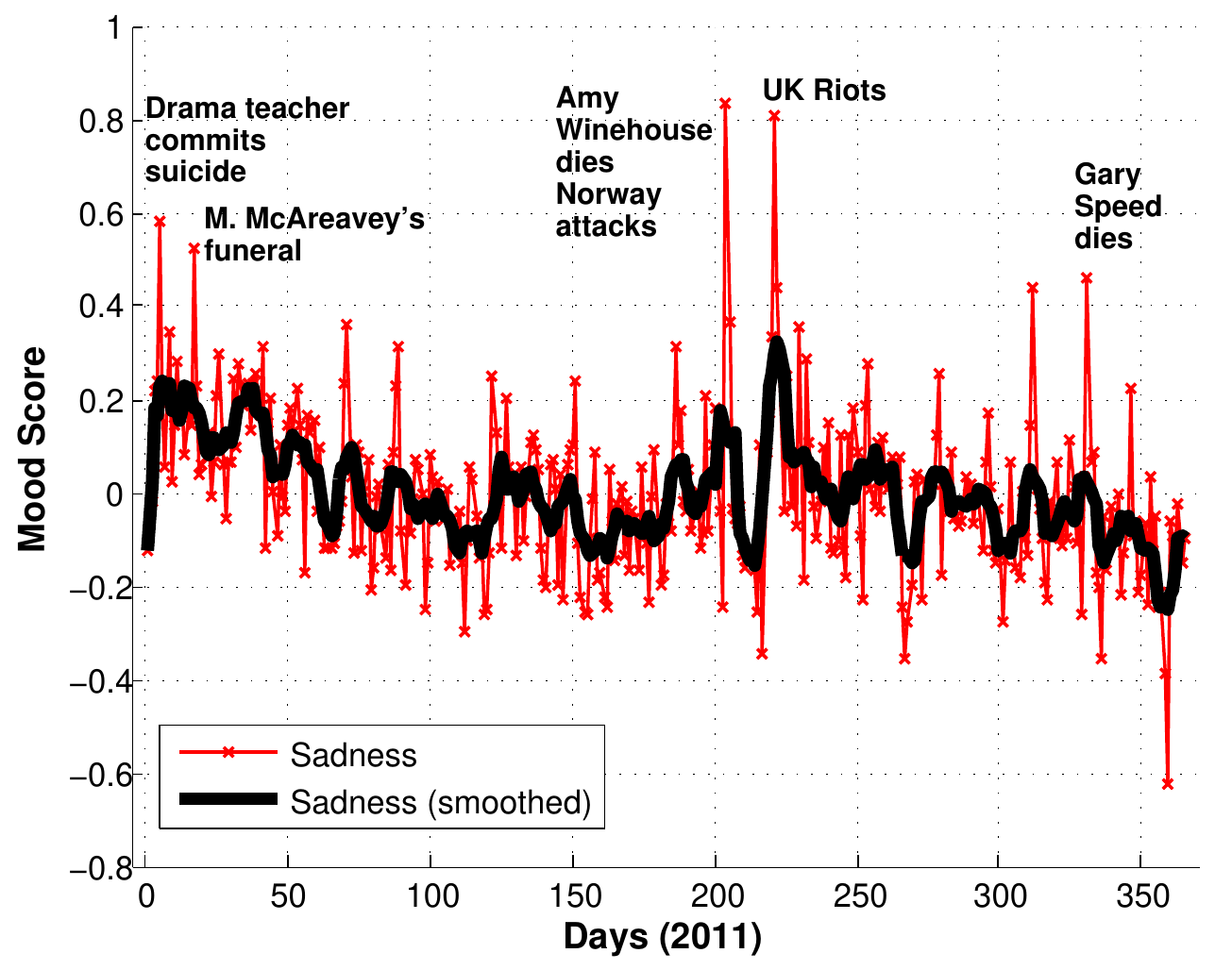}
    \label{fig_mood_sadness_2011_MSFMS}}
    \end{center}
    \caption{Daily Scores for Sadness (exact and smoothed) in 2011 based on Twitter content geolocated in the UK using MFMS and MSFMS.}
    \label{fig_mood_sadness_2011}
\end{figure*}

Figures \ref{fig_mood_anger_2011}, \ref{fig_mood_fear_2011}, \ref{fig_mood_joy_2011} and \ref{fig_mood_sadness_2011} show the extracted daily mood scores under MFMS and MSFMS for anger, fear, joy and sadness respectively. In each plot, apart from the exact scores we have also included their smoothed version (7-point moving average). The exact values allow us to retrieve the dates of peaks in the emotional signal, whereas the smoothed time series reveal a weekly trend of the emotional level and thus, can be used to extract periods with alleviated or aggravated affect. In this section, we try to explain those time series based on the assumption that the affective norms\index{affective norm} on Twitter might be emerging from events in real-life. It turns out that significant events -- as perceived by the common sense -- can be tracked and identified within those time series.

\begin{table}
\caption{Correlations between the derived daily mood patterns under MFMS and MSFMS. Correlations with an asterisk ($\ast$) are not statistically significant.}
\label{table_mood_scoring_corr_daily_2011}
\renewcommand{\arraystretch}{1.2}
\setlength\tabcolsep{1mm}
\newcolumntype{C}{>{\centering\arraybackslash} m{4cm} }
\newcolumntype{V}{>{\arraybackslash} m{3cm} }
\centering
\small
\(\begin{tabular}{V|C|C|}
\cline{2-3}
                              & \textbf{MFMS vs. MSFMS} & \textbf{Smoothed} (7-point MA)\\\hlinewd{2pt}
\multicolumn{1}{|l|}{\textbf{Anger}}                & 0.0588$^{\ast}$         & 0.096$^{\ast}$    \\\hline
\multicolumn{1}{|l|}{\textbf{Fear}}                 & 0.8085                  & 0.7888            \\\hline
\multicolumn{1}{|l|}{\textbf{Joy}}                  & $-$0.0494$^{\ast}$      & $-$0.1699         \\\hline
\multicolumn{1}{|l|}{\textbf{Sadness}}              & 0.5984                  & 0.5663            \\\hline
\multicolumn{1}{|l|}{\textbf{NA minus PA}}       & 0.5814                  & 0.4617               \\\hline
\end{tabular}\)
\end{table}

\textbf{Anger}. In Figure \ref{fig_mood_anger_2011_MFMS} (MFMS) we can see two distinctive periods of anger, emerging in October and July respectively. Anger is peaking on the 9th of October, something that could be explained by a series of events that happened on that day, including the death of many civilians in clashes in Egypt, the announcement that the number of jobless people in the UK reached a new record high, Vettel's win\footnote{ This win secured the Formula 1 (F1) title for S. Vettel, a fact that probably displeased the majority of F1 fans in the UK.} and Liam Fox's security scandal. The other peak on the 15th and 16th of July could have been emerged due to the News International phone hacking scandal.\footnote{ Wikipedia's page on this topic, \url{http://goo.gl/47q2Y}.} Two additional distinctive moments -- not periods -- in terms of anger, were the 23rd of January (possibly related with the case of Jo Yaetes' murder), and the 29th of April, the day of the Royal Wedding. However, the signal derived by applying MSFMS is quite different; indeed, the linear correlation of the two signals is almost non existent (see Table \ref{table_mood_scoring_corr_daily_2011}). Yet, this uncorrelated signal uncovers even more interesting moments in 2011. The most `angry' period is now August 2011, with peaks on the 9th and 10th of this month; this is when the riots took place all over the UK.\footnote{ UK riots time line of events, \url{http://goo.gl/hWF2P}.} The rest distinctive peaks happen on the 18th of January (Jo Yaetes' murder case), 22nd of September (Troy Davis' execution) and the 12th of May (many victims in Syria, Yemen and Libya).

\textbf{Fear}. Fear is the only emotion, where both scoring schemes produce equivalent outputs; the linear correlation between MFMS and MSFMS time series is equal to 0.8085 (see Table \ref{table_mood_scoring_corr_daily_2011}). The two most distinctive fearful moments in both signals (see Figure \ref{fig_mood_fear_2011}) happen during the UK riots (peaking on the 9th of August) and the 9M earthquake in Japan which was followed by a tsunami.\footnote{ More information is available at \url{http://goo.gl/C4iHy}.} Fear is also high on the day of Amy Winehouse's death (23rd of July) which co-occurred with Ander Breivik's attacks in Norway.\footnote{ Wikipedia's page on `2011 Norway Attacks', \url{http://goo.gl/C4iHy}.} As a side effect of the wordings used to describe Halloween's traditions, we see that the period of Halloween is also characterised by -- possibly -- artificial fear.

\textbf{Joy.} Similarly to anger, MFMS (Figure \ref{fig_mood_joy_2011_MFMS}) and MSFMS (Figure \ref{fig_mood_joy_2011_MSFMS}) produce totally different scores for joy and therefore, uncorrelated time series (see Table \ref{table_mood_scoring_corr_daily_2011}). MFMS has the interesting attribute of identifying all significant -- at least for the population of Twitter users -- events throughout the year; the peak happens on the Christmas Day, and the rest significant days -- listed in descending score order -- are New Year's Day, Christmas Eve, New Year's Eve, Valentine's Day, Easter Sunday, Father's Day, St. Andrew's Day and Mother's Day. Since MFMS is based on the assumption that the relative importance of each term in the final mood score is defined by its frequency in the corpus, it is made clear that during traditional celebrations the same set of `joyful words' is overused simplifying their identification. On the other side, MSFMS shows peaks on very different moments, which are much more difficult to explain using calendar events or news outlets. Apart from the peak that clearly correlates with the UK riots (providing evidence that some part of the Twitter population received them positively), the rest could only be explained by major sport events (such as the Cricket World Cup 2011 and the Wimbledon Grand Slam Tennis Tournament), events of international scale such as the Egyptian Revolution or the death of Libyan warlord Muammar Gaddafi, and -- possibly -- weather conditions.

\textbf{Sadness.} According to Figure \ref{fig_mood_sadness_2011_MFMS} (MFMS), the most sad moments occurred on the 5th of May, a date marked by Osama Bin Laden's death, and the 23rd of July (Winehouse's death co-occurring with Anders Breivik's attacks). The rest significant moments in this time series also relate to unexpected deaths (Gary Speed's as well as the beheading of a British woman in Tenerife, Spain). There is also a short peak during the riots in August. MSFMS time series has a correlation of 0.5984 with the result derived from MFMS (see Table \ref{table_mood_scoring_corr_daily_2011}); in those time series, sadness peaks on the 23rd of July, followed by the UK riots period and Gary Speed's death date (27th of November). There are also some other less strong peaking moments that could be matched with sudden or tragic deaths. Overall, in both extracted signals, the main characteristic is that significant moments are usually matched with tragic and/or sudden deaths.

\begin{figure*}
    \begin{center}
    \subfigure[NA -- PA, \ie negative mood types minus joy (MFMS)]{\includegraphics[width=4.2in]{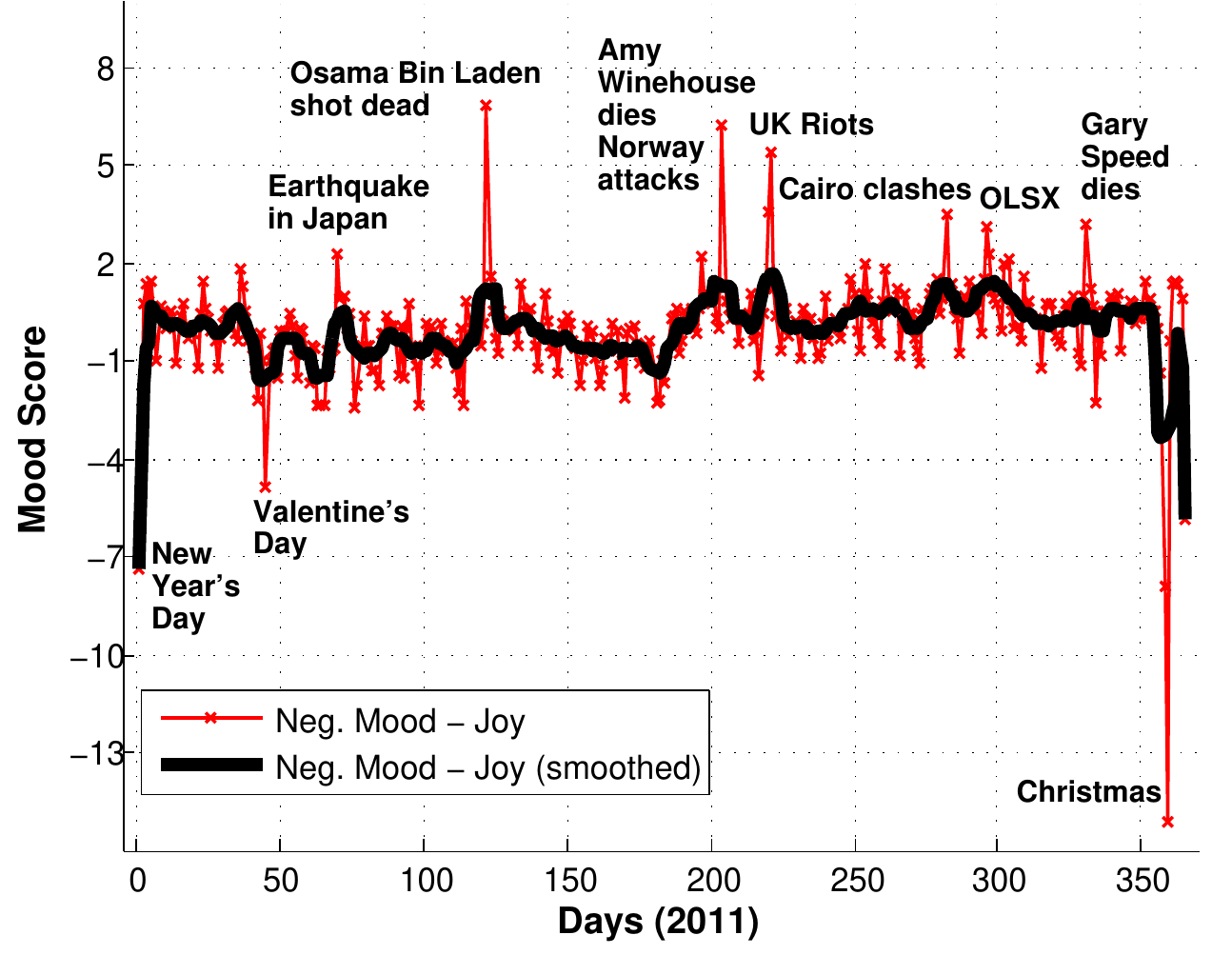}
    \label{fig_mood_neg_2011_MFMS}}
    \hfil
    \subfigure[NA -- PA, \ie negative mood types minus joy (MSFMS)]{\includegraphics[width=4.2in]{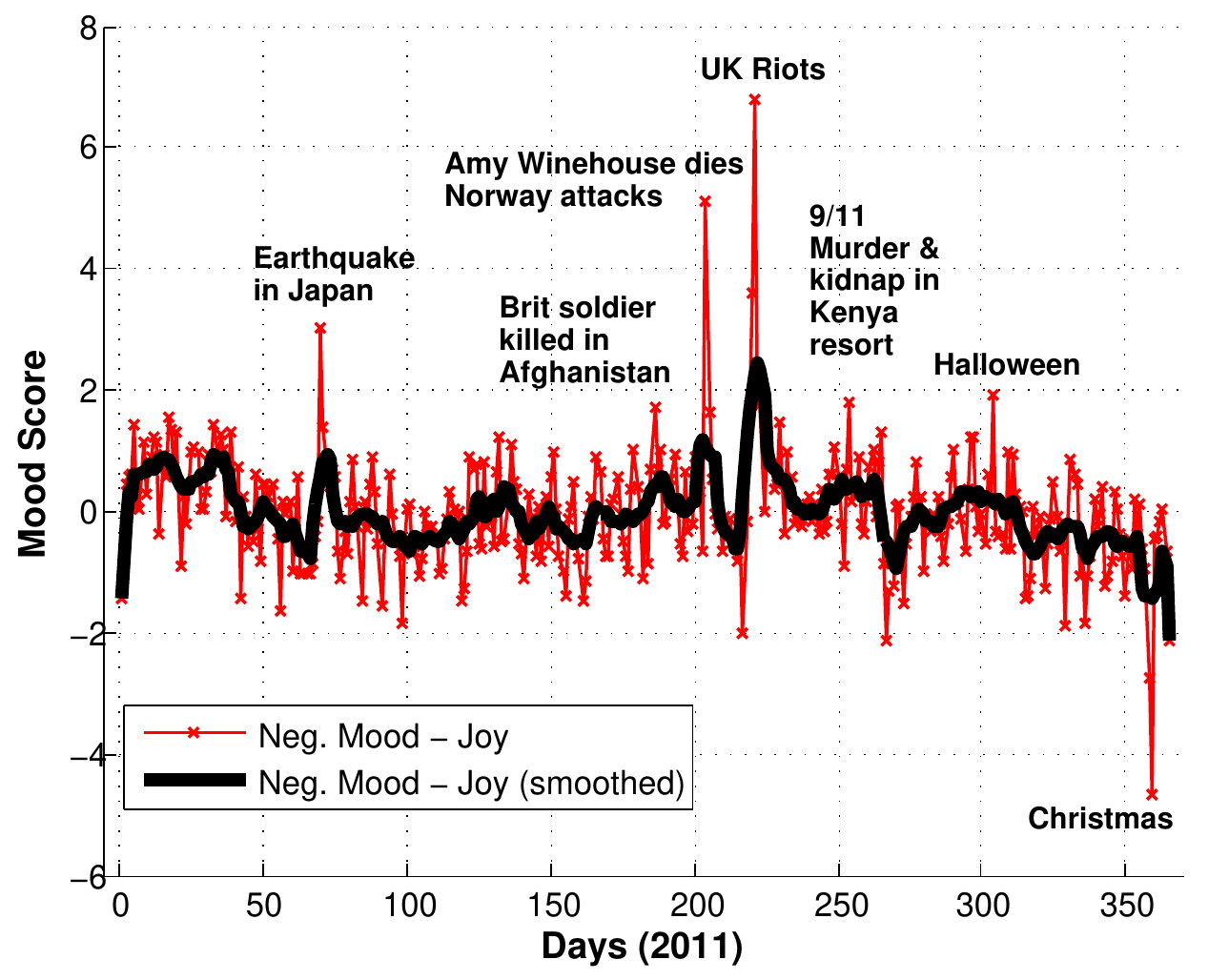}
    \label{fig_mood_neg_2011_MSFMS}}
    \end{center}
    \caption{Daily Scores for Negative Affectivity (anger, fear, sadness) minus Positive Affectivity (joy) in 2011 based on Twitter content geolocated in the UK using MFMS and MSFMS.}
    \label{fig_mood_neg_2011}
\end{figure*}

\textbf{NA minus PA.} Figure \ref{fig_mood_neg_2011} shows the time series of the average NA\index{NA} (the average of anger, fear and sadness scores) minus the score for joy (PA\index{PA}) under MFMS and MSFMS. In that way, we aim to detect days in the year 2011, which triggered all negative emotions of Twitter users. Osama Bin Laden's and Amy Winehouse's deaths together with the UK riots are the most distinguishable events based on the score peaks when applying MFMS; we also see high levels of negative emotion during the earthquake in Japan, the `Occupy London Stock Exchange' movement, Gary Speed's death and so on. The most non negative day -- hence positive -- is the Christmas Day. MSFMS time series has a correlation of 0.5814 (see Table \ref{table_mood_scoring_corr_daily_2011}) with the ones of MFMS; the most distinctive dates are now characterised by the UK riots, Amy Winehouse's death and the earthquake in Japan; again the most positive day in the year is Christmas.

%comment on the tables
\textbf{Linear correlations amongst mood types.} Table \ref{table_mood_types_corr_daily_2011} shows the linear correlations across the different mood types for the two scoring schemes (below and above the main diagonal for MFMS and MSFMS respectively). We can observe that under MFMS, correlations follow common logic, \ie joy is negatively correlated or uncorrelated with the other emotions; positive correlations exist among anger, fear and sadness, the highest of which occurs between fear and sadness (0.4068). On the contrary, MSFMS shows a very high correlation (0.5528) between joy and anger; still the stronger correlation between negative emotions happens between fear and sadness. Looking at the correlations between the smoothed time series of each emotion (Table \ref{table_mood_types_smoothed_corr_daily_2011}), which reveal similarities in weekly tendencies of each signal, we see that under MFMS the negative correlations of joy with fear and sadness are increasing, whereas the most correlated negative emotions are now anger and fear. MSFMS shows that anger is highly correlated with joy (0.8199) as well as joy with sadness (0.5835); this -- at first sight -- might seem to contradict with common sense, however those emotional levels do not express the affective norms of an individual, but of a possibly non uniform set of Twitter users. Therefore, our data set can include moments, where opposing emotions may coexist in that population.

\begin{table}
\caption{Correlations amongst the daily scores of different mood types for MFMS and MSFMS (below and above table's main diagonal respectively). Correlations with an asterisk ($\ast$) are not statistically significant.}
\label{table_mood_types_corr_daily_2011}
\renewcommand{\arraystretch}{1.2}
\setlength\tabcolsep{1mm}
\newcolumntype{C}{>{\centering\arraybackslash} m{2.3cm} }
\newcolumntype{V}{>{\arraybackslash} m{1.8cm} }
\centering
\small
\(\begin{tabular}{|V|C|C|C|C|}
\hline
\textbf{MSFMS}/ \textbf{MFMS}  & \textbf{Anger}     & \textbf{Fear} & \textbf{Joy}  & \textbf{Sadness}\\\hlinewd{2pt}
\textbf{Anger}                 & --                 & 0.415        & 0.5528         & 0.4686\\\hline
\textbf{Fear}                  & 0.2654             & --           & 0.1104         & 0.4772\\\hline
\textbf{Joy}                   & $-$0.0838$^{\ast}$ & $-$0.1322    & --             & 0.2534\\\hline
\textbf{Sadness}               & 0.2169             & 0.4068       & $-$0.1537      & --\\\hline
\end{tabular}\)
\end{table}

\begin{table}[tp]
\caption{Correlations amongst the daily smoothed scores (using a 7-point moving average) of different mood types for MFMS and MSFMS (below and above table's main diagonal respectively). Correlations with an asterisk ($\ast$) are not statistically significant.}
\label{table_mood_types_smoothed_corr_daily_2011}
\renewcommand{\arraystretch}{1.2}
\setlength\tabcolsep{1mm}
\newcolumntype{C}{>{\centering\arraybackslash} m{2.3cm} }
\newcolumntype{V}{>{\arraybackslash} m{1.8cm} }
\centering
\small
\(\begin{tabular}{|V|C|C|C|C|}
\hline
\textbf{MSFMS}/ \textbf{MFMS}  & \textbf{Anger}     & \textbf{Fear} & \textbf{Joy}   & \textbf{Sadness}\\\hlinewd{2pt}
\textbf{Anger}                 & --                 & 0.3684        & 0.8199         & 0.6428\\\hline
\textbf{Fear}                  & 0.321              & --            & 0.0835$^{\ast}$& 0.4266\\\hline
\textbf{Joy}                   & $-$0.0796$^{\ast}$ & $-$0.2149     & --             & 0.5835\\\hline
\textbf{Sadness}               & 0.1185             & 0.2203        & $-$0.1812      & --\\\hline
\end{tabular}\)
\end{table}

\begin{figure*}
    \begin{center}
    \subfigure[Anger -- 1-day: 0.3677, 7-day: 0.3444]{\includegraphics[width=2.5in]{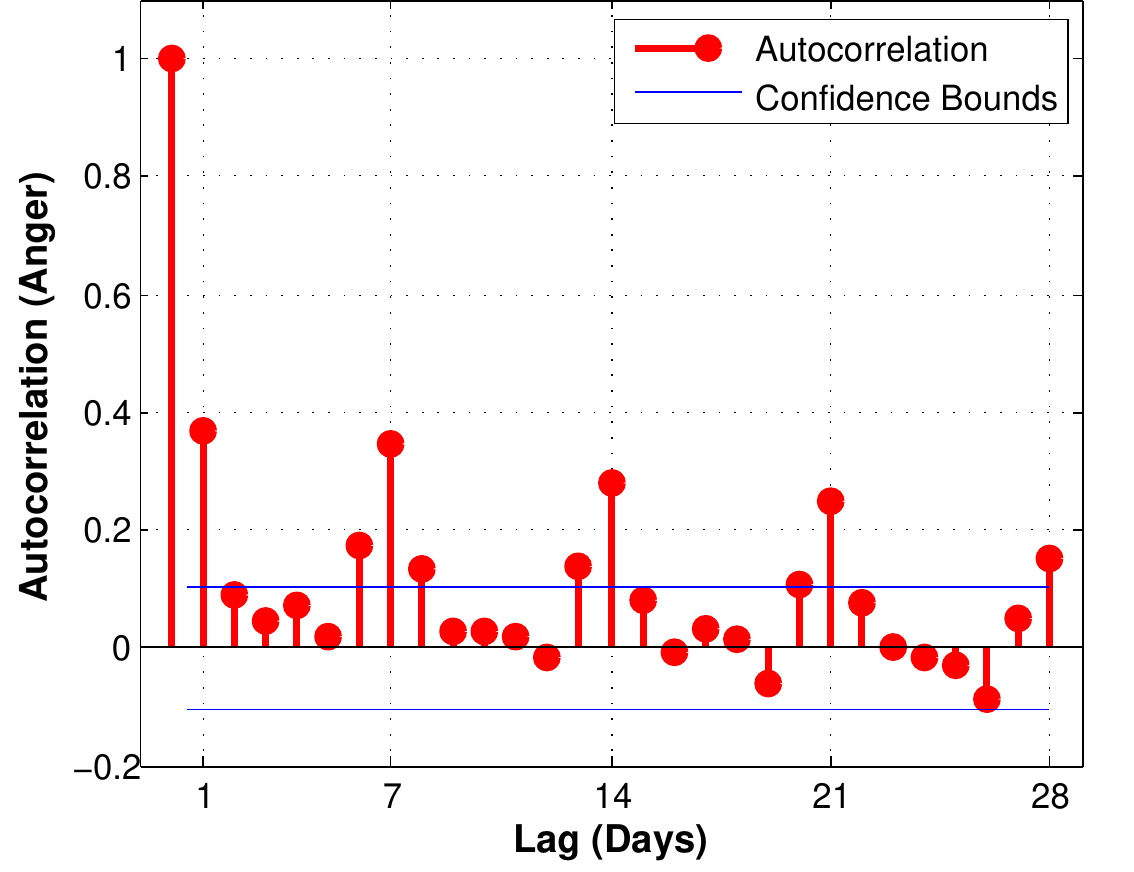}
    \label{fig_mood_autocorr_anger_NoStd}}
    \hfil
    \subfigure[Fear -- 1-day: 0.5335, 7-day: 0.2904]{\includegraphics[width=2.5in]{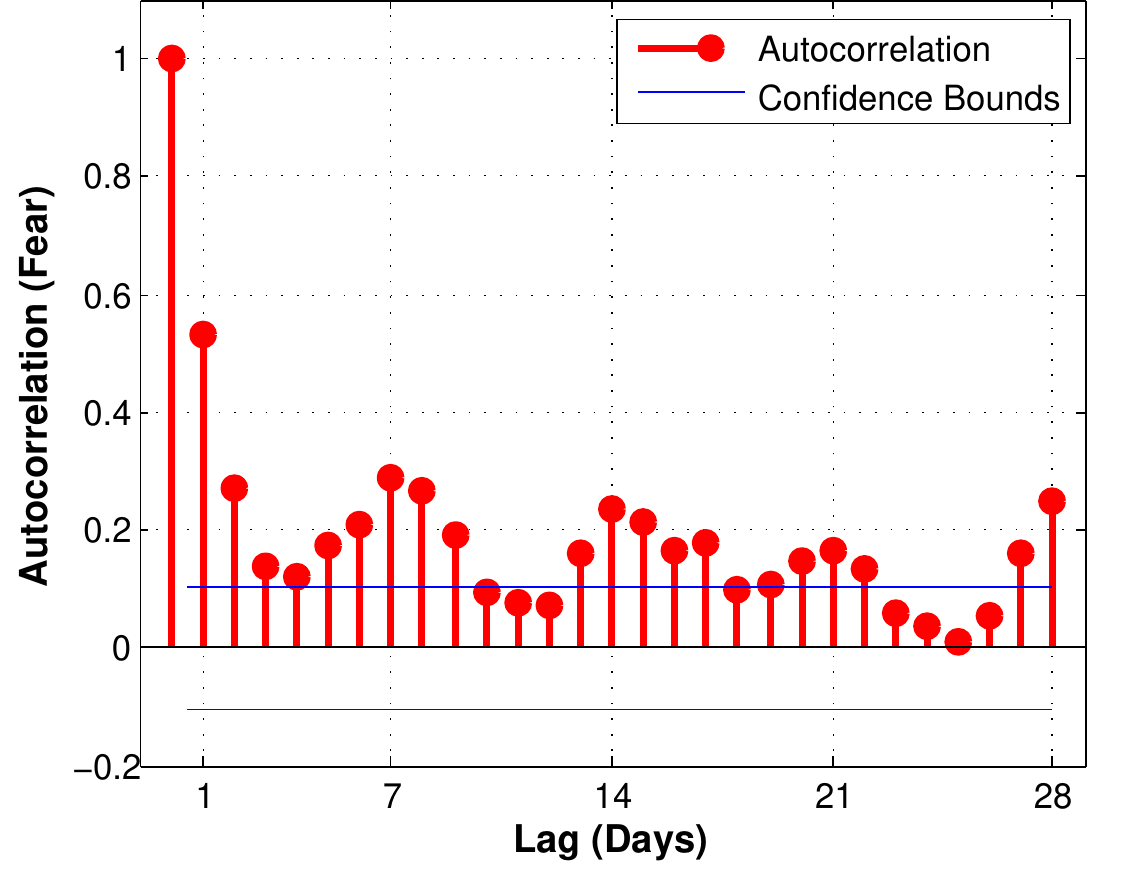}
    \label{fig_mood_autocorr_fear_NoStd}}
    \hfil
    \subfigure[Joy -- 1-day: 0.3892, 7-day: 0.1342]{\includegraphics[width=2.5in]{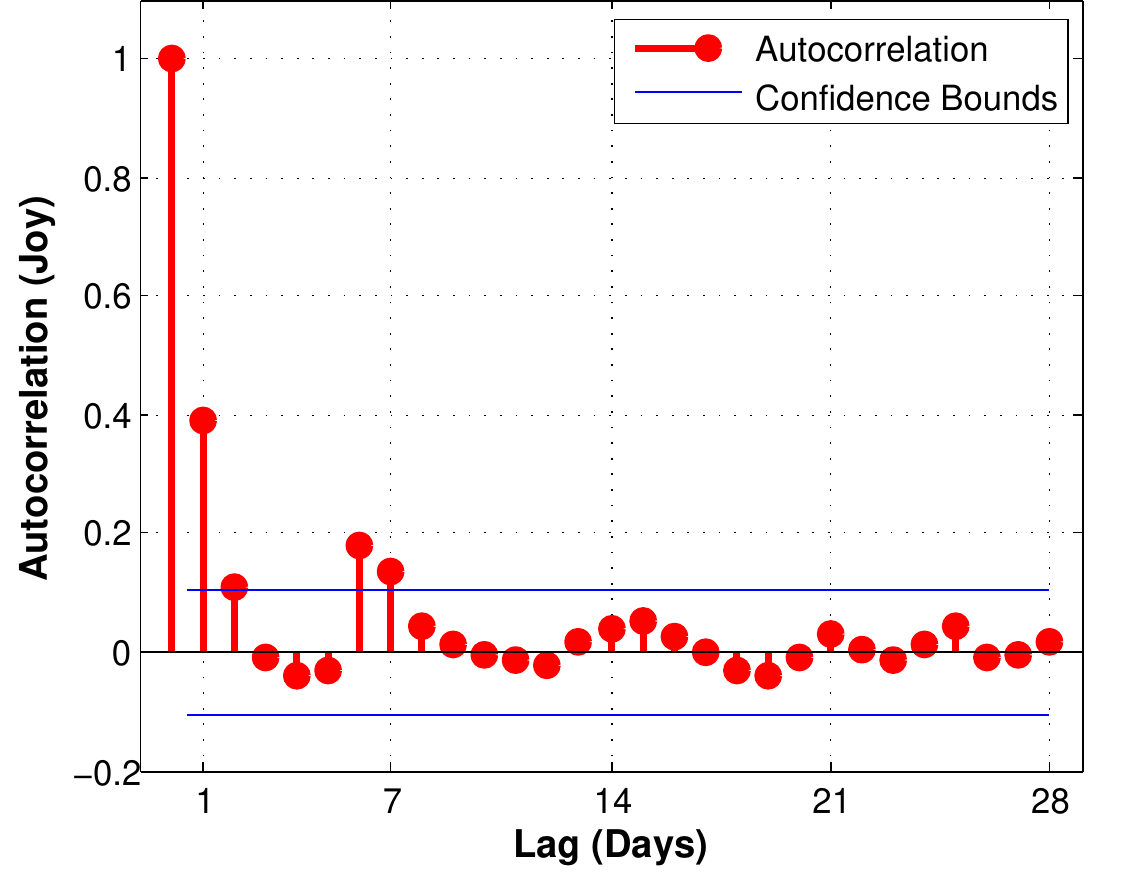}
    \label{fig_mood__autocorr_joy_NoStd}}
    \hfil
    \subfigure[Sadness -- 1-day: 0.315, 7-day: 0.1485]{\includegraphics[width=2.5in]{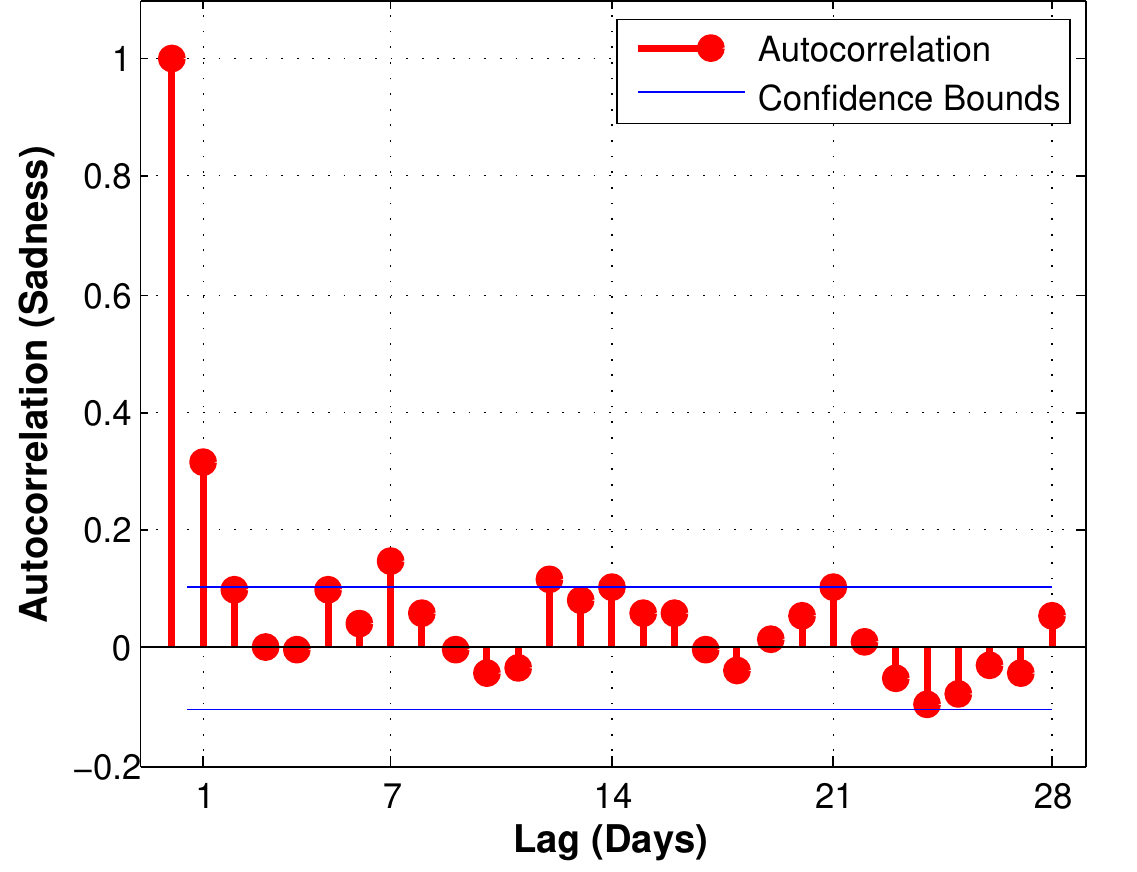}
    \label{fig_mood__autocorr_sadness_NoStd}}
    \end{center}
    \caption{Autocorrelation figures for all mood types under Mean Frequency Mood Scoring (MFMS). Correlations outside the confidence bounds are statistically significant.}
    \label{fig_mood_autocorr_2011_MFMS}
\end{figure*}

\begin{figure*}
    \begin{center}
    \subfigure[Anger -- 1-day: 0.5253, 7-day: 0.414]{\includegraphics[width=2.5in]{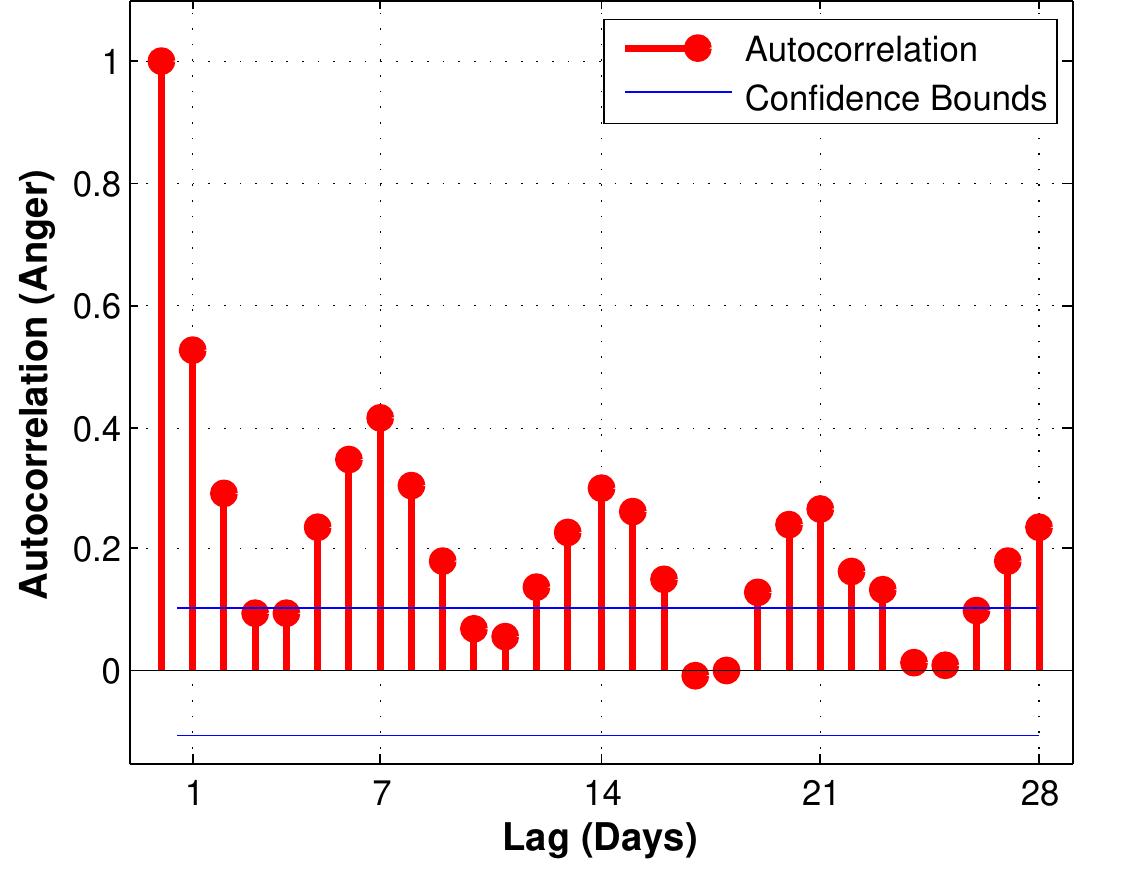}
    \label{fig_mood_autocorr_anger}}
    \hfil
    \subfigure[Fear -- 1-day: 0.4113, 7-day: 0.205]{\includegraphics[width=2.5in]{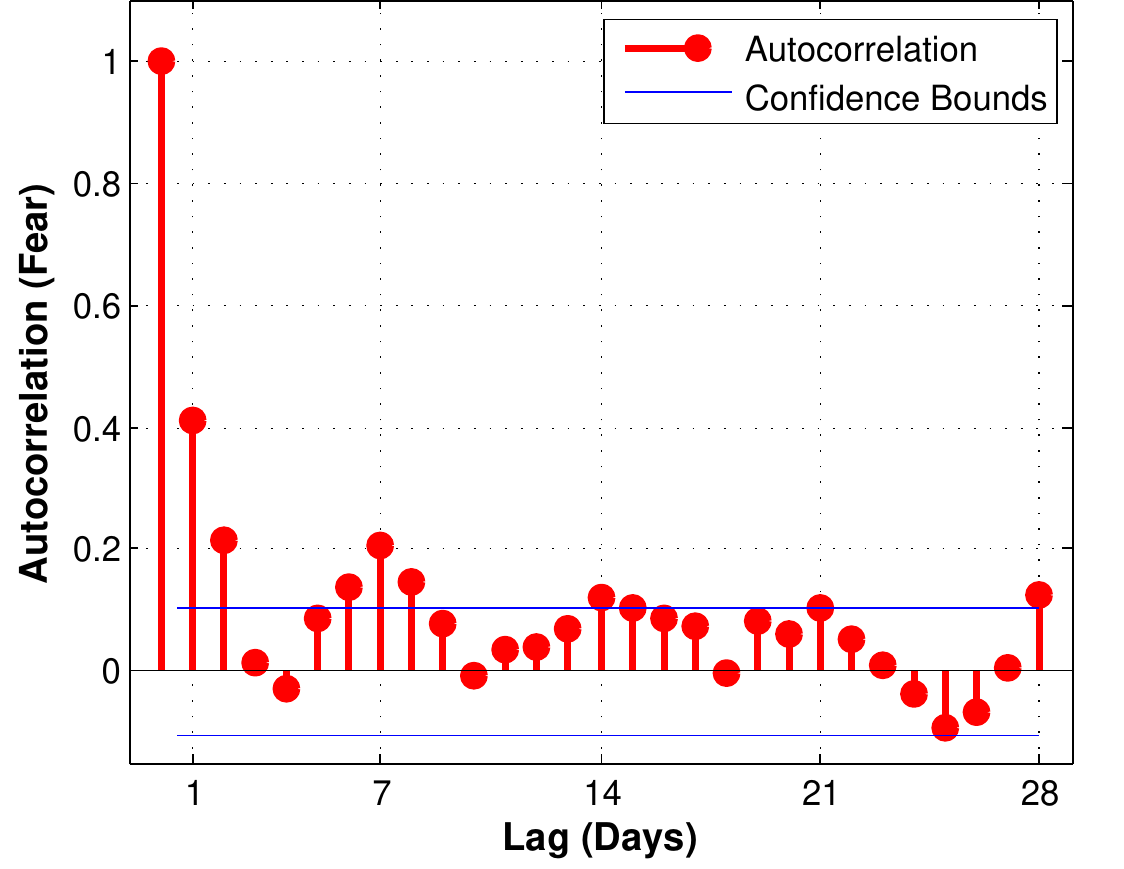}
    \label{fig_mood_autocorr_fear}}
    \hfil
    \subfigure[Joy -- 1-day: 0.5684, 7-day: 0.561]{\includegraphics[width=2.5in]{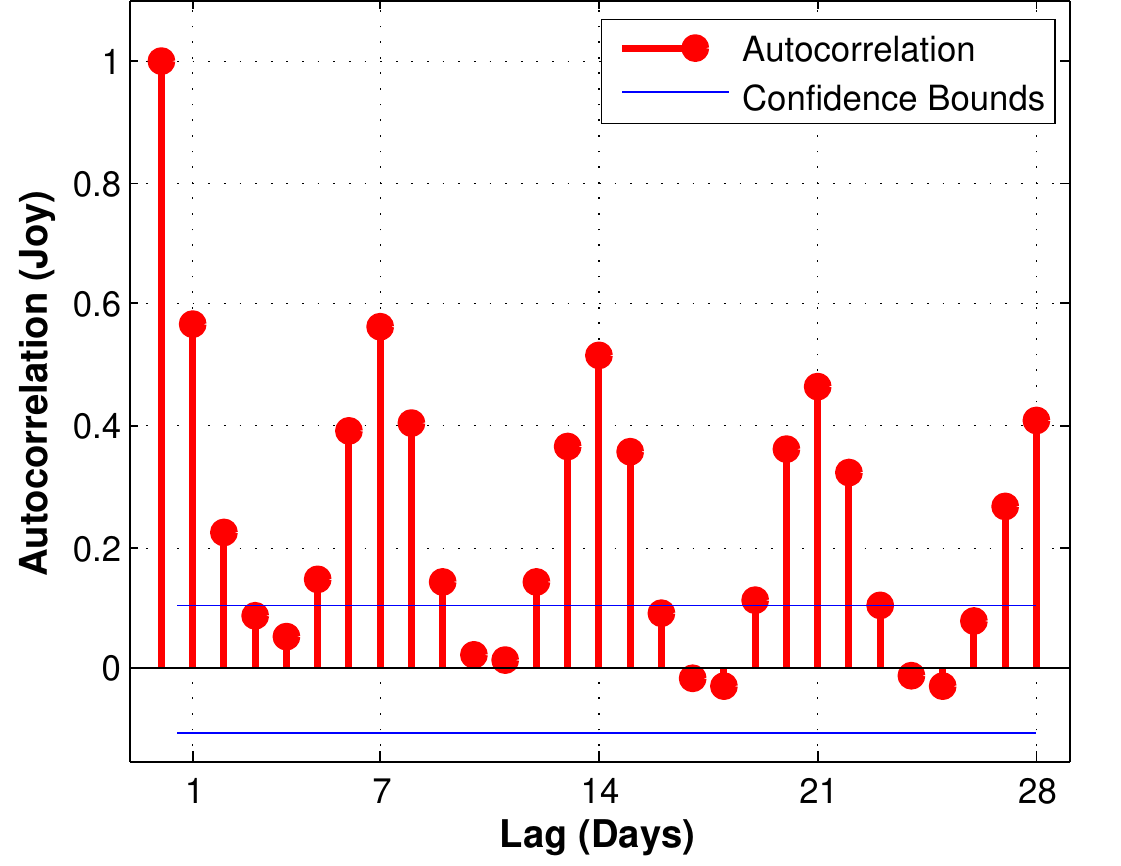}
    \label{fig_mood__autocorr_joy}}
    \hfil
    \subfigure[Sadness -- 1-day: 0.4121, 7-day: 0.2173]{\includegraphics[width=2.5in]{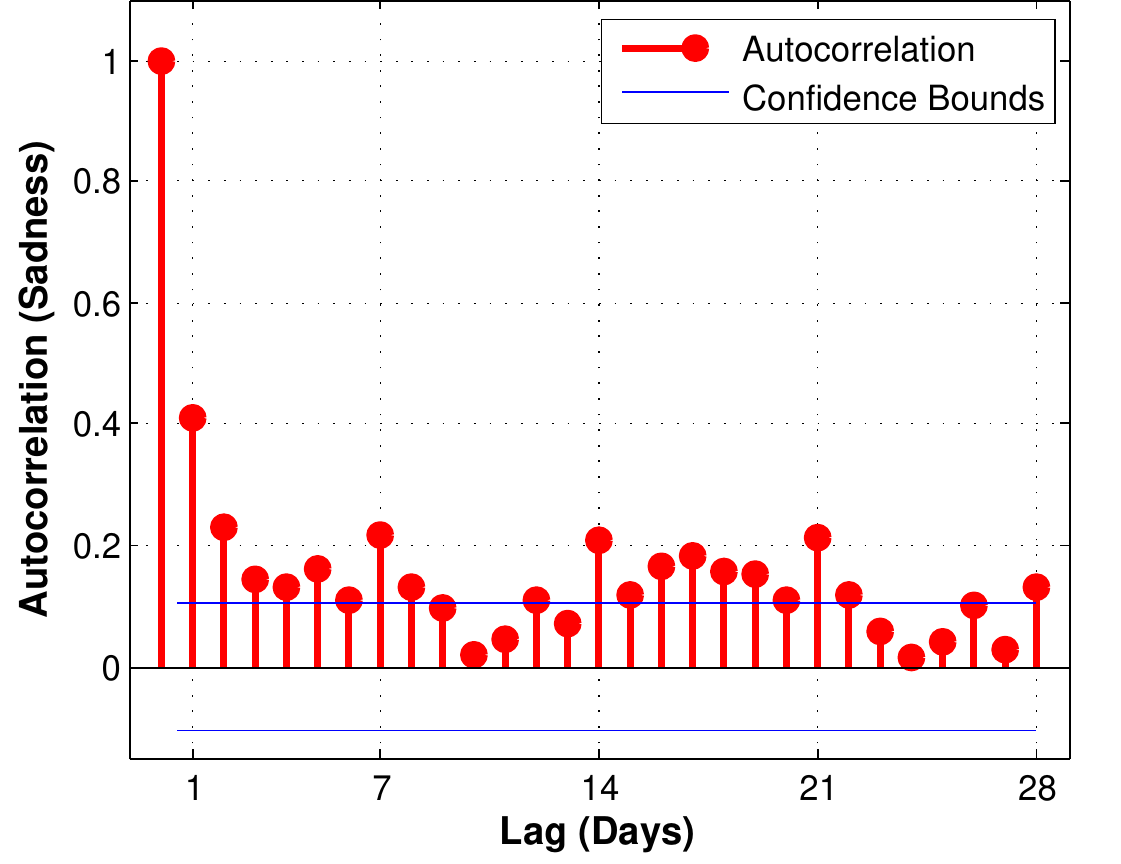}
    \label{fig_mood__autocorr_sadness}}
    \end{center}
    \caption{Autocorrelation figures for all mood types under Mean Standardised Frequency Mood Scoring (MSFMS). Correlations outside the confidence bounds are statistically significant.}
    \label{fig_mood_autocorr_2011_MSFMS}
\end{figure*}

\textbf{Autocorrelation of mood types.} Figures \ref{fig_mood_autocorr_2011_MFMS} and \ref{fig_mood_autocorr_2011_MSFMS} show autocorrelation\index{autocorrelation} figures for up to a 28-day lag ($\ell$) (\ie 4 weeks) under MFMS and MSFMS respectively. Correlations inside the region defined by the positive and negative confidence bounds are not statistically significant. Autocorrelation is used as an experimental indication for the existence of periodic patterns in the data (similarly to the experimental process in Section \ref{section:circadian_mood_periodicity}). All emotion types and under both scoring schemes show the highest autocorrelations for $\ell =$ 1; this may be interpreted by the fact that emotion does not usually change rapidly between consecutive days. MFMS gives the highest 1-day lag autocorrelation to the emotion of fear (0.5335), whereas MSFMS to the emotion of joy (0.5684). The most interesting result here considers the autocorrelations for $\ell =$ 7 (the length of a week) as they reveal that mood has a periodic weekly behaviour. Indeed, under MFMS anger, fear and sadness autocorrelations for $\ell =$ 7 have the second largest value in the 28 extracted ones; the most strong weekly pattern belongs to the emotion of anger (0.3444). An exception is joy's strongest autocorrelation which happens for $\ell =$ 6; this might be the result of having fixed annual celebration dates which distort the overall periodical pattern. We retrieve very similar results under MSFMS; now all emotions (including joy) have a strong 7-day period with the highest belonging to joy (0.561).

%\begin{figure*}
%    \begin{center}
%    \subfigure[Clusters of months in 2011 using MFMS]{\includegraphics[width=3.2in]{Emotion_Clusters_PCA_Months_2011_NoStd}
%    \label{fig_mood_clusters_months_2011_MFMS}}
%    \hfil
%    \subfigure[Clusters of months in 2011 using MSFMS]{\includegraphics[width=3.2in]{Emotion_Clusters_PCA_Months_2011_std}
%    \label{fig_mood_clusters_months_2011_MSFMS}}
%    \end{center}
%    \caption{Clusters of months in 2011 based on four mood types (anger, fear, joy and sadness) by applying PCA.}
%    \label{fig_mood_clusters_months_2011}
%\end{figure*}

\begin{figure*}
    \begin{center}
    \subfigure[Clusters of weekdays using MFMS]{\includegraphics[width=2.85in]{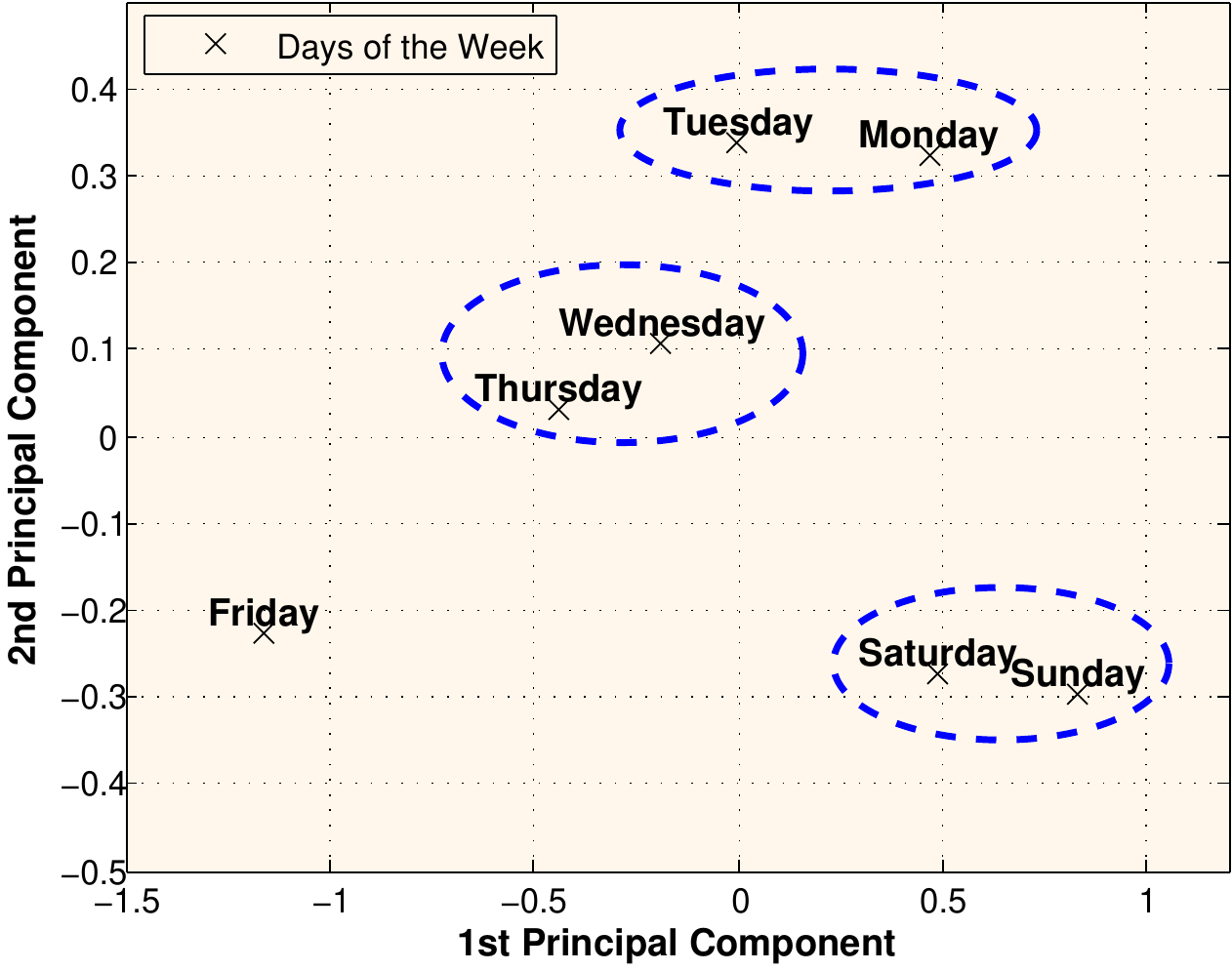}
    \label{fig_mood_clusters_weekdays_2011_MFMS}}
    \hfil
    \subfigure[Clusters of weekdays using MSFMS]{\includegraphics[width=2.85in]{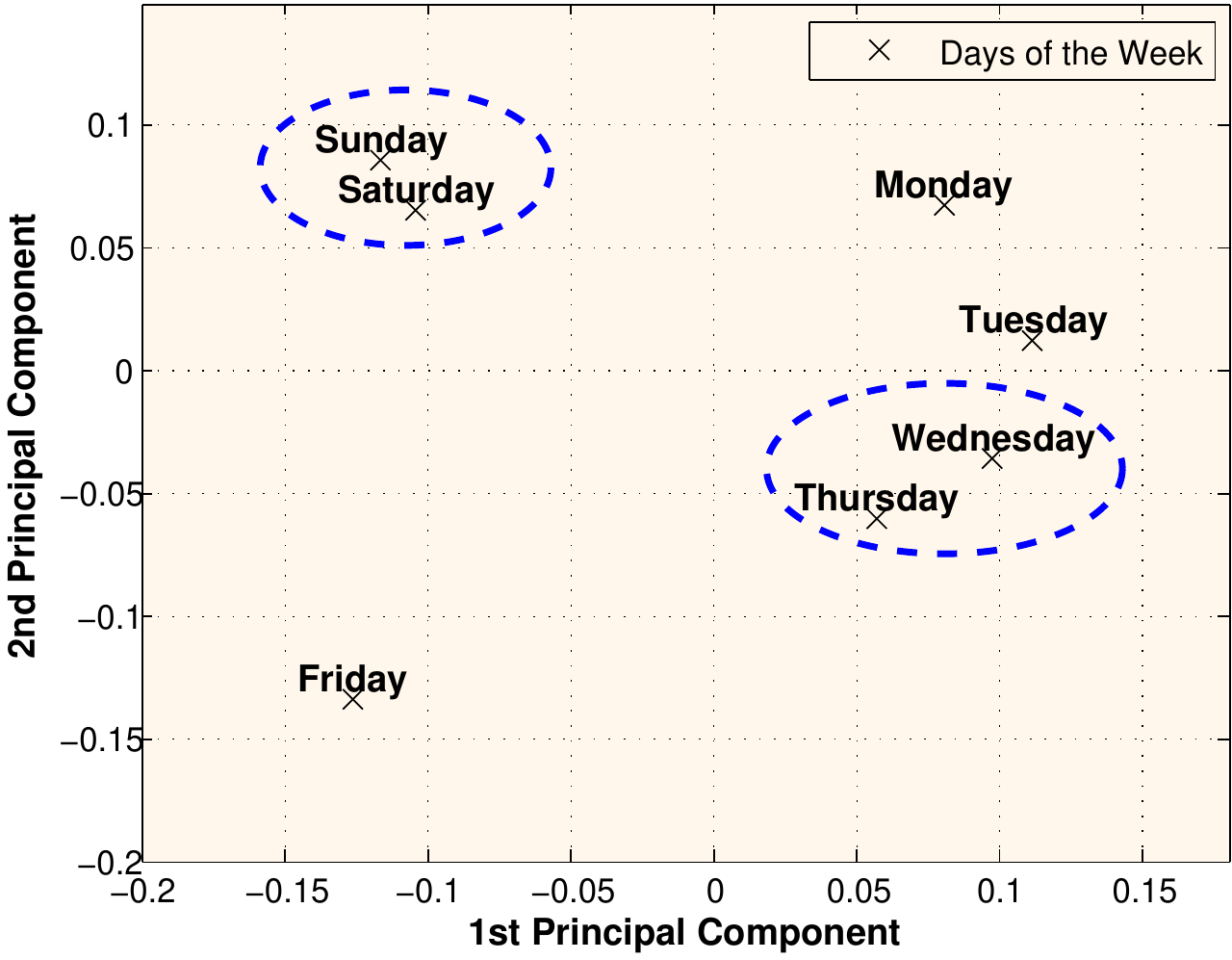}
    \label{fig_mood_clusters_weekdays_2011_MSFMS}}
    \end{center}
    \caption{Clusters of weekdays based on four mood types (anger, fear, joy and sadness) by applying PCA.}
    \label{fig_mood_clusters_weekdays_2011}
\end{figure*}

\begin{figure*}
    \begin{center}
    \subfigure[Clusters of days in 2011 using MFMS]{\includegraphics[width=2.85in]{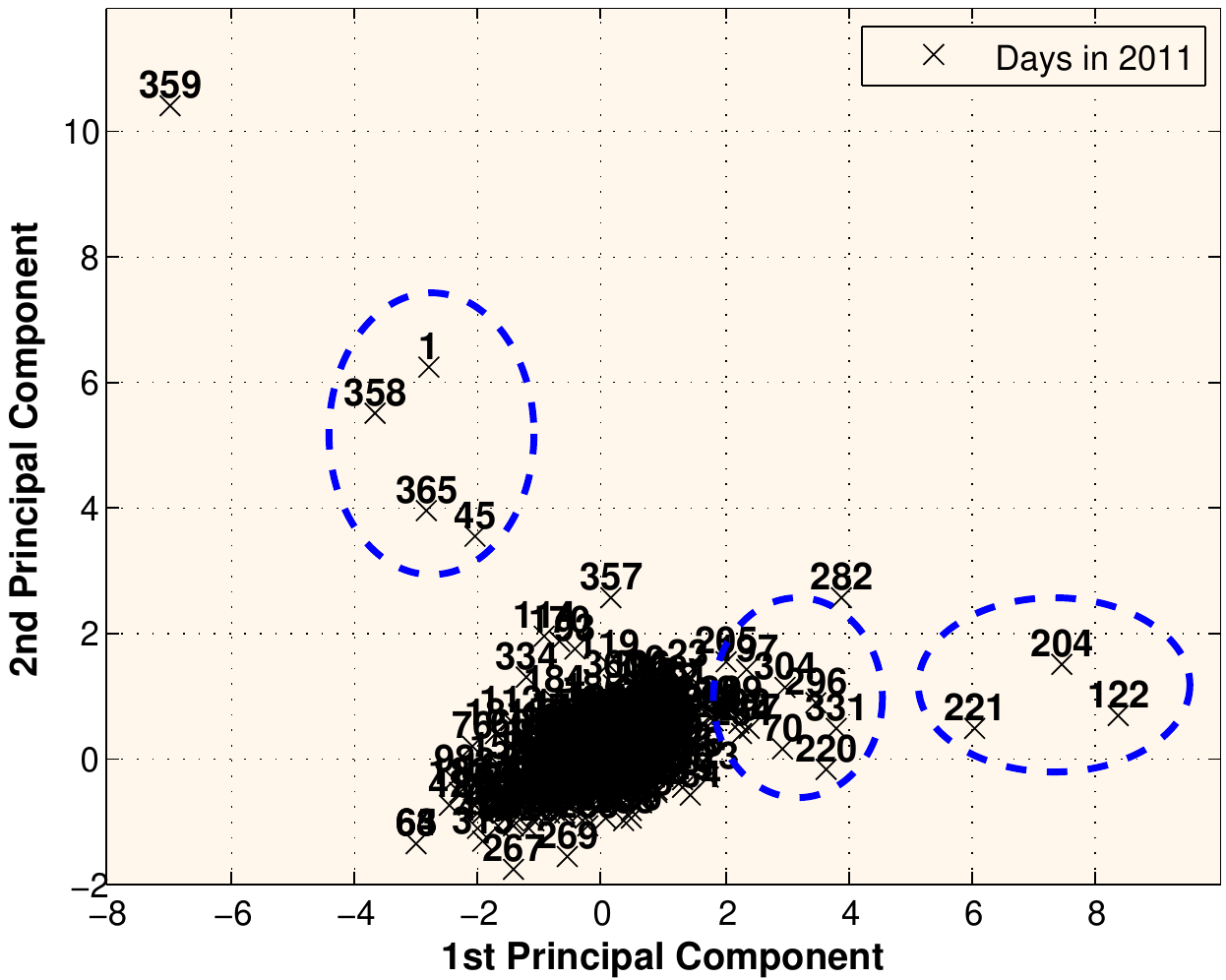}
    \label{fig_mood_clusters_days_2011_MFMS}}
    \hfil
    \subfigure[Clusters of days in 2011 using MSFMS]{\includegraphics[width=2.85in]{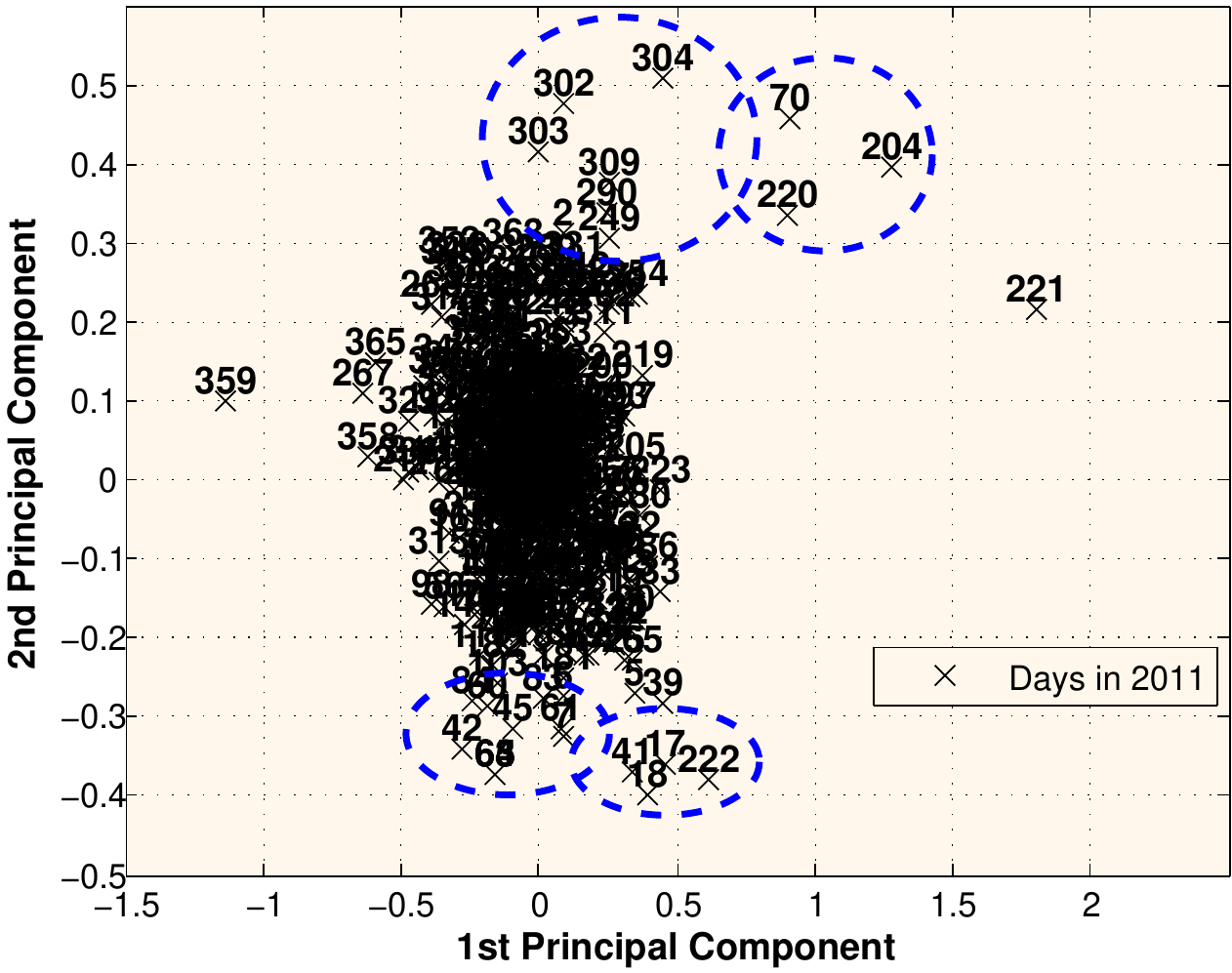}
    \label{fig_mood_clusters_days_2011_MSFMS}}
    \end{center}
    \caption{Clusters of days in 2011 based on four mood types (anger, fear, joy and sadness) by applying PCA.}
    \label{fig_mood_clusters_days_2011}
\end{figure*}

\textbf{Emotion clusters of weekdays and days in 2011.} We average the four mood scores over each weekday in the data set and obtain a 7 (weekdays) $\times$ 4 (emotions) matrix, on which PCA\index{PCA} is applied. The data set is then projected on the first two principal components and the result for both scoring schemes is visualised in Figure \ref{fig_mood_clusters_weekdays_2011}. Under MFMS we see very clear clusters of consecutive days, $\ie$ Monday-Tuesday, Wednesday-Thursday and Saturday-Sunday. Friday seems to be a `special day' as it has been separated from all other days. The same pattern is evident under MSFMS with the only difference that Tuesday is closer to Wednesday not Monday; this gives an indication that under MSFMS Mondays might also be represented as days of distinctive emotional patterns. The second task is to cluster single days in 2011 and therefore, PCA is applied on a 365 (days) $\times$ 4 (emotions) matrix. The derived clusters (see Figure \ref{fig_mood_clusters_days_2011}) under MFMS and MSFMS -- as expected -- are not entirely similar. In Figure \ref{fig_mood_clusters_days_2011_MFMS} we see that the most characteristic moments in 2011 (days 122, 204, and 221 relating to Osama Bin Laden's death, Amy Winehouse's death -- Anders Breivik's attacks in Norway and the UK Riots respectively) have been clustered together. Christmas Day (359) is entirely separated from the rest and some popular days of fixed celebrations (New Year's Day and Eve, Christmas Eve, Valentine's Day) have been also grouped together. We also see that the dates of Gary Speed's death, Japan's earthquake and Halloween are also separated from the main cluster of days. Under MSFMS, we observe that Christmas Day (359) and probably the most distinctive day of the UK riots (221) are entirely separated from the rest and have also been placed on opposite points of the considered 2-dimensional space. A distinctive cluster -- close to the point `221' -- is formed by another date within the UK riots (220) together with the dates of the earthquake in Japan (70) as well as Winehouse's death combined with Breivik's attacks (204). The remaining clusters do not deviate much from the main space where the majority of the points lies.

\subsection{Further discussion of the results}
\label{section_mood_daily_patterns_discussion}
As a general conclusion, we have seen that emotional affect\index{emotional affect} extracted from Twitter's textual stream shows a correlation with real-life events emerging in the UK as well as on an international level. This is something that occurs under both scoring schemes; recall that MFMS is based on the assumption that the frequency of an emotional term defines its importance, whereas MSFMS applies a standardised weighting scheme. Emotional peaks are also present during annual calendar celebrations such as Christmas, Valentine's Day, Halloween and so on. The most significant events of 2011 that concerned the UK, in terms of their correlation with the elevated negative affect on Twitter and under both scoring schemes, are the riots in August, Amy Winehouse's death (combined with the massacre in Norway) and the severe earthquake in Japan.

The two scoring schemes gave rise to quite different events, especially for the emotions of anger and joy; fear and sadness are much more consistent across the proposed schemes. The autocorrelation figures under both scoring schemes indicate that all emotions correlate between consecutive days (lag equal to 1), but when the lag is increased to 2, autocorrelations decrease significantly. This is an indicator pointing out that the state of mood might have a relative 1-day consistency, but usually changes after 2 days have passed. This perhaps is due to `sudden' events which force emotional reconfigurations. Interestingly, autocorrelations -- for all mood types -- increase again, when the lag is equal to 7 revealing a weekly periodic pattern. As a result, there is statistical proof that mood is also affected by the day of the week, \ie there is a common emotional ground for each weekday over time. Both scoring schemes indicate that anger might have the strongest weekly autocorrelation (1st under MFMS, 2nd under MSFMS); based solely on MSFMS only, joy has the highest weekly period.

By clustering weekdays based on the two principal components of the 4-dimensional daily affective norm, we discovered that -- under both scoring schemes -- consecutive days, and especially weekends, tend to group together in a common cluster, apart from `Fridays' which seem to be unique days by displaying a distinctive emotional behaviour. Finally, by extracting clusters of single dates in 2011, we can observe that all important events are positioned on identifiable areas of the 2-dimensional space and sometimes are clustered together.

Similarly to the previous section, those results face the same limitations. Most importantly, we acknowledge the fact that the users of Twitter might be a biased sample of the population; by using a large data set, we try to overcome or smooth such biases. In contrast to the previous section, where no actual ground truth indications existed (apart from fundamental affective norms extracted by psychiatrists), here we show some evidence -- coming primarily by events reported in the news --, which can partly justify moments in the mood signals.

\section{Summary of the chapter}
\label{section_summary_chapter_mood}
In this chapter, we investigated seasonal circadian and daily patterns of emotional affect\index{emotional affect} on Twitter content. We considered four mood types, three expressing NA\index{NA}, \ie anger, fear and sadness, and one expressing PA\index{PA}, \ie joy. Two methods have been applied in order to extract mood scores; the first one (MFMS) weighs each emotional term proportionally to its frequency in the corpus, whereas the second one (MSFMS) removes this factor by considering a standardised version of each term's frequency.

Firstly, we extracted seasonal diurnal mood rhythms based on Twitter content published within the geographical region of the UK. The derived results are partly in an agreement with other psychiatric studies, but contradict with the ones presented in a recent similar work \cite{Golder2011}. Among other interesting findings, we acquired a clear indication that -- within our data set and considering the inevitable biases -- PA is stronger than NA during the daytime and vice versa during the night hours. Furthermore, all mood patterns show a level of periodicity; the most common values for the period were 24 and 168 hours (lengths of a day and a week respectively) and the less periodic emotion was the one of anger.

%Under MFMS, circadian patterns for the negative mood types show an increase during the day and peak during midnight. However, under MSFMS -- which also shows a higher statistical significance --, anger and fear have also an intermediate peak in afternoon and morning respectively, whereas sadness peaks in the morning only. Joy peaks approximately at the same morning hourly intervals in both scoring schemes, but under MSFMS the scores on the left and right apart from being symmetrical, they are also much lower compared to the ones retrieved by MFMS. One more mention worthy observation is that the circadian patterns of fear and sadness have the highest correlation under both scoring schemes and therefore, we a connection between those two emotions is evident. By decoupling the signal into a winter and a summer one, we can spot some small deviations and differences at specific time intervals, but the general pattern remains similar with an average linear correlation of 0.825 (and $\geq$ 0.74) across all schemes. The main common pattern here was that negative emotions tend to be higher in the winter mornings than during summer; joy displayed an increased correlation between winter and summer (above 0.9 in both scoring schemes).

By analysing daily mood patterns, we gained evidence supporting that real-life events affect the mood of Twitter users. By reversing this relationship, mood figures could be used to track significant events emerging in real life; for example during 2011, we saw that the UK riots, Winehouse's death combined with the Norwegian massacre as well the Japanese earthquake or Osama Bin Laden's death affected the emotional output of Twitter users. Among other relevant discoveries, we could also track annual celebrations and calendar events such as Christmas, the New Year's Day, Valentine's, Halloween, the Father's Day and so on. The two different scoring schemes give very different and uncorrelated results for the emotions of anger and joy; however, for fear and sadness -- two emotions that are also correlated to each other --, scoring schemes produced quite similar results. Nevertheless, in the analysis of daily mood patterns, we show that the output of the two scoring schemes can be combined as some events identified in one scheme are missed by the other one.

Finally, we showed that all the investigated affective types have a periodic behaviour for lags equal to 1 or 7 days, meaning firstly that consecutive days are more likely to be emotionally correlated, and secondly, that mood might also have a dependency on the specific day of the week. In this direction, we clustered all weekdays based on their mood scores showing that consecutive days are positioned closer on a 2-dimensional space with the exception of `Fridays'; we also showed that dates with exceptionally negative or positive mood are also identifiable on this space. 

%% file: Chapters/Chapter7.tex
\chapter{Pattern Discovery Challenges in User Generated Web Content}
\label{chapter:pattern_discovery}

\rule{\linewidth}{0.5mm}
In this chapter, we present some additional but preliminary work that is mainly based on Pattern Discovery and NLP\index{NLP} methodologies and proposes ways to explore further the rich information on the Social Web and in particular, Twitter. At first, we investigate spatiotemporal content relationships between locations in the UK by forming networks based on similarities between their published tweets. We show that those networks are stable over time and therefore, form a trustful description of how content is shared across UK's microblogging space. Then, we briefly examine posting time patterns on Twitter. We show that those patterns are slightly different between weekdays and weekends -- something explained by the different characteristics of those time periods -- and then use posting time data as features in order to form clusters of days. Finally, a preliminary method is presented where we aim to address the interesting problem of voting intention inference based on inputs from the Social Media; UK 2010 General Election is used as a case study.\footnote{ This work has been published in a technical report titled as ``On voting intentions inference from Twitter content: a case study on UK 2010 General Election'' \cite{Lampos2012b}.} At this point, we would like to remind the reader that the work presented in this chapter is considered as work-in-progress; still, some results were interesting and could serve as the basis for future research directions.
\newline \rule{\linewidth}{0.5mm}
\newpage

%%%%%%%%%%%%%%%%%%%%%%%%%%%%%%%%%%%%%%%%%%%%%%%%%%%%%%%%%%%%%%%%%%
\section{Spatiotemporal relationships of Twitter content}
\label{section:time-spatial}
In the following sections, we investigate spatiotemporal relationships of the content published on Social Networks such as Twitter. One of our aims is to study if and how the geographical distance between locations influences the respective similarity between tweets geolocated in their vicinity. Additionally, we look into whether content originated in different locations but at the same time window has significant similarities, and if such similarities exist, how they are affected by the length of the time window. The latter question is answered with the formation of networks that show how content is shared among locations.

\subsection{Data}
For the following experiments, we are using a set of 71 million tweets published on Twitter during the second semester of 2010 (from July to the end of December) and geolocated in 54 urban centres in the UK. The process of data collection is similar to the one carried out in our previous experiments and has already been described in Section \ref{section:crawlers_data_collection_storage}. As a reminder, the radius of each urban centre or location is equal to 10Km, which -- in turn -- means that all tweets geolocated within this area are mapped to the corresponding location.

We have conducted the same experimental process under two configurations of the settings. In the first experiment, a document is formed by the set of tweets geolocated in one of the urban centres during a day. Since the data set is comprised by 184 days, the total number of documents considered is 184 days $\times$ 54 locations $=$ 9,936. In the second experiment, the time span of a document is reduced to 10 minutes, but we also reduce the total number of days to one month (December). Therefore, the total number of documents becomes 31 days $\times$ 144 documents per day $\times$ 54 locations $=$ 241,056.

\subsection{Content correlation vs. geographical distance}
\label{section_content_correlation_vs_geographical_distance}
Is location proximity proportional to content similarities on Twitter? To answer this question we investigate the relationship between those two quantities, \ie the pairwise geographical distance between locations and the respective similarity of their textual content. The pairwise geographical distance is approximated by using the longitude and latitude of the urban centres, and the content correlation is computed using the cosine similarity\index{cosine similarity} metric (see Appendix \ref{Ap:Cosine_Similarity}).

We denote the set of the 54 urban centres with $\mathcal{U} = \{u_i\}$, $i\in\{$1, ..., 54$\}$ and the considered time intervals as $t_j$, where $j\in\{$1, ..., 184$\}$ in the first experiment and $j\in\{$1, ..., 31 $\times$ 144 = 4464$\}$ in the second. A document (set of tweets) geolocated in $u_i$ during $t_j$ is denoted as $\mathcal{D}(i,j)$. The VSR of $\mathcal{D}(i,j)$ is retrieved by applying TF-IDF\index{TF-IDF} on the entire collection of documents -- after removing stop words and stemming -- and is denoted by $d_{i,j}$.

The pairwise cosine similarity of all the documents published during $t_j$ is computed by:
\begin{equation}
\text{Cosine-Similarity}(i,k,j) = \frac{d_{i,j} \cdot d_{k,j}}{\|d_{i,j}\|_{\ell_2}\|d_{k,j}\|_{\ell_2}}\text{, }i\neq k.
\end{equation}
In total we have ${\text{54} \choose \text{2}} =$ 1431 location pairs; after removing 15 pairs with geographical distance smaller than 20Km,\footnote{ Those location pairs might share identical tweets given that the radius of each urban centre is 10Km.} we end up with 1416 pairs which are denoted by $\mathcal{L}_p$.

For each $\{u_i,u_k\}\in\mathcal{L}_p$ we compute the average cosine similarity over all time intervals $t_j$:
\begin{equation}
\text{Average-Similarity}(i,k) = \frac{1}{n} \sum_{j=1}^{n} \text{Cosine-Similarity}(i,k,j),
\end{equation}
where $n$ is the number of time intervals.

\begin{figure*}
    \begin{center}
    \subfigure[Document's time span: 1 day] {\includegraphics[width=4.5in]{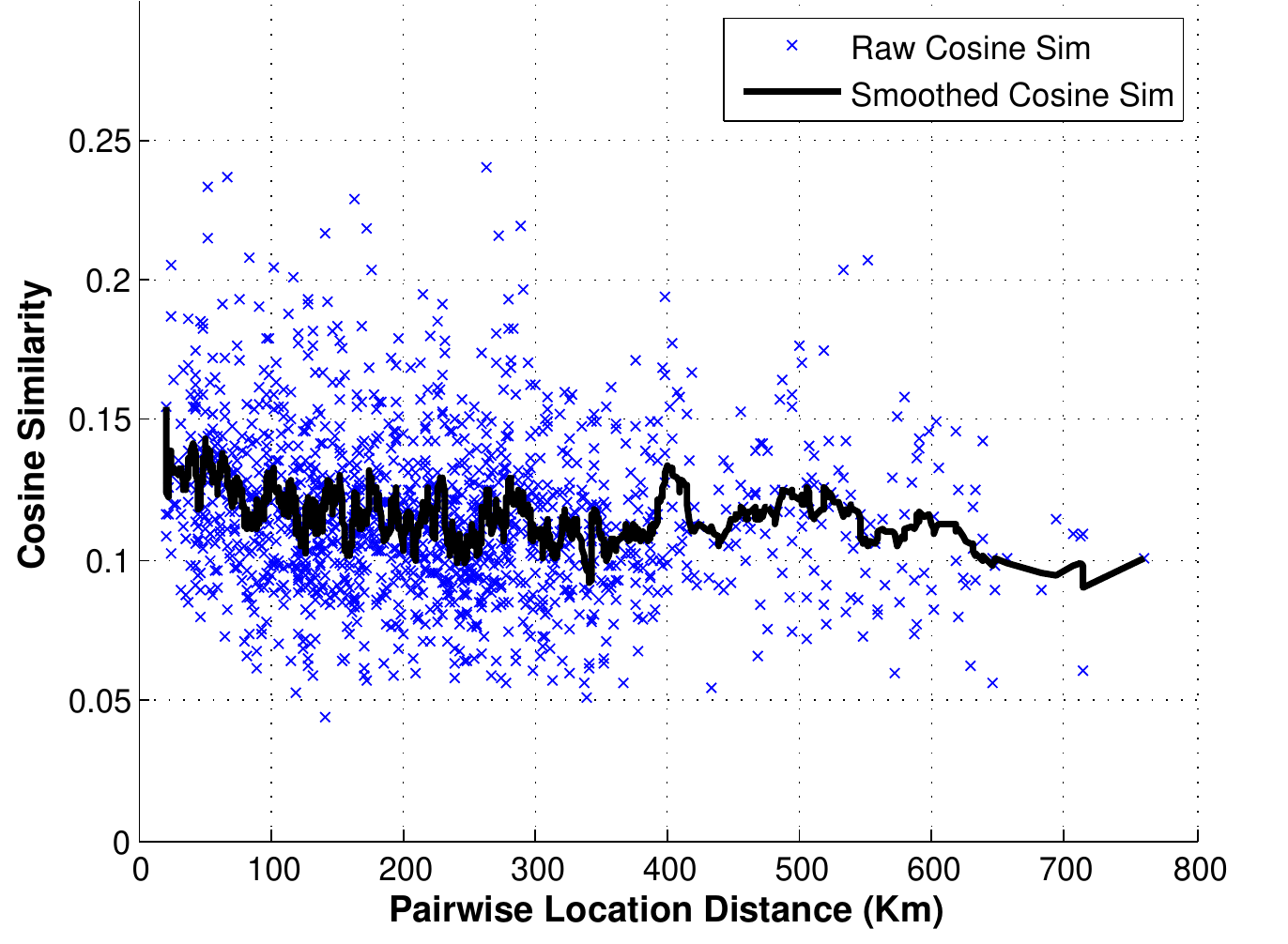}
    \label{fig_sim_vs_dist_experiment1}}
    \hfil
    \subfigure[Document's time span: 10 minutes]{\includegraphics[width=4.5in]{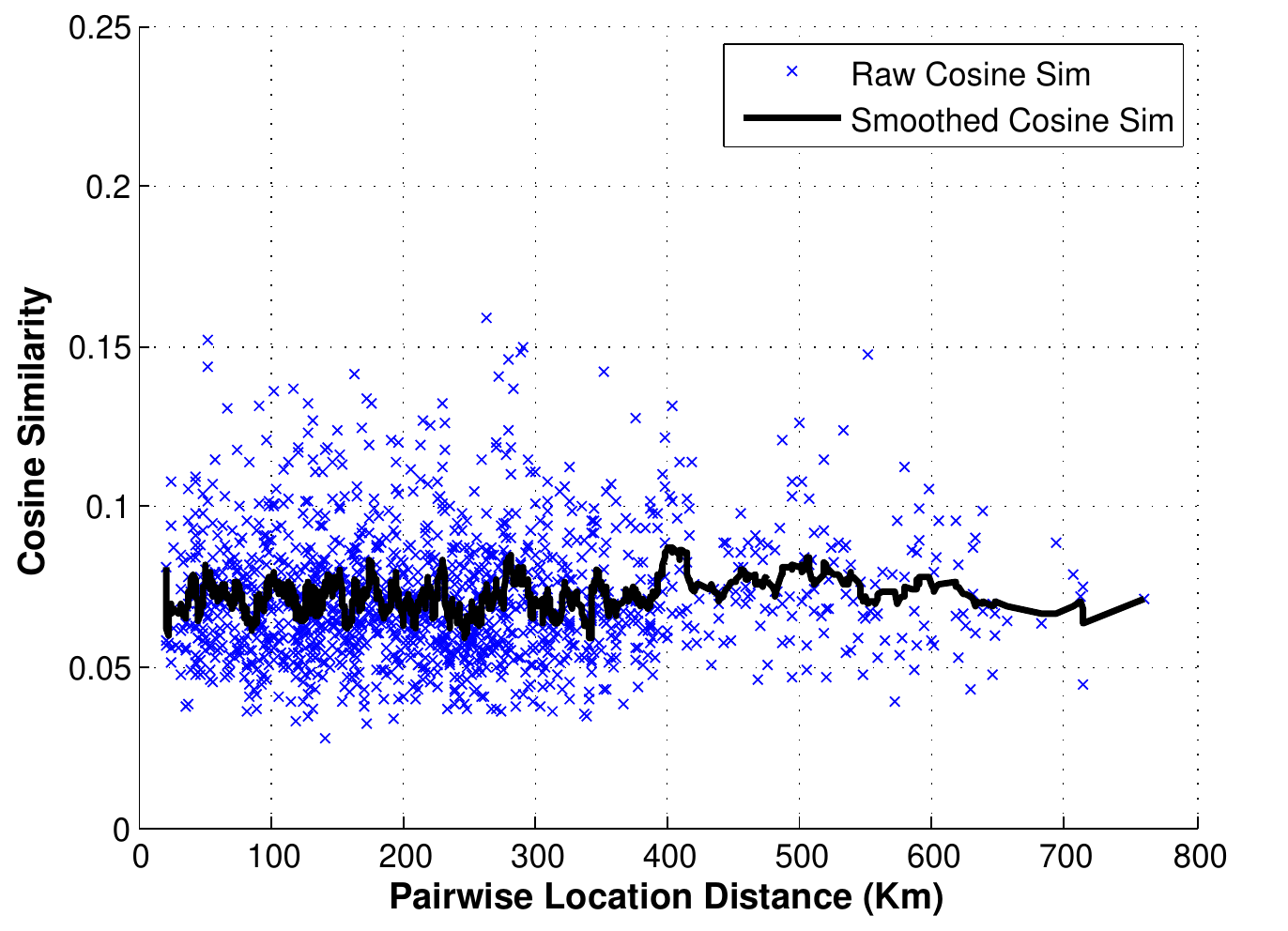}
    \label{fig_sim_vs_dist_experiment3}}
    \end{center}
    \caption{The average pairwise cosine similarity for all location pairs plotted against their respective geographical distance. The x's denote the raw points, whereas the black line is their smoothed equivalent (20-point moving average).}
    \label{fig_sim_vs_dist}
\end{figure*}

The average cosine similarity for both experimental settings is depicted in Figures \ref{fig_sim_vs_dist_experiment1} and \ref{fig_sim_vs_dist_experiment3}. By looking at the smoothed equivalents, we see that no obvious relation exists between geographical distance and content similarity; in the first experiment there is a slight decrease in similarity as the distance increases, but this is relatively small to be significant, whereas in the second experiment similarity seems to be unaffected by the distance. In both figures, we can observe high content correlations for urban centres located 500 Km from each other. In addition, the linear correlation between the two results is equal to 0.9185 with a p-value $\ll$ 0.001, showing that the time windows used do not affect much the final pattern. However, it also becomes apparent that a decreased time window produces lower cosine similarities.

Therefore, an answer to one of our primary questions is that location proximity does not necessarily affect content similarity on social networks such as Twitter. Of course, looking at Twitter data on a global level, this pattern is very likely to change. We cannot generalise this result due to the fact that we are using Twitter content geolocated in the UK only. On the other hand, this could also be one of the main characteristics of Twitter and microblogging in general: the ability to quickly spread and share information between places regardless of their actual geographical proximity, at least within the well defined space of a country or a unitary state (such as the UK).

\begin{figure*}
    \begin{center}
    \subfigure[Document's time span: 1 day] {\includegraphics[width=2.75in]{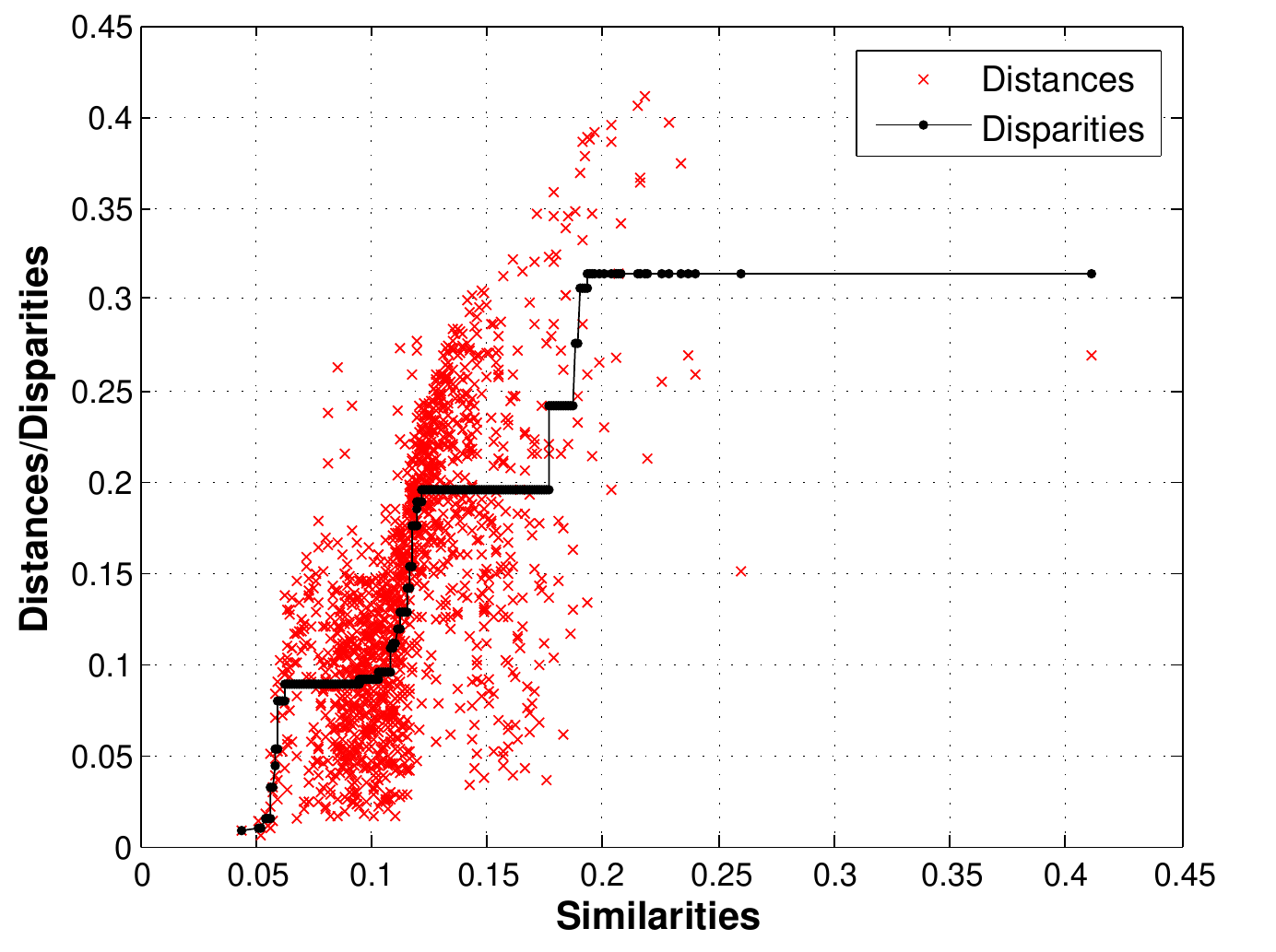}
    \label{fig_sim_MDS_disp_experiment1}}
    \hfil
    \subfigure[Document's time span: 10 minutes]{\includegraphics[width=2.75in]{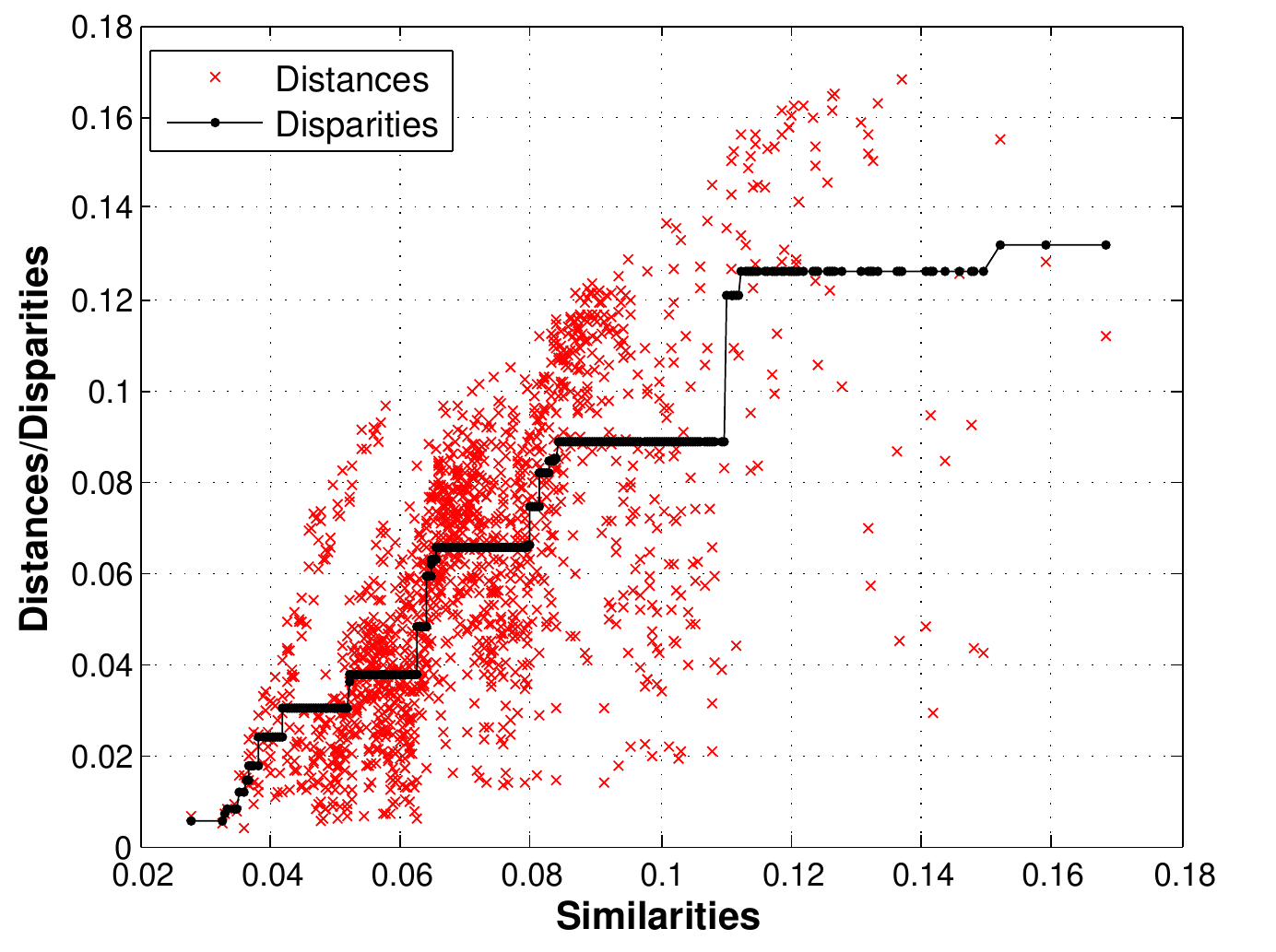}
    \label{fig_sim_MDS_disp_experiment3}}
    \end{center}
    \caption{Shepard plots showing the pairwise similarities and their monotonic transformation (disparities) when applying nonmetric MDS.}
    \label{fig_sim_MDS_disp}
\end{figure*}

\begin{figure*}
    \begin{center}
    \subfigure[Document's time span: 1 day] {\includegraphics[width=5.5in]{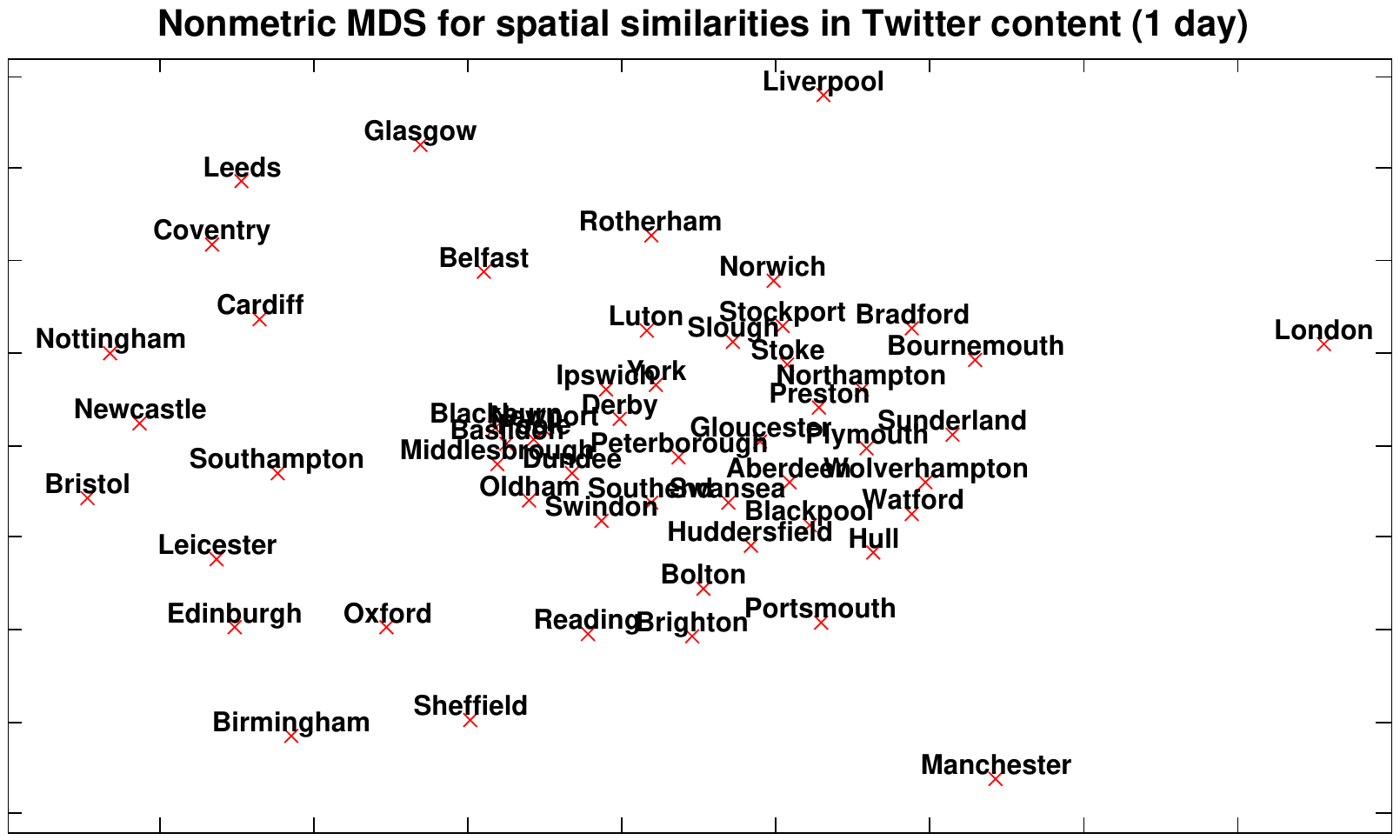}
    \label{fig_sim_MDS_experiment1}}
    \hfil
    \subfigure[Document's time span: 10 minutes]{\includegraphics[width=5.5in]{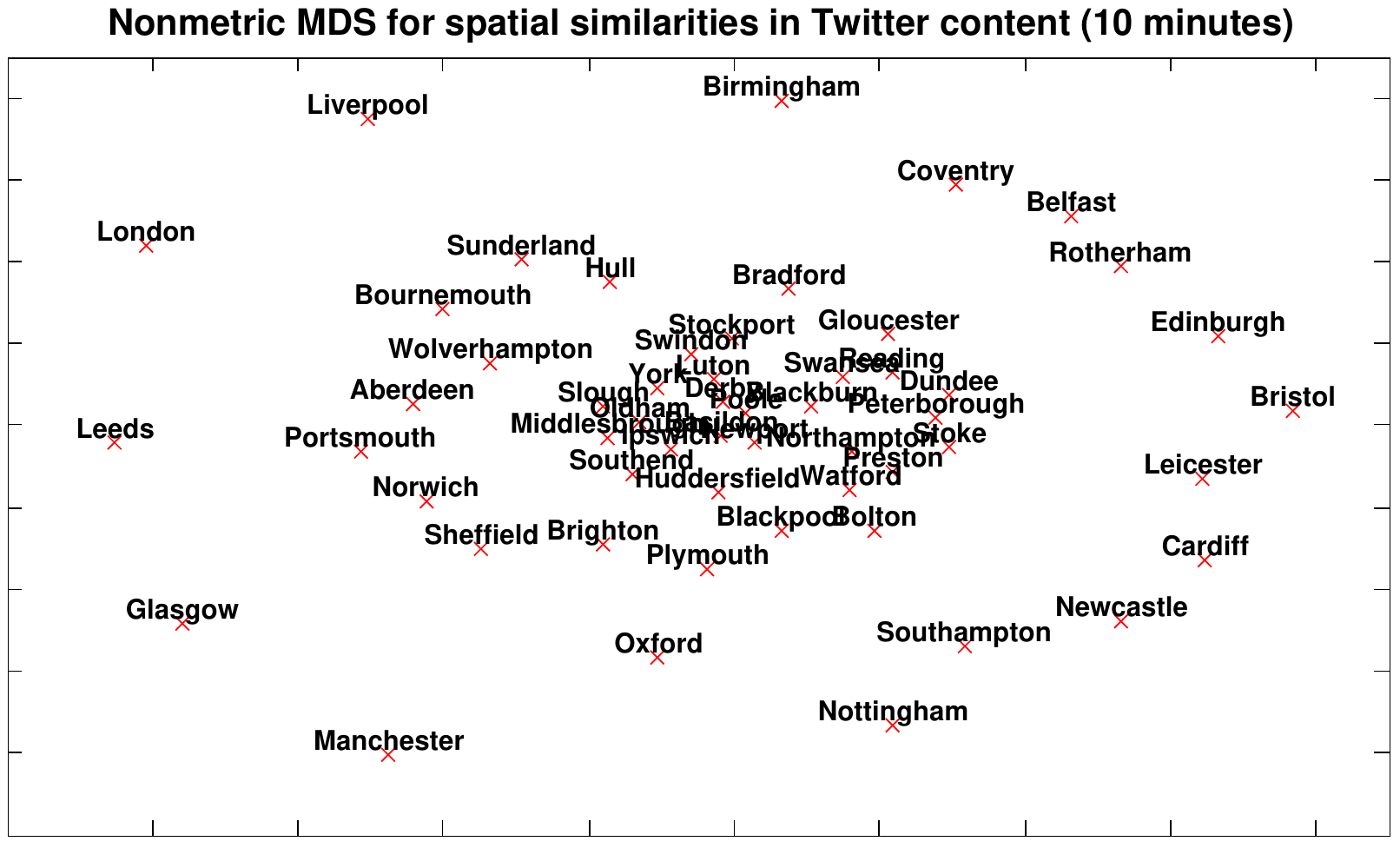}
    \label{fig_sim_MDS_experiment3}}
    \end{center}
    \caption{The considered 54 locations of the UK plotted on a 2-dimensional space after applying nonmetric MDS on their pairwise cosine similarities.}
    \label{fig_sim_MDS}
\end{figure*}

To strengthen our argument and visualise this result better, we apply Multidimensional Scaling\index{Multidimensional Scaling|see{MDS}} \index{MDS} (\textbf{MDS}) on our data sets, aiming to plot the considered locations on a 2-dimensional space based on their pairwise cosine similarities. MDS is a well-established method for visualising patterns in proximities among a set of variables; after performing several trials on our data set, the best solution was retrieved by performing nonmetric scaling, which is a form of MDS, where the computed solution represents ordinal properties of the data \cite{Borg2005}. Figures \ref{fig_sim_MDS_disp_experiment1} and \ref{fig_sim_MDS_disp_experiment3} are the Shepard plots\index{Shepard plot} showing the pairwise similarities as well as their monotonic transformations induced by nonmetric MDS, known as disparities. From those figures, we see that the disparities are a fairly good nonlinear approximation of the pairwise-cosine similarities. Hence, we can plot the configuration of points that we retrieved by applying MDS on a 2-dimensional space expecting that this representation reflects the original similarities at a satisfactory level (see Figures \ref{fig_sim_MDS_disp_experiment1} and \ref{fig_sim_MDS_disp_experiment3}). For both time spans (1 day and 10 minutes), we can see that spatial proximities do not seem to influence the clusters of locations; there exist some nearby locations which are clustered together but at the same time very distant locations might be members of the same cluster as well. An additional observation visualised in both figures is that the major UK cities (such as London, Manchester, and Liverpool) are situated on the outer space of the 2-dimensional map, whereas locations with smaller population figures have been gathered at centre of this map. In the next section, we go a step further, trying to understand how exactly content is shared among those locations.

%%%%%%%%%%%%
\subsection{Forming networks of content similarity}
Based on the previous experimental results, which revealed that proximity does not have a significant influence on content similarity within a country, we are taking a step further concentrating our efforts on explaining spatiotemporal\index{spatiotemporal} content relations. In particular, we want to examine how Twitter content is shared among different urban centres by proposing a preliminary method for forming networks\index{network}, which try to capture and describe those relationships.

For a pre-specified set of time intervals $\mathcal{T} = \{t_j\}$, $j\in\{$1, ..., m$\}$, we compute the average cosine similarity of all location pairs $\{u_i,u_k\}\in\mathcal{L}_p$ in the same manner as it has already been described in the previous section. $\mathcal{T}$'s length is an arbitrary choice that defines the overall time period encapsulated in the network.

Every location $u_k$ retrieves an Impact Score (\textbf{IS}), a quantity expressing the degree of shared content of $u_k$ during $\mathcal{T}$. IS($u_k$,$\mathcal{T}$) is defined as the average cosine similarity over all the location pairs in which $u_k$ participates and is computed using:
\begin{equation}
\text{IS}(u_k) = \frac{1}{m}\sum_{\substack{
            i=1\text{, }i \neq k,\\
            \{u_i,u_k\} \in \mathcal{L}_p}}^{|\mathcal{U}|} \text{Average-Similarity}(k,i),
\end{equation}
where $m \leq |\mathcal{U}|$ is the number of times $u_k$ occurs in a location pair of $\mathcal{L}_p$.

The average pairwise cosine similarities Average-Similarity$(i,k)$ for all location pairs $\{u_i,u_k\}$ $\in$ $\mathcal{L}_p$ are ranked in a decreasing order and the $\alpha$ top ones are being selected. Those $\alpha$ location pairs will form the edges of the inferred spatiotemporal network. The directionality of each edge is decided by the impact scores of the participants: if $\text{IS}(u_i) > \text{IS}(u_j)$, then $u_i \longrightarrow u_j$, else $u_i \longleftarrow u_j$. The numerical value of $\alpha$ is an arbitrary choice; in our experiments $\alpha =$ 100 and therefore we expect to infer networks comprised by 100 edges.

\begin{figure}[!t]
\centering
\includegraphics[width=5in]{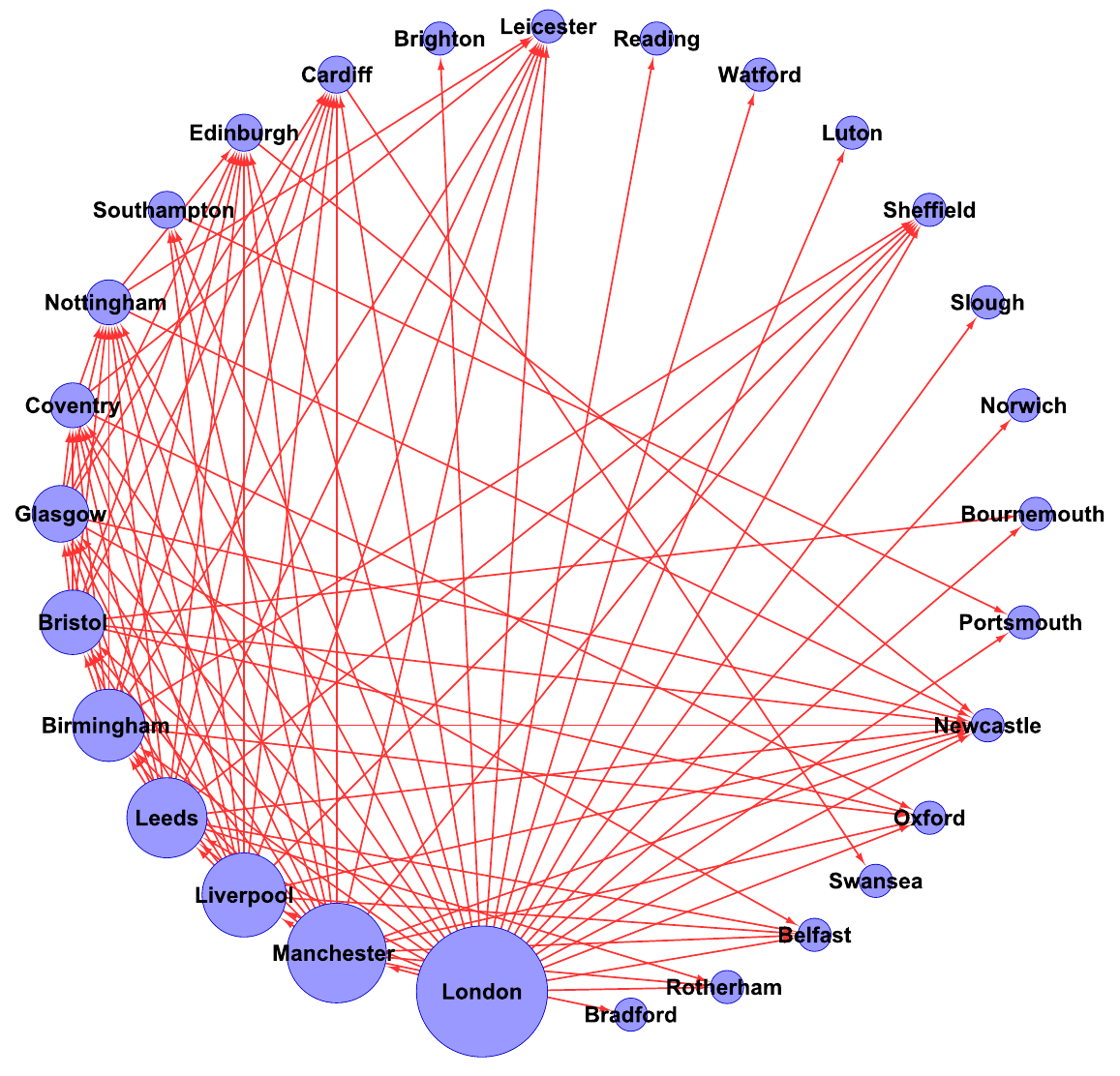}
\caption{Content similarity network for $\mathcal{T}$ equal to 6 months (entire data set) and $t_j$ equal to 1 day. The size of each node is proportional to its out-degree.}
\label{fig_spatial_relationships_experiment1_100pairs}.
\end{figure}

\begin{figure}[!t]
\centering
\includegraphics[width=5in]{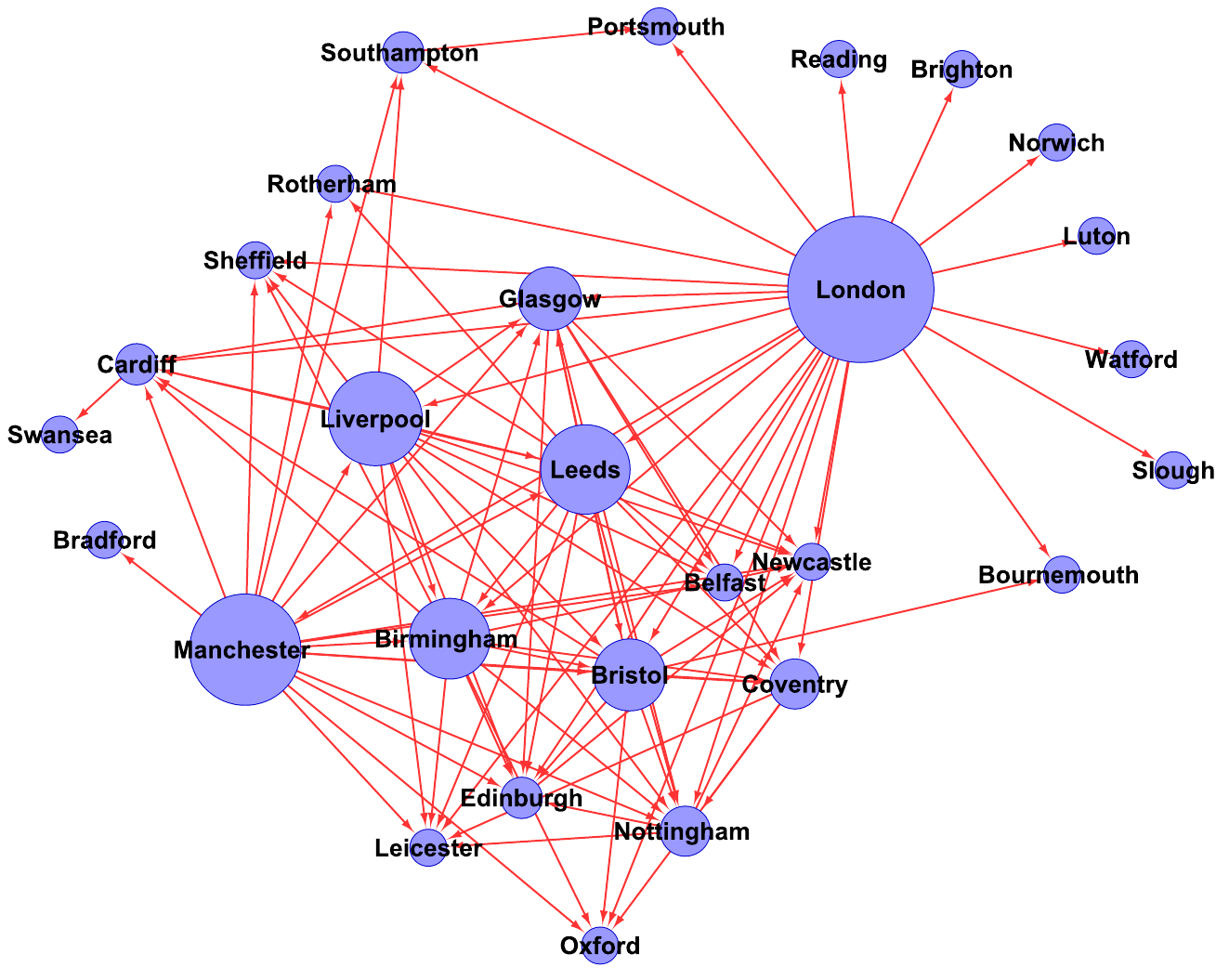}
\caption{The graph of Figure \ref{fig_spatial_relationships_experiment1_100pairs} in a spring-embedded layout.}
\label{fig_spatial_relationships_experiment1_100pairs_alt_view}.
\end{figure}

First, we set $\mathcal{T}$ equal to the entire 6-month time span of the data set and $t_j$ equal to one day. This will give us an average picture of the network over a 6-month period based on one-day long content similarities between the considered locations. The result is depicted in Figures \ref{fig_spatial_relationships_experiment1_100pairs} and \ref{fig_spatial_relationships_experiment1_100pairs_alt_view}.

The network in Figure \ref{fig_spatial_relationships_experiment1_100pairs} starts from the node with the highest out-degree (London) and then -- in a clockwise manner -- lists the other nodes in decreasing out-degree order. London is shown to be the most central location in the topic space, followed by Manchester and Liverpool. It is apparent that the locations with the highest out-degrees are the ones with the highest populations; in general, this was something expected to happen on average during a day. Figure \ref{fig_spatial_relationships_experiment1_100pairs_alt_view} depicts the same network by using a spring-embedded layout that makes the edges of the network more visible; we can directly observe that London shares content with nearby cities such as Luton and Reading as well as with places that are far away, such as Liverpool and Manchester. Likewise, there is a connection between Cardiff and Swansea, but also between Cardiff and Glasgow.

\begin{figure}[!t]
\centering
\includegraphics[width=5in]{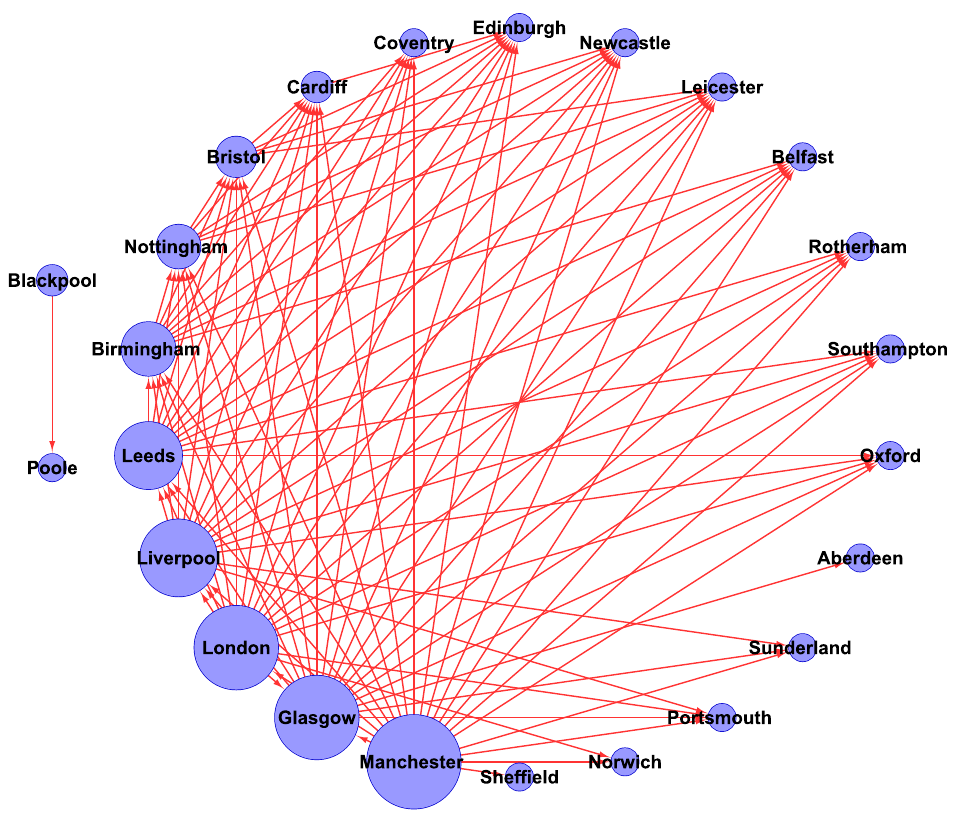}
\caption{Content similarity network for $\mathcal{T}$ equal to 1 month (December, 2010) and $t_j$ equal to 10 minutes. The size of each node is proportional to its out-degree.}
\label{fig_spatial_relationships_experiment3_100pairs}
\end{figure}

In the second experiment, we reduce $\mathcal{T}$ to one month (December, 2010) and $t_j$ is set to 10 minutes. Similarly to the previous experiment, Figure \ref{fig_spatial_relationships_experiment3_100pairs} shows an average picture of the inferred network over December based on 10-minute long content similarities between the considered locations. This network is not directly comparable with the previous one, since it encapsulates data from December 2010 only and not the entire 6-month period; still, we can spot major differences, such as the fact that Manchester is now the most central location in the topic space, followed by Glasgow and London.

\begin{figure*}
    \begin{center}
    \subfigure[Content similarity network for $\mathcal{T}$ equal to 1 month (December, 2010) and $t_j$ equal to 1 day] {\includegraphics[width=4.2in]{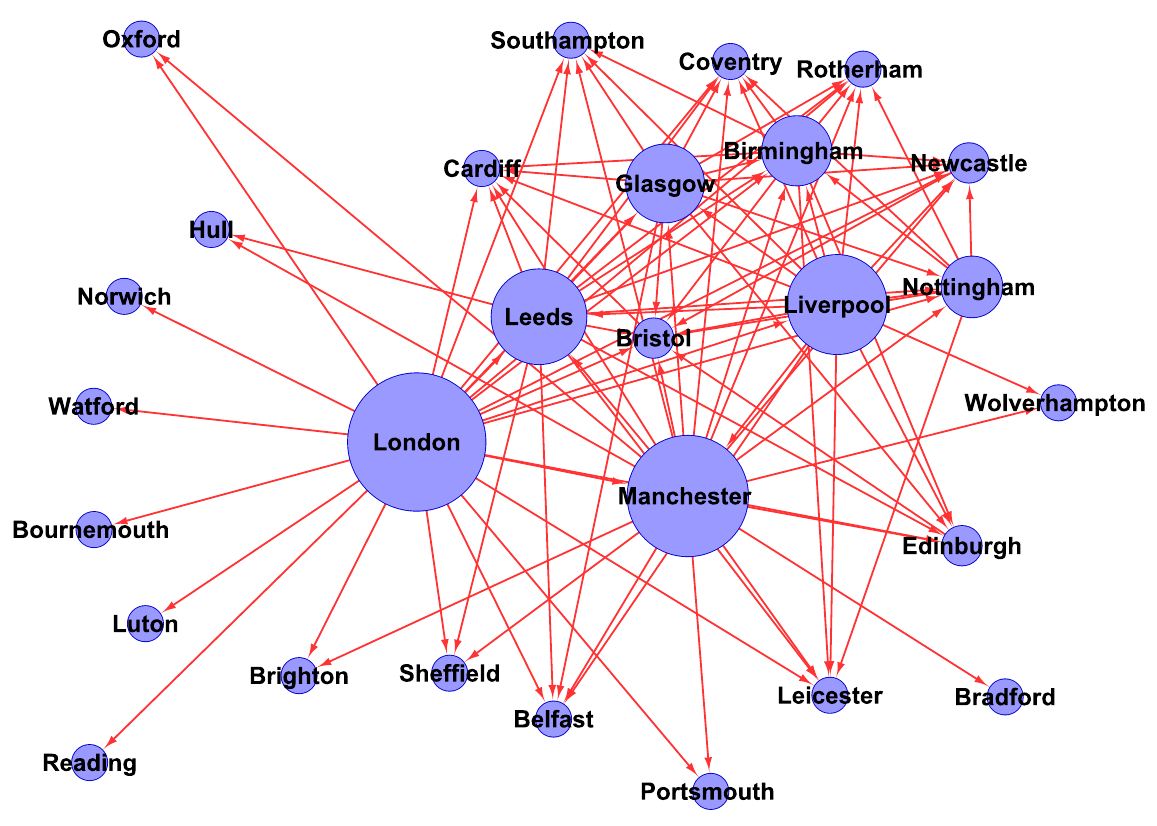}
    \label{fig_spatial_relationships_experiment1_100pairs_December}}
    \hfil
    \subfigure[Content similarity network for $\mathcal{T}$ equal to 1 month (December, 2010) and $t_j$ equal to 10 minutes] {\includegraphics[width=4.2in]{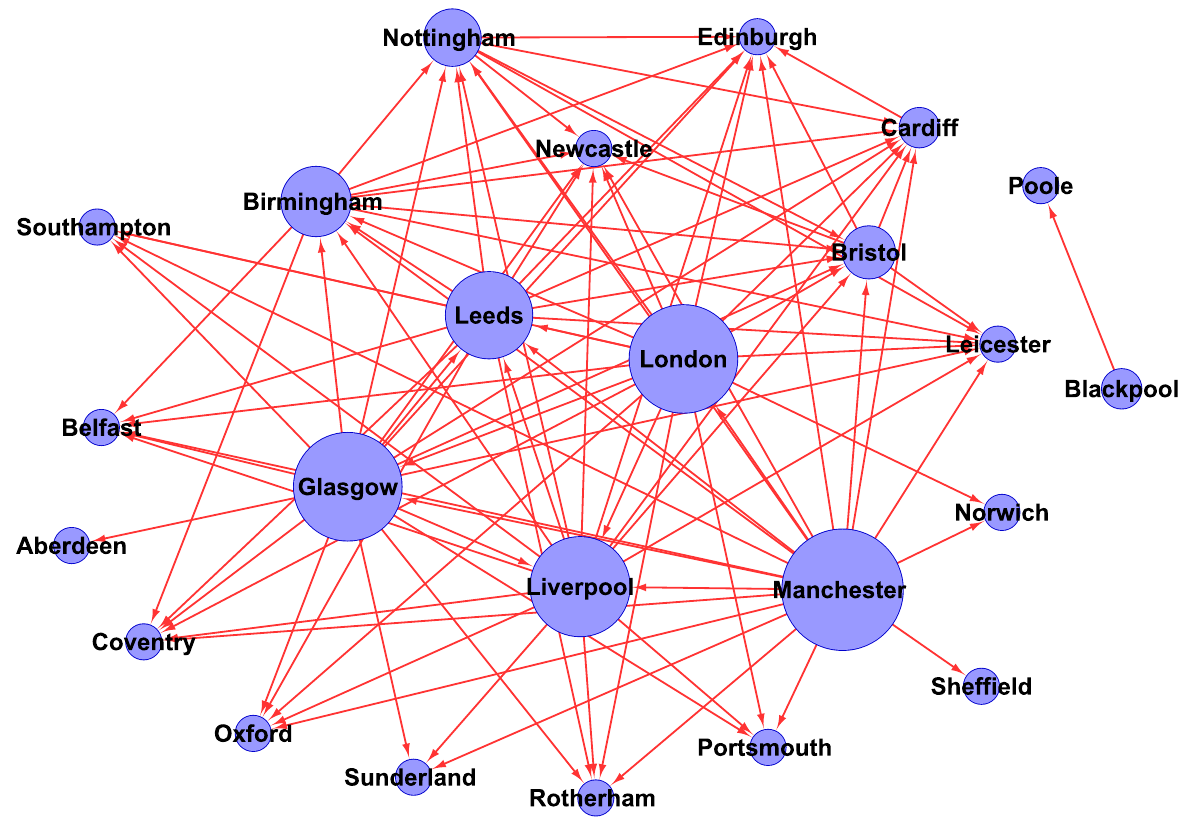}
    \label{fig_spatial_relationships_experiment3_100pairs_alt_view}}
    \end{center}
    \caption{Comparing networks encapsulating the same month but using different content similarity time spans.}
    \label{fig_spatial_rel_net_comparison}
\end{figure*}

Figures \ref{fig_spatial_relationships_experiment1_100pairs_December} and \ref{fig_spatial_relationships_experiment3_100pairs_alt_view} show two comparable networks. Figure \ref{fig_spatial_relationships_experiment1_100pairs_December} is a reduced version of the network in the first experiment, where we have set $\mathcal{T}$ to one month (December, 2010) and $t_j$ to one day, whereas Figure \ref{fig_spatial_relationships_experiment3_100pairs_alt_view} is identical to the network inferred in the second experiment, but this time displayed using a spring-embedded layout.

The inferred networks under different values of $t_j$ are quite dissimilar. By examining them more closely, we can observe that the one based on short-term (\ie 10-minute) content similarities between locations not only has a different major `player' (Manchester), but also distributes the out-degrees in a more balanced manner among locations; London's nearby locations are not included in this network and we also see that Glasgow has a significantly increased level of shared content. Additionally, it also reveals a `strange' connection between Poole and Blackpool -- a connection that does not appear in the former network, where $t_j$ was set to 1 day. By applying a network similarity scoring function (introduced in the next section and ranging from 0 to 1 -- see Equation \ref{equation_similarity_score}) on those two networks, their similarity score is equal to 0.5152; this is a numerical proof on the observable dissimilarity of the two inferred networks.

%\begin{figure}[!t]
%\centering
%\includegraphics[width=5in]{experiment1_top100_December}
%\caption{Content similarity and influence network for $\mathcal{T}$ equal to 1 month (December, 2010) and $t_j$ equal to 1 day. The size of %each node is proportional to its out-degree.}
%\label{fig_spatial_relationships_experiment1_100pairs_December}
%\end{figure}

%%%%%
\subsection{Network stability}
\label{section_net_stability}
In the previous section, we presented a method for forming content similarity networks among locations over two different time intervals. In this section, we show that the inferred networks are stable\index{network stability} over time and therefore, good and well defined descriptions of how different locations share content over time on Twitter. A similar approach to the one introduced in \cite{Flaounas2010} is followed; to prove network stability over time we first form a time series of networks, then apply a similarity metric to compare consecutive networks in this time series, and finally -- using a null hypothesis -- show that those similarities are statistically significant.

Given that the maximum number of nodes in our networks is known (and is equal to the total number of urban centres considered $|\mathcal{U}|$), we are using a simple similarity measure that compares the directed edges of the networks, known as the Jaccard Distance (\textbf{JD})\index{Jaccard Distance}. For two sets of edges $\mathcal{E}_a$ and $\mathcal{E}_b$, JD is defined as:
\begin{equation}
\text{JD}(\mathcal{E}_a,\mathcal{E}_b) = \frac{|\mathcal{E}_a \bigcup \mathcal{E}_b| - |\mathcal{E}_a \bigcap \mathcal{E}_b|}{|\mathcal{E}_a \bigcup \mathcal{E}_b|}.
\end{equation}
JD ranges in [0,1] and is a numerical indication of the dissimilarity between $\mathcal{E}_a$ and $\mathcal{E}_b$. By reversing this notion, we derive the Similarity Score (\textbf{SS})\index{Similarity Score} that is used in our experiments:
\begin{equation}
\label{equation_similarity_score}
\text{SS}(\mathcal{E}_a,\mathcal{E}_b) = 1 - \text{JD}(\mathcal{E}_a,\mathcal{E}_b) = \frac{\mathcal{E}_a \bigcap \mathcal{E}_b}{\mathcal{E}_a \bigcup \mathcal{E}_b}.
\end{equation}
SS ranges also in [0,1], but its value is proportional to the similarity of the networks under comparison.

Following similar approaches presented in \cite{Milo2002, Flaounas2010}, we base our null hypothesis model on a randomisation strategy able to generate networks with the same degree distribution as the original ones, but randomised topology. This randomisation strategy starts from a given graph and randomly switches edges $n$ times. For example, if edges $u_v \rightarrow u_x$ and $u_y \rightarrow u_z$ are present in the original network, they will be replaced by $u_v \rightarrow u_z$ and $u_x \rightarrow u_y$. In our experiments, when the similarity between two consecutive in time versions of the inferred network -- say $\mathcal{E}_{t_1}$ and $\mathcal{E}_{t_2}$ -- is tested for statistical significance, randomised switching is performed on $\mathcal{E}_{t_1}$. After $n =$ 1,000 random swaps, the randomised network $\mathcal{E}_{t_1}^r$ is compared to the original $\mathcal{E}_{t_1}$. By counting the times SS$(\mathcal{E}_{t_1},\mathcal{E}_{t_1}^r)$ is higher than or equal to SS$(\mathcal{E}_{t_1},\mathcal{E}_{t_2})$, we compute the p-value of the null hypothesis (number of successes divided by the total number of random swaps). P-values lower than or equal to 0.05 justify statistical significance and hence, network stability.

\begin{figure*}
    \begin{center}
    \subfigure[$\mathcal{T}$'s length is equal to a month and time intervals $t_j$ are one hour long] {\includegraphics[width=3in]{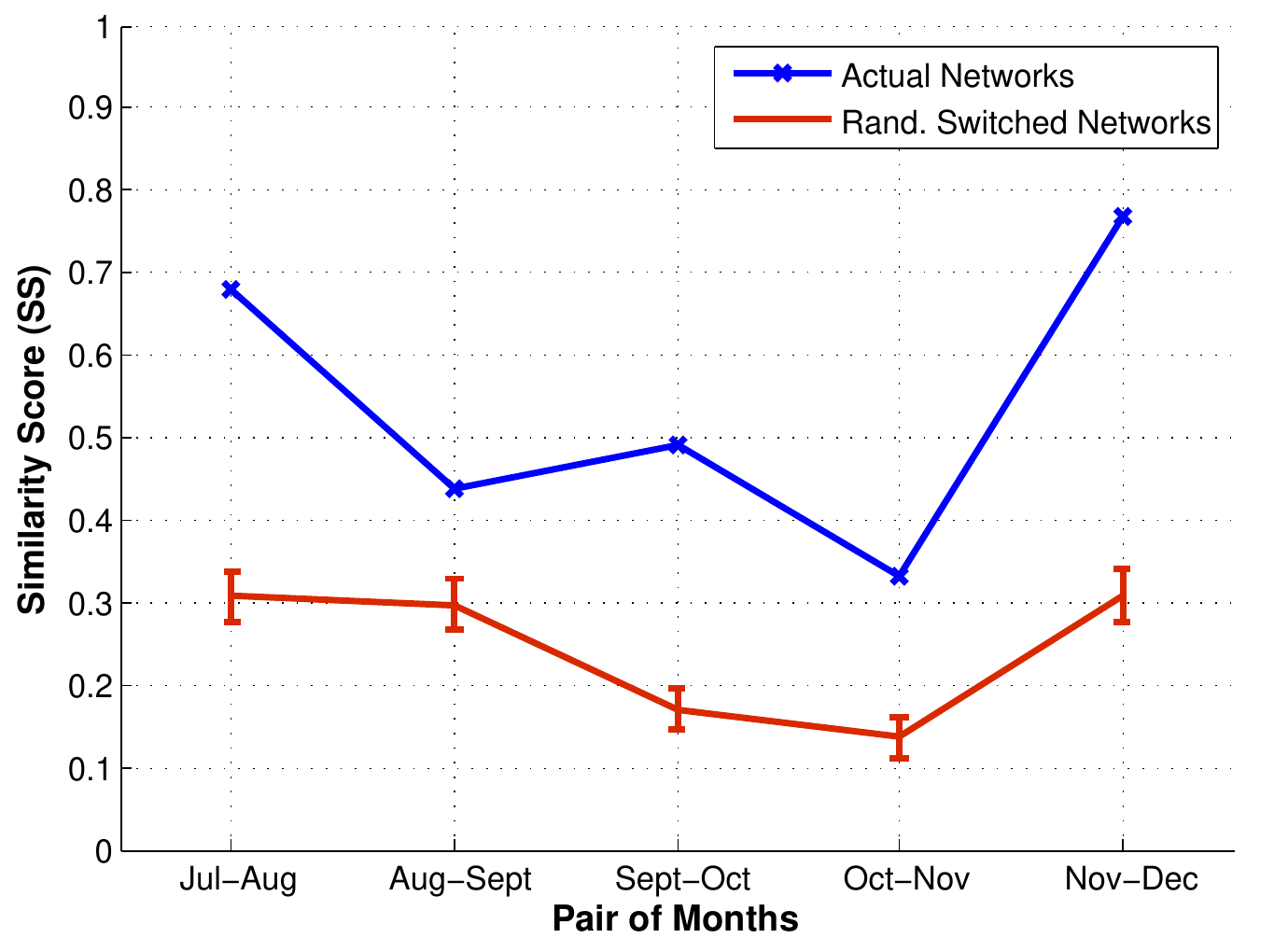}
    \label{fig_stability_experiment1}}
    \hfil
    \subfigure[$\mathcal{T}$'s length is equal to one day and time intervals $t_j$ are 10 minutes long]{\includegraphics[width=3in]{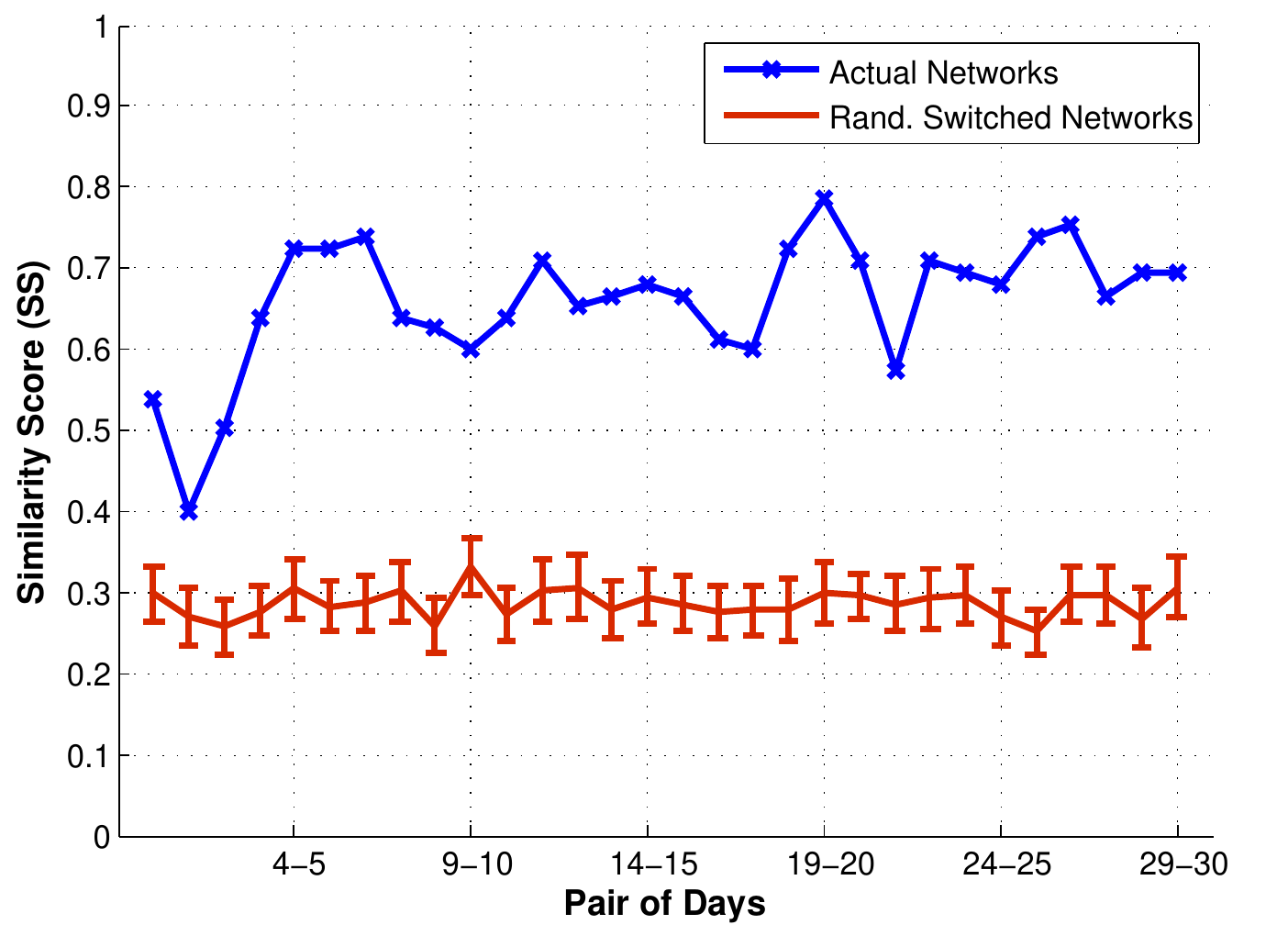}
    \label{fig_stability_experiment3}}
    \end{center}
    \caption{Stability of content similarity networks derived by measuring the network pairwise similarity score over consecutive months (a) and days (b).}
    \label{fig_net_stability}
\end{figure*}

We study the stability of the network over two cases. In the first one, we set $\mathcal{T}$ equal to one month and $t_j$ to one day. In this sense, we are examining whether networks based on one day long content similarities are stable on a monthly basis. In the second, by setting $\mathcal{T}$ to one month (December, 2010) and $t_j$ to 10 minutes, we are investigating whether networks based on 10-minute long content similarities are stable on a day-to-day basis. The results have been plotted in Figures \ref{fig_stability_experiment1} and \ref{fig_stability_experiment3} respectively. From those figures, it becomes apparent that the networks are stable on a monthly basis in the first case and on a daily basis in the second, as the similarity score between their consecutive instances is always higher than the one derived by the randomised switching technique (with p-values always smaller than 0.001). As it might have been expected by common logic, the network fluctuates more on a monthly basis, and thus the similarity scores for the first case are lower than the ones observed in the second.

\begin{figure}
\centering
\includegraphics[width=3.5in]{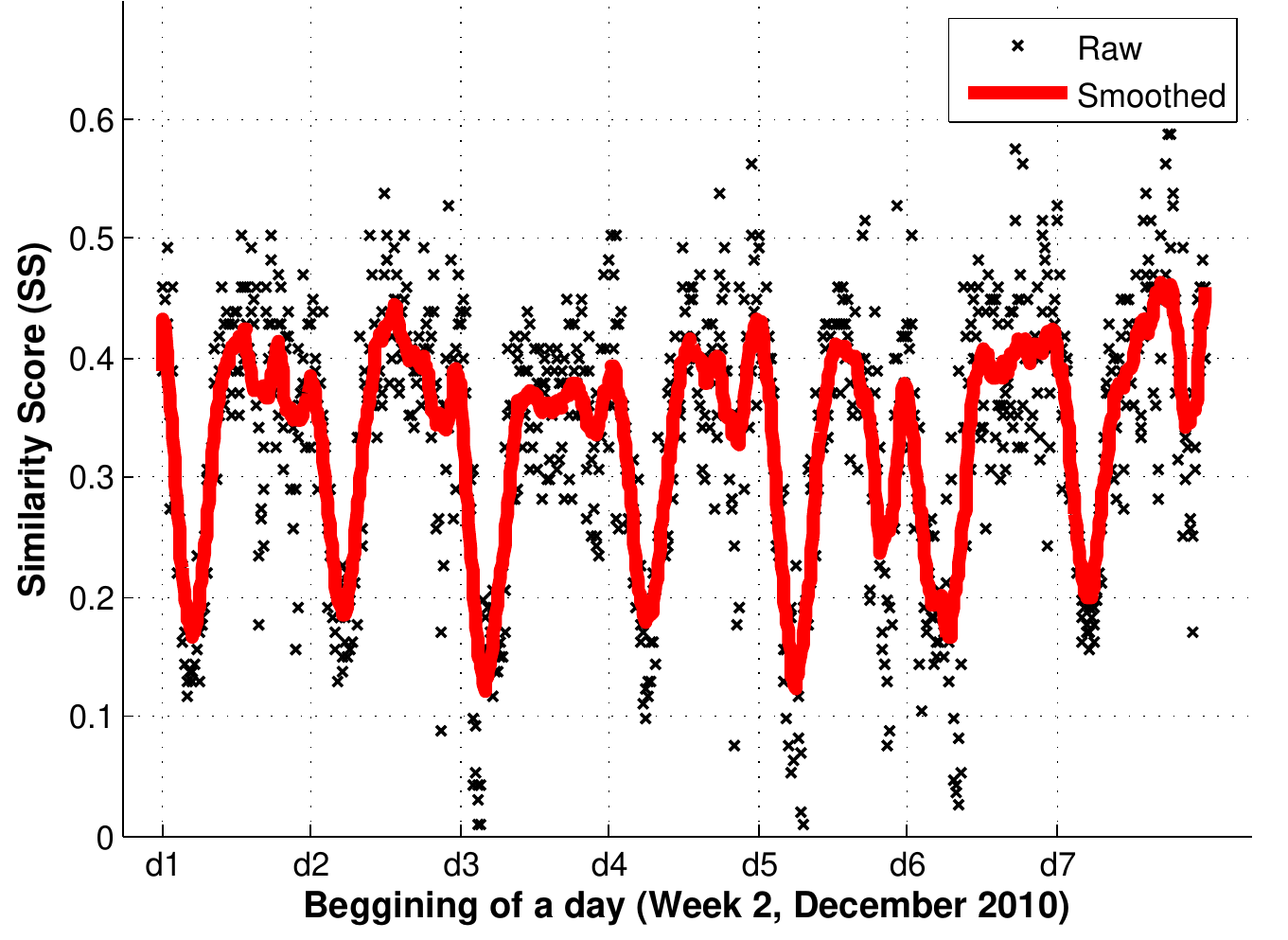}
\caption{Similarity Score time series for consecutive networks with $\mathcal{T}$ and $t_j$ equal to 10 minutes. The red line is a smoothed version of the original plot that uses a 24-point moving average.}
\label{fig_spatial_relationships_experiment3_100pairs_7days_10minInt}
\end{figure}

At last, using the data of the second experiment, we investigate how the inferred networks evolve on a 10 minute basis. Following the notation of this section, this is achieved by setting both $\mathcal{T}$ and $t_j$ to 10 minutes. Figure \ref{fig_spatial_relationships_experiment3_100pairs_7days_10minInt} shows the derived result. It is a time series of similarity scores drawn on 7-day period (second week of December 2010). The mean similarity score is equal to 0.334 with a standard deviation of 0.11; it is lower -- as expected -- compared to ones derived by comparing networks of larger time spans. After smoothing with a 24-moving point average, so to extract a 4 hour trend (24 points $\times$ 10 minutes per point), an interesting pattern is derived. Similarity scores reach a minimum every day after midnight and during the long night hours and then rise back again, indicating a possible periodic bias. Of course, this could be due to the reduced use of Twitter during the night by individual users but most importantly by professional news and media agencies. This latter observation served as a motivation for the next section, where we study posting time patterns of users on Twitter.

%%%%%%%%%%%%%%%%%%%%%%%%%%%%%%%%%%%%%%%%%%%%%%%%%%%%%%%%%%%%%%%%%%%%%%%%%%%%%%%%%%%
%%%%%%%%%%%%%%%%%%%%%%%%%%%%%%%%%%%%%%%%%%%%%%%%%%%%%%%% posting patterns %%%%%%%%%
\section{Analysing posting time patterns on Social Media}
\label{section:analysing_posting_patterns}
\begin{figure}[!t]
\centering
\includegraphics[width=5in]{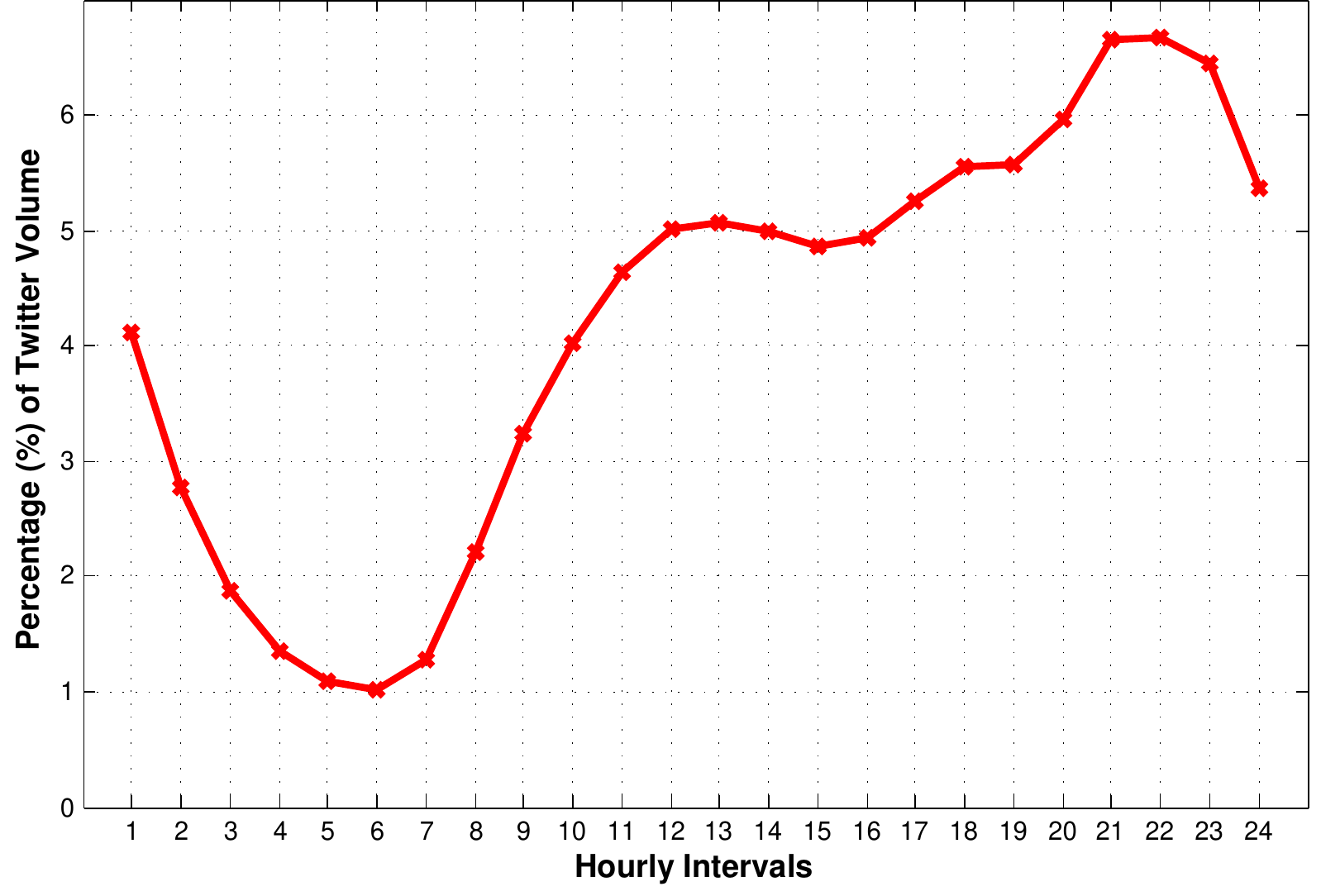}
\caption{Percentage of daily tweets posted per hourly interval based on Twitter content posted between July and December 2010 and geolocated in the UK}
\label{fig_posting_patterns_1}
\end{figure}

In this section, we are investigating posting time patterns\index{posting patterns} on Twitter. For this purpose, we use approx. 71 million tweets posted from July to December, 2010 and geolocated in the UK. Initially, we show the relative volume (percentage) of tweets per hour by aggregating all days in the data set (see Figure \ref{fig_posting_patterns_1}). The three most distinctive time intervals in terms of posting volume are 12--1p.m. (lunch time), 6--7p.m. (end of working hours for the majority of occupations) and 9--10p.m., which is the hour with the highest number of tweets. We also see that Twitter volume goes significantly down during the late night hours and that Twitter is becoming `alive' again at approx. 7a.m. Therefore, this simple statistical analysis confirms common sense and indicates that Twitter activity might be highly correlated with people's real life activities.

\begin{figure}[!t]
\centering
\includegraphics[width=5in]{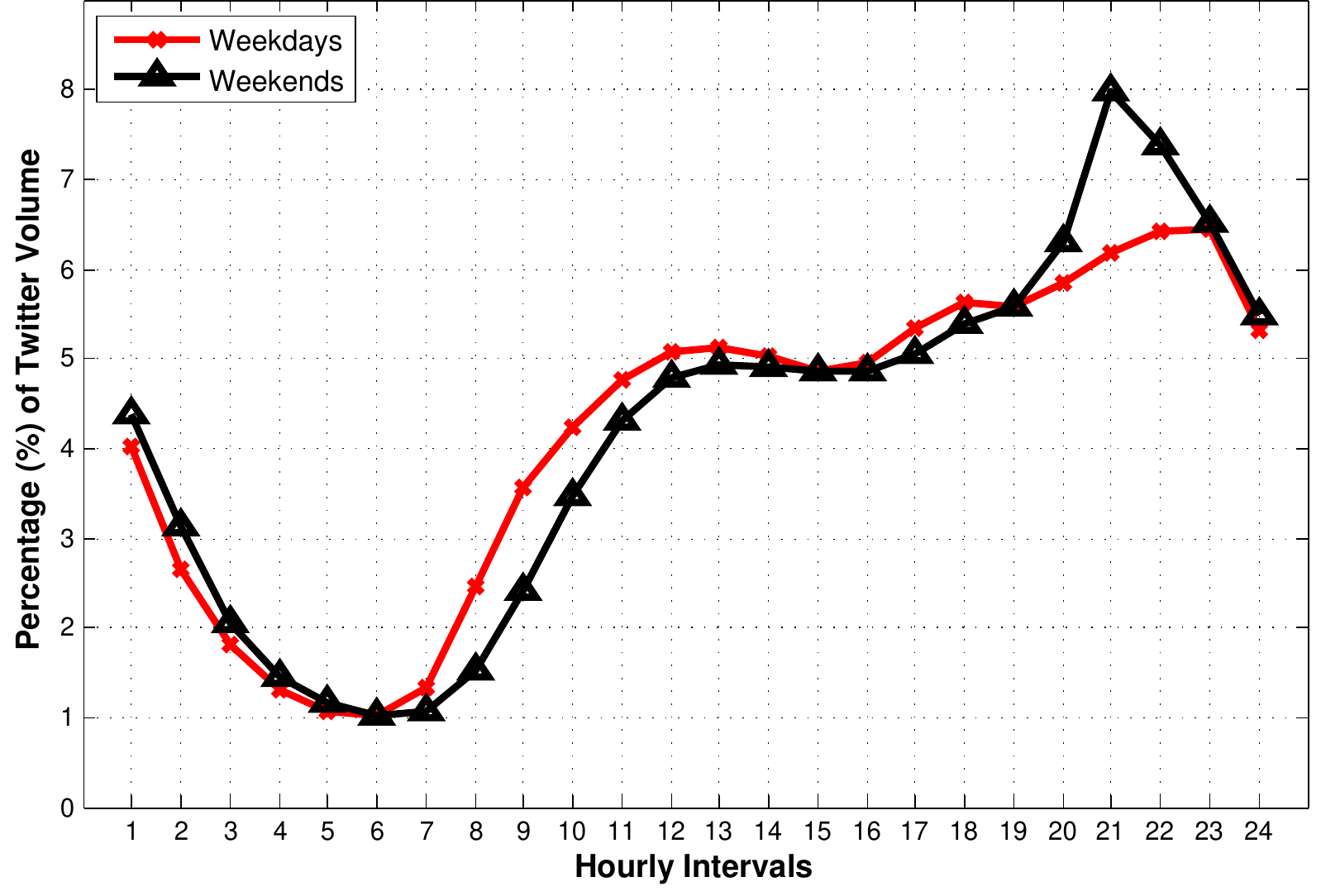}
\caption{Percentage of tweets posted per hourly interval -- Week days vs. Weekends}
\label{fig_posting_patterns_3}
\end{figure}

We test the stability of this pattern ($\bar{V}$) by performing the following basic statistical significance test. For every day $d_i$ in our data set we compute the percentages of Twitter volume for each hourly interval ($V_{d_i}$) and obtain their linear correlation coefficient $\rho(\bar{V},V_{d_i})$ with the general volume pattern. We also compute the linear correlation between $\bar{V}$ and 1,000 randomly permuted versions of $V_{d_i}$ and count how many times -- say $N$ -- the correlation of a pseudo-randomised pair was higher or equal to $\rho(\bar{V},V_{d_i})$. We perform this operation for all 184 days in our data set; hence, the p-value is equal to $N/($1,000$\times$184$)$. A p-value $\leq$ 0.05 justifies statistical significance and shows that the extracted pattern is stable. The p-value for the posting times pattern presented in the previous paragraph is equal to 0.004 and therefore, we can assume that it is a stable representation of the hourly volume percentages.

% The result depicted in Figure \ref{fig_posting_patterns_1} was computed by averaging over the percentages of hourly volume for the six months considered. To increase our certainty that this result is a stable pattern, we are also performing a basic statistical significance test. We select one month $M$ that is present in our data set and first compute the average percentage of hourly Twitter volume without considering this month; then, for each day in October (d1 to d31), we compute its percentages for each hourly interval and compare it with the average percentages using the MSE loss function. For each day, we also form a randomly permuted version of the hourly percentages and compare it to the average percentages using MSE. We repeat this process for 1,000 times - the fraction of the randomly permuted versions with lower or equal MSE than the non permuted ones is the p-value. Figure \ref{fig_posting_patterns_2} shows the result we retrieving by performing the aforementioned process -- the error on the randomly permuted data is always higher and the pattern is statistically significant with a p-value of 0.

%\begin{figure}
%\centering
%\includegraphics[width=5in]{MainPatternSS_vs_October}
%\caption{MSE of the randomly and non-randomly permuted hourly intervals of October 2010 against the average hourly Twitter volume}
%\label{fig_posting_patterns_2}
%\end{figure}

Moving on into this basic statistical analysis, we divide our data set into two subsets: one containing the Twitter volume for week days and one for weekends. If Twitter follows real-life patterns, then we expect an altered behaviour in the posting trends during weekends. Indeed, on weekends Twitter volume peaks in a slightly different time interval (8--9p.m.); however, the initial pattern is still there and the two signals (for weekdays and weekends) are very similar -- their linear correlation is equal to 0.9594 (p-value: 1.43e-013). More interesting are the deviations occurring during the late night hours showing that Twitter users tend to post more during weekends -- possibly because they sleep later than the other days of the week --, as well as the ones in the early morning hours, where a higher volume of tweeting is observed during weekdays. By repeating the previous statistical test, we show that both patterns are stable; the p-values are equal to 0.0052 and 0.001 for weekdays and weekends respectively.

\begin{figure*}
    \begin{center}
    \subfigure[Shepard plot showing goodness of fit for the MDS solution] {\includegraphics[width=2.5in]{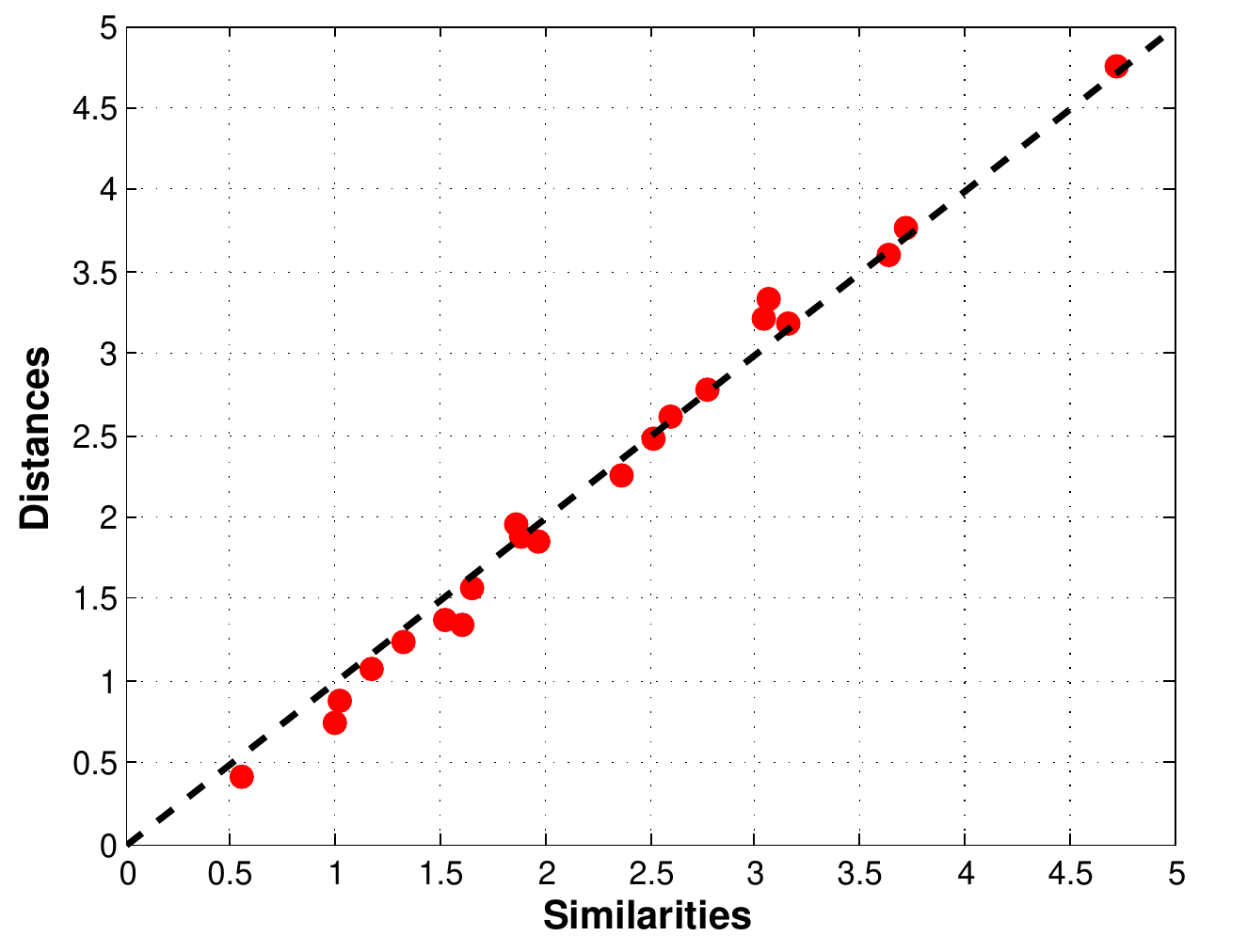}
    \label{fig_twittervol_MDS_Shepard}}
    \hfil
    \subfigure[Clusters of days retrieved by applying MDS]{\includegraphics[width=3in]{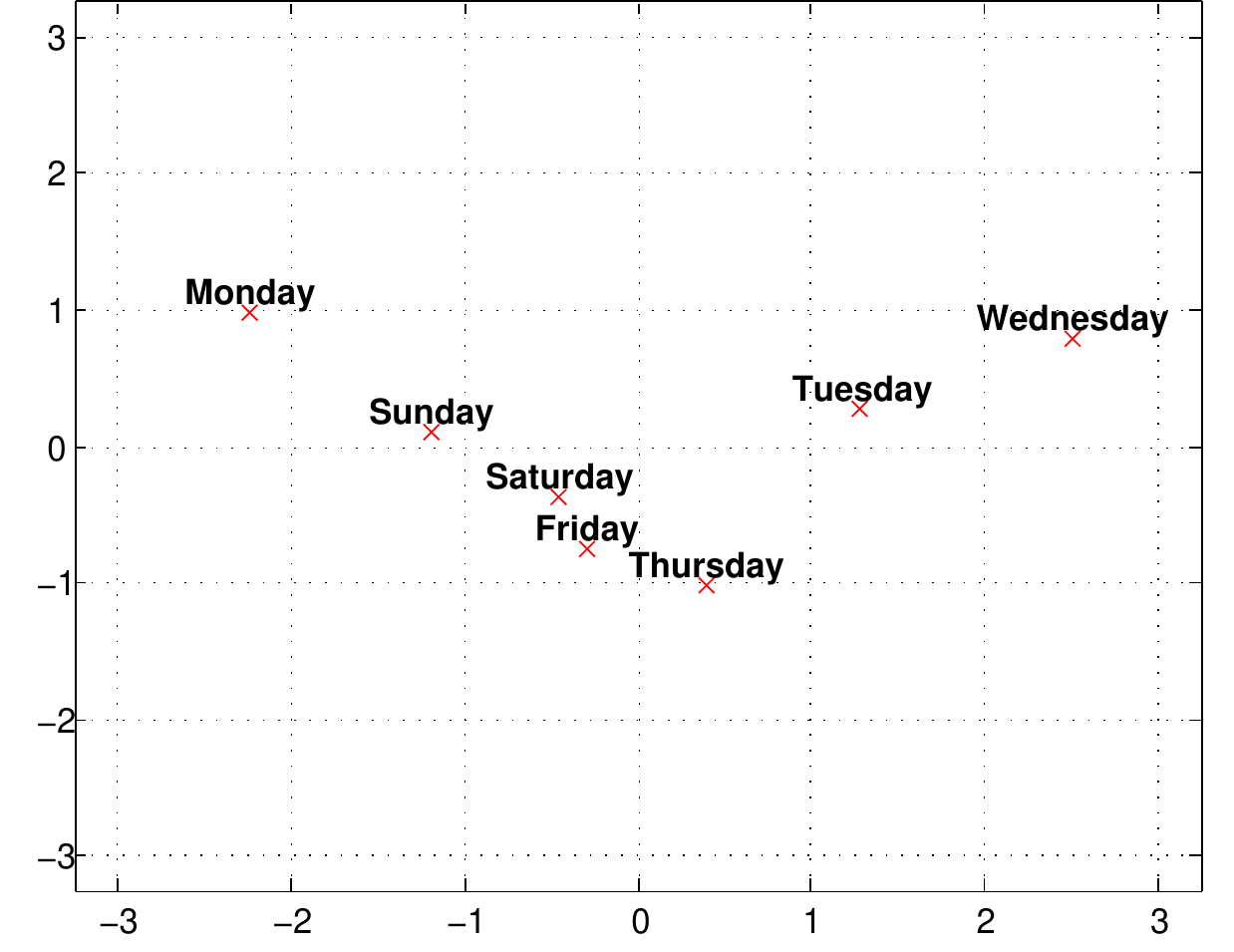}
    \label{fig_twittervol_MDS_days}}
    \end{center}
    \caption{Clustering of days by applying Multidimensional Scaling (MDS) based on their posting time patterns.}
    \label{fig_twittervol_MDS}
\end{figure*}

So far it has been shown that there exist at least two slightly different posting time patterns -- one for weekdays and another one for weekends. To investigate whether more posting patterns are present and visualise the result, we performed a simple clustering method. Each day is represented by a vector of 24 features, \ie the Twitter volume percentage for each hourly interval. Similarly to Section \ref{section_content_correlation_vs_geographical_distance}, we apply MDS\index{MDS}; the Shepard plot\index{Shepard plot} (Figure \ref{fig_twittervol_MDS_Shepard}) indicates that the MDS solution presented in Figure \ref{fig_twittervol_MDS_days} is a good fit. From this figure, we can derive that Monday is the most distinctive day of the week (in terms of its posting volume pattern); we can also observe that consecutive days have been placed in nearby positions of this 2-dimensional space. Interestingly, Friday is more close to Saturday than Thursday and Saturday is more close to Friday than Sunday.

%we performed a simple clustering method to extract clusters with similar behaviours on a daily basis. We first compute the mean normalised percentages for 24 the time intervals of each week day in our data set; then we retrieve their pairwise Euclidean distances and finally create a hierarchical cluster tree (a dendrogram) based on Ward's linkage algorithm \cite{Ward1963}. The result is depicted in Figure \ref{fig_posting_patterns_4}, where we can observe three main clusters formed by the days \{Monday, Tuesday, Wednesday\}, \{Thursday, Friday\} and \{Saturday, Sunday\}. Consecutive days have stronger similarities; the two clusters consisted of week days are closer to each other and they are clearly distinct to the weekend cluster. The dendrogram makes evident that there exists an evolving pattern throughout the week and also pinpoints the `special' behaviour of Mondays.

%\begin{figure}
%\centering
%\includegraphics[width=3.5in]{dendrogram}
%\caption{A dendrogram showing the day clusters of posting patterns on Twitter}
%\label{fig_posting_patterns_4}
%\end{figure}

As a general comment, this simple -- but also probably interesting -- statistical analysis offers some additional proof for the common assumption that Social Media\index{Social Media} might follow patterns emerging in real life; since the web is an important tool for the majority of people, it is perhaps not easy to differentiate it from their real-life trends.

%%%%%%%%%%%%%%%%%%%%%%%%%%%%%%%%%%%%%%%%%%%%%%%%%%%%%%%%%%%%%%%%%%%%%% voting intentions %%%%%
%%%%%%%%%%%%%%%%%%%%%%%%%%%%%%%%%%%%%%%%%%%%%%%%%%%%%%%%%%%%%%%%%%%%%%%%%%%%%%%%%%%%%%%%%%%%%%
\section{A preliminary method for inferring voting intentions from Twitter content}
\label{section:voting_intentions}
In this section, we present a preliminary method for extracting voting intention\index{voting intention} figures from Twitter. The case study used to verify our findings is the 2010 General Election in the United Kingdom (UK).\footnote{United Kingdom General Election 2010, \url{http://en.wikipedia.org/wiki/United_Kingdom_general_election,_2010}.} In the recent past, a few papers have been published on this topic \cite{Lui2010,tumasjan2010predicting,Gayo-Avello2011,Metaxas2011a,OConnor2010}, offering preliminary solutions or discussing the limitations that several approaches might have; we refer to and discuss them at the end of this section.

We consider only the three major parties in the UK; namely the Conservative Party (\textbf{CON}), the Labour Party (\textbf{LAB}) and the Liberal Democrat Party (\textbf{LIBDEM} or \textbf{LIB}). Overall, we are using three techniques for extracting positive and negative sentiment from tweets and then two different methods to map this sentiment to voting intention percentages.

Ground truth is acquired by YouGov's Published Results\footnote{YouGov Archive, \url{http://labs.yougov.co.uk/publicopinion/archive/}.} and consists of 68 voting intention polls dated from January to May 2010. Polls usually refer to a pair of days (in our data set only 3 of the them are 1-day polls) and indicate an expectation for the voting intention percentage per political party. As we move closer to the election day (6th of May), they become more dense; there is a new poll published every day. Tweets are drawn from the same period of time, \ie January to May; their total number is greater than 50 million, but not all of them are used as it will become apparent in the following sections. After applying some filtering to keep tweets regarding politics, we end up with 300,000 tweets, \ie approximately 100,000 tweets per political party.

\subsection{Extracting positive and negative sentiment from tweets}
\label{section:positive_negative_sentiment}
The common characteristic in all three approaches is that at first we retrieve a set of tweets regarding each political party by using a handmade set of keywords (see Appendix \ref{Ap:voting_intention}). The keywords are different per party, not many (approx. 60 per party) and as simple as possible; we use the names of the top politicians in each party, the name of the party and so on. The search is case sensitive when the keyword is an 1-gram and case insensitive when it is an n-gram, with $n >$ 1. In the latter case, we are not looking for an exact match either; we are just searching for tweets that contain all 1-grams of the target n-gram. Character `\#' in front of an 1-gram denotes that we are searching for a Twitter topic. Since those keywords are based mainly on names or entities, one could argue that they could also be created in an automatic manner or extracted from a repository and hence, the human involvement could become insignificant.

The three approaches build on each other; each one is a more elaborate version of its precedent. In the first approach, we are using a stemmed version of SentiWordNet 3.0 to extract positive and negative sentiment\index{sentiment} from tweets without taking into consideration the different parts of speech (verb, noun, adjective and so on). SentiWordNet is the result of automatically annotating all WordNet synsets according to their degrees of positivity, negativity and neutrality \cite{Baccianella2010,Esuli2006a}. Stemming is performed by applying Porter's algorithm \cite{porter1980} and Part-Of-Speech\index{part of speech|see{POS}}\index{POS} (\textbf{POS}) tagging is `skipped' by computing the average positive and negative sentiment weight for each stem over all possible POS that it might appear in. Stems with equal positive and negative sentiment are not considered. The positive and negative sentiment scores of a tweet are retrieved by computing the sum of positive and negative sentiment weights of the words it contains. It is noted that there might exist tweets with no words listed in SentiWordNet; those tweets have zero positive and negative sentiment and therefore are ignored. The acronym \textbf{SnPOS} (Sentiment no POS tagging) is used to denote this approach. The motivation behind the removal of POS tagging is the assumption that Twitter language might not follow the norms of formal scripts and therefore POS taggers -- trained on more `mainstream' types of text -- might create inaccurate results. However, as we will see in practice in the following sections, this is not always the case, probably because tweets that refer to politics have a much better structure than the more casual ones.

The second approach is based on the same basic principles as SnPOS, only this time POS tagging is applied. When one stem is mapped to a particular POS more than once (this can happen due to stemming), the average positive and negative sentiment weights are assigned to it. POS tagging of tweets is carried out by using Stanford POS Tagger, which is a Java implementation of the log-linear POS taggers described in \cite{Toutanova2000,Toutanova2003}. By summing over the sentiment weights of a tweet's terms, we retrieve its sentiment score. This method is denoted as \textbf{SPOS} (Sentiment with POS tagging).

Finally, we extend SPOS, by incorporating the core word senses of WordNet \cite{Miller1995}, a semi-automatically compiled list of 5,000 terms.\footnote{Available at \url{http://wordnet.princeton.edu/wordnet/download/standoff/}.} WordNet gives a set of synonyms for each term in this list and we use those synonyms to extend the content of a tweet. That is, if a tweet contains a word listed in WordNet's core terms, the tweet is extended by attaching the synonyms of this word to it. Again, the sentiment score of a tweet is computed by summing over the sentiment weights of its POS tagged terms. This method is identified by the acronym \textbf{SPOSW} (Sentiment with POS tagging and WordNet's core terms). The main motivation behind extending the content of a tweet with synonyms is the fact that the short length of a tweet might reduce expressiveness and therefore, adding more words could enhance its semantic orientation. Furthermore, SentiWordNet does not include all English words and by attaching synonyms, we achieve to compute pseudo-sentiment for a greater number of tweets.

\subsection{Inferring Voting Intentions}
By applying a method from the ones described in the previous section, we compute the positive and negative sentiment scores for a set of $t$ tweets, which in turn has been originally extracted by using the set of keywords for a political party. The next task is to turn those scores into a percentage which will represent the voting intention for this particular party. In this section, we describe two methods that address this task.

Optionally, one could first remove tweets with almost equal positive and negative sentiments. The semantic orientation of those tweets is unclear and therefore might not always help. Later on, when we present our experimental findings, for every experimental setting we test, we are also replicating the same experiment by introducing a threshold $\delta$, which removes the top 20,000 (this is equal to approx. 6.67\% of the entire number of tweets) most `semantically unclear' tweets. Learning the optimal value of $\delta$ from data is possible, but also dependent to the existence of a larger sample.

For the remaining $m \leq t$ tweets we compute their mean positive and negative sentiment scores, say $\mu_{\text{pos}_m}$ and $\mu_{\text{neg}_m}$ respectively. The sentiment score assigned to this set of tweets is derived by simply subtracting those two quantities
\begin{equation}
\label{eq:senti_mts}
\text{Senti}(m) = \mu_{\text{pos}_{m}} - \mu_{\text{neg}_{m}}.
\end{equation}
%where $c$ is a predefined constant, so that Senti$(m) > 0$. If Senti$(m)$ contains negative values, we set $c$ equal to $c = |\min(\text{Senti})| + \sigma(\text{Senti}(m))$, where $\min(\text{Senti})$ is a global minimum ensuring that the addition of its absolute value to the original sentiment score will result to a number $\geq 0$, and $\sigma(\text{Senti}(m))$ is the standard deviation of the sentiment score in the $m$ tweets.

Suppose that we have computed the sentiment scores for a set of $n$ time instances (a time instance is equal to one or two days based on the target voting intention poll) for all three parties, $\text{Senti}_{\text{CON}}^{(n)}$, $\text{Senti}_{\text{LAB}}^{(n)}$ and $\text{Senti}_{\text{LIB}}^{(n)}$. To calibrate the relation between sentiment scores and voting intention percentages, we regress each one of these vectors with the corresponding ground truth using OLS and compute the three weights $w_{\text{CON}}$, $w_{\text{LAB}}$ and $w_{\text{LIB}}$. For example,
\begin{equation}
w_{\text{CON}} = \argmin_{w}\sum_i^n \left(\text{Poll}_{\text{CON}}^{(i)} - \text{Senti}_{\text{CON}}^{(i)}w\right)^2,
\end{equation}
where $\text{Poll}_{\text{CON}}^{(i)}$ denotes the official voting intention poll percentage for time instance $i$. We do not introduce bias terms in OLS regression, as during our experimental process it became evident that they receive large values, reducing significantly the freedom of our model and causing overfitting.

In this case study, we are only considering the percentages of the three major parties. After the inference process, the results are normalised in order to represent valid percentages (having a sum equal to 1); in the special (and also rare) case, where inferences are lower than 0, negative results are thresholded to 0 and normalisation does not take place, unless the nonzero percentages sum up to a value greater than 1. Official voting intentions, however, take into consideration the existence of other political `forces' as well as people's desire not to vote and so on. To create an equivalent and comparable representation, we have normalised the official voting intention percentages as well -- based on the fact that the sum of voting intention percentages for the three major parties is on average equal to approx. 90\%, this normalisation does not change the picture much (and we can always normalise the results back to a 90\% level, if required).

Voting intention inferences are made on unseen data. Again, we have a set of $k$ time instances, which do not overlap with the $n$ time instances used for learning the calibration weights, and we first compute the sentiment scores for each party per time instance. Then, those sentiment scores are multiplied by the corresponding calibration weight; for example the inferred score for the Conservative Party will be equal to:
\begin{equation}
\text{CON}^{(k)} = \text{Senti}_{\text{CON}}^{(k)} \times w_{\text{CON}}.
\end{equation}
After computing the triplet of inferences, $\text{CON}^{(k)}$, $\text{LAB}^{(k)}$ and $\text{LIB}^{(k)}$, we normalise them so that their sum is equal to 1, for example:
\begin{equation}
\text{CON}^{(k)}_{\text{norm}} = \frac{\text{CON}^{(k)}}{\text{CON}^{(k)} + \text{LAB}^{(k)} + \text{LIB}^{(k)}},
\end{equation}
represents the $k$ normalised inferences of the voting intention for the Conservative Party. We denote the method described above as Mean Thresholded Sentiment (\textbf{MTS}).

The second method is identical to MTS, but uses a different function to compute Senti$(m)$ (Equation \ref{eq:senti_mts}). After retrieving the $m$ tweets that satisfy a threshold $\delta$, we count how many of them have a positive sentiment score that is higher than the negative and vice versa. The overall sentiment score of those tweets is then computed by
\begin{equation}
\text{Senti}(m) = \frac{\#\{\text{pos} > \text{neg}\} - \#\{\text{neg} > \text{pos}\}}{m},
\end{equation}
where $\#\{\text{pos} > \text{neg}\}$ is the number of tweets with a positive sentiment greater than the negative one and $\#\{\text{neg} > \text{pos}\}$ the number of tweets with a negative sentiment greater than the positive. We refer to this method as Dominant Sentiment Class (\textbf{DSC}). %Similarly to MTS, if Senti$(m)$ contains negative values, $c$ is set equal to $c = |\min(\text{Senti})| + \sigma(\text{Senti}(m))$, otherwise $c$ is 0.

\subsection{Experimental process and results}
We measure the performance of our methods using two loss functions. Primarily, we use the Mean Absolute Error (\textbf{MAE})\index{Mean Absolute Error|see{MAE}}\index{MAE} (and its standard deviation) between the inferred and the target voting intention percentages. This metric allows for an easier interpretation because it can be read as a percentage, \ie MAE has the same units as the inferred values. In addition, aiming to assess how good is the ranking of the political parties based on the inferred voting intention percentages, we are measuring a Mean Ranking Error\index{Mean Ranking Error|see{MRE}}\index{MRE} (\textbf{MRE}). For each triplet of voting intention percentages, the ranking error is defined as the sum over all three parties of the distance between the correct and the inferred ranking. For example, if there exists one incorrect ranking of size 1 (where size measures the difference between a correct and an inferred position), then -- since we are dealing with 3 variables only -- either there exists one more incorrect ranking of size 1 and therefore the triplet's total error is equal to 2, or two more ranking errors of size 1 and 2, which will make the triplet's total error equal to 4. Obviously, a triplet with correct rankings has an error equal to 0 and the maximum error ($\max(\text{RE})$) per triplet is equal to 4. MRE ranges in [0,1] and is computed by:
\begin{equation}
\text{MRE}(n) = \frac{1}{n\times\max(\text{RE})} \sum_{i = 1}^{n} \text{RE}(i).
\end{equation}

In total, by combining the three methods for extracting positive and negative sentiment from tweets (SnPOS, SPOS and SPOSW) and the two methods for converting sentiment scores to voting intentions (MTS and DSC), we come up with six experimental setups. We run those experiments for $\delta = 0$, \ie for all tweets with a sentiment score, but also for the value of $\delta$ which removes the top 20,000 most `semantically unclear' tweets.

\begin{figure*}
    \begin{center}
    \subfigure[First 34 time instances]{\includegraphics[width=6in]{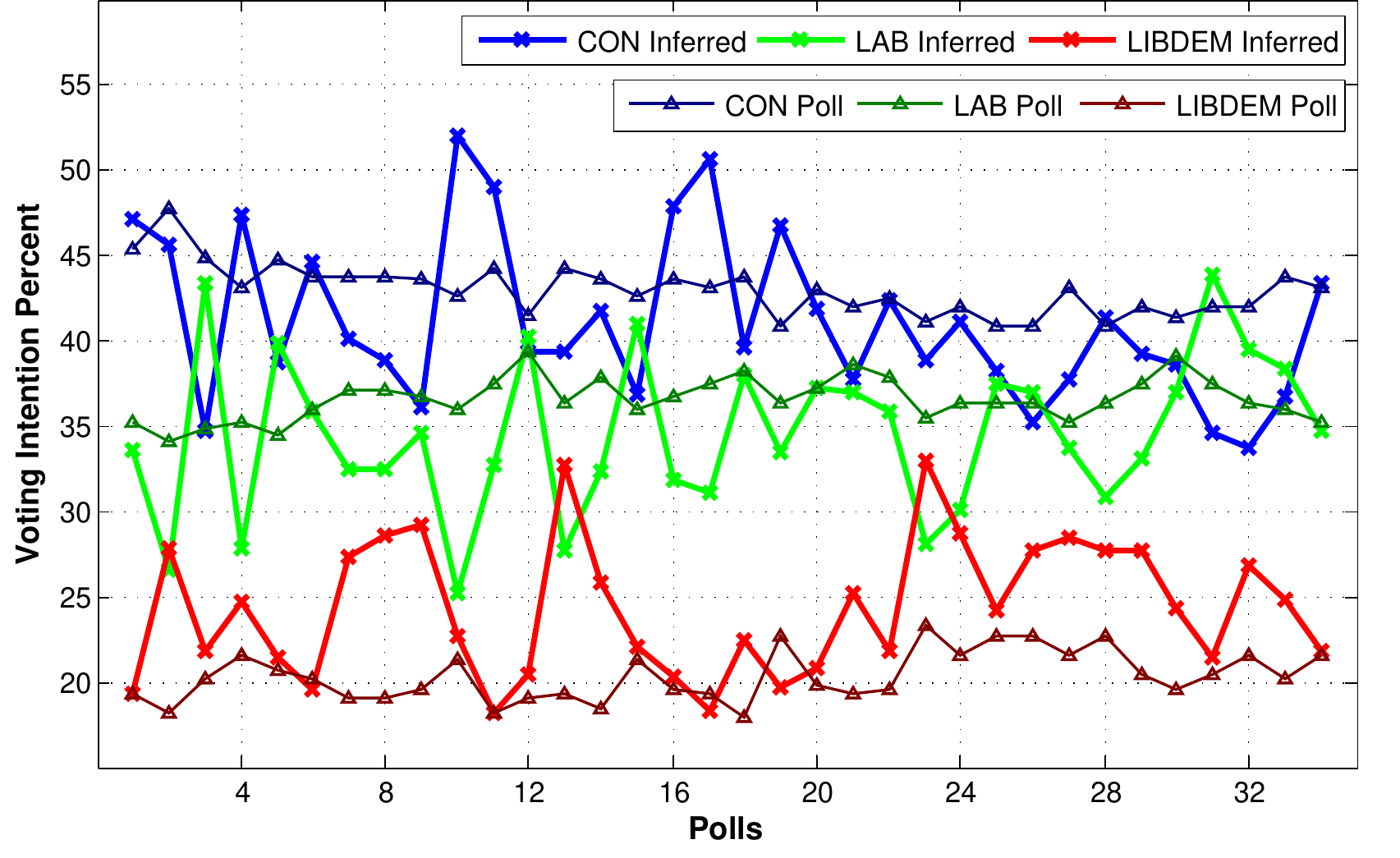}
    \label{fig_vi_l1out_MTS_SPOWS_part1}}
    \hfil
    \subfigure[Last 34 time instances]{\includegraphics[width=6in]{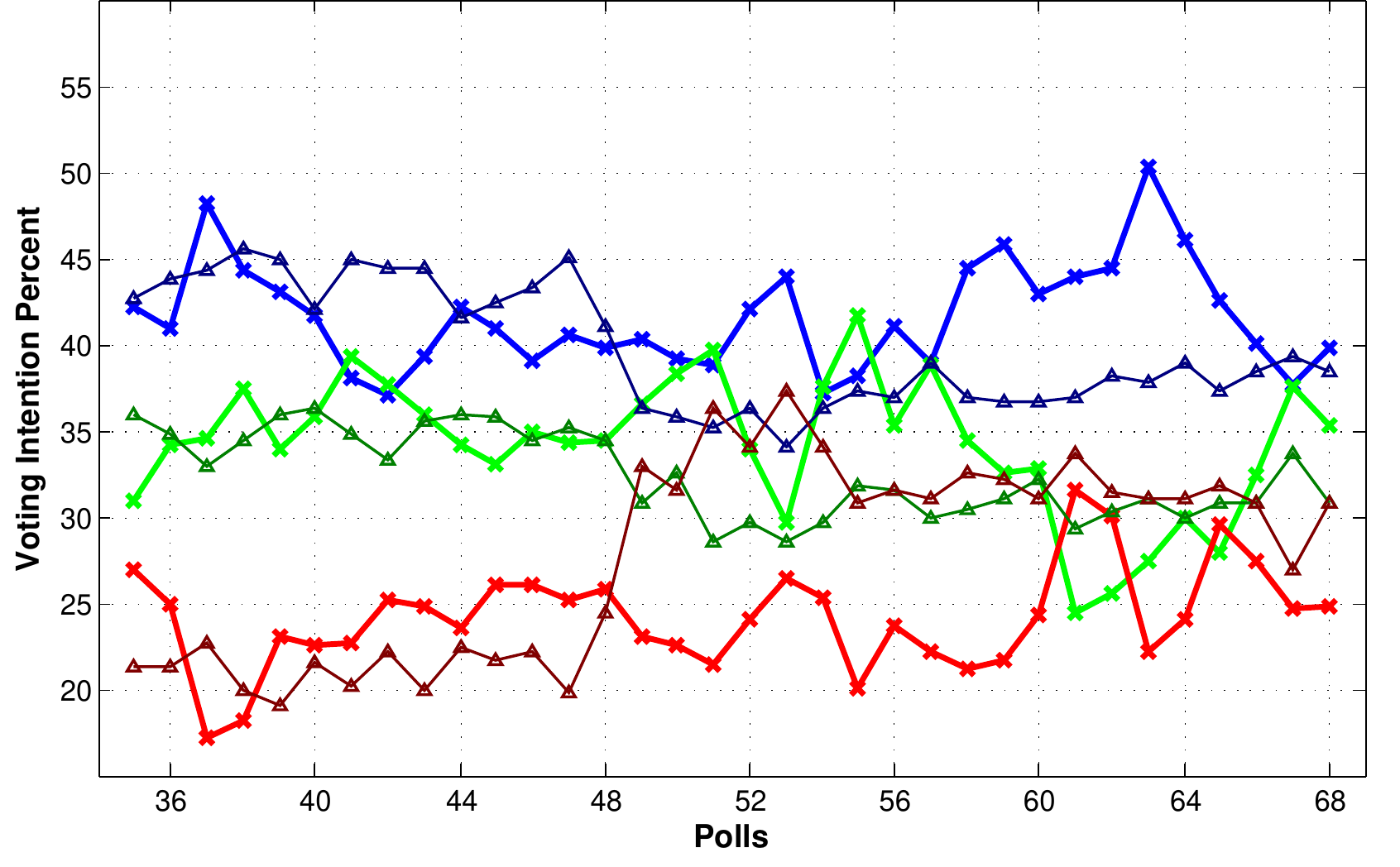}
    \label{fig_vi_l1out_MTS_SPOSW_part2}}
    \end{center}
    \caption{Leave-one-out cross validation by applying SPOSW and MTS thresholded (in two parts for a better visualisation).}
    \label{fig_vi_l1out_MTS_SPOSW}
\end{figure*}

We retrieve the first performance figures by performing leave-one-out cross validation on the aforementioned experimental setups. The performance results for MTS and DSC are presented in Tables \ref{table_vi_l1out_MTS} and \ref{table_vi_l1out_DCS} respectively. Under both methods, we see that thresholding tends -- in most occasions -- to improve the inference performance in terms of MAE and MRE. SnPOS performs significantly better than SPOS, which also fails to rank voting intentions properly under MTS. Using DCS results in a better performance on average, but does not deliver the overall best performance, which is derived by SPOSW under MTS using a thresholded data set. Figure \ref{fig_vi_l1out_MTS_SPOSW} depicts those best performing inferences against the corresponding ground truth; in this case, MAE across all parties is equal to 4.34 with a standard deviation of 2.13 and MRE is 0.1912.

\begin{table}
\caption{MAE $\pm$ MAE's standard deviation and MRE for Mean Thresholded Sentiment (\textbf{MTS}) by performing leave-one-out cross validation. $\delta$ denotes the minimum distance between the positive and negative sentiment score for a tweet in order to be considered.}
\label{table_vi_l1out_MTS}
\footnotesize
\renewcommand{\arraystretch}{1.2}
\setlength\tabcolsep{1mm}
\centering
\(\begin{tabular}{cc|ccc|c|c}
& $\delta$ & \textbf{CON} & \textbf{LAB}             & \textbf{LIBDEM}           & \textbf{All Parties}     & \textbf{MRE}\\\hline
\textbf{SnPOS} & 0      & 12.24 $\pm$ 10.66 & 10.66 $\pm$ 10.39 & 10.1  $\pm$ 11.54 & 11    $\pm$ 8.31 & 0.4559\\
\textbf{SnPOS} & 0.025  & 12.17 $\pm$ 10.63 & 10.54 $\pm$ 10.47 & 10.11 $\pm$ 11.44 & 10.94 $\pm$ 8.27 & 0.4779\\\hline
\textbf{SPOS}  & 0      & 36.06 $\pm$ 7.09  & 17.86 $\pm$ 11.29 & 15.12 $\pm$ 11.65 & 23.02 $\pm$ 5.23 & 0.9412\\
\textbf{SPOS}  & 0.0072 & 35.99 $\pm$ 7.17  & 17.88 $\pm$ 11.26 & 15.06 $\pm$ 11.65 & 22.98 $\pm$ 5.3  & 0.9412\\\hline
\textbf{SPOSW} & 0      & 4.22  $\pm$ 2.78  & 3.75  $\pm$ 2.86  & 5.1   $\pm$ 3.76  & 4.36  $\pm$ 2.13 & 0.2059\\
\textbf{SPOSW} & 0.0238 & 4.17  $\pm$ 2.88  & 3.76  $\pm$ 2.82  & 5.07  $\pm$ 3.7   & 4.34  $\pm$ 2.13 & 0.1912\\
\end{tabular}\)
\end{table}

\begin{table}
\caption{MAE $\pm$ MAE's standard deviation and MRE for Dominant Class Sentiment (\textbf{DCS}) by performing leave-one-out cross validation.}
\label{table_vi_l1out_DCS}
\footnotesize
\renewcommand{\arraystretch}{1.2}
\setlength\tabcolsep{1mm}
\centering
\(\begin{tabular}{cc|ccc|c|c}
& $\delta$ & \textbf{CON} & \textbf{LAB}             & \textbf{LIBDEM}           & \textbf{All Parties}     & \textbf{MRE}\\\hline
\textbf{SnPOS} & 0      & 9.68  $\pm$ 8.82  & 9.26 $\pm$ 8.76 & 9.34 $\pm$ 9.23 & 9.42  $\pm$ 6.66 & 0.4338\\
\textbf{SnPOS} & 0.025  & 8.23  $\pm$ 6.83  & 8.66 $\pm$ 8.34 & 9.28 $\pm$ 9.12 & 8.72  $\pm$ 6.39 & 0.375\\\hline
\textbf{SPOS}  & 0      & 11.61 $\pm$ 11.99 & 7.29 $\pm$ 5.27 & 7.9  $\pm$ 6.06 & 8.93  $\pm$ 6.21 & 0.5074\\
\textbf{SPOS}  & 0.0072 & 11.62 $\pm$ 11.89 & 7.52 $\pm$ 5.34 & 8.09 $\pm$ 6.33 & 9.08  $\pm$ 6.46 & 0.5074\\\hline
\textbf{SPOSW} & 0      & 5.5   $\pm$ 4.12  & 4.19 $\pm$ 3.35 & 5.52 $\pm$ 4.82 & 5.07  $\pm$ 3.04 & 0.2647\\
\textbf{SPOSW} & 0.0238 & 5.47  $\pm$ 3.9   & 3.95 $\pm$ 3.34 & 5.49 $\pm$ 4.49 & 4.97  $\pm$ 2.88 & 0.2279\\
\end{tabular}\)
\end{table}

%\begin{figure*}
%    \begin{center}
%    \subfigure[MTS for SnPOS, $\delta = 0.025$]{\includegraphics[width=5in]{MeanThres_Fig1_SnPOS}
%    \label{fig_vi_meanthres_1}}
%    \hfil
%    \subfigure[MTS for SPOS, $\delta = 0.0072$]{\includegraphics[width=5in]{MeanThres_Fig2_SPOS}
%    \label{fig_vi_meanthres_2}}
%    \hfil
%    \subfigure[MTS for SPOSW, $\delta = 0.0238$]{\includegraphics[width=5in]{MeanThres_Fig3_SPOSW}
%    \label{fig_vi_meanthres_3}}
%    \end{center}
%    \caption{Voting Intention Inferences for Mean Thresholded Sentiment (\textbf{MTS})}
%    \label{fig_vi_meanthres}
%\end{figure*}
%
%\begin{figure*}
%    \begin{center}
%    \subfigure[DCS for SnPOS, $\delta = 0$]{\includegraphics[width=5in]{DomClass_Fig1_SnPOS}
%    \label{fig_vi_domclass_1}}
%    \hfil
%    \subfigure[DCS for SPOS, $\delta = 0$]{\includegraphics[width=5in]{DomClass_Fig2_SPOS}
%    \label{fig_vi_domclass_2}}
%    \hfil
%    \subfigure[DCS for SPOSW, $\delta = 0.0238$]{\includegraphics[width=5in]{DomClass_Fig3_SPOSW}
%    \label{fig_vi_domclass_3}}
%    \end{center}
%    \caption{Voting Intention Inferences for Dominant Class Sentiment (\textbf{DCS})}
%    \label{fig_vi_domclass}
%\end{figure*}

Since our main data set is of small size and based on the fact that the oscillations in the actual voting intention signals are not major (something that might affect the training and testing processes), we also perform a more focused testing. From the 68 polls, we use 1--30 and 47--58 for training and we perform testing on the remaining 26 ones (31--46 and 59--68). The number of tweets that we retrieve using our search terms is increasing as we get nearer to the election day. This is why the training set has been sliced into two parts, one `way' before the election day (05/01 to 24/03) and another much closer (13/04 to 25/04). In this experiment, we additionally retrieve a statistical significance indication for our inferences. To do that, we randomly permute the outcomes of MTS or DSC and come up with a randomised training and test set; we repeat this process for 1,000 times and count how many times a model that is based on randomly permuted data delivers a better inference performance than the actual one -- this fraction gives the p-value. We consider that a p-value lower than 0.05 indicates statistical significance.

\begin{table}
\caption{MAE $\pm$ MAE's standard deviation and MRE for Mean Thresholded Sentiment (\textbf{MTS}). $\delta$ denotes the minimum distance between the positive and negative sentiment score for a tweet in order to be considered.}
\label{table_vi_MTS}
\footnotesize
\renewcommand{\arraystretch}{1.2}
\setlength\tabcolsep{1mm}
\centering
\(\begin{tabular}{cc|ccc|c|c|c}
& $\delta$ & \textbf{CON} & \textbf{LAB}             & \textbf{LIBDEM}           & \textbf{All Parties}     & \textbf{MRE} & \textbf{p-value} \\\hline
\textbf{SnPOS} & 0      & 12.97 $\pm$ 10.89 & 10.56 $\pm$ 10.05 & 6.66  $\pm$ 7.08  & 10.06 $\pm$ 9.72  & 0.4038 & 0.222\\
\textbf{SnPOS} & 0.025  & 12.98 $\pm$ 10.92 & 10.51 $\pm$ 10.03 & 6.7   $\pm$ 7.19  & 10.06 $\pm$ 9.74  & 0.4038 & 0.259\\\hline
\textbf{SPOS}  & 0      & 35.03 $\pm$ 6.96  & 21.31 $\pm$ 9.4   & 12.71 $\pm$ 7.16  & 23.02 $\pm$ 12.11 & 0.9231 & 0.531\\
\textbf{SPOS}  & 0.0072 & 35.14 $\pm$ 6.82  & 21.33 $\pm$ 9.41  & 12.77 $\pm$ 7.13  & 23.08 $\pm$ 12.1  & 0.9231 & 0.527\\\hline
\textbf{SPOSW} & 0      & 4.44  $\pm$ 3.18  & 2.66  $\pm$ 1.85  & 3.74  $\pm$ 2.86  & 3.61  $\pm$ 2.76  & 0.1731 & 0.019\\
\textbf{SPOSW} & 0.0238 & 4.44  $\pm$ 3.31  & 2.65  $\pm$ 1.8   & 3.84  $\pm$ 2.71  & 3.64  $\pm$ 2.75  & 0.1346 & 0.006\\
\end{tabular}\)
\end{table}

\begin{figure}[!t]
\centering
\includegraphics[width=6in]{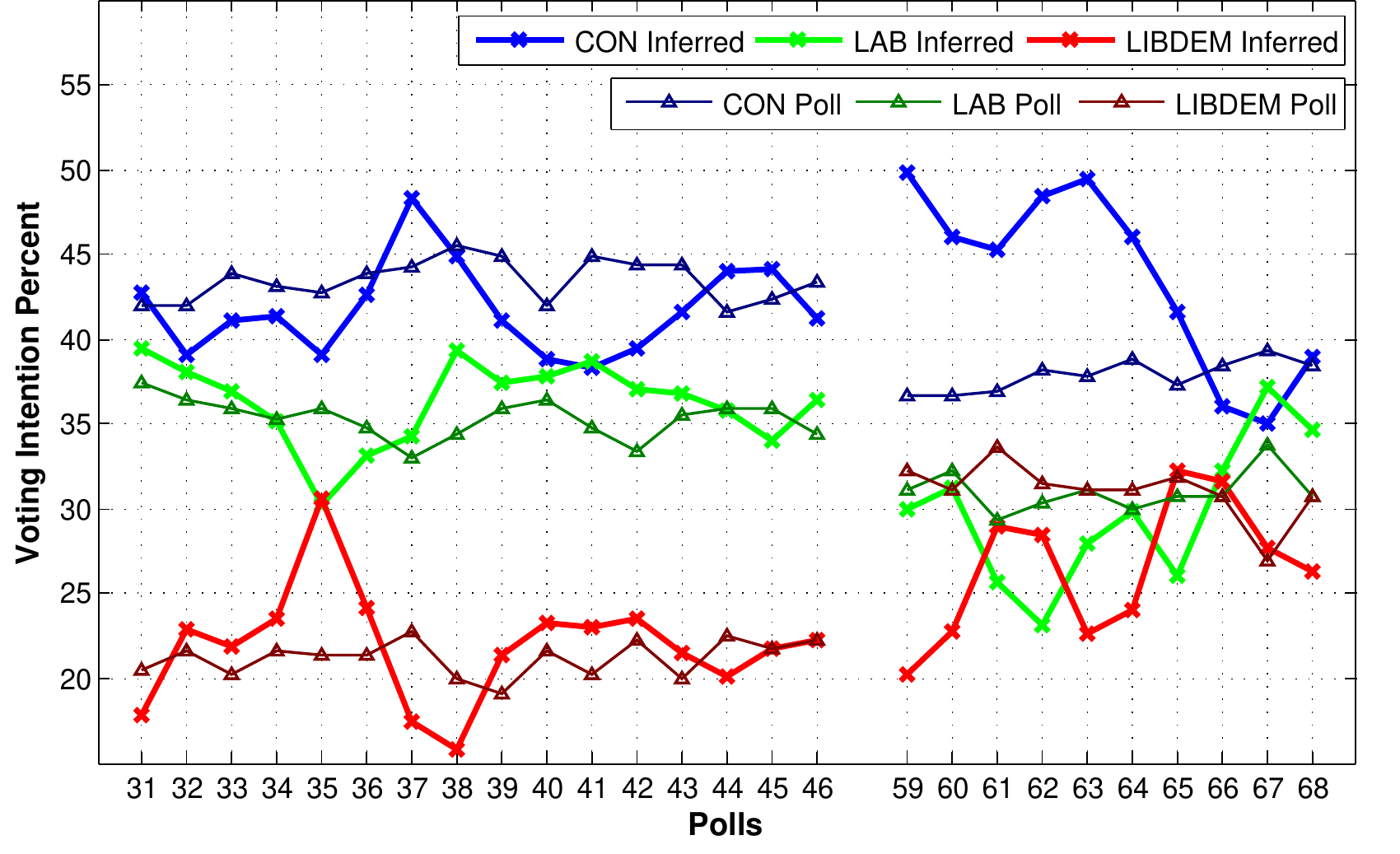}
\caption{SPOSW under thresholded DCS.}
\label{fig_vi_example_DCS_SPOSW}
\end{figure}

Inference results for MTS and DCS are presented in Tables \ref{table_vi_MTS} and \ref{table_vi_DCS} respectively. Again, it becomes apparent that combining SentiWordNet with POS tagging and extending tweets with WordNet's core senses (SPOSW) gives out the best inference performance as well as that DCS performs on average better than MTS. SnPOS and SPOS not only show a fairly poor performance in terms of MAE and MRE, but also do not deliver statistically significant results. On the contrary, SPOSW's inferences are shown to be statistical significant; its best performance is now achieved under DCS by performing thresholding reaching an MAE of 3.49 $\pm$ 2.98 and an MRE of 0.0962 (see Figure \ref{fig_vi_example_DCS_SPOSW}).

\begin{table}
\caption{MAE $\pm$ MAE's standard deviation and MRE for Dominant Class Sentiment (\textbf{DCS}).}
\label{table_vi_DCS}
\footnotesize
\renewcommand{\arraystretch}{1.2}
\setlength\tabcolsep{1mm}
\centering
\(\begin{tabular}{cc|ccc|c|c|c}
& $\delta$ & \textbf{CON}    & \textbf{LAB}             & \textbf{LIBDEM}           & \textbf{All Parties}       & \textbf{MRE} & \textbf{p-value}\\\hline
\textbf{SnPOS}  & 0      & 10.56 $\pm$ 6.75 & 9.82 $\pm$ 9.18 & 7.69 $\pm$ 9.88 & 9.36  $\pm$ 8.68 & 0.4038 & 0.467\\
\textbf{SnPOS}  & 0.025  & 9.66  $\pm$ 6.89 & 9.46 $\pm$ 9.05 & 7.25 $\pm$ 9.04 & 8.79  $\pm$ 8.35 & 0.3654 & 0.529\\\hline
\textbf{SPOS}   & 0      & 10.63 $\pm$ 8.94 & 8.09 $\pm$ 6.37 & 6.12 $\pm$ 5.12 & 8.28  $\pm$ 7.14 & 0.3846 & 0.238\\
\textbf{SPOS}   & 0.0072 & 10.51 $\pm$ 9.14 & 7.95 $\pm$ 6.18 & 6.08 $\pm$ 5.5  & 8.18  $\pm$ 7.26 & 0.4038 & 0.149\\\hline
\textbf{SPOSW}  & 0      & 4.51  $\pm$ 3.45 & 2.87 $\pm$ 2.06 & 3.53 $\pm$ 3.29 & 3.64  $\pm$ 3.04 & 0.1154 & 0\\
\textbf{SPOSW}  & 0.0238 & 4.49  $\pm$ 3.49 & 2.46 $\pm$ 1.81 & 3.51 $\pm$ 3.14 & 3.49  $\pm$ 2.98 & 0.0962 & 0\\
\end{tabular}\)
\end{table}

\subsection{Related work and further discussion of the results}
\label{section_voting_intention_discussion}
The topic of voting intention\index{voting intention} or electoral result inference from the content of Twitter is quite new in the scientific literature. Tumasjan \etal have published a paper, which after providing proof that Twitter is a platform where political discussions are conducted, proposes a method for predicting the result of the German elections in 2009 \cite{tumasjan2010predicting}. Their method uses the Linguistic Inquiry and Word Count (LIWC2007) for semantic analysis \cite{Pennebaker2007a}; this tool produces a set of 12 dimensions -- not just positive or negative sentiment -- and therefore in their method they have to introduce an averaging technique in order to acquire 1-dimensional representations. Their method like ours uses sets of keywords (party and politician names) to select politically oriented tweets and then proposes a model for matching the Twitter traffic per party to the final result of the elections. Another paper \cite{OConnor2010} presents a different method for tracking voting intention polls based on the ratio of positive versus negative sentiment per tweet; their learning approach has some similarities to ours (it is based on linear regression), but only deals with bivariate problems, \ie polls with two outcomes. Nevertheless, a paper published after indicated that the methods in \cite{tumasjan2010predicting} and \cite{OConnor2010} were problem specific and not very generic, as their application was proven to not do better than chance in predicting the result of the US congressional elections in 2009 \cite{Gayo-Avello2011}. Another paper showed that the popular tool of Google Trends has a limited capacity in predicting election results \cite{Lui2010}.

Moreover, an interesting paper conducted a further analysis of the aforementioned methodologies and proposed a triplet of necessary standards that any theory aiming to provide a consistent prediction of elections using content from the Social Media should follow \cite{Metaxas2011a}. The authors recommend that firstly the prediction theory should be formed as a well defined algorithm, secondly the analysis should be aware of the different characteristics arising in the Social Web and thirdly that the method must include experimental justification on why it works.

Based on those propositions, we tried to formalise the steps in our methodology, we provided statistical significance figures for our results and finally by forming the three different schemes (SnPOS, SPOS and SPOSW) for extracting sentiment from tweets, we tried to encounter some of the special characteristics of Twitter's textual stream. Particularly in SPOSW, where we enriched tweets by adding synonymous words in order to enhance their semantic interpretation, we observed a significant improvement on inference performance.

Similarly to the aforementioned works, the first two sentiment extraction methods that we tried (SnPOS and SPOS) showed poor performance and were not statistically significant. On average, modelling sentiment with DCS performed better than MTS. This is quite interesting as -- by definition -- DCS compresses more information than MTS; one might argue that this compression results in a much clearer signal, \ie removes some noisy observations, especially when applied on extended tweets. The contribution of thresholding in our experiments is ambiguous; sometimes it improves the inference performance and in most occasions reduces the MRE. As we mentioned in the beginning, the optimal value as well as the contribution of thresholding should be investigated further.

We remind the reader that the presented methods for modelling voting intention polls in this section are preliminary and that several aspects are a matter of future research. Primarily, our methods should be applied to other data sets as well, to come up with experimental proof about their capability to generalise. Another important factor is the choice of keywords that are used to select tweets relevant to the task at hand. We argued that those keywords can be selected from topic-related repositories; still, the influence of each keyword should be modelled. Ideally, an automatic mechanism or algorithm should be compiled in order not only to select an optimal subset of keywords, but also to quantify the contribution of each term. Moreover, possible biases introduced by the use of different sentiment analysis tools should also be considered; tools that incorporate special emotional expressions used in instant messaging and Twitter such as `;-)' or `:D' might achieve a better performance.

A general conclusion that could be extracted or reconfirmed from the presented work is that Social Media do encapsulate content related to public's political opinion; extending tweets with synonymous terms probably assists in the amplification of this signal. However, it is very important to contemplate that the validity of voting intention polls is questionable. Are those polls a good representation of what is actually happening? Usually, polls from different agencies tend to have significant differences in their percentages and quite often those differences exceed, for example, the lowest total MAE derived in our experimental process (3.49\%). Therefore, the formation of a consistently good ground truth is another issue that approaches like this one must resolve.% Nevertheless, methods like ours can be used as an additional indication for the professionals and be applied in conjunction with the original voting intention polls to improve the accuracy of their findings.

\section{Summary of the chapter}
\label{section_summary_chapter_patterns}
In this chapter, we presented some additional but preliminary work trying to exploit further the rich content of Twitter. Firstly, we investigated spatiotemporal relationships based on the textual stream. We showed that content correlation does not necessarily depend on geographical distance within the space of a country. That led to the proposal of a preliminary method capable of forming networks of content similarity among the considered UK locations. Those networks were proven to be stable over time and therefore, a good description of how content is shared across the UK.

Secondly, we focused on extracting posting time patterns on Twitter. The main result showed that Twitter users tend to tweet more as the day evolves with a peak in tweeting between 8 and 9p.m.; as expected, the lowest volume rates occur during the late night hours. More interestingly, this temporal pattern changes during the weekends, where we see that the special characteristics of those days (\eg no work, increased sleeping time in the mornings etc.) can explain the observed deviations from the weekday pattern. By forming clusters of weekdays based on our data set, we also showed that consecutive days are clustered together; the only exception is Monday, which seems to have a more `independent' behaviour.

Finally, we presented a preliminary approach for addressing the task of inferring voting intention polls from the content of Twitter, something that could lead to the approximate inference of an election result. Our approach -- similarly to other approaches on the topic (\eg \cite{tumasjan2010predicting}) -- selects tweets that might refer to politics by using lists of relevant keywords. We have formed three ways for converting those tweets to positive and negative sentiment scores; the first method (SnPOS) applies an averaged version of SentiWordNet \cite{Baccianella2010} disregarding the different parts of speech, the second one (SPOS) considers the different parts of speech by incorporating Stanford POS Tagger \cite{Toutanova2000}, and the third one (SPOSW) builds on the second by extending the selected tweets with synonyms from WordNet's core terms \cite{Miller1995}. We proposed two ways for converting the sentiment scores of a set of tweets into a voting intention percentage (MTS and DSC); the experimental results indicated that SPOSW (with either MTS or DSC) achieves the best performance. 

%% file: Chapters/Chapter8.tex
\chapter{Theory in Practice -- Applications Driven by our Theoretical Derivations}
\label{chapter:theory_in_practice}
%\lhead{Chapter 7. Theory in Practice -- Real-Time Applications driven by our Theoretical Derivations}

\rule{\linewidth}{0.5mm}
In this chapter, we present two applications which have been created in our effort to showcase our findings to the academic community and the general public. \textbf{Flu Detector} uses the methodology presented in Chapter \ref{Chapter_Nowcasting_Events_From_The_Social_Web} to infer influeza-like illness rates in several UK regions. \textbf{Mood of the Nation} is an application of a mood scoring method (MFMS) presented in Chapter \ref{chapter:detecting_temporal_mood_patterns} and displays the levels of four types of emotion (anger, fear, joy and sadness) for the same set of UK regions. Parts in this chapter are based on our publication ``Flu Detector -- Tracking epidemics on Twitter'' \cite{Lampos2010}.\footnote{Note that since this chapter is short, self-consistent and does not introduce new theory, a summary at the end was considered as redundant and hence, it has been skipped.}
\newline \rule{\linewidth}{0.5mm}
\newpage

\section{Flu Detector}
\label{section:Flu_Detector}
Monitoring the diffusion of an epidemic disease such as seasonal influenza is a very important task. Various methods are deployed by the health sector in order to detect and constrain epidemics, such as counting the consultation rates of GPs \cite{fleming2007lessons}, school or workforce absenteeism figures \cite{neuzil2002illness} and so on. The need of a proper infrastructure and the time delays due to the necessary data processing are the main drawbacks of those methodologies.

\begin{figure}
\centering
\includegraphics[width=5.2in]{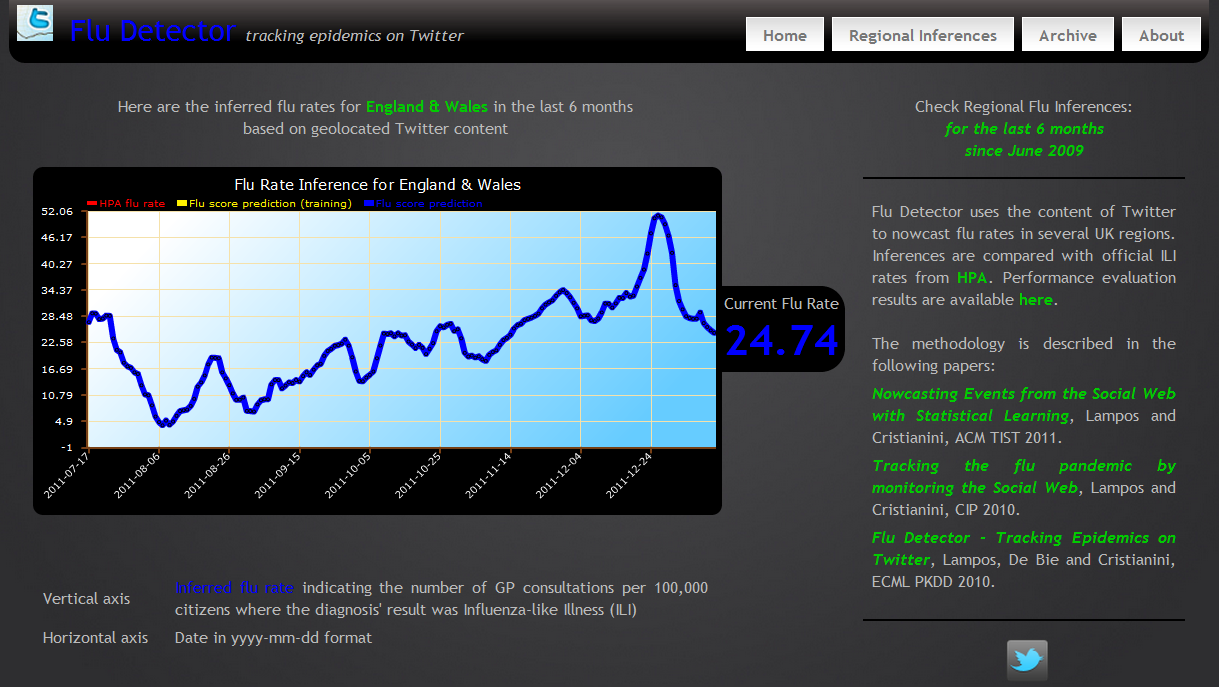}
\caption{The front page of Flu Detector's website. The inferred flu scores for the region of England \& Wales are displayed.}
\label{fig_flu_detector_1}
\end{figure}

We argue that information available on the web can provide an additional means for tackling this problem. In \cite{ginsberg2008detecting,polgreen2008using} it has been demonstrated that user queries on web search engines can be used to provide an early warning for an epidemic. Furthermore, recent work \cite{Lampos2010f,Asur2010a} has shown that the social web media have a predictive power on different domains. In particular, in article \cite{Lampos2011a} we have presented a method for inferring ILI rates for several regions in the UK by using data from Twitter (see also Chapter \ref{Chapter_Nowcasting_Events_From_The_Social_Web}). The core of the method performed feature selection by applying soft Bolasso with consensus threshold validation. In this section, we present a complete pipelined application implementing our theoretical findings titled as the \textbf{Flu Detector}\index{Flu Detector} (\url{http://geopatterns.enm.bris.ac.uk/epidemics/} -- see Figure \ref{fig_flu_detector_1}).

Flu Detector is an automated tool with a web interface that is used for tracking the prevalence of ILI\index{ILI} in several regions of the UK using the contents of Twitter's microblogging service. The data used to train and validate the models behind this tool are similar to the ones described in Section \ref{section_nowcasting_flu}. In particular, we use approx. 90 million tweets collected during a period of 392 days (from 21/06/2009 to 25/04/2010 and from 04/10/2010 to 26/12/2010) and geolocated within a 10Km range from the 54 most populated urban centres in the UK. Official ILI rates from the HPA\index{HPA} form our ground truth.

\subsection{Summary of applied methodology}
\label{section_flu_detector_methodology}
The applied methodology has been already described in detail in Section \ref{section_bolasso_methodology}; here we give a short summary of this process. The considered candidate features are formed by 1-grams and 2-grams; the feature extraction process has been described in Section \ref{section_nowcasting_flu}. Algorithm \ref{algorithm_soft_bolasso_with_CT_validation} is applied on the data set comprised by 1-grams as well as on the one of 2-grams and features (1-grams and 2-grams) are selected for all CTs. Then, by applying Algorithm \ref{algorithm_hybrid_combination_1grams_2grams_I} we decide the optimal CT and select a combined set of 1-grams and 2-grams (feature class \emph{H}). We learn the weight of each n-gram by applying OLS regression. Finally, by combining the selected features, their weights and the daily Twitter content geolocated within a UK region, we are able to compute a daily flu score for this region.

Since an inferred flu score can sometimes be a negative number, something which is not a realistic value given that flu rates should always larger than or equal to zero, we threshold negative inferences to zero. Additionally, based on the fact that flu rates show a smooth behaviour and therefore, no sudden peaks are expected, we also smooth each inference with the non smoothed version of the inferences in the past 6 days. In that way, we also maintain a weekly trend in the displayed scores.

\subsection{Back-end operations and web interface}
\label{section_flu_detector_operations}
Section \ref{section:crawlers_data_collection_storage} has already presented the data collection process. In short, we focus our data collection on tweets geolocated within a 10Km radius from the 54 most populated urban centres in the UK and query Twitter's Search API periodically in order to get the most recent tweets per urban centre. The posts are retrieved in Atom format, parsed using the ROME Java API and stored in a MySQL database.

\begin{figure*}
    \begin{center}
    \subfigure[Flu scores in the last 6 months for several UK regions. In this picture we see inferences for Scotland.]{\includegraphics[width=5.7in]{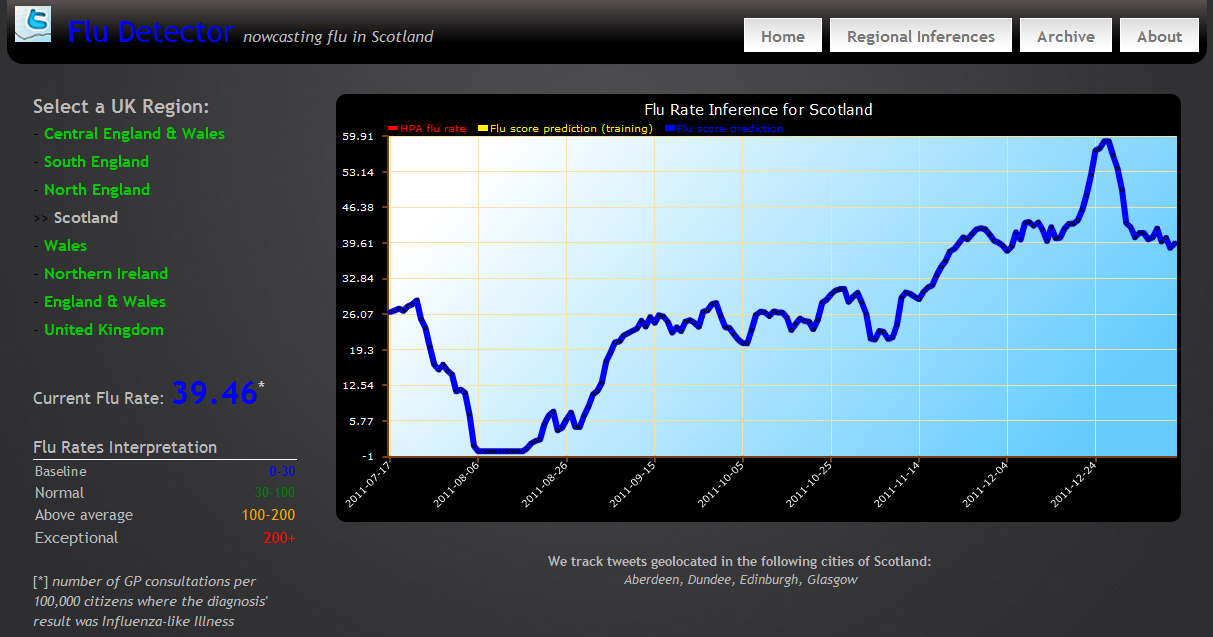}
    \label{fig_flu_detector_2}}
    \hfil
    \subfigure[Archive of flu scores (since June, 2009) for several UK regions. In this picture we see inferences for the entire UK.]{\includegraphics[width=5.7in]{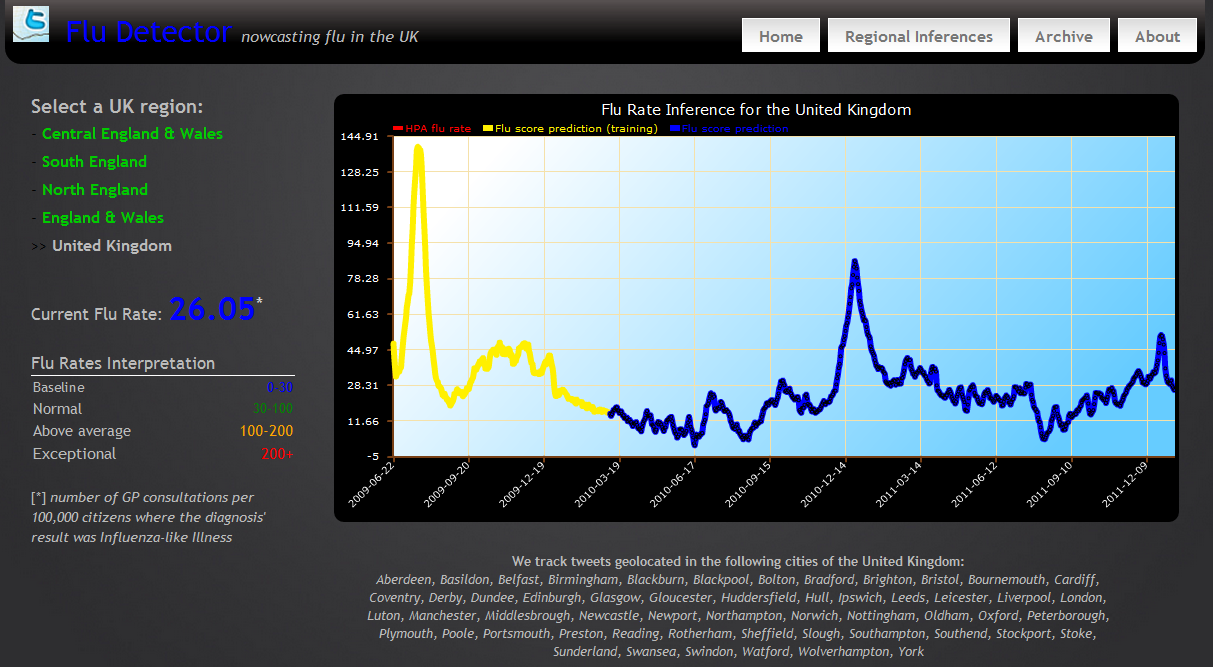}
    \label{fig_flu_detector_3}}
    \end{center}
    \caption{Sample pages of Flu Detector's website.}
    \label{fig_flu_detector_2_3}
\end{figure*}

The learning process -- described in the previous section -- is performed offline. An automated procedure updates the flu-score inferences on a daily basis and uploads them on the website. Apart from the main three regions (North England, South England, Central England \& Wales) on which the training procedure was based, we also display flu score projections for Northern Ireland, Scotland, Wales, England \& Wales and the entire UK.

Flu Detector's website is running on an Apache server using common web technologies (HTML, CSS and PHP). The charts are Flash applications based on the Open Flash Chart API.\footnote{ Open Flash Chart API, \url{http://teethgrinder.co.uk/open-flash-chart/}.} Figure \ref{fig_flu_detector_2} shows the page of the website where inferred flu scores of the last 6 months are displayed for the aforementioned regions (accessible via a navigation menu). The inferred flu rate of the previous day, \ie the current flu score, is also displayed on a separate box on the left. Recall that a flu score denotes the number of ILI-diagnosed patients in a population of a 100,000 citizens; rates between 0 and 30 are considered as baseline, between 30 and 100 as normal, 100 and 200 as above average and if they exceed 200, then they are characterised as exceptional. Finally, Figure \ref{fig_flu_detector_3} shows the page of the website, where the entire flu rate time series is displayed (since June 2009).

%%%%%%%%%%%%%%%%%%%%%%%%%%%%%%%%%%%%%%%%%%%%%%%
\section{Mood of the Nation}
\label{section_mood_of_the_nation}
Social media can be perceived as another way to look into the society \cite{Savage2011}. In particular, work has shown that the `mood' of tweets is able to predict the stock market \cite{Bollen2011} as well as detect significant events or situations emerging in real-life (\cite{Lansdall-Welfare2012} and Chapter \ref{chapter:detecting_temporal_mood_patterns}). In this section, we present a complete pipelined application implementing some of our theoretical findings titled as \textbf{Mood of the Nation}\index{Mood of the Nation} (\url{http://geopatterns.enm.bris.ac.uk/mood/} -- see Figure \ref{fig_mood_of_nation_1}).

\subsection{Summary of applied methodology}
\label{section_mood_of_the_nation_method}
The method that we applied in order to compute the displayed mood scores has been already described in Sections \ref{section_MFMS_circadian} and \ref{section:data_methods_mood_daily} in detail. Mood of the Nation displays the scores for four emotions, namely anger, fear, joy and sadness, based on the affective terms included in WordNet Affect \cite{Strapparava2004} (see also Appendix \ref{Ap:detecting_temporal_mood_patterns}). In particular, we have applied the Mean Frequency Mood Scoring (MFMS) scheme which is based on the assumption that the importance of each term is proportional to its frequency in the corpus.

We do not display raw mood scores but their z-scores, \ie from each score we subtract the mean and then divide the remainder with the standard deviation of the raw mood score. In that way, all mood scores are comparable to each other and can be plotted in a single figure. A positive score $m\in\mathbb{R}$ indicates that this mood type is $m$ standard deviations larger than its mean value; the opposite interpretation stands for a negative mood score. We have chosen to multiply all standardised mood scores (positive or negative) with a constant number equal to 10, to allow for a better presentation of the final results to the end user. Another choice of ours was to maintain a dynamic mean and standard deviation for each mood type, which means that every day -- as new data comes -- those values are automatically adapted.\footnote{ In our future plans is to allow the users to create and try different means directly from the website's interface.} As a final note, in contrast with Flu Detector's rates, here smoothing does not take place as it might `hide' sudden emotional peaks which are of interest.

\subsection{Back-end operations and web interface}
\label{section:mood_of_nation_backend_interface}
Data collection and other back-end operations are carried out in a similar manner as in Flu Detector. We kept the same skeleton for the website with a small amount of changes and also maintained the considered UK regions. Of course, we might be using the same Twitter data set, but now the core computation is different. Therefore, MFMS scheme is implemented in Java and updates the scores for the four mood types on a daily basis.

\begin{figure*}[tp]
    \begin{center}
    \subfigure[Front page of Mood of Nation where the inferred scores for Anger, Fear, Joy and Sadness in the entire UK are displayed.]{\includegraphics[width=5.5in]{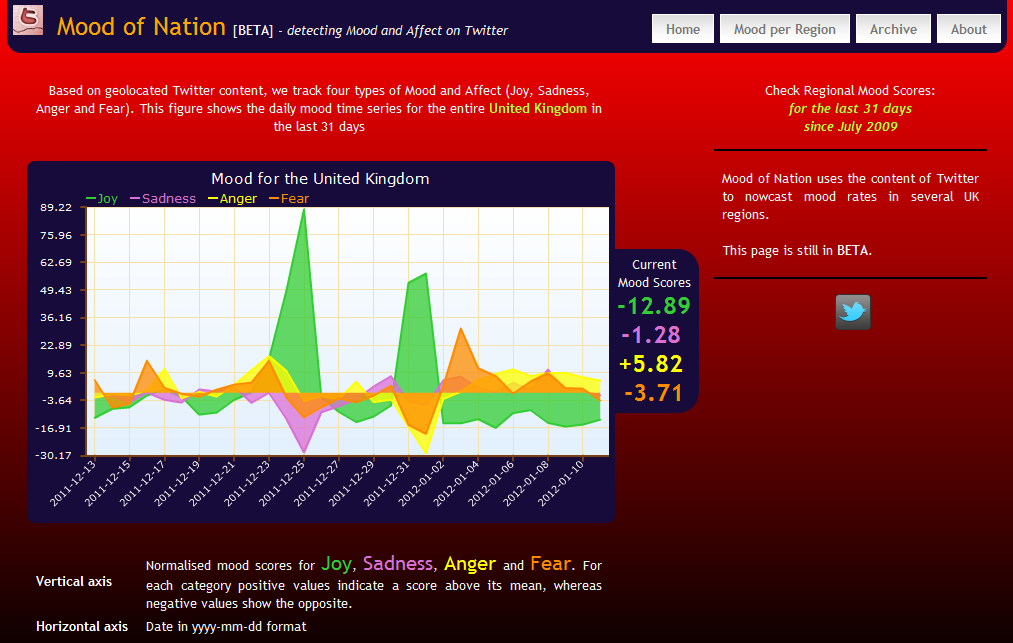}
    \label{fig_mood_of_nation_1}}
    \hfil
    \subfigure[Mood scores for the last 30 days for several UK regions. In this picture we see inferences for the region of South England.]{\includegraphics[width=5.5in]{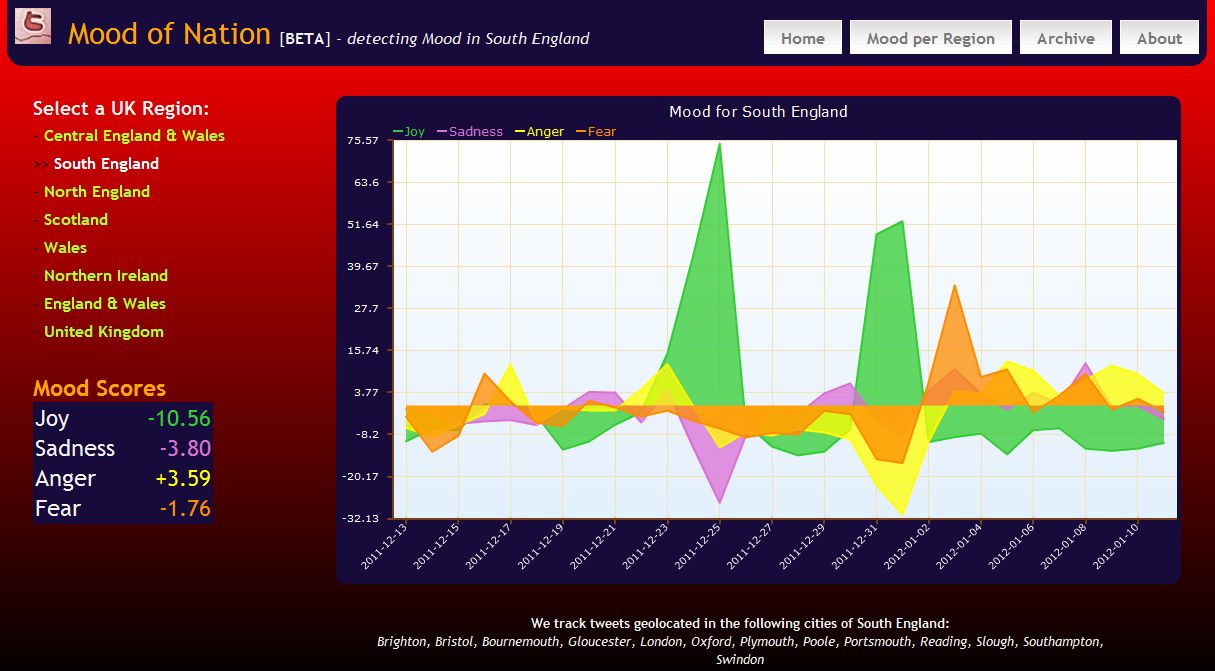}
    \label{fig_mood_of_nation_2}}
    \end{center}
    \caption{Sample pages of Mood of the Nation website.}
    \label{fig_mood_of_nation}
\end{figure*}

Every page of Mood of the Nation shows one chart that includes semi-transparent plots of all four mood types (coloured in yellow, orange, green and pink for anger, fear, joy and sadness respectively). Figure \ref{fig_mood_of_nation_1} shows the main page of the website where mood scores for the entire UK are displayed, whereas Figure \ref{fig_mood_of_nation_2} is the page where the mood scores of the last 30 days are displayed per region. Another interesting page is the one that displays historical mood data (starting from June 2009),\footnote{ No figure is available for this page, but it can be found at \url{http://geopatterns.enm.bris.ac.uk/mood/twitter-mood-archive.php?region=UK}.} where for example one can observe -- among more interesting observations discussed in Chapter \ref{chapter:detecting_temporal_mood_patterns} -- peaks of `joy' during the Christmas day for 2009, 2010 and 2011.

%\section{Summary of the chapter}
%\label{section_summary_chapter_apps} 

%% file: Chapters/Chapter9.tex
\chapter{Conclusions of the Thesis}
\label{Chapter_Conclusions}

\rule{\linewidth}{0.5mm}
This chapter summarises and closes the Thesis. We start by providing the reader with a chapter-by-chapter synopsis of the script. Then, we discuss the outcomes and conclusions of this work from a more general perspective, referring also to its impact on the academic community. At the end of this chapter, we propose further research directions that might be interesting to explore in the future.
\newline \rule{\linewidth}{0.5mm}
\newpage

\section{Summary of the Thesis with hindsight}
In \textbf{Chapter \ref{chapter:theoretical_background}}, we summarised the theoretical background behind our research. The core scientific field of our work is primarily defined by Statistical Machine Learning, but we also apply methods from the discipline of IR. Amongst other theoretical notions, this chapter also introduced the reader to regularised regression, explained how CARTs operate and introduced various vector space models.

\textbf{Chapter \ref{chapter:data_characterisation_collection}} gave a characterisation of Twitter, the social network that is used as the primary source of information for this project, firstly by describing its main attributes and then by referring to several `Twitter-driven' studies proving the significant value of this content in practice. In the second part of this chapter, we explained how we crawl, store and then process Twitter data, and also described the information sources that served as ground truth in our experiments.

\textbf{Chapter \ref{Chapter_first_steps}} presented the first and preliminary method for tracking a flu epidemic or in general the diffusion of ILI in the population by exploiting the content of Twitter \cite{Lampos2010f}. Starting from a manually formed set of illness related keywords, which were proven to have well correlated frequencies with the official flu rates, but then moving on and proposing a fully automated methodology that used a sparse regressor, the LASSO \cite{tibshirani1996regression}\index{regression!LASSO}, we were able not only to extract textual features (1-grams) that had a close semantic connection with the underlying topic, but also to infer accurately the flu rates for several regions in the UK during the H1N1\index{H1N1} epidemic (2009). However, this approach faced some limitations mainly because LASSO is not a model-consistent learner \cite{Lv2009, Zhao2006}, 1-grams did not always encapsulate a valid semantic description of the topic, and no proof was provided for a possible generalisation of the experimental results.

In \textbf{Chapter \ref{Chapter_Nowcasting_Events_From_The_Social_Web}}, we presented a more generic methodology for inferring the occurrence and magnitude of an event or phenomenon by exploring the rich amount of unstructured textual information on the social part of the web -- in particular, Twitter -- \cite{Lampos2011a}, by improving on the methods presented in the previous chapter. Linear (bootstrap LASSO with consensus threshold validation combined with OLS regression) and nonlinear (ensemble of CARTs\index{CART}) learning functions have been applied in the core of our methodology and the inference performance has been compared to other techniques used to address similar problems \cite{ginsberg2008detecting,Culotta2010}. In this study, we have used two types of features, 1-grams and 2-grams, and have also investigated several hybrid combinations of those feature classes in our effort to nowcast two phenomena emerging in real life: flu and rainfall rates. Whereas the importance and usefulness of inferring flu rates is a clear fact, rainfall rates inference mainly served as a benchmark and much harder problem that tested the capacity of our proposed inference schemes.

In \textbf{Chapter \ref{chapter:detecting_temporal_mood_patterns}}, we explored the same data source (\ie Twitter content) for a different problem: the extraction of seasonal circadian \cite{Lampos2012p} and daily mood patterns \cite{Lansdall-Welfare2012}. Our analysis was based on two methodologies, one (MFMS) that assigns emotion-triggering words with an importance weight analogous to their frequency in the text stream and another one (MSFMS) which standardises term frequencies and therefore removes biases induced by frequent emotional markers. We investigated four mood types, namely anger, fear, joy and sadness. The inferred circadian patterns were partly confirmed by research results from the field of Psychiatry, but contradicted with the ones presented in a similar recent work \cite{Golder2011}. In general, PA\index{PA} was shown to be stronger that NA\index{NA} during the daytime and vice versa during the night hours. Looking into daily mood patterns, we gained evidence supporting that real-life events affect the mood of Twitter users; in 2011, for example, the UK riots, Winehouse's death or the earthquake in Japan caused an observable increase in the NA of Twitter users geolocated in the UK. Both circadian and daily mood patterns have some periodical features; the most common period for all mood signals is the length of a week.

\textbf{Chapter \ref{chapter:pattern_discovery}} presented three additional Pattern Discovery\index{Pattern Discovery} studies on Twitter content that could serve as basis for future research directions. The first one investigated spatiotemporal relationships in the textual stream. Within the limits of a country, we showed that similarities in Twitter content do not depend on the distance between the locations, where the published content is authored; then, we proposed a preliminary method for forming networks of content similarity among the considered UK locations and demonstrated that the inferred networks are stable over time. Secondly, we focused on extracting posting time patterns on Twitter distinguishing between weekdays and weekends. Interestingly, those patterns seem to confirm common sense as they match with behavioural norms drawn from real-life. Finally, we reported on a preliminary set of methods for addressing the problem of voting intention inference from Twitter content using the 2010 UK General Election as a case study.

Lastly, in \textbf{Chapter \ref{chapter:theory_in_practice}} we presented two online tools, which showcase some of our findings to the academic community and also serve as dynamic applications of our theoretical derivations. `Flu Detector'\footnote{ Flu Detector, Link: \url{http://geopatterns.enm.bris.ac.uk/epidemics/}.}\index{Flu Detector} detects influenza-like illness for several UK regions \cite{Lampos2010}, whereas `Mood of the Nation'\footnote{ Mood of the Nation, Link: \url{http://geopatterns.enm.bris.ac.uk/mood/}.}\index{Mood of the Nation} displays mood scores (anger, fear, joy and sadness) for the same set of UK regions; both tools use Twitter content as their input information.
%In this chapter, we present two applications which have been created in our effort to showcase our findings to the academic community and the general public. \textbf{Flu Detector} uses the methodology presented in Chapter \ref{Chapter_Nowcasting_Events_From_The_Social_Web} to infer influeza-like illness rates in several UK regions. \textbf{Mood of the Nation} is an application of one of the methods presented in Chapter \ref{chapter:detecting_temporal_mood_patterns} and displays the scores of four mood types again for several regions in UK. Parts in this chapter are based on our publication ``Flu Detector -- Tracking epidemics on Twitter'' \cite{Lampos2010}.

\section{General conclusions}
\label{section:conc_general_conclusions}
%\textbf{Does useful information exist in the `open' text streams lying on the web?}
The work presented in this Thesis is an additional and consistent proof that information of high value does exist in unstructured, user-generated web text streams. This information reflects on various aspects of real-life already and as the number of people using and conducting parts of their life through the Social Media\index{Social Media} constantly increases, it is logically expected to reflect on even more.

%\textbf{Supposing that such information does exist, how can it be successfully extracted from the text stream?}
Throughout this Thesis we demonstrated several ways for extracting types of information from the social web and especially from Twitter's microblogging service. More specifically, we focused on extracting identifiable moments emerging in real-life, commonly referred to as events, as well as more ambiguous trends, which are encapsulated in the general task of Pattern Discovery\index{Pattern Discovery}. Our case studies on flu and rainfall rates inference (or nowcasting) showed that events, which tend to oscillate between active and inactive moments and -- at the same time -- may be a topic of discussion in the Social Media, could be successfully detected and quantified. Additional work showed that the textual stream encloses affective norms of the general population and delivered common life patterns, such as weekly periodic behaviours or distinctive emotional norms during weekends. Content similarities through time were used to form stable spatiotemporal networks of urban centres which reveal how information is shared among them. By combining statistical learning with NLP\index{NLP}, we were in the position to produce a function, which turns Twitter content to voting intention figures and therefore propose a preliminary method for this interesting task.

%\textbf{Is this extracted information a good representation of -- at least -- a real life's fraction?}
An important and crucial part of our work was the verification of our methodologies. In a supervised learning scenario, this is usually part of the problem's formulation and standard loss functions or similarity metrics can be applied, based on the fact that a ground truth is available to compare with. In an unsupervised scenario, one has to verify the acquired results with facts -- numerical, statistical or more abstract -- which might not be directly comparable with them. Based on the performance of our supervised methods for nowcasting events or inferring voting intention polls, we can strongly argue that information on the social web is a good representation of real life. This conclusion is becoming stronger by observing how well the results of mood analysis fit with real-life happenings such as sudden deaths of famous personas, celebrations or public unrest.

%\textbf{Formulate generic methodologies.}
We made an effort to develop as generic and generalisable methods or algorithms as possible. On the one side, the entirety of the proposed methodologies is directly applicable to other textual content streams. But most importantly, by investigating distinct case studies, we also acquired experimental proof -- both in terms of semantic relativity and measurable distance from a target topic or variable -- that those techniques would be able to extract several types of signals; not only the specific ones that we were in the position to test. Our experiments might have been based on one social platform only, but this, of course, should not be considered as a significant limitation as we avoided to use the specific structural features of Twitter.

%\textbf{Improve various subtasks.}
Wherever `early' success was achieved,\footnote{ Early acquires a relative definition which is decided by the strict time limits that a Ph.D. project introduces.} we also tried to extend and improve our methodologies. The use of 2-grams, for example, increased the semantic connection between the automatically selected terms and the target topic, and the hybrid combinations of 1-grams and 2-grams improved further the inference performance. LASSO's model selection inconsistencies have been resolved by bootstrapping and nonlinearities have been explored by using CARTs. Lastly, experimental proof indicated that enriching the content of tweets with synonyms might improve the performance of sentiment analysis on them.

\section{Work impact on academic research}
\label{section_conclusions_work_impact}
Our publications were well-received by the academic community and several new papers referred to or built on their basic ideas; in this short paragraph we refer to some of them. The detection of flu or other infectious diseases using the contents of the social streams together with geolocation properties -- in some occasions -- became a research subject in many conference or journal publications \cite{Culotta2010, Paul2011, Signorini2011, Paul2010, Gomide2011, Krieck2011, Mooney2011, Collier2011} as well as the main theme of a recent book \cite{Vora2012} serving the newly introduced topic of `Infoveillance' \cite{Eysenbach2009}. Another research subject with increasing interest is the extraction of voting intentions by applying sentiment analysis on tweets \cite{Lui2010, Metaxas2011a}. Finally, the detection of environmental incidents and possible hazards by observing social networks and especially Twitter has been another important topic \cite{Doan2011,Smid2011}.\footnote{ At the moment (April 2012), our work has been cited in more than 40 scientific publications.}
%-- MOA-tweetreader: real-time analysis in twitter streaming data \cite{Bifet2011}
%-- measuring issue salience: using supervised machine learning to generate data from free responses to the ``most important problem'' question
%-- information credibility on twitter \cite{Castillo2011}
%-- vocal minority versus silent majority: discovering the opinions of the long tail
%-- automatic discovery of patterns in media content \cite{Cristianini2011}
%-- predicting asset value through twitter buzz \cite{Zhang2012a}
% real-time query systems for complex data sources
% understanding and exploiting information spreading and integrating technologies
% syndromic classification of twitter messages
% a weighted multi-factor algorithm for microblog search \cite{Zhao2011}
% the migraine radar-a medical study analyzing twitter messages?

\section{Research directions for future work}
\label{section:conc_limitations_future_work}
This work could be extended in several directions and we are glad to report that many of them are already explored by a growing academic community, something that can be easily proven by the increasing number of EU-funded projects and scientific publications on Social Media analysis.

A research direction is always the one of improving the core theory of the proposed methods, algorithms and techniques. More specifically, one could investigate the application of even more sophisticated linear, nonlinear or generative learners. The experimentation with additional case studies might also give out new information about the key properties that the learning function should hold. In addition, the mixture of linear with nonlinear learning functions could also be of interest. As far as the feature space is concerned and based on the fact that 2-grams showed a much better semantic correlation with the target topics, one could also investigate this further using n-grams with $n >$ 2 or a formalised set of regular expressions aimed to look for, identify and match with specific textual characteristics.

%Furthermore, one could also turn the event detection to a classification task by forming discrete classes; in some occasions, where we do not want to quantify an event but only detect wheather it happens or not, a classifier might provide us with ....
%Looking at the social web differently, one could also this information to solve this rich and large-scale amount of information could also serve as an excellent data set for testing the performance of the hundreds statistical or natural language tools that have been proposed by the academic community throughout

Another interesting aspect for future exploration is the incorporation of mood or sentiment analysis in the event detection methodology. It is rational to consider that events emerging in real life affect the mood in Social Media as we already have proof for this. Therefore, this relationship needs to be formalised, so that those affective norms could assist in the detection or quantification of at least a class of events in the effort of detecting or preventing potential social dangers.

Our methods have been concentrated on textual content written in English language only, hence porting our methods to other languages by applying Machine Translation frameworks could be of interest. Going a step further, an exciting challenge would be to successfully fuse different types of information; in this work we only used textual information, but the web offers much more. Information in images, podcasts and videos should be explored in combination with the signals from the textual stream.

Important challenges also emerge in combination with other scientific fields. This vast amount of information provides for example the social scientists or psychiatrists with the opportunity to answer questions, which before have been considered as experimentally infeasible. On the other hand, results derived by a statistical analysis usually need the interpretation of experts from other disciplines. Therefore, interdisciplinary research paths should naturally be explored.

As a final but equally important note, we have to warn the academic community for privacy issues imposed by all those new developments. Those issues should be taken very carefully in consideration, as all the proposed methods, could indirectly expose and stigmatise users and therefore, people. Computer scientists must work closely with legislators to reassure that all this automation will improve the quality of life and will not restrict society's freedom of speech and right to privacy. 

%% file: Appendices/AppendixA.tex
% Appendix A
\chapter{Basic Theoretical Notions}
\label{AppendixA}
%\lhead{Appendix A. \emph{Basic Theoretical Notions}}

\section{Arithmetic Mean, (Co)Variance, Standard Deviation and Z-score}
\label{Ap:Mean_Std_Zscore}
%\footnote{ There exist several statistical quantities called means, \emph{e.g.} the harmonic or the geometric mean, but in this script we refer to and use the arithmetic mean.}
The arithmetic mean for an $N$-sized sample of a discrete variable $X = \{x_1, x_2, ..., x_N\}$ is given by:
\begin{equation}
\label{eq:mean}
\mu_{X} = \frac{1}{N}\sum_{i=1}^{N}x_i.
\end{equation}

The variance $\sigma_{X}^{2}$ of the above sample is defined as
\begin{equation}
\label{eq:variance}
\sigma_{X}^{2} = \frac{1}{N}\sum_{i=1}^{N}(x_i-\mu_X)^2,
\end{equation}
whereas the standard deviation $\sigma_{X}$ is equal to $\sqrt{\sigma_{X}^{2}}$.

The covariance of two variables $X$ and $Y$ of size $N$ is given by:
\begin{equation}
\label{eq:covariance}
\text{cov}_{XY} = \sum_{i = 1}^{N}\frac{(x_i - \mu_X)(y_i - \mu_Y)}{N}
\end{equation}
A zero covariance means that $X$ and $Y$ are uncorrelated; if $\text{cov}_{XY} > 0$, then as $X$ increases, $Y$ tends to increase and if $\text{cov}_{XY} < 0$, then $Y$ decreases, as $X$ increases.

In the above formulas, we are making the assumption that the samples are independent and identically distributed (\emph{i.i.d.} samples).

Z-score\index{z-score} is a method used for standardising the samples of a variable, using its sample mean and standard deviation. The numerical result of this standardisation shows how many standard deviations away from the mean is the value of each sample. In our example, $x_i$ will be transformed to:
\begin{equation}
\label{eq:zscore}
x_{i}^{(z)} = \frac{x_i - \mu_X}{\sigma_X}\text{, }i\in\{1,2, ..., N\}.
\end{equation}
%
%\section{Root Mean Square Error and Root Mean Square Deviation}
%\label{Ap:RMSE}
%Root Mean Square Error (\textbf{RMSE}) is simply the square root of the Mean Square Error (\textbf{MSE}); suppose we want to measure the error for an inferred vector $x$ from response vector $y$, then
%\begin{equation}
%\text{RMSE}(x,y) = \sqrt{\text{MSE}(x,y)} = \sqrt{\sum_{i=1}^{n} (x_i - y_i)^2},
%\end{equation}
%where $n$ is the size of $x$ and $y$. Many times RMSE is a preferred (to MSE) way for presenting performance results as the error figures have the same units with the responses.
%
%Together with RMSE, we often use a Normalised Root Mean Square Deviation (\textbf{NRMSD}), which is defined as:
%\begin{equation}
%\text{NRMSD}(x,y) = \frac{\text{RMSD}(x,y)}{x_{\text{max}} - x_{\text{min}}},
%\end{equation}
%where $x_{\text{max}}$ and $x_{\text{min}}$ are the max and min values of vector $x$ respectively.
%
%define root mean square
%say why it is used
%then also define root mean square deviation

\section{Variance and bias of a predictor}
\label{Ap:variance_bias_predictor}

Suppose that $y = f(X) + \varepsilon$, where $y$ is the target variable, $X$ denotes a set of observations, $f(\bullet)$ is the predictor's function and $\varepsilon \sim \mathcal{N}(0,\sigma^{2}_{\varepsilon})$. For an observation instance $X = x_i$, we compute the mean square error loss of the prediction $\hat{f}(x_i)$:
\begin{equation}
\renewcommand{\arraystretch}{1.4}
\begin{array}{clc}
\text{MSE}(x_i) & = \text{E}\left[\left(\hat{f}(x_i) - y\right)^2\right]_{X=x_i} & \\
                & = \text{E}\left[\left(\hat{f}(x_i) - f(x_i) - \varepsilon\right)^2\right] & \text{Let }\mu = \text{E}\left(\hat{f}(x_i)\right)\\
                & = \text{E}\left[\left(\hat{f}(x_i) - \mu + \mu - f(x_i) - \varepsilon\right)^2\right] & \\
                & =
                    \underbrace{\text{E}\left(\hat{f}(x_i) - \mu\right)^2}_{\text{variance}} +
                    \underbrace{\text{E}\Bigl(\mu - f(x_i)\Bigl)^2}_{\text{bias squared}} +
                    \text{E}\Bigl(\varepsilon - 0\Bigl)^2
                        & \text{(rest coefficients are equal to 0)}\\
                & = \text{var}\left(\hat{f}(x_i)\right) + \text{bias}\left(\hat{f}(x_i)\right)^2 + \sigma^{2}_{\varepsilon}. &\\
\end{array}
\end{equation}

In that way, the MSE is broken down into two main components: the variance of the prediction and the squared bias showing the amount by which the average of our estimate differs from the target value. The third component is the variance of the target around its true mean something that cannot be avoided unless $\sigma^{2}_{\varepsilon}$ is equal to 0. This process is known as bias-variance decomposition\index{bias-variance decomposition}. Depending on their properties learning models face bias-variance tradeoffs. Commonly, increasing the complexity of a model, results to a lower bias but higher variance. See \cite{hastie2005elements} for an extended reference on bias-variance decomposition.

\section{Cosine Similarity}
\label{Ap:Cosine_Similarity}
For two vectors $x$ and $y$ of the same size, their cosine similarity is simply the cosine\index{cosine similarity} of their angle $\theta_{xy}$. Therefore
\begin{equation}
\text{Sim}(x,y) = \cos(\theta_{xy}) = \frac{x \cdot y}{\|x\|_{\ell_2} \|y\|_{\ell_2}}.
\end{equation}
Sim$(x,y)$ ranges from -1 to 1 and intuitively the higher it is, the stronger the similarity between the vectors under comparison.

%\section{Covariance and Linear Correlation Coefficient
%\label{Ap:Pearson_Correlation}
%A mainstream way to test whether two variables are linearly correlated is the linear -- or Pearson's -- correlation coefficient [reference]. It is defined as:
%\begin{equation}
%\label{eq:corr_coef}
%\rho_{XY} = \frac{\text{cov}_{XY}}{\sigma_{X}\sigma_{Y}}
%\end{equation}
%The linear correlation coefficient -- also known as Pearson Correlation [reference]--

\section{Moving Average}
\label{Ap:MovingAverage}
Given a series of $k$ points $X$ = $\{x_1, x_2, ..., x_k\}$, the $n$-point moving average\index{moving average} is an equally sized series $X^{(n)}$ = $\{x_1^*, x_2^*, ..., x_k^*\}$, where
\begin{equation}
x_j^* = \frac{x_{j - \lfloor n/2 \rfloor} + ... + x_{j-1} + x_j + x_{j+1} + ... + x_{j + \lfloor n/2 \rfloor} }{n},
\end{equation}
and $n$ is an odd number. When $j \leq \lfloor n/2 \rfloor$ or $j > k - \lfloor n/2 \rfloor$, the enumerator of the above equation can have $2j - 1$ or $k - j + 1$ elements respectively.

\begin{figure}[!t]
\centering
\includegraphics[width=5in]{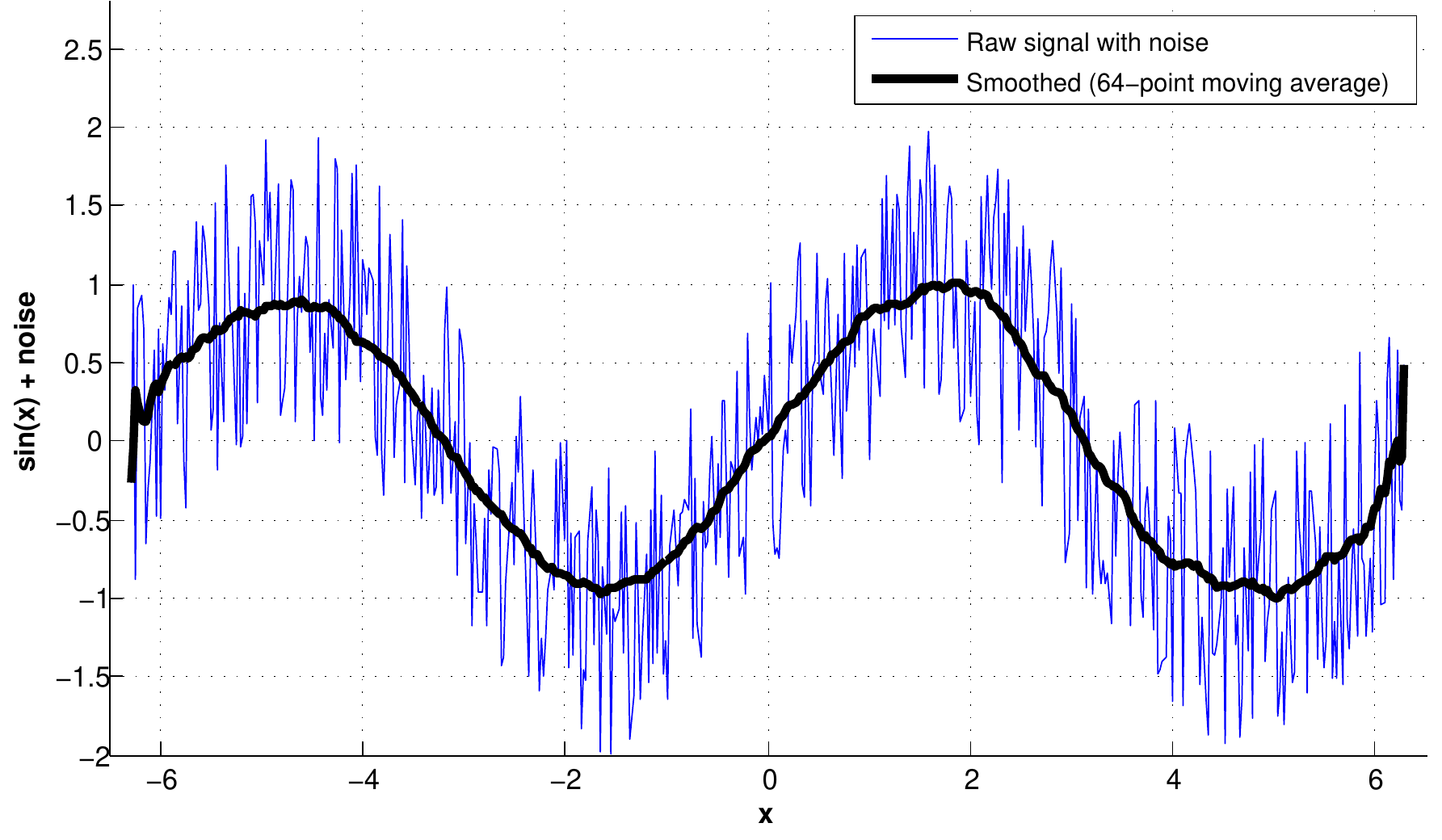}
\caption{Applying moving average to smooth and remove noise from a $\sin(x)$ signal.}
\label{fig_smoothing_example}
\end{figure}

In our work, we are applying the moving average method as a smoothing technique. A simple example follows. Suppose that we are adding random noise $\in$ [-1, 1] to the $\sin(x)$ function, where $x \in$ [-2$\pi$, 2$\pi$]. In figure \ref{fig_smoothing_example}, we have plotted a 513-point sample of the $\sin(x)$ $+$ \textbf{noise} signal, as well as its 64-point moving average. We see that by smoothing most of the (artificial) random noise has been removed recovering the original $\sin(x)$ signal.

\section{Null hypothesis, p-value and statistical significance}
\label{Ap:Null_pvalue}
A \textbf{null hypothesis}\index{null hypothesis} is an assertion, which is made on the expectation to be rejected due to the lack of statistical proof in the data that will support it. By rejecting a null hypothesis, we are aiming to support a hypothesis of an opposite nature, if such hypothesis can be defined.

\textbf{P-value}\index{p-value} $\rho$ is the probability that the null hypothesis is true. Given a set of $n$ trials, $\rho$ is equal to:
\begin{equation}
\rho = \frac{\#\{\text{Null hypothesis is true}\}}{n},
\end{equation}
\ie the number of times that the null hypothesis is true divided by the number of trials.

If a p-value is below a threshold $\alpha$ -- in most occasions $\alpha =$ 0.05 --, then the null hypothesis can be rejected due to the fact that statistics indicate it is unlikely to hold. This usually serves as an indication that a hypothesis of an opposite nature or a related assumption, which contradicts with the null hypothesis, might be \textbf{statistically significant}\index{statistical significance}.

\section{Principal Component Analysis}
\label{Ap:PCA}
A widely used technique with a clear application to feature extraction is Principal Component Analysis\index{Principal Component Analysis|see{PCA}} (\textbf{PCA}\index{PCA}). Given $X$, a set of $n$ $D$-dimensional observations $\{x_1, ..., x_n\}$, PCA can be defined as the orthogonal projection of $X$ onto an $M$-dimensional space, where $M \leq D$, with respect to maximising the variance of the projected data \cite{hotelling1933analysis}.\footnote{ PCA was first introduced by Pearson in \cite{pearson1901onlines}.} Generally, it can be expressed as the following constrained maximisation problem:
\begin{equation}
\renewcommand{\arraystretch}{1.5}
\begin{array}{lc}
\displaystyle\max_{w_1} & w_{1}^{\text{T}} S w_1\\
\text{s.t.} & w_{1}^{\text{T}}w_1 \leq 1,
\end{array}
\end{equation}
where $w_1$ denotes the direction of the variance and $S$ is the covariance matrix of the $X$, \ie
\begin{equation}
S = \frac{1}{n} \sum_{i=1}^{n}(x_i - \bar{x})(x_i - \bar{x})^{\text{T}}.
\end{equation}
It is shown that the solution of this optimisation problem is derived by solving a simple eigenvalue problem:
\begin{equation}
Sw_1 = \lambda_{1}w_{1}.
\end{equation}
The eigenvector $w_1$ is known as the principal component of $X$; additional principal components can be defined by computing the $M$ eigenvectors $w_1$, ..., $w_M$ of $S$ corresponding to the $M$ largest eigenvalues $\lambda_1$, ..., $\lambda_M$ \cite{Bishop2006}.

%\section{Bayes Rule}
%not sure maybe

%% file: Appendices/AppendixB.tex
% Appendix B
\chapter{Complementary Tables and Data}
\label{AppendixB}
%\lhead{Appendix B. Complementary Tables and Data}

\section{Urban Centres Per Region}
\label{Ap:Urban_centres_per_region}
Twitter data collection has been focused on a set of 54 UK urban centres. We collect tweets geolocated within a 10Km range from each centre. When inferring flu rates, we are using sets or groups of locations that belong in one geographic region. Here we list all the considered geographic regions as well as the locations belonging to each one of them.
\begin{itemize}
  \item \textbf{South England}: Brighton, Bristol, Bournemouth, Gloucester, London, Oxford, Plymouth, Poole, Portsmouth, Reading, Slough, Southampton, Swindon
  \item \textbf{Central England}: Basildon, Birmingham, Coventry, Derby, Ipswich, Leicester, Luton, Northampton, Norwich, Nottingham, Peterborough, Southend, Stoke, Watford, Wolverhampton
  \item \textbf{North England}: Blackburn, Blackpool, Bradford, Bolton, Huddersfield, Hull, Leeds, Liverpool, Manchester, Middlesbrough, Newcastle, Oldham, Preston, Rotherham, Sheffield, Stockport, Sunderland, York
  \item \textbf{Scotland}: Aberdeen, Dundee, Edinburgh, Glasgow
  \item \textbf{Wales}: Cardiff, Newport, Swansea
  \item \textbf{Northern Ireland}: Belfast
\end{itemize}

\section{An important observation -- 41 flu markers}
\label{Ap:41markers}
This is the list of terms (Table \ref{table_41_markers}) used in one of our very first experiments aiming to extract flu rates from Twitter content (Section \ref{subsection:gtruth_expansion}). It is a manually formed list with 41 textual markers (or n-grams) that we thought they might be able to describe an illness (in that particular case a flu-like illness) and its symptoms.

\begin{table}[h]
\caption{41 hand-picked flu markers}
\label{table_41_markers}
\renewcommand{\arraystretch}{1.2}
\small
\setlength\tabcolsep{10mm}
\centering
\begin{tabular}{ccc}
\toprule
aches           & fatigue           & pain \\
back pain       & feel sick	        & pneumonia\\
blood           & feeling bad       & runny nose\\
catarrh	        & feeling unwell	& shivers\\
cold            & fever	            & sick\\
colic           & flu               & sneezes\\
coma            & gastroenteritis	& sore throat\\
cough           & headache          & spasm\\
cramps          & infection         & stomach ache\\
croups          & inflammation      & temperature\\
diarrhea        & irritation        & tiredness\\
discharge       & muscles ache      & tonsils\\
dizzy           & muscles pain      & vomit\\
fainting        & nausea            &       \\
\bottomrule
\end{tabular}
\end{table}

\section{Detecting temporal mood patterns -- 1-grams from WordNet Affect}
\label{Ap:detecting_temporal_mood_patterns}
Tables with the terms used to track affective norms for anger (Table \ref{table_anger_words}), fear (Table \ref{table_fear_words}), joy (Table \ref{table_joy_words}) and sadness (Table \ref{table_sadness_words}). Those terms have been retrieved from WordNet Affect \cite{Strapparava2004} first by removing the very few existing 2-grams and then by stemming all 1-grams.

\begin{table}[h]
\caption{List of stemmed words used to detect \textbf{Anger} -- 146 terms}
\label{table_anger_words}
\renewcommand{\arraystretch}{0.8}
\setlength\tabcolsep{0.2mm}
\scriptsize
\centering
\begin{tabular}{cccccccccccc}
\toprule
abhor   & abhorr    & abomin    & aggrav    & aggress   & amok      & amuck     & anger     & angri     & angrili   & animos    & animu\\
annoi   & annoy     & antagon   & avarici   & baffl     & balk      & bedevil   & begrudg   & belliger  & bitter    & bother    & bothersom\\
brood   & chafe     & choler    & covet     & crucifi   & dander    & despis    & despit    & detest    & devil     & discourag & disdain\\
displeas & displeasur & dudgeon & dun & enmiti & enrag & envi & enviabl & enviou & envious & evil & exacerb\\
exasper     & execr & fit & frustrat & furi & furiou & furious & gall & grabbi & grasp & gravel & greedi\\
grievanc & grizzl & grudg & hackl & harass & harri & hate & hatr & heartburn & hostil & huffi & huffili\\
incens & indign & indignantli & infuri & ir & irasc & irrit & jealou & jealous & jealousi & livid & loath\\
mad & madden & malefic & malevol & malic & malici & malign & misanthrop & misanthropi & misogyn & misogyni & mison\\
murder & nettl & nettlesom & odium & offend & offens & oppress & outrag & peev & persecut & peski & pester\\
pestifer & piqu & pout & prehensil & provok & rag & rage & rancor & resent & resentfulli & rile & roil\\
scene   & score  & scorn    & sore      & spite & spleen        & stew  & stung & sulk  & sulki     & tantal    & tantrum\\
teas    & temper & torment  & umbrag    & veng  & vengefulli    & venom & vex   & vexat & vexati    & vindict   & warpath\\
wrath   & wroth  & & & & & & & & & &\\
\bottomrule
\end{tabular}
\end{table}

\begin{table}[h]
\caption{List of stemmed words used to detect \textbf{Fear} -- 92 terms}
\label{table_fear_words}
\renewcommand{\arraystretch}{0.8}
\setlength\tabcolsep{0.2mm}
\scriptsize
\centering
\begin{tabular}{cccccccccccc}
\toprule
affright & afraid & alarm & alert & anxiou & anxious & appal & apprehens & atroci & aw & awfulli & bashfulli\\
bode & browbeaten & bulli & chill & constern & cow & cower & crawl & creep & cring & cruel & cruelli\\
cruelti & dash & daunt & diffid & dire & dismai & dread & dreadfulli & fawn & fear & fearfulli & fearsom\\
forebod & fright & frighten & frighteningli & grovel & hangdog & hardheart & heartless & heartlessli & hesit & hesitantli & hesitatingli\\
hideou & hideous & horrend & horribl & horrid & horrif & horrifi & horrifyingli & horror & hyster & hysteria & intimid\\
merciless & monstrous & outrag & pall & panic & panick & panicki & pitiless & premonit & presag & presenti & ruthless\\
scare & scari & scarili & shadow & shiveri & shyli & shyness & suspens & terribl & terrifi & terror & timid\\
timidli & timor & trepid & ugli & unassert & uneasili & unkind & unsur & & & &\\
\bottomrule
\end{tabular}
\end{table}

\begin{table}[h]
\caption{List of stemmed words used to detect \textbf{Joy} -- 224 terms}
\label{table_joy_words}
\renewcommand{\arraystretch}{0}
\setlength\tabcolsep{0.1mm}
\scriptsize
\centering
\begin{tabular}{ccccccccccc}
\toprule
admir & ador & affect & affection & amatori & amic & amor & anticip & appreci & approb & approv\\
ardor & attach & avid & bang & banter & barrack & beam & beguil & belong & benef & benefic\\
benefici & benevol & bewitch & blith & bonheur & brotherhood & brotherli & buoyanc & captiv & captur & care\\
carefre & catch & chaff & charg & charit & charm & cheer & cheerfulli & cheeri & close & comfort\\
commend & compat & complac & concern & congratul & consol & content & crush & delight & devot & eager\\
eagerli & ebulli & elan & elat & embolden & emot & empath & empathet & empathi & enamor & enamour\\
enchant & endearingli & enjoi & enthral & enthusiasm & enthusiast & entranc & esteem & euphor & euphoria & exalt\\
excit & exhilar & exhort & expans & expect & exuber & exuberantli & exult & exultantli & fanci & fascin\\
favor & favour & festal & festiv & flush & fond & fondli & fratern & friendli & fulfil & gai\\
gaieti & gala & glad & gladden & glee & gleeful & gleefulli & gloat & good & goodwil & gratifi\\
gratifyingli & great & gusto & happi & happili & heart & hearten & hilar & hilari & identif & identifi\\
impress & infatu & insouci & inspir & intimaci & intox & jocular & jocund & joi & jolli & jolliti\\
jovial & joy & joyfulli & joyou & joyous & jubil & jubilantli & keen & kick & kid & kindheart\\
kindli & laudabl & lighthearted & lightsom & likabl & like & likeabl & love & loving & lovingli & loyalti\\
merri & merrili & merriment & mirth & mirthfulli & near & nigh & occupi & partial & penchant & perki\\
predilect & preen & prefer & pride & protect & proudli & rapport & recreat & regard & rejoic & relish\\
respect & revel & riotous & romant & rush & satisfact & satisfactori & satisfactorili & satisfi & satisfyingli & schadenfreud\\
scream & smug & softheart & solac & sooth & stimul & strike & sunni & suspens & sympathet & sympathi\\
tast & teas & tender & thrill & tickl & titil & tranc & triumph & triumphal & triumphant & triumphantli\\
unworri & uplift & uproari & urg & wallow & warm & warmheart & warmhearted & warmth & weak & worri\\
worship & zeal & zealou & zest & & & & & & &\\
\bottomrule
\end{tabular}
\end{table}

\begin{table}[h]
\caption{List of stemmed words used to detect \textbf{Sadness} -- 115 terms}
\label{table_sadness_words}
\renewcommand{\arraystretch}{0.8}
\setlength\tabcolsep{0.1mm}
\scriptsize
\centering
\begin{tabular}{ccccccccccc}
\toprule
aggriev & attrit & bad      & bereav    & bereft   & blue         & bore      & cheerless & compunct  & contrit     & dark\\
deject  & demor  & deplor   & depress   & desol    & despairingli & despond   & dingi     & disconsol & discourag   & dishearten\\
dismai  & dismal & dispirit & distress  & dole     & dolefulli    & dolor     & downcast  & downheart & downtrodden & drab\\
drear   & dreari & dysphor  & dysphoria & execr    & forlorn      & forlornli & gloom     & gloomi    & gloomili    & glum\\
godforsaken  & grief      & griev     & grievou    & grievous & grim    & guilt     & guilti    & hangdog    & hapless & harass\\
heartach     & heartbreak & heartrend & heartsick  & helpless & joyless & joylessli & lachrymos & laden      & lament  & loneli\\
lorn         & low        & melanchol & melancholi & miser    & miseri  & misfortun & mourn     & mournfulli & oppress & pathet\\
penanc       & penit      & penitenti & persecut   & piteou   & piti    & pitiabl   & plaintiv  & poor       & regret  & remors\\
remorsefulli & repent     & rue       & rueful      & ruefulli & ruth   & sad       & sadden    & sadli      & shame   & shamefac\\
somber       & sorri      & sorrow    & sorrowfulli & suffer   & tear   & tyrann    & unhappi   & weep       & weepi   & weight\\
woe          & woebegon   & woeful    & woefulli    & wretch   & & & & & &\\
\bottomrule
\end{tabular}
\end{table}

\section{Inferring Voting Intentions from Twitter Content -- Search Terms}
\label{Ap:voting_intention}
Terms used to select tweets related to the major political parties in the UK. Character `\_' denotes an empty space and `\#' denotes a Twitter topic. Tables \ref{table_conservatives}, \ref{table_labour} and \ref{table_libdem} hold the terms for the Conservative (52 terms), Labour (57 terms) and Liberal Democrat party (62 terms) respectively.

\begin{table}[h]
\caption{Terms for the Conservative Party (\textbf{CON}) -- 52 terms}
\label{table_conservatives}
\renewcommand{\arraystretch}{1.2}
\footnotesize
\centering
\begin{tabular}{cccc}
\toprule
\#conservative       & CONSERVATIVE          & Eric Pickles      & Margaret Thatcher\\
\#Tories             & Conservative          & Francis Maude     & Michael Gove\\
\#tories             & Conservative Party    & George Osborne    & Oliver Letwin\\
\#Tory               & Crispin Blunt         & Gerald Howarth    & Owen Paterson\\
\#tory               & David Cameron         & Grant Shapps      & Patrick McLoughlin\\
\_TORIES             & David Davis           & Hugo Swire        & Philip Hammond\\
\_tories             & David Mundell         & Iain Smith        & President Cameron\\
Alan Duncan          & David Willetts        & Jeremy Hunt       & Theresa May\\
Andrew Lansley       & debate Cameron        & John Redwood      & Tories\\
Andrew Mitchell      & Dominic Grieve        & Ken Clarke        & vote Cameron\\
Boris Johnson        & Duncan Smith          & Lady Warsi        & vote conservative\\
Caroline Spelman     & Edward Leigh          & Leader Cameron    & vote Tory\\
Cheryl Gillan        & Edward Vaizey         & Liam Fox          & William Hague\\
\bottomrule
\end{tabular}
\end{table}

\begin{table}[h]
\caption{Terms for the Labour Party (\textbf{LAB}) -- 57 terms}
\label{table_labour}
\renewcommand{\arraystretch}{1.2}
\footnotesize
\centering
\begin{tabular}{cccc}
\toprule
\#labour            & debate Brown          & LABOUR                & Rosie Winterton\\
\_Ed \_Balls        & Diane Abbott          & Labour                & Sadiq Khan\\
Alan Johnson        & Douglas Alexander     & Labour Party          & Shaun Woodward\\
Alastair Campbell   & Ed Miliband           & Leader Brown          & Stella Creasy\\
Andy Burnham        & Gordon Brown          & Liam Byrne            & Tessa Jowell\\
Angela Eagle        & Government Brown      & Luciana Berger        & Tom Harris\\
Ann McKechin        & Harriet Harman        & Maria Eagle           & Tom Watson\\
Ben Bradshaw        & Hilary Benn           & Mary Creagh           & Tony Blair\\
Caroline Flint      & Ivan Lewis            & Meg Hillier           & Tony Lloyd\\
Charles Clarke      & Jim Murphy            & MILIBAND              & vote Brown\\
Chris Bryant        & John Denham           & Miliband              & vote labour\\
Chuka Umunna        & John Healey           & Peter Hain            & Yvette Cooper\\
David Blunkett      & John Prescott         & Peter Mandelson       & \\
David Lammy         & Jon Trickett          & President Brown       & \\
David Miliband      & Kerry McCarthy        & Prime Minister Brown  & \\
\bottomrule
\end{tabular}
\end{table}

\begin{table}[h]
\caption{Terms for the Liberal Democrat Party (\textbf{LIB}) -- 62 terms}
\label{table_libdem}
\renewcommand{\arraystretch}{1.2}
\footnotesize
\centering
\begin{tabular}{cccc}
\toprule
\#Clegg                  & Fiona Hall       & LIBERAL               & Norman Backer\\
\#clegg                  & Greg Mulholland  & Liberal               & Norman Lamb\\
\#liberal                & Jenny Willott    & Liberal Democrat      & Paul Burstow\\
\_Ed Davey               & Jeremy Browne    & Lord McNally          & President Clegg\\
\_Jo Swinson             & John Hemming     & Lord Shutt            & Sarah Teather\\
Alistair Carmichael     & John Pugh         & Lorely Burt           & Simon Hughes\\
Andrew Stunell          & John Thurso       & Lynne Featherstone    & Steve Webb\\
Charles Kennedy         & Julian Huppert    & MP Clegg              & Tessa Munt\\
Chris Huhne             & Leader Clegg      & Mark Hunter           & Tim Farron\\
Dan Rogerson            & Lembit Opik       & Martin Horwood        & Tom McNally\\
Danny Alexander         & Lib\_ \_Dem       & Michael Crockart      & Vince Cable\\
David Heath             & LIBDEM            & Mike Crockart         & Vincent Cable\\
David Shutt             & LibDem            & Mike Hancock          & vote Clegg\\
debate Clegg            & Libdem            & Nicholas Clegg        & vote liberal\\
Duncan Hames            & libDem            & Nick Clegg            &\\
Edward Davey            & libdem            & Nick Harvey           &\\
\bottomrule
\end{tabular}
\end{table}